\UseRawInputEncoding
\documentclass[10pt]{report}

\usepackage[table]{xcolor}
\usepackage{mathrsfs,amssymb,amsmath}
\usepackage{graphicx}
\usepackage{hyperref} 
\usepackage{multirow} 
\usepackage{xspace}
\usepackage{booktabs}
\usepackage{enumitem} 
\usepackage{subcaption} 
\usepackage{longtable}
\usepackage[htt]{hyphenat} 
\usepackage{dirtree}
\usepackage{eurosym}

\hypersetup{colorlinks,linkcolor={gray!50!black},citecolor={blue!50!black},urlcolor={blue!80!black}}

\usepackage[toc,page]{appendix}
\usepackage[top=1.5in, bottom=1.5in, left=1.41in, right=1.41in]{geometry}
\usepackage{titlesec}
\newcommand{\chapnumfont}{\usefont{T1}{pnc}{b}{n}\fontsize{57}{100}\selectfont}
\colorlet{chapnumcol}{gray!75}  
\titleformat{\chapter}[display]{\filleft\bfseries}{\filleft\chapnumfont\textcolor{gray!75}{\thechapter}}{-24pt}{\Huge}
\setlist{topsep=2.2pt,itemsep=0.5pt} 
\usepackage{etoolbox}
\makeatletter
\patchcmd{\ttlh@hang}{\parindent\z@}{\parindent\z@\leavevmode}{}{}
\patchcmd{\ttlh@hang}{\noindent}{}{}{}
\makeatother

\usepackage{tikz}
\usetikzlibrary{shapes,calc,positioning,automata,arrows,trees}
\usepackage[tikz]{bclogo}
\renewcommand\logowidth{15pt}

\newcommand\bcludii{\includegraphics[width=\logowidth]{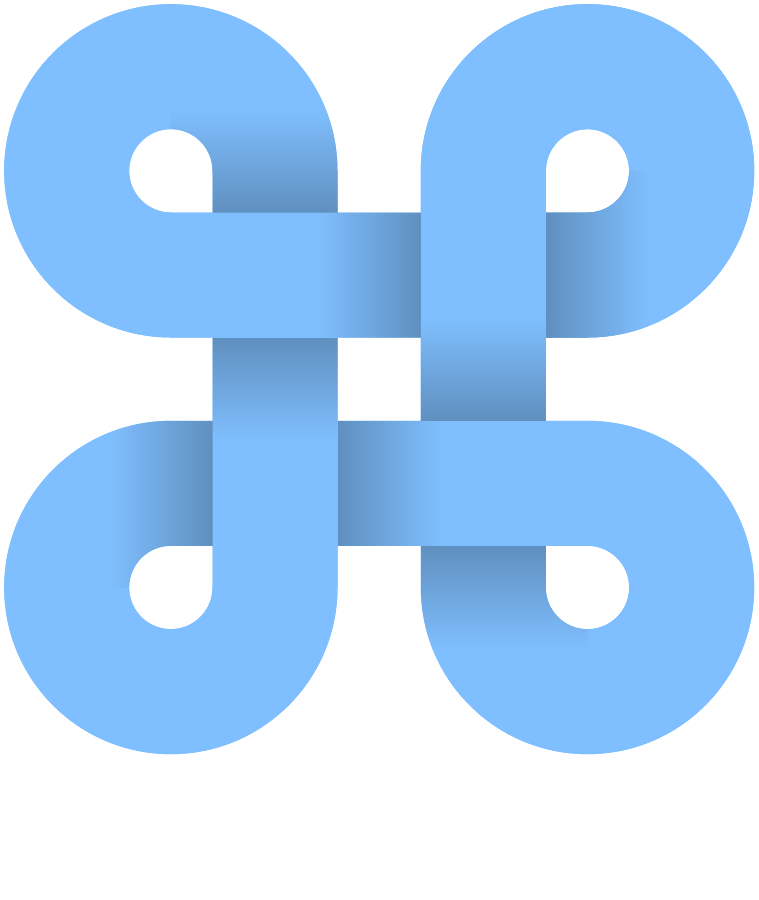}} 
\newcommand\bcjava{\includegraphics[width=\logowidth]{figs/java-icon-2b.pdf}} 
\newcommand\bcgrammar{\includegraphics[width=\logowidth]{figs/folder-icon-3.png}} 
\usepackage[skins,breakable,xparse]{tcolorbox} 
\tikzset{
  dirtree/.style={
    grow via three points={one child at (0.8,-0.7) and two children at (0.8,-0.7) and (0.8,-1.45)}, 
    edge from parent path={($(\tikzparentnode\tikzparentanchor)+(.4cm,0cm)$) |- (\tikzchildnode\tikzchildanchor)}, growth parent anchor=west, parent anchor=south west},
}
\usepackage{tikz-uml}
\usepackage{tikz-er2}

\usepackage{listingsutf8}

\lstdefinelanguage{ludii}{
  keywords={aPI,card,component,die,piece,domino,path,tile,board,boardless,surakartaBoard,hashiBoard,puzzleBoard,track,container,deck,dice,hand,equipment,item,dominoes,hints,map,regions,all,allDiceEqual,allDiceUsed,allPassed,allType,allClaimed,baseBooleanFunction,booleanConstant,booleanFunction,can,canMove,canType,all,allDifferent,allPuzzleType,forAll,isUnique,is,isPuzzleGraphType,isPuzzleRegionResultType,isPuzzleSimpleType,isCount,isSum,allConnected,isProduct,isShape,sameParity,isSolved,isThreatened,isWithin,isBlocked,isConnected,isCrossing,isLastFrom,isLastTo,isIn,isInvisible,isMasked,isVisible,isAnyDie,isEven,isFlat,isOdd,isPipsMatch,isSidesMatch,isVisited,is,isComponentType,isConnectType,isEdgeType,isGraphType,isIndexPlayerType,isIntegerType,isInType,isLineType,isLoopType,isPathType,isPlayerType,isRegularGraphType,isRelationType,isSimpleType,isSiteType,isStringType,isTargetType,isTreeType,isLine,isLoop,isPath,isEnemy,isFriend,isMover,isNext,isPrev,isTriggered,isRegularGraph,isRelated,isCycle,isFull,isPending,isEmpty,isOccupied,isDecided,isProposed,isTarget,isCaterpillerTree,isSpanningTree,isTree,isTreeCentre,isPlanarGraph,isPlanarGraph2,and,equals,ge,gt,if,le,lt,not,notEqual,or,xor,no,noMoves,noType,was,wasPass,wasType,directions,directionsFunction,if,baseFloatFunction,floatConstant,floatFunction,baseGraphFunction,basis,brick,brickShapeType,diamondOrPrismOnBrick,spiralOnBrick,squareOrRectangleOnBrick,celtic,customOnHex,diamondOnHex,hex,hexagonOnHex,hexShapeType,rectangleOnHex,starOnHex,triangleOnHex,customOnMesh,mesh,morris,quadhex,customOnSquare,diagonalsType,diamondOnSquare,rectangleOnSquare,square,squareShapeType,tiling,tiling31212,tiling333333_33434,tiling33336,customOn33344,tiling33344,tiling33434,customOn3464,diamondOn3464,hexagonOn3464,parallelogramOn3464,rectangleOn3464,starOn3464,tiling3464,tiling3464ShapeType,triangleOn3464,customOn3636,tiling3636,tiling4612,customOn488,squareOrRectangleOn488,tiling488,tilingType,customOnTri,diamondOnTri,hexagonOnTri,rectangleOnTri,starOnTri,tri,triangleOnTri,triShapeType,circle,rectangle,repeat,shape,shapeTypeStar,spiral,wedge,graphFunction,add,clip,complete,dual,hole,intersect,keep,makeFaces,merge,remove,renumber,rotate,scale,shift,skew,splitCrossings,subdivide,trim,union,baseIntArrayFunction,intArrayConstant,intArrayFunction,difference,degrees,groupSizes,rotations,baseIntFunction,ahead,centrePoint,column,coord,cost,handSite,id,layer,mapEntry,phase,row,where,card,cardSimpleType,cardSiteType,cardTrumpSuit,cardRank,cardSuit,cardTrumpRank,cardTrumpValue,groupProduct,tree,define,option,ruleset,rulesets,between,edge,from,hint,level,site,to,track,var,countPieces,countPips,count,countComponentType,countSimpleType,countSiteType,countSpaceType,countStepsType,countActive,countCells,countColumns,countEdges,countMoves,countMovesThisTurn,countPhases,countPlayers,countRows,countTrials,countTurns,countVertices,countAdjacent,countDiagonal,countNeighbours,countNumber,countOff,countOrthogonal,countSites,countGroups,countLiberties,countSteps,countDegreeProduct,countLeaves,countNonLeaves,countNonLeavesDegree,countTreeDepth,countTreeSize,face,pips,intConstant,intFunction,last,lastFrom,lastTo,lastType,matchScore,abs,add,div,if,max,min,mod,mul,pow,sub,sizeGroup,sizeTerritory,sizeStack,size,sizeGroupType,sizeSiteType,sizeTerritoryType,topLevel,amount,counter,mover,next,previous,score,state,what,who,pathExtent,trackSiteMove,trackSiteEndTrack,trackSite,trackSiteMoveType,trackSiteType,valuePiece,valuePlayer,valuePending,value,valueComponentType,valuePlayerType,valueSimpleType,tree,triangleGroupCount,baseRangeFunction,exact,max,min,range,rangeFunction,baseRegionFunction,forEach,difference,expand,if,intersection,union,regionConstant,regionFunction,sitesAround,sitesCoords,sitesCrossing,sitesCustom,sitesDirection,sitesDistance,sitesAngled,sitesAxial,sitesHorizontal,sitesSlash,sitesSlosh,sitesVertical,sitesGroup,sitesIncident,sitesCell,sitesColumn,sitesEdge,sitesEmpty,sitesPhase,sitesRow,sitesState,sitesLineOfSight,sitesOccupied,sitesEquipmentRegion,sitesHand,sitesInvisible,sitesMasked,sitesStart,sitesTrack,sitesVisible,sitesWinning,sitesSide,sitesBoard,sitesBottom,sitesCentre,sitesConcaveCorners,sitesConvexCorners,sitesCorners,sitesHint,sitesInner,sitesLastFrom,sitesLastTo,sitesLeft,sitesLineOfPlay,sitesMajor,sitesMinor,sitesOuter,sitesPending,sitesPlayable,sitesRight,sitesToClear,sitesTop,sites,sitesAroundType,sitesCrossingType,sitesDirectionType,sitesDistanceType,sitesEdgeType,sitesGroupType,sitesIncidentType,sitesIndexType,sitesLineOfSightType,sitesOccupiedType,sitesPlayerType,sitesSideType,sitesSimpleType,sitesTrackType,sitesWalk,borderSites,game,games,match,subgame,mode,player,players,baseEndRule,byScore,end,endRule,forEach,if,result,automove,meta,metaRule,noRepeat,swap,nextPhase,phase,baseMoves,decisions,move,moveBetType,moveFromToType,moveHopType,moveLeapType,moveMessageType,movePromoteType,moveRemoveType,moveSelectType,moveSetType,moveShootType,moveSimpleType,moveSiteType,moveSlideType,moveStepType,add,apply,bet,claim,effects,fromTo,hop,leap,note,attract,custodial,intervene,deal,directionCapture,opposite,before,after,enclose,flip,allCombinations,and,append,if,or,push,roll,setDirection,setCounter,setVar,setNextPlayer,setPending,setScore,setValue,set,setDirectionType,setNextPlayerType,setPendingType,setPlayerType,setRegionType,setSiteType,setTrumpType,setValueType,setCount,setState,setInvisible,setMasked,setVisible,setTrumpSuit,sow,addScore,moveAgain,rememberState,shiftPlayers,swapPlayers,swapPieces,swap,swapPlayersType,swapSitesType,surrounded,takeControl,takeDomino,take,takeControlType,takeSimpleType,then,trigger,pass,playCard,promote,propose,remove,avoidStoredState,do,firstMoveOnTrack,maxDistance,max,maxDistanceType,maxMovesType,maxCaptures,maxMoves,priority,satisfy,select,shoot,slide,step,vote,forEachDie,forEachDirection,forEach,forEachDieType,forEachDirectionType,forEachPieceType,forEachPlayerType,forEachSiteType,forEachPiece,forEachPlayer,forEachSite,moves,play,rule,rules,deal,set,placeItem,place,placeRandomType,placeStackType,placeRandom,placeCustomStack,placeMonotonousStack,setAmount,setScore,setTeam,set,setStartGraphType,setStartPlayersType,setStartPlayerType,setStartSitesType,allInvisible,setCost,setCount,setPhase,setSite,split,splitType,start,startRule,basisType,landmarkType,puzzleElementType,regionTypeDynamic,regionTypeStatic,relationType,shapeType,siteType,stepType,tilingBoardlessType,trackType,cardType,dealableType,suitType,dummy,modeType,repetitionType,resultType,roleType,whenType,gameType,noType,absoluteDirection,compassDirection,direction,directionType,directionUniqueName,relativeDirection,rotationalDirection,spatialDirection,dummy,score,card,hint,region,values,edge,face,graph,graphElement,itemScore,measureGraph,perimeter,poly,properties,radial,radials,step,steps,trajectories,vertex,count,pair,between,flips,from,to,what,who,ai,aIItem,features,featureSet,heuristics,centreProximity,cornerProximity,currentMoverHeuristic,heuristicTerm,lineCompletionHeuristic,material,mobilitySimple,ownRegionsCount,playerRegionsProximity,playerSiteMapCount,regionProximity,score,sidesProximity,divNumBoardSites,divNumInitPlacement,heuristicTransformation,logisticFunction,tanh,bestAgent,pair,board,boardBooleanType,boardColourType,boardShapeType,boardStyleThicknessType,boardStyleType,boardCheckered,boardColour,boardShape,boardStyle,boardStyleThickness,graphics,graphicsItem,noAnimation,noBoard,noCurves,noHandScale,noMaskedColour,no,noBooleanType,adversarialPuzzle,stackType,suitRanking,pieceColour,pieceColourFromState,pieceFamilies,pieceBackground,pieceForeground,pieceAddStateToName,pieceExtendName,pieceRename,piece,pieceColourFromStateType,pieceColourType,pieceFamiliesType,pieceGroundType,pieceNameType,pieceReflectType,pieceScaleByType,pieceScaleType,pieceStyleType,pieceReflect,pieceScale,pieceScaleByValue,pieceStyle,playerColour,player,playerColourType,regionColour,region,regionColourType,showCost,showPits,showPlayerHoles,showRegionOwner,showCheck,showPieceState,showPieceValue,showEdges,showScore,show,showBooleanType,showCheckType,showComponentDataType,showComponentType,showEdgeType,showScoreType,showSiteDataType,showSiteType,showSymbolType,showSitesAsHoles,showSitesShape,showSymbol,boardGraphicsType,colour,colourRoutines,userColourType,componentStyleType,containerStyleType,controllerType,edgeInfoGUI,edgeType,lineStyle,metadataFunctions,metadataImageInfo,pieceStackType,valueLocationType,whenScoreType,aliases,author,classification,credit,date,description,origin,publisher,rules,source,version,info,infoItem,metadata,metadataItem},
  basewidth  = {.6em,0.6em},
  keywordstyle=\color{mblue}\bfseries,
  ndkeywords={Off,End,Undefined,DiceUsed,DiceEqual,Passed,Result,type,Move,Different,Unique,Count,Sum,Solved,Threatened,Within,Connected,Blocked,Crossing,LastFrom,LastTo,Visible,Masked,Invisible,Odd,Even,Visited,SidesMatch,PipsMatch,Flat,AnyDie,In,Line,Loop,Path,Mover,Next,Prev,Friend,Enemy,Triggered,RegularGraph,Related,Cycle,Pending,Full,Empty,Occupied,Proposed,Decided,Target,Tree,SpanningTree,CaterpillerTree,TreeCentre,Moves,Pass,Square,Rectangle,Diamond,Prism,Spiral,Limping,NoShape,Square,Rectangle,Diamond,Triangle,Hexagon,Star,Limping,Prism,Implied,Solid,Alternating,Concentric,Radiating,NoShape,Square,Rectangle,Diamond,Limping,Custom,Square,Rectangle,Diamond,Prism,Triangle,Hexagon,Star,Limping,T31212,T3464,T488,T33434,T33336,T33344,T3636,T4612,T333333_33434,NoShape,Square,Rectangle,Diamond,Triangle,Hexagon,Star,Limping,Prism,Star,TrumpSuit,Rank,Suit,TrumpValue,TrumpRank,Pieces,Pips,Rows,Columns,Turns,Moves,Trials,MovesThisTurn,Phases,Vertices,Edges,Cells,Players,Active,Sites,Adjacent,Neighbours,Orthogonal,Diagonal,Off,Groups,Liberties,Steps,To,From,Group,Stack,Territory,Move,EndSite,Piece,Player,Pending,Around,Crossing,Direction,Distance,Axial,Horizontal,Vertical,Angled,Slash,Slosh,Group,Incident,Row,Column,Phase,Cell,Edge,State,Empty,LineOfSight,Occupied,Hand,Start,Track,Winning,Visible,Masked,Invisible,Side,Board,Top,Bottom,Left,Right,Inner,Outer,Corners,ConcaveCorners,ConvexCorners,Major,Minor,Centre,Hint,ToClear,LineOfPlay,Pending,Playable,LastTo,LastFrom,Track,Bet,FromTo,Hop,Leap,Propose,Vote,Promote,Remove,Select,Set,Shoot,Pass,PlayCard,Add,Claim,Slide,Step,Direction,NextPlayer,Pending,Value,Score,Visible,Masked,Invisible,Count,State,TrumpSuit,Counter,Var,Players,Pieces,Control,Domino,Distance,Moves,Captures,Die,Direction,Piece,Player,Site,Random,Stack,AllInvisible,Team,Amount,Score,Count,Cost,Phase,Deck,NoBasis,Triangular,Square,Hexagonal,T33336,T33344,T33434,T3464,T3636,T4612,T488,T31212,T333333_33434,SquarePyramidal,HexagonalPyramidal,Wheel,Circle,Spiral,Dual,Brick,Mesh,Morris,Celtic,QuadHex,CentreSite,LeftSite,RightSite,Topsite,BottomSite,FirstSite,LastSite,Cell,Edge,Vertex,Hint,Empty,NotEmpty,Own,NotOwn,Enemy,NotEnemy,AllPlayers,Rows,Columns,AllDirections,HintRegions,Layers,Diagonals,SubGrids,Regions,Vertices,Corners,Sides,SidesNoCorners,AllSites,Touching,Orthogonal,Diagonal,Off,Adjacent,All,NoShape,Custom,Square,Rectangle,Triangle,Hexagon,Cross,Diamond,Prism,Quadrilateral,Rhombus,Wheel,Circle,Spiral,Wedge,Star,Limping,Polygon,Vertex,Edge,Cell,F,B,L,R,Square,Triangular,Hexagonal,Track,Joker,Ace,Two,Three,Four,Five,Six,Seven,Eight,Nine,Ten,Jack,Queen,King,Dominoes,Cards,Clubs,Spades,Diamonds,Hearts,Alternating,Simultaneous,InTurn,InGame,Positional,Situational,Infinite,Win,Loss,Draw,Tie,Abandon,Crash,Neutral,P1,P2,P3,P4,P5,P6,P7,P8,P9,P10,P11,P12,P13,P14,P15,P16,Team1,Team2,Team3,Team4,Team5,Team6,Team7,Team8,Team9,Team10,Team11,Team12,Team13,Team14,Team15,Team16,Each,Shared,All,Any,Mover,Next,Prev,NonMover,Enemy,Ally,NonAlly,Partner,NonPartner,NonNeutral,Player,StartOfMove,EndOfMove,StartOfTurn,EndOfTurn,StartOfRound,EndOfRound,StartOfPhase,EndOfPhase,StartOfGame,EndOfGame,StartOfMatch,EndOfMatch,StartOfSession,EndOfSession,Sites,Moves,All,Angled,Adjacent,Axial,Orthogonal,Diagonal,Off,SameLayer,Upward,Downward,Rotational,N,E,S,W,NE,SE,NW,SW,NNW,WNW,WSW,SSW,SSE,ESE,ENE,NNE,CW,CCW,In,Out,U,UN,UNE,UE,USE,US,USW,UW,UNW,D,DN,DNE,DE,DSE,DS,DSW,DW,DNW,N,NNE,NE,ENE,E,ESE,SE,SSE,S,SSW,SW,WSW,W,WNW,NW,NNW,N,NNE,NE,E,SSE,SE,S,SSW,SW,W,NW,NNW,WNW,ENE,ESE,WSW,CW,Out,CCW,In,UNW,UNE,USE,USW,DNW,DNE,DSE,DSW,U,UN,UW,UE,US,D,DN,DW,DE,DS,Forward,Backward,Rightward,Leftward,Forwards,Backwards,Rightwards,Leftwards,FL,FLL,FLLL,BL,BLL,BLLL,FR,FRR,FRRR,BR,BRR,BRRR,SameDirection,OppositeDirection,Out,CW,In,CCW,D,DN,DNE,DE,DSE,DS,DSW,DW,DNW,U,UN,UNE,UE,USE,US,USW,UW,UNW,Checkered,Colour,Shape,StyleThickness,Style,Board,Animation,HandScale,Curves,MaskedColour,ColourFromState,Colour,Families,Background,Foreground,Rename,ExtendName,AddStateToName,Reflect,ByValue,Scale,Style,Colour,Colour,Pits,PlayerHoles,RegionOwner,Cost,Check,State,Value,Piece,Edges,Score,Shape,AsHoles,Sites,Cell,Symbol,InnerEdges,OuterEdges,Phase0,Phase1,Phase2,Phase3,Symbols,Vertices,White,Black,Grey,LightGrey,VeryLightGrey,DarkGrey,VeryDarkGrey,Dark,Red,Green,Blue,Yellow,Pink,Cyan,Brown,DarkBrown,Purple,Turquoise,Orange,DarkOrange,LightRed,DarkRed,LightGreen,DarkGreen,LightBlue,VeryLightBlue,DarkBlue,IceBlue,Gold,Silver,Bronze,GunMetal,HumanLight,HumanDark,Cream,DeepPurple,PinkFloyd,BlackSabbath,KingCrimson,MoodyBlues,TangerineDream,Piece,Tile,Card,Die,Domino,LargePiece,ExtendedShogi,Board,Hand,Deck,Dice,Boardless,ConnectiveGoal,Mancala,PenAndPaper,Pyramidal,Spiral,Isometric,Puzzle,GraphPuzzle,LineSegment,PuzzlePenAndPaper,RegionPuzzle,Agon,Backgammon,Chess,ChineseCheckers,Connect4,Goose,Go,Graph,HoundsAndJackals,Janggi,Lasca,Pachisi,Ploy,Scripta,Shogi,SnakesAndLadders,Surakarta,Tafl,Xiangqi,UltimateTicTacToe,Futoshiki,Hashi,Kakuro,Sudoku,BasicController,PyramidalController,All,Inner,Outer,Interlayer,Thin,Thick,ThinDotted,ThickDotted,ThinDashed,ThickDashed,Hidden,Default,Ground,Reverse,Fan,None,Backgammon,Ring,None,Corner,Middle,Always,Never,AtEnd},
  ndkeywordstyle=\color{dviolet}\bfseries,
  identifierstyle=\color{black},
  sensitive=true,   
  comment=[l]{//},
  commentstyle=\color{dred}\ttfamily,
  stringstyle=\color{dgreen}\ttfamily,
  morestring=[b]',
  morestring=[b]",
  escapechar=@,
  showstringspaces=false,
  xleftmargin=1pt,xrightmargin=1pt,
  breaklines=true,basicstyle=\ttfamily\small,backgroundcolor=\color{colorex},inputencoding=utf8/latin9,texcl
}

\lstdefinelanguage{syntax}{
  keywords={game,metadata,control,rules,keyword,text,time,play,player,equipment,board,square,moves,to,indexOf,empty,end,line,result,index,mode,modeType,compassDirection,tiling,directions,directionChoice,playout,model,addToEmpty,byPiecePlayout,
  get,item,component,container,disc,ball,colour,hand,owner,numItems,modify,wheel,hexHex,rect,spokes,aligned,dim,cols,join,cut,remove,cellA,
  cellB,list,end,start,place,who,count,target,posn,fromTo,or,action,regionFunction,sites,resultType,line,not,occupied,store,where,what,class,
  arg,subClass,terminal,add,roleType,knight,cross,bishop,king,queen,pawn,rook,piece,card,die,letter,number,tile,chess,timeType,basis,label,
  rows,players,name,and,dirn,dirnType,length,sitesFrom,sitesTo,functions,cont,site,role,num,types,cell,alternatingMove,realTime,simultaneousMove,map,regions,track,deck,dominoes,cardType,byPiece,byPieceType,captureByApproach,captureByWithdraw,checkmate,checkMove,hop,leap,pass,promotion,select,shoot,slide,step,constraints,flanked,flip,forDirn,if,observe,pending,priority,replay,roll,setCounter,setDouble,setOwner,setScore,setState,sow,surrounded,then,top,variableType,region,included,exception,recursive,allSites,around,regionType,borderRegion,bottom,centre,column,corners,difference,directionRegion,edge,edgeFace,edgeVertex,exteriors,facesEdge,hintRegion,own,phase,playable,row,set,stateRegion,union,verticeFace,walk,direction,stepType,other,hint,values,puzzleConstraints,allDifferent,excepts,except,connected,areaType,crossing,equalParities,forEach,mult,shape,shapeType,solved,sum,unique,decision,decisionPuzzle,large,domino,largePiece,lPiece,animal,arrow,checkers,dominoHalf,dot,hands,hex,pill,senetPiece,shogi,pieceState,stratego,tafl,triangle,urPiece,xiangqi,arrow,camel,cat,cow,dog,elephant,goat,horse,leopard,rabbit,sheep,tiger,wolf,dame,man,manStar,mann,paper,rock,scissors,fuhyo,ginsho,hisha,kakugyo,keima,kinsho,kyosha,narigin,narikei,narikyo,osho,ryuma,ryuo,tokin,state,flipDisc,realTimePiece,real_time,bike,bomb,captain,colonel,flag,general,lieutenant,major,marshal,miner,scout,sergeant,spy,jarl,thrall,jiang,ju,ma,pao,shi,xiang,zu,alquerqueBoard,arimaaBoard,backgammonBoard,boardless,cardBoard,chessBoard,chineseCheckersBoard,connect4Board,goBoard,graphBoard,halmaBoard,hexBoard,hexYBoard,mancalaBoard,morrisBoard,peralikatumaBoard,reversiBoard,senetBoard,shogiBoard,snakesAndLaddersBoard,solitaireBoard,strategoBoard,surakartaBoard,taflBoard,urBoard,wheelBoard,xiangqiBoard,puzzleBoard,joinDiago,joinOrtho,modifyType,cutAll,cutDiago,cutOrtho,futoshikiBoard,graphPuzzleBoard,kakuroBoard,lineSegmentBoard,nonogramBoard,regionPuzzleBoard,sudokuBoard,dotBoard,hashiBoard,handDice,startRule,deal,fill,fogOfWar,hints,initScore,placeRandomly,setCount,byScore,abs,counter,directionSite,double,forwardOnTrack,from,height,lastEdgeMove,lastFromMove,lastToMove,level,max,min,mod,mover,next,posPiece,previous,replayCount,score,size,turn,value,le,adjacent,allPass,canMove,configuration,connect,contains,encircle,equal,even,full,ge,gt,in,isCheckmate,isEnemy,isFriend,isFriendAt,isMover,isNext,isPending,isPiece,isPrev,isState,know,le,lt,odd,reachedRegion,stalemated,threatened,visited,sub
},
  keywordstyle=\color{dblue}\slshape,ndkeywords={String,int,boolean,string,ints,booleans,integer,Integer,Boolean},
  ndkeywordstyle=\color{dviolet}\bfseries,
  basewidth  = {.5em,0.5em},
  escapechar=@,
  xleftmargin=1pt,xrightmargin=1pt,
  breaklines=true,basicstyle=\ttfamily\linespread{1.0}\small,backgroundcolor=\color{colorsy},inputencoding=utf8/latin9,texcl
}

\definecolor{javared}{rgb}{0.6,0,0} 
\definecolor{javagreen}{rgb}{0.25,0.5,0.35} 
\definecolor{javapurple}{rgb}{0.5,0,0.35} 
\definecolor{javadocblue}{rgb}{0.25,0.35,0.75} 
 
\newcommand{\core}[1]{ 
  \medskip \begin{tcolorbox}[
    enhanced,
    breakable,
    before={},
    boxsep=0pt,top=0pt,bottom=0pt,left=7mm,right=1mm,
    toprule=0.1mm,leftrule=0.1mm,rightrule=0.25mm,bottomrule=0.25mm,shadow={0.2mm}{-0.2mm}{0mm}{dgray},
    overlay unbroken and first={\node (logo) at ([xshift=6mm,yshift=-5mm]frame.north west) {#1}; 
    },
    colframe=colorex,titlerule=-0.2mm,toptitle=3mm,coltitle=black,fonttitle=\bfseries,
    lines before break=1000
}

\newcommand{\corenonbreakable}[1]{ 
  \medskip \begin{tcolorbox}[
    enhanced,
    before={},
    boxsep=0pt,top=0pt,bottom=0pt,left=7mm,right=1mm,
    toprule=0.1mm,leftrule=0.1mm,rightrule=0.25mm,bottomrule=0.25mm,shadow={0.2mm}{-0.2mm}{0mm}{dgray},
    overlay unbroken and first={\node (logo) at ([xshift=6mm,yshift=-5mm]frame.north west) {#1}; 
    },
    colframe=colorex,titlerule=-0.2mm,toptitle=3mm,coltitle=black,fonttitle=\bfseries,
}

\newcounter{cntJa}
\newcounter{cntEx}
\newcounter{cntSy}

\ifx\bw\undefined
  \lstnewenvironment{java}[1][]{\lstset{language=java,#1}}{}
  \lstnewenvironment{ludii}[1][]{\lstset{language=ludii,#1}}{} 
  \lstnewenvironment{syntax}{\lstset{language=syntax}}{} 
  
  \newenvironment{boxex}
    {\stepcounter{cntEx} \core{\bcludii} ,colback=colorex,title style={color=colorex},title=~ Ludeme \thecntEx]}
    {\end{tcolorbox}} 
    
  \newenvironment{boxexnonbreakable}
    {\stepcounter{cntEx} \corenonbreakable{\bcludii} ,colback=colorex,title style={color=colorex},title=~ Ludeme \thecntEx]}
    {\end{tcolorbox}} 
    
\definecolor{v2lgray}{gray}{0.85}
\definecolor{dgray}{rgb}{0.4,0.4,0.4}
\definecolor{dblue}{RGB}{0,0,99}
\definecolor{dred}{RGB}{150,6,54}
\definecolor{dgreen}{RGB}{47,135,7}
\definecolor{dviolet}{RGB}{102,0,153}
\definecolor{mblue}{RGB}{0,0,180}
\definecolor{colorja}{RGB}{255,248,220}
\definecolor{colorex}{HTML}{DFEFFF}  
\definecolor{colorsy}{HTML}{F2F2F2}
\definecolor{grey1}{rgb}{0.9,0.9,0.9}

\definecolor{grey}{rgb}{0.75,0.75,0.75}

\usepackage{caption} 
\usepackage{listings}
\usepackage[export]{adjustbox}
\usepackage{cleveref}
\usepackage{subcaption}
\usepackage{apacite}
\usepackage{hyperref}

\usepackage{longtable}

\usepackage{chngcntr}
\counterwithout{footnote}{chapter}

\newcolumntype{M}[1]{>{\centering\arraybackslash}m{#1}}

\usepackage{tikz-uml}

\begin{document}

\thispagestyle{empty}

\begin{centering}

\includegraphics[scale=0.35]{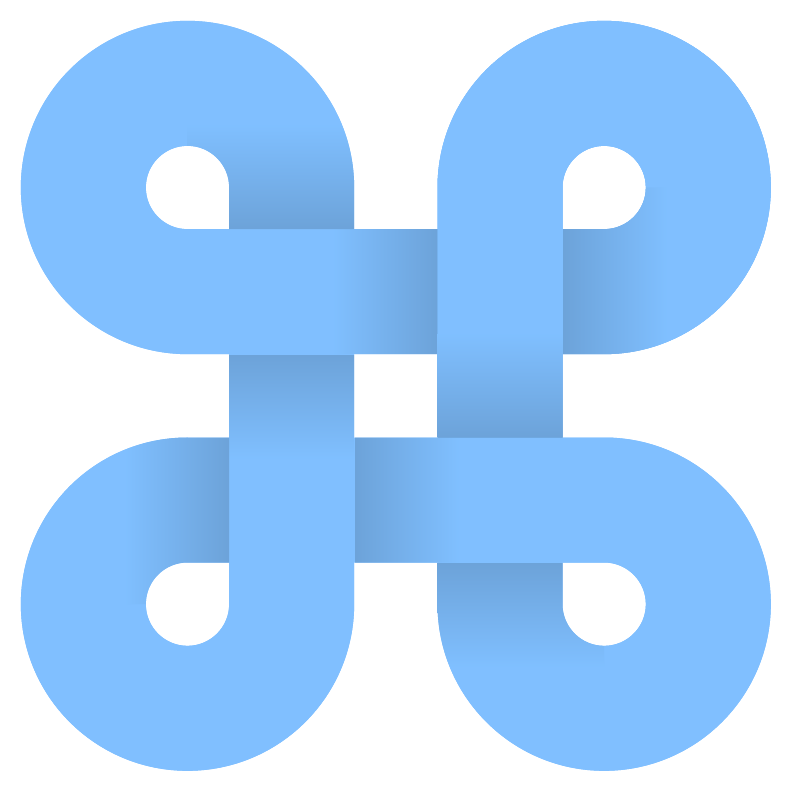}

\vspace{30mm}
\noindent\rule{14.5cm}{0.5pt}

\vspace{5mm}
{\Huge \bf Ludii Game Logic Guide}
 
\vspace{4mm}
{\Large \bf Ludii Version 1.3.4}

\vspace{2mm}
\noindent\rule{14.5cm}{0.5pt}
 
\vspace{10mm}
{\Large {\'E}ric Piette, Cameron Browne and Dennis J. N. J. Soemers}
 
\vspace{10mm}
{\large Department of Data Science and Knowledge Engineering (DKE)}
 
\vspace{1mm}
{\large Maastricht University}

\vspace{1mm}
{\large Maastricht, the Netherlands}

\vspace{10mm}
{\large \today}

\pagebreak

\end{centering}

\begin{centering}


{\Large \bf Introduction}

\end{centering}

\phantom{}

\noindent
This technical report outlines the fundamental workings of the game logic behind Ludii -- a ``general game system'' that can be used to play a wide variety of games. Ludii is a program developed for the ERC-funded Digital Ludeme Project, in which mathematical and computational approaches are used to study how games were played, and spread, throughout history. The source code of Ludii is available at \url{https://github.com/Ludeme/Ludii} 

This report explains how general game states and equipment are represented in Ludii, and how the rule {\it ludemes}\footnote{Ludemes are ``game memes'' or units of game-related information~\cite{Parlett2016}.} dictating play are implemented ``behind the scenes'', giving some insight into the core game logic behind the Ludii general game player. 
This guide is intended to help game designers using the Ludii game description language~\cite{LLR:2020} to understand it more completely and make fuller use of its features when describing their games. 

Up-to-date news on Ludii and the Digital Ludeme Project can be found on the relevant websites and Twitter pages:
\begin{itemize}
    \item \textbf{Ludii website}: \hfill \url{http://ludii.games/}
    \item \textbf{Ludii Twitter}: \hfill \url{https://twitter.com/ludiigames}
    \item \textbf{Digital Ludeme Project website}: \hfill \url{http://www.ludeme.eu/} 
    \item \textbf{Digital Ludeme Project Twitter}: \hfill \url{https://twitter.com/archaeoludology/}
\end{itemize}

\vfill

\noindent
Please contact the authors with any comments or corrections at: {\tt ludii.games@gmail.com}

\pagebreak

\tableofcontents

\pagebreak


\chapter{\textcolor{gray!95}{Overview}}\label{Chapter:Overview}


Ludii~\cite{Piette2020Ludii} is a complete general game system implemented in Java. 
The objective of this report is to bring to light the core elements of Ludii used for  game representation, including the code logic of each class in a game description. 

Ludii uses a {\it class grammar} approach~\cite{Browne2016ClassGrammar} to give a direct correspondence between the keywords in its game descriptions and the underlying Java code that implements them, using Java Reflection. 
The game description language is generated from the constructors in the class hierarchy of the Ludii source code. Game descriptions expressed in the grammar are automatically instantiated back into the corresponding library code for compilation, giving a guaranteed 1:1 mapping between the source code and the grammar. 

The implementation details are hidden from the user, who only sees the simplified grammar, which summarises the code to be called. The full grammar--with full documentation--is available in the Ludii Language Reference, which may be downloaded from \url{https://ludii.games/downloads/LudiiLanguageReference.pdf}. 


\chapter{\textcolor{gray!95}{Game Representation}}\label{Chapter:GameRepresentation}


\section{A Ludemic Approach}

Ludii uses a {\it ludemic} approach to game design, first outlined in~\cite{Browne2009}. 
The core of Ludii is a ludeme library, consisting of a number of classes, each implementing a specific ludeme. 
A {\it ludeme} is a conceptual unit of game-related information, which we use to define both the {\it form} of the game (its rules and equipment) and its {\it function} (legal moves and outcomes for the end state).

The game description for a Ludii implementation of a game is written in a .lud file, which is a tree of ludemes. Each term in the .lud file corresponds to a class in the core of Ludii or to a parameter used in a class such as a string, an integer, an array or an enumeration type. The language used to describe games for Ludii is automatically derived from the ludeme classes available in Ludii's .jar file. \textbf{Ludeme 1} below is an example of the complete Ludii game description for Amazons\footnote{Amazons: \href{https://en.wikipedia.org/wiki/Game_of_the_Amazons}{Wikipedia}} in the EBNF-style grammar used by Ludii.

\begin{boxex}
\begin{ludii}
(game "Amazons"  
    (players 2)  
    (equipment { 
        (board (square 10)) 
        (piece "Queen" Each (move Slide (then (moveAgain))))
        (piece "Dot" Neutral)
    })  
    (rules 
        (start { 
            (place "Queen1" {"A4" "D1" "G1" "J4"})
            (place "Queen2" {"A7" "D10" "G10" "J7"})
        })
        
        (play 
            (if (is Even (count Moves))
                (forEach Piece)
                (move Shoot (piece "Dot0"))
            )
        )
        
        (end (if (no Moves Next)  (result Mover Win) ) )  
    )
)
\end{ludii} 
\end{boxex}

Each term after an opening parenthesis in this tree of ludemes (for example, players, board and count) is a ludeme implemented in the core of Ludii by a corresponding Java class (for example, Players.java, Board.java, and Count.java, respectively). Terms between quotes (such as ``Queen1" or ``Queen2") are string parameters, in this case for the ludeme (place ...). The numbers 2 and 10 are integer parameters for ludemes (players ...) and (square ...), respectively. Terms between curly braces '\{' and '\}' are elements of an array, which can also be a parameter of a ludeme.

Every java class corresponding to a ludeme implements the interface Ludeme.java, which is at the top of the ludeme class hierarchy. They are organised into the following categories:




\begin{center}
\begin{tikzpicture}

\umlinterface[rectangle split parts=1,]{Ludeme}{}{}
\umlclass[rectangle split parts=1,y = 2]{Game}{}{}
\umlclass[rectangle split parts=1,x =-3, y = 2]{Mode}{}{}
\umlclass[rectangle split parts=1,x = 3, y = 2]{Players}{}{}
\umlclass[rectangle split parts=1,x = -4]{Equipment}{}{}
\umlclass[rectangle split parts=1,x =-3, y = -2]{Rules}{}{}
\umlclass[rectangle split parts=1,x = 3, y = -2]{Types}{}{}
\umlclass[rectangle split parts=1,x = 4]{Functions}{}{}
\umlclass[rectangle split parts=1,y = -3]{Auxiliary}{}{}

\umlinherit[geometry=-|]{Game}{Ludeme}
\umlinherit[geometry=|-]{Equipment}{Ludeme}
\umlinherit{Rules}{Ludeme}
\umlinherit[geometry=-|-]{Functions}{Ludeme}
\umlinherit{Types}{Ludeme}
\umlinherit[geometry=-|]{Auxiliary}{Ludeme}
\umlinherit{Players}{Ludeme}
\umlinherit{Mode}{Ludeme}
\end{tikzpicture}
\end{center}

\begin{itemize}
    \item \textbf{Game}: The root ludeme of any tree of ludemes.
    \item \textbf{Players and Mode}: These two ludemes are direct members of game.
    \item \textbf{Equipment}: Any ludeme that is part of the game, such as the game board, pieces, dice, and cards.
    \item \textbf{Rules}: The ludemes representing starting, playing and ending rules. For example, the Moves ludemes are playing rules.
    \item \textbf{Functions}: The ludemes representing functions used in a game description, such as Boolean functions, Graph functions, Integer functions, Directions functions and Region functions.
    \item \textbf{Types}: This includes ludemes that are enumeration types such as Compass Direction (N, S, W, E ...), Role types (Mover, Next, P1, ...) or  Site types (Vertex, Edge, Cell).
    \item \textbf{Auxiliary}: This includes ludemes such as pair used in the Map ludeme to associate two elements or the from/between/to ludeme structures used in many Moves ludemes.
\end{itemize}

The Types ludemes are described in the Ludii Language Reference \cite{LLR:2020}. 
All the other ludemes are described below in chapters \ref{Chapter:Equipment}, \ref{Chapter:Rules} and \ref{Chapter:Functions}.


\section{Context}

In Ludii, a single object called $\it{Context}$ is used to store the game $\mathcal{G}$, its game state representation $s$, and the sequence of moves played so far called the trial $\tau$.

\begin{center}
\begin{tikzpicture}
\begin{umlpackage}{Core}

\begin{umlpackage}{util}
\umlclass[rectangle split parts=1]{Context}{}{}
\umlclass[rectangle split parts=1, x = 2, y = 1]{Trial}{}{}
\umlclass[rectangle split parts=1, x = 2, y = -1]{State}{}{}
\end{umlpackage}

\umlclass[rectangle split parts=1, y = -3]{Game}{}{}

\end{umlpackage}
\umlunicompo[geometry=-|]{Context}{Trial}
\umlunicompo[geometry=-|]{Context}{State}
\umlunicompo[geometry=|-]{Context}{Game}
\end{tikzpicture}
\end{center}

For any operation such as computing the graph of a container, computing the initial state $s_0$, or computing the legal moves for a state $s$, Ludii evaluates a tree of ludemes by calling a method $\it eval(Context$ $context)$ to evaluate it according to the current state. The $\it{Context}$ also contains the random number generator used for any stochastic operations in the corresponding trial and the value of model (Alternating, Simultaneous, or Simulation) used to apply moves or compute the legal moves in a specific game state.




The following subsections describe how a game is represented internally in Ludii. 


\section{Game}
In Ludii, a game is defined by a quintuple of ludemes $\mathcal{G} = \langle Name, Players, Mode, Equipment, Rules \rangle$. The Game ludeme (game ...) is the root of a tree of ludemes and corresponds internally in Ludii to the class $\it Game.java$.
The main members of this class correspond to each element of the tuple defining a game:

\begin{center}
\begin{tikzpicture}
\begin{umlpackage}{game}
\umlclass[rectangle split parts=2]{Game}{String name}{}
\umlclass[rectangle split parts=1,x = -2, y = 2]{Players}{}{}
\umlclass[rectangle split parts=1,x = 2, y = 2]{Mode}{}{}
\umlclass[rectangle split parts=1,x = -2, y = -2]{Equipment}{}{}
\umlclass[rectangle split parts=1,x = 2, y = -2]{Rules}{}{}
\end{umlpackage}

\umlunicompo{Game}{Players}
\umlunicompo{Game}{Mode}
\umlunicompo{Game}{Equipment}
\umlunicompo{Game}{Rules}
\end{tikzpicture}
\end{center}

\begin{itemize}
    \item \textbf{Name}: The name of the game.
    \item \textbf{Players}: Defines all information about the players of the game.
    \item \textbf{Mode}: Defines the game control mode.
    \item \textbf{Equipment}: Defines the items for the game.
    \item \textbf{Rules}: Defines the starting, playing and ending rules for the game.
\end{itemize}

Each Game instance also contains values for the following Boolean variables, which may activate different meta rules:
\begin{itemize}
    \item \textbf{automove}: Activates the meta rule to apply automatically all the legal moves only applicable to a single site (e.g. Trax\footnote{Trax: \href{https://en.wikipedia.org/wiki/Trax_(game)}{Wikipedia}}).
    \item \textbf{gravity}: Activates the meta rule to apply a specific gravity (e.g. Shibumi board\footnote{Shibumi games: \href{http://cambolbro.com/games/shibumi/shibumi-rule-book-v1.1.pdf}{Shibumi Rule Book}}).
    \item \textbf{pin}: Activates the meta rule to remove from the legal moves the ones which can remove a piece which is supported by more than one piece (e.g. Shibumi board\footnote{Shibumi games: \href{http://cambolbro.com/games/shibumi/shibumi-rule-book-v1.1.pdf}{Shibumi Rule Book}}).
    \item \textbf{repetitionType}: Activates a specific no-repetition rule (e.g. Go\footnote{Go: \href{https://en.wikipedia.org/wiki/Go_(game)}{Wikipedia}}).
    \item \textbf{usesSwapRule}: Activates the Swap rule for the game (e.g. Hex\footnote{Hex: \href{https://en.wikipedia.org/wiki/Hex_(board_game)}{Wikipedia}}).
\end{itemize}

After compilation of the game, when all the ludemes are instantiated, Ludii computes the game flags by recursively calling the method $\it computeGameFlags()$ on all the ludemes used to describe a game $\mathcal{G}$ and stores them in a long variable called $gameFlags$. 
These flags are used for a variety of purposes, such as changing how some ludemes are evaluated, calculating the correct state for a game, and optimising compilation time.
\pagebreak

Here is the list of all the game flags:
\renewcommand{\arraystretch}{1.3}
\arrayrulecolor{white}
\rowcolors{2}{gray!25}{gray!10}
\begin{longtable}{@{}p{.22\textwidth} | p{.68\textwidth}@{}}
\rowcolor{gray!50}
\textbf{Game Flag} & \textbf{Description} \\
UsesFromPositions & "On" if this game may generate moves using different 'from' and 'to' positions. \\
SiteState & "On" if the game involves a state value for a site. \\
Count & "On" if the game involves a count for a site. \\
HiddenInfo &"On" if the game has hidden info. \\
Stacking & "On" if the game is a stacking game. \\
Boardless & "On" if the game is a boardless game. \\
Stochastic & "On" if the game involves stochasticity. \\
DeductionPuzzle &"On" if the game is a deduction puzzle. \\
Score & "On" if the game involves a score. \\
Visited & "On" if the game stores all the visited sites in the same turn. \\
Simultaneous & "On" if the game has simultaneous actions. \\
ThreeDimensions & "On" if this game is a 3D game. \\
NotAllPass & "On" if the game does not end if all the players pass. \\
Card & "On" if the game uses cards. \\
LargePiece & "On" if the game has one or more large pieces. \\
SequenceCapture & "On" if the game allows multiple captures in sequence. \\
Track &"On" if the game uses tracks. \\
Rotation & "On" if the game uses rotations. \\
Team & "On" if the game has teams. \\
Bet & "On" if the game has some betting actions. \\
HashScores & "On" if the game has some hash scores. \\
HashAmounts & "On" If the game has some hash amounts. \\
HashPhases & "On"if the game has some hash phases. \\
Graph & "On" if the game is a graph game. \\
Vertex & "On" if the game is played on the vertices. \\
Cell & "On" if the game is played on the cells. \\
Edge & "On" if the game can be played on the edges. \\
Dominoes & "On" if the game has dominoes. \\
LineOfPlay & "On" if the game has a line of play used. \\
MoveAgain & "On" if the game has some replay actions. \\
Value & "On" if the game has some components with a value. \\
Vote &"On" if the game has some vote actions. \\
Note & "On" if the game has some Note actions. \\
Loops & "On" if the game involves a loop. \\
StepAdjacentDistance & "On" if the adjacent distance between sites is needed. \\
StepOrthogonalDistance &"On" if the orthogonal distance between sites is needed. \\
StepDiagonalDistance &"On" if the diagonal distance between sites is needed. \\
StepOffDistance & "On" if the off diagonal distance between sites is needed. \\
StepAllDistance & "On" if the distance between all sites is needed. \\
InternalLoopInTrack &"On" if the tracks on the game have an internal loop. \\
UsesSwapRule & "On" if the game uses a swap rule. \\
RepeatInGame & "On" if the game checks the repetition in the game. \\
RepeatInTurn & "On" if the game checks the repetition in the turn. \\
PendingValues & "On" if the game uses some pending states/values. \\
MapValue & "On" if the game uses some maps to values. \\
RememberingValues & "On" if the game uses some values to remember. \\
\end{longtable}
\rowcolors{0}{}{}


The Game object also contains two variables: maxTurnLimit and maxMovesLimit, used as a limit for the number of turns and a limit for the number of moves of a game, respectively. 
Currently the limit of moves used in Ludii is 10,000 and the limit of turns is 1,250 multiplied by the number of players. When one of these limits is reached but the game is not over, the game ends in a draw for the remaining players.


\section{State}

A game state $s_i$ corresponds to all the information of the game after $i$ moves played from the initial state $s_0$. 
According to the flags of the game $\mathcal{G}$, Ludii automatically selects the appropriate state type to ensure the most suitable representation.

\subsection{Container State}\label{section:ContainerState}

Each container of a game $\mathcal{G}$ is modelled by a graph defined by a set $C$ of cells, a set $V$ of vertices and a set $E$ of edges. We specify \textit{locations} $loc = \langle c_i, t_i, s_i, l_i \rangle$ by their container $c = \langle C, V, E \rangle$, a type of site $t_i$ $\in$ \{Cell, Vertex, Edge\}, a site $s_i \geq 0$, and a \textit{level} $l_i \geq 0$. Every location specifies a specific type of site in a specific container at a specific level.

Each container of the game has its own state representation called {\tt ContainerState}. 
Different representations are implemented in order to minimise the memory footprint and to optimise the time needed to access necessary data for reasoning on any game:
\begin{itemize}
    \item \textbf{ContainerFlatState}: For games played only on the cells.
    \item \textbf{ContainerGraphState}: For games played on the vertices or the edges.
    \item \textbf{ContainerStateStacks}: For stacking games played only on the cells.
    \item \textbf{ContainerGraphStateStacks}:  For stacking games played on the vertices or the edges.
    \item \textbf{ContainerStateCards}: For card games.
    \item \textbf{ContainerDeductionPuzzleState}: For deduction puzzles.
    \item \textbf{ContainerDeductionPuzzleStateLarge}: For deduction puzzles with more than 32 possibles values per variable.
\end{itemize}

\begin{center}
\begin{tikzpicture}

\begin{umlpackage}{util}
\begin{umlpackage}{state}
\umlclass[rectangle split parts=1, x = -5]{State}{}{}
\umlinterface[rectangle split parts=1, x = -1]{ContainerState}{}{}
\umlclass[rectangle split parts=1, x = -5, y = -3]{DeductionPuzzle}{}{}
\umlclass[rectangle split parts=1, x = -5, y = -6]{DeductionPuzzleLarge}{}{}
\umlclass[rectangle split parts=1, x = -1, y = -3]{Flat}{}{}
\umlclass[rectangle split parts=1, x = -1, y = -6]{Graph}{}{}
\umlclass[rectangle split parts=1, x = 2, y = -3]{Stacks}{}{}
\umlclass[rectangle split parts=1, x = 2, y = -6]{GraphStacks}{}{}
\umlclass[rectangle split parts=1, x = 4, y = -3]{Cards}{}{}
\end{umlpackage}
\end{umlpackage}

\umlaggreg[geometry=-|]{ContainerState}{State}
\umlinherit{Flat}{ContainerState}
\umlinherit[geometry=-|]{Graph}{Flat}
\umlinherit{DeductionPuzzle}{ContainerState}
\umlinherit[geometry=-|]{DeductionPuzzleLarge}{DeductionPuzzle}
\umlinherit{Stacks}{ContainerState}
\umlinherit[geometry=-|]{GraphStacks}{Stacks}
\umlinherit{Cards}{ContainerState}
\end{tikzpicture}
\end{center}

A {\tt ContainerState} $cs$ is implemented as a collection of data vectors for each playable site. Each vector is defined using a custom {\tt BitSet} class, called {\tt ChunkSet}, that compresses the required state information into a minimal memory footprint, based on each game's definition. For example, if an $N$-player game involves no more equipment than a board and uniform pieces in $N$ colours, then the container state is described by a {\tt ChunkSet} subdivided into chunks of $B$ bits per board cell, where $B$ is the lowest power of 2 that provides enough bits to represent every possible state per playable site (including state 0 for the empty sites).\footnote{Chunk sizes are set to the lowest power of 2 to avoid issues with chunks straddling consecutive {\tt long} values.}

The different possible data vectors are:
\begin{itemize}
    \item \textbf{what}: The index of a component at a specific location, $0$ if no component.
    \item \textbf{who}: The index of the owner of a component at a specific location, $0$ if no component.
    \item \textbf{count}: The number of the same component at a specific location.
    \item \textbf{state}:  The value of the state at a specific location.
    \item \textbf{rotation}: The value for the rotation direction at a specific location, $0$ corresponding to the first supported direction by the board in starting with North then incrementing clockwise. 
    \item \textbf{value}:  The piece value at a specific location.
    \item \textbf{playable}: For boardless games, returning if a location is playable ($1$) or not ($0$).
    \item \textbf{hidden[]}: For each player, if the location is invisible.
    \item \textbf{hiddenWhat[]}: For each player, if the piece index on the location is hidden.
    \item \textbf{hiddenWho[]}: For each player, if the piece owner on the location is hidden.
    \item \textbf{hiddenCount[]}: For each player, if the number of pieces on the location is hidden.
    \item \textbf{hiddenState[]}: For each player, if state of the location is hidden.
    \item \textbf{hiddenRotation[]}: For each player, if piece rotation is hidden.
    \item \textbf{hiddenValue[]}: For each player, if the piece value is hidden.
\end{itemize}

These data vectors are instantiated or not according to the game flags. For example the different vectors about hidden information are instantiated only if the game involves hidden information. 

On the following picture, the different effects of using the hidden information is shown: \\

\begin{minipage}{0.25\textwidth}
\begin{center}
\includegraphics[width=0.6\linewidth]{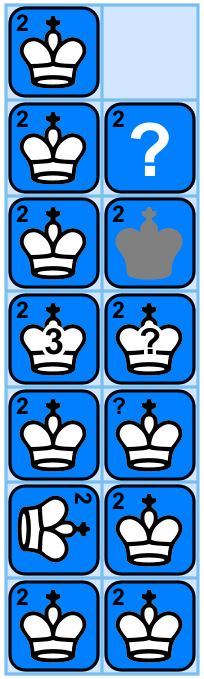}
\end{center}
\end{minipage}
\begin{minipage}{0.68\textwidth}
For each line, a king tile owned by the white player is placed on the two cells with the same data. The number in the top left of the tile is the value of the piece. In the second column of each line, a different type of hidden information is shown:
\begin{itemize}
    \item line 1: All the data are invisible to the white player.
    \item line 2: The piece index is hidden to the white player.
    \item line 3: The piece owner is hidden to the white player.
    \item line 4: The number of pieces is hidden to the white player.
    \item line 5: The piece value is hidden to the white player.
    \item line 6: The piece rotation is hidden to the white player.
    \item line 7: The state of the site is hidden to the white player.
\end{itemize}
\end{minipage}

Each of these data has a maximum value. The maximum value of the data vector 'what' corresponds to the number of pieces and the maximum value of the data vertor 'who' corresponds to the number of players + 2 (included a neutral player and a shared player). The maximum value of the data vectors 'count', 'state', 'rotation' and 'value' are computed when the game is computed. The rotation maximum is fixed and corresponds to the maximum number of possible directions of each playable site. The three other data are computed thanks to the initial state but can also be adapted in the piece definition:
\begin{itemize}
    \item count: The default maximum value is the maximum between the number of playable sites corresponding to the default site type of the board and the sum of all the pieces placed in the initial state. That value can be modified with the (piece ...) ludeme in specifying the parameter "maxCount:".
    \item state: The default maximum value is the maximum between the number of players, the maximum local state of a piece placed in the initial state, the maximum number of states of a die component and the maximum states of a large piece. That value can be modified with the (piece ...) ludeme in specifying the parameter "maxState:".
    \item value: The default maximum value is the maximum between the maximum value of a piece placed in the initial state and 128 (corresponding to $2^7$). That value can be modified with the (piece ...) ludeme in specifying the parameter "maxValue:".
\end{itemize}

\subsection{Other data}\label{Section:OtherStateData}

The information about the player who moved in the previous state ($prev(s)$), the player who has to move in the current state ($mover(s)$) and the player who has to move in the next state ($next(s)$) are stored in 3 variables: prev, mover, next.

Each state $s$ encodes also the following structured data values used to model the game:
\begin{itemize}
    \item \textbf{amount}: Amount, for example of money, belonging to each player. 
    \item \textbf{counter}: Variable, incremented by one after each move played. 
    \item \textbf{currentDice}: The current dice values of each set of dice. 
    \item \textbf{currentPhase}: Current Phase for each player. 
    \item \textbf{diceAllEqual}: True if all the values of the dice are equal. 
    \item \textbf{isDecided}: The decision taken after voting. 
    \item \textbf{notes}: The notes sent to each player during the game. 
    \item \textbf{numTurn}: The number of turns played so far. 
    \item \textbf{numTurnSamePlayer}: The number of turns played successively by the same player. 
    \item \textbf{pendingValues}: Values stored in a set which are in pending between two states. 
    \item \textbf{pieceToRemove}: For sequence of capture to remove (e.g. some draughts games\footnote{International Draughts: \href{https://en.wikipedia.org/wiki/International_draughts}{Wikipedia}}). 
    \item \textbf{playerOrder}: Gives the current game state player order initially indexed by the original game start player order (which may be different after swaps).. 
    \item \textbf{propositions}: The list of propositions done in the previous state. 
    \item \textbf{pot}: The value of the pot, for example of money, of the game. 
    \item \textbf{remainingDominoes}: All the remaining dominoes.
    \item \textbf{rememberingValues}: All the values stored in the state by previous state in calling the move (remember Value ...).
    \item \textbf{stalemated}: For every player, a bit indicating whether they are stalemated. 
    \item \textbf{sumDice}: The sum of all dice. 
    \item \textbf{teams}: Team to which each player belongs, if any. 
    \item \textbf{triggered}: For every player, a bit indicating whether some conditions are triggered (e.g. checkmate). 
    \item \textbf{trumpSuit}: The index of the suit which is the trump in a card game. 
    \item \textbf{valueMap}: Values set to custom variables in the description. 
    \item \textbf{valuesPlayer}: Values associated with each player. 
    \item \textbf{visited}: BitSet used to store all the locations already visited (from and to) by each move done by the player in a sequence of turns played by the same player. 
    \item \textbf{votes}: The list of votes taken for the propositions. 
\end{itemize}





\subsection{Deduction puzzle}\label{section:DeductionPuzzle}

In case of a deduction puzzle, the state $s$ of Ludii is modelled as a Constraint Satisfied Problem ($CSP$) \cite{Piette2019Puzzle}. A $CSP$ consists of a set of variables -- each associated with a domain of possible values -- and a set of constraints that link the variables and define allowed combinations of values among them. As an example of a deduction puzzle, we provide the complete description of the 8 Queens Puzzle.\footnote{8 Queens Puzzle: \href{https://en.wikipedia.org/wiki/Eight_queens_puzzle}{Wikipedia}}

\begin{boxex}
\begin{ludii}
(game "N Queens"
    (players 1)  
    (equipment  { 
        (puzzleBoard (square 8) (values Cell (range 0 1)))
        (regions {AllDirections})
    })
    
    (rules
        (play
            (satisfy {
                (is Count (sites Board) of:1 8)
                (all Different except:0)
            })
        )
        (end (if (is Solved) (result P1 Win)))
    )
)
\end{ludii} 
\end{boxex}

Each site on the board is seen as a variable and the domain of each variable is defined thanks to the ludeme $(values$ $...)$. In the case of the 8 Queens Puzzle, 64 variables represent the cells of the board with each of them a domain of two values $\{0$ $1\}$. 
The constraints of the puzzle are defined by the ludeme (satisfy ...).

The dedicated container state for deduction puzzles called $ContainerDeductionPuzzleState$ models each variable as a {\tt BitSet} where each bit corresponds to one single possible value for this variable. Initially all these bits are set to true to mean they are still possible values for this variable. To remove a value from the set of possible values, the bit must be false. To set a value to a variable, only one single bit has to be true.
When all the variables are set to a value, the ludeme (is Solved ...) checks if all the constraints are satisfied. If this is the case the player wins.

The main methods of this container state are:
\begin{itemize}
    \item \textbf{bit(var,value)}: True if the value is a possible value for the variable $var$.
    \item \textbf{set(var,value)}: To keep only the bit corresponding to that value true and set all the others to false for the variable $var$.
    \item \textbf{toggle(var,value)}: To toggle the bit corresponding to that value for the variable $var$.
    \item \textbf{reset(var)}: To reset all the bits of each value of a variable $var$ to true.
    \item \textbf{isResolved(var)}: True if a value is set to the variable $var$.
    \item \textbf{what(var)}: The value set to the variable $var$. If the value is not set, it returns 0.
\end{itemize}


\section{Trials}

A Ludii successor function is given by $\mathcal{T}:(S\setminus S_{ter}, \mathcal{A}) \mapsto S,$
where $S$ is the set of all the Ludii game states, $S_{ter}$ the set of all the terminal states, and $\mathcal{A}$ the set of all possible lists of actions. 

Given a current state $s \in S \setminus S_{ter}$, and a list of actions $A = \left[ a_i \right] \in \mathcal{A}$, $\mathcal{T}$ computes a successor state $s' \in S$. Intuitively, a complete list of actions $A$ can be understood as a single ``move'' selected by a player, which may have multiple effects on a game state (each implemented by a different primitive action).

A trial $\tau$ is a sequence of states $s_i$ and action list $A_i$: $\langle s_0, A_1, s_1, \ldots, s_{f-1}, A_f, s_f \rangle$ such that $f \geq 0$, and for all $i \in \{1, \ldots f\}$,
\begin{itemize}
    \item the played action list $A_i$ is legal for the $mover(s_{i-1})$
    \item states are updated: $s_i = \mathcal{T}(s_{i-1}, A_{i})$
    \item only $s_f$ may be terminal: $\{s_0, \ldots, s_{f-1}\} \cap S_{ter} = \emptyset$
\end{itemize}

$\tau$ is stored in a {\tt Trial} object, providing a complete record of a game played from start to end, including all the moves made stored in a list. Thanks to that any reasoning on any game can be parallelised using separate trials per thread. All the data members of the {\tt Game} object are constant and can therefore be shared between threads. A thread will be able to use a {\tt Trial} object to compute any playouts from any state. 

A trial is over when the variable status is set to the index of the winner of the game (or to $0$ in case of a draw) and the outcome of the game corresponding to the ranking of the players is stored in an array of double values called ranking.
The trial also stores the list of all the locations of each component placed in the initial state in order to be able to access them with the ludeme (sites Start ...) and a list of long values encoding the previous states thanks to a Zobrist hashing in order to be able to compare efficiently two states in case of meta rules such as no repetition.

When a new state $s_{i+1}$ is reached after applying a move selected from the list of legal moves for a state $s_i$, Ludii computes the new list of legal moves of $s_{i+1}$ and stores them in the trial in order for any feature to access them quickly without needing to compute them again. 

\subsection{Move and Actions}\label{Section:Move}
As described in the previous section, the transition between two successive states $s_i$ and $s_{i+1}$ is possible thanks to a sequence of actions $A_i$ applied on $s_i$. In Ludii, such a sequence is modelled as a Move object.

\begin{center}
\begin{tikzpicture}

\begin{umlpackage}{util}
\umlclass[rectangle split parts=1]{Move}{}{}

\begin{umlpackage}{action}
\umlinterface[rectangle split parts=1, x = 3]{Action}{}{}
\end{umlpackage}
\end{umlpackage}

\umlaggreg[geometry=-|]{Action}{Move}
\end{tikzpicture}
\end{center}

In Ludii, an atomic action is the only way to modify the state after its creation. Consequently, when a player selects their move $m$ in the list of legal moves available in the trial, Ludii applies successively each atomic action in the list of actions composing the move $m$.

Each action and move are associated with their own data:
\begin{itemize}
    \item \textbf{from}: The 'from' site. If the action is not related to a site, that value is equal to -1.
    \item \textbf{levelFrom}: The level of the 'from' site. If the game is not a stacking game, that value is equal to 0.
    \item \textbf{fromType}: The graph type of the 'from' site (Cell, Vertex or Edge).
    \item \textbf{to}: The 'to' site. If the action is not related to a site, that value is equal to -1. Moreover, if the action is relative to only one site, that value is equal to the 'from' site.
    \item \textbf{levelTo}: The level of the 'to' site. If the game is not a stacking game, the value is equal to 0.
    \item \textbf{toType}: The graph type of the 'to' site (Cell, Vertex or Edge).
    \item \textbf{stacking}: True if the move is related to a stack.
    \item \textbf{oriented}: True if the action is oriented.
    \item \textbf{what}: The index of the component moved.
    \item \textbf{who}: The index of the player making a move.
    \item \textbf{state}: The new state value of the site. If the action is not modifying the state value, the value is equal to -1.
    \item \textbf{rotation}: The new rotation value of the site. If the action is not modifying the rotation value, the value is equal to -1.
    \item \textbf{value}: The new piece value of the site. If the action is not modifying the piece value, the value is equal to -1.
\end{itemize}

Ludii makes a clear difference between a decision move and an effect move. A decision move is the move decided by the player as opposed to an effect move, which is applied following a decision.
In each sequence of actions $A$ composing a move $m$ in the list of legal moves, only one action $a_d$ is a decision: $\exists!$ $a_d$ $\in$ $A$. For a specific move $m$ all the data defined previously correspond to the data of $a_d$.

Here is the list of all the current atomic actions designed in Ludii:

\renewcommand{\arraystretch}{1.3}
\arrayrulecolor{white}
\rowcolors{2}{gray!25}{gray!10}
\begin{longtable}{@{}p{.26\textwidth} | p{.64\textwidth}@{}}
\rowcolor{gray!50}
\textbf{Action} & \textbf{Description} \\
ActionAdd & Adds one or more piece(s) to a location. \\
ActionAddPlayerToTeam & Adds a player to a team. \\
ActionBet & Makes a bet for a player. \\
ActionCopy & Copies a component from a location to another. \\
ActionForfeit & Forfeits a player. \\
ActionForgetValue & Forget a value previously remembered. \\
ActionInsert & Inserts a component inside a stack. \\
ActionNextInstance & Moves the state on to the next instance in a Match. \\
ActionNote & Sends a message to a player. \\
ActionMove & Moves a piece from a location to another (only one piece or one full stack). \\
ActionMoveN & Moves many pieces from a location to another (but not in a stack). \\
ActionPass & Pass the turn.  \\
ActionPromote & Promotes a piece to another piece. \\
ActionPropose & Proposes a subject to vote. \\
ActionRememberValue & Remember a value. \\
ActionRemove & Removes one or more component(s) from a location. \\
ActionReset & Resets all the values of a variable to not set in a deduction puzzle. \\
ActionSelect & Selects a 'from' and 'to' site. \\
ActionSet & Sets a value to a variable in a deduction puzzle. \\
ActionSetAmount & Sets a player's amount. \\
ActionSetCost & Sets the cost of a graph element. \\
ActionSetCount & Sets the count of a location. \\
ActionSetCounter & Sets the counter of the state. \\
ActionSetDiceAllEqual & Specifies for the state that all dice are equal. \\
ActionSetHidden & Makes a site invisible or not to a player. \\
ActionSetHiddenCount & Makes the number of pieces on a site hidden or not to a player. \\
ActionSetHiddenRotation & Makes the piece rotation on a site hidden or not to a player. \\
ActionSetHiddenState & Makes the state of the site hidden or not to a player. \\
ActionSetHiddenValue & Makes the piece value on a site  hidden or not to a player. \\
ActionSetHiddenWhat & Makes the piece index on a site  hidden or not to a player. \\
ActionSetHiddenWho & Makes the piece owner on a site hidden or not to a player. \\
ActionSetNextPlayer & Sets the next player. \\
ActionSetPending & Sets a value to pending in the state. \\
ActionSetPhase & Sets the phase of a graph element. \\
ActionSetPot & Sets the pot. \\
ActionSetRotation & Sets the rotation value of a location. \\
ActionSetScore & Sets the score of a player. \\
ActionSetState & Sets the state of a location. \\
ActionSetTrumpSuit & Sets the trump suit for card games. \\
ActionSetValueOfPlayer & Sets the value of a player in the state. \\
ActionSetVar & Sets a var in the state. \\
ActionStackMove &  Moves a part of a stack. \\
ActionStoreStateInContext & Stores the current state of the game. \\
ActionSwap & Swap two players. \\
ActionToggle & Excludes a value from the possible values of a variable in a deduction puzzle. \\
ActionTrigger & Triggers an event for a player. \\
ActionUpdateDice & Sets the state of the site where the die is and updates the value of the die. \\
ActionUseDie &  Uses a die and removes it from the current dice of the context. \\
ActionVote & Votes on a proposition done previously. \\
\end{longtable}
\rowcolors{0}{}{}


\chapter{\textcolor{gray!95}{Equipment}}\label{Chapter:Equipment}

All the items used in a game are defined in the equipment ludeme. Here is the description of an example equipment:

\begin{boxex}
\begin{ludii}
(equipment {
        (board 
            (square 5)
            {
            (track "Track1" "2,E,N,W,S,E1,N3,E2" P1 directed:true)
            (track "Track2" "22,W,S,E,N,W1,S3,W2" P2 directed:true)
            }
        )
        (piece "Ball" Each)
        (map "StartingPoint" {(pair 1 2) (pair 2 22)})
        (dice d:6 num:2)
        (regions P1 (sites Bottom))
        (regions P2 (sites Top))
    })
\end{ludii} 
\end{boxex}

The equipment is defined by a list of item objects, which can be of different types:

\begin{center}
\begin{tikzpicture}

\begin{umlpackage}{game}
\begin{umlpackage}{equipment}
\umlclass[rectangle split parts=1, y = -6]{Equipment}{}{}
\umlinterface[rectangle split parts=1]{Item}{}{}
\umlclass[rectangle split parts=1, x = 7]{Dominoes}{}{}
\umlclass[rectangle split parts=1, x = -7]{Hints}{}{}
\umlinterface[rectangle split parts=1, x = -2, y = -3]{Component}{}{}
\umlinterface[rectangle split parts=1, x = 2, y = -3]{Container}{}{}
\umlclass[rectangle split parts=1, x = -5, y = -2]{Map}{}{}
\umlclass[rectangle split parts=1, x = 5, y = -2]{Regions}{}{}

\end{umlpackage}
\end{umlpackage}

\umlaggreg[geometry=|-]{Component}{Equipment}
\umlaggreg[geometry=|-]{Container}{Equipment}
\umlaggreg[geometry=|-]{Regions}{Equipment}
\umlaggreg[geometry=|-]{Map}{Equipment}
\umlaggreg[geometry=|-]{Hints}{Equipment}
\umlinherit{Dominoes}{Item}
\umlinherit{Hints}{Item}
\umlinherit{Component}{Item}
\umlinherit{Container}{Item}
\umlinherit{Map}{Item}
\umlinherit{Regions}{Item}

\end{tikzpicture}
\end{center}

\begin{itemize}
    \item \textbf{Container}: Containers of the game such as the board or the hands of the players.
    \item \textbf{Component}: Components of the game such as the pieces or the tiles.
    \item \textbf{Map}: List of pairs associating two integers.
    \item \textbf{Regions}: Specific regions of the board possibly associated with a name or a player. 
    \item \textbf{Dominoes}: Set of dominoes.
    \item \textbf{Hints}: List of hints used for deduction puzzles.
\end{itemize}

Each type of item is stored in a separate list in the Equipment class in order to access them efficiently. Each item has a name, an owner (equals to -1 if not necessary) and an index used to identify them in the appropriate list in the equipment. In this chapter, we dedicate a section to describe each type of Item object.


\section{Container}

Any game compiled by Ludii has at least a board and potentially other containers used to define the playable sites of the game and the size of each {\tt ChunkSet} used to defined the container state (see Section \ref{section:ContainerState}). Four container types exist:

\begin{center}
\begin{tikzpicture}

\begin{umlpackage}{equipment}
\begin{umlpackage}{container}
\umlinterface[rectangle split parts=1]{Container}{}{}
\umlclass[rectangle split parts=2, x = -2, y = -2]{Board}{Track[] tracks}{}
\umlclass[rectangle split parts=1, x = 2, y = -2]{Hand}{}{}
\umlclass[rectangle split parts=1, x = -4]{Dice}{}{}
\umlclass[rectangle split parts=1, x = 4]{Deck}{}{}

\end{umlpackage}
\end{umlpackage}

\umlinherit{Board}{Container}
\umlinherit{Hand}{Container}
\umlinherit[geometry=|-]{Dice}{Container}
\umlinherit[geometry=|-]{Deck}{Container}

\end{tikzpicture}
\end{center}

\begin{itemize}
    \item \textbf{Board}: The main board of the game.
    \item \textbf{Hand}: A hand of a player designed as a succession of cells generally used to place the pieces belonging to the player but not currently on the board.
    \item \textbf{Dice}: The set of dice for the game.
    \item \textbf{Deck}: The deck of cards for the game. 
\end{itemize}

Each container of a game $\mathcal{G}$ is modelled by a graph defined by a set $C$ of cells, a set $V$ of vertices and a set $E$ of edges. A Topology object is stored in each Container object and contains a separate list for each graph element type stored as a TopologyElement list (cells, vertices and edges). The topology also contains many pre-generated sets of sites for each graph element type. Here is the list of all the pre-generated sets:

\renewcommand{\arraystretch}{1.3}
\arrayrulecolor{white}
\rowcolors{2}{gray!25}{gray!10}
\begin{longtable}{@{}p{.25\textwidth} | p{.65\textwidth}@{}}
\rowcolor{gray!50}
\textbf{Pre-generated sets} & \textbf{Description} \\
Major & The list of major generator sites. \\
Minor & The list of minor generator sites. \\
Corners & The list of corner sites. \\
Convex Corners & The list of convex corner sites. \\
Concave Corners & The list of concave corner sites. \\
Outer & The list of outer sites. \\
Perimeter & The list of perimeter sites. \\
Inner & The list of inner sites. \\
interlayer & The list of interlayer sites. \\
top & The list of sites in the top of the container. \\
bottom & The list of sites in the bottom of the container. \\
left & The list of sites in the left of the container. \\
right & The list of sites in the right of the container. \\
centre & The list of sites in the centre of the container. \\
columns & The list of each column. \\
rows & The list of each row. \\
phases & The list of each set of sites in the same phase. \\
sides & The list of set of sites for each side of the board indexing by direction. \\
layers & The list of set of sites in the same layer (for 3D games). \\
axials & The list of axial edges. \\
horizontal & The list of horizontal edges. \\
vertical & The list of vertical edges. \\
angled & The list of angled edges. \\
slash & The list of angled edges as a slash. \\
slosh & The list of angled edges as a slosh. \\
\end{longtable}
\rowcolors{0}{}{}

The geometry of each board is described in the next section.

\subsection{Board Geometry}

The board shared by all players is represented internally as a graph, in which the vertices, edges and faces represent the playable {\it sites} at which components can be placed. The edges represent adjacency between neighbouring vertices and faces. 
To generate the graph defining the topology of each container, Ludii uses a set of functions called GraphFunctions.

\subsubsection{GraphFunction}

The simplest graph function is the {\tt (graph ...)} ludeme shown below, which allows the user to specify a Cartesian (x,y) position for each vertex, and then specify the edges of the graph as a list of pairs of adjacent vertex indices. 
Users can also optionally define cells in this step as a list of lists of vertex indices.

\begin{minipage}{0.5\textwidth}
\begin{boxex}
\begin{ludii}
(board 
   (graph
      vertices:{{0 0}{0 1}
         {1 1}{1 0}{0.5 1.5}}
      edges:{{0 1}{1 2}{2 3}
         {3 0}{1 4}{4 2}}
   )
   use:Vertex
)
\end{ludii} 
\end{boxex}
\end{minipage}
\begin{minipage}{0.5\textwidth}
\begin{center}
\includegraphics[width=0.5\linewidth]{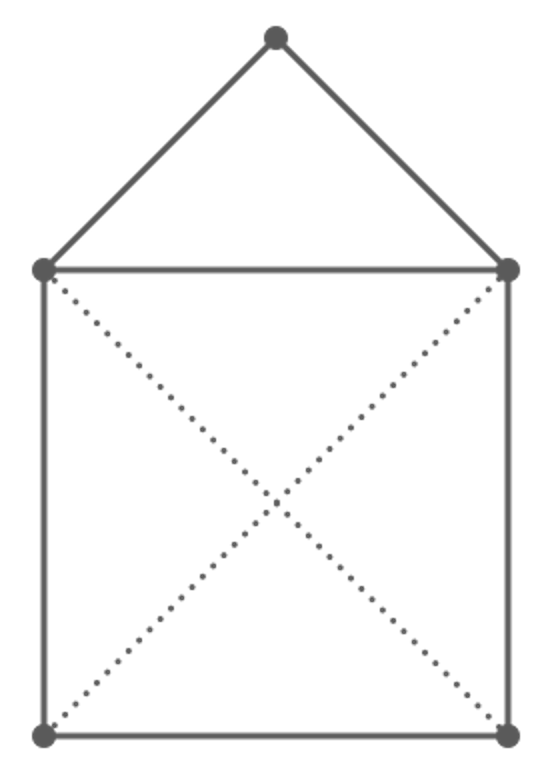}
\end{center}
\end{minipage}

The {\tt (board ...)} ludeme shown above generates a board on which pieces are played on the vertices, due to the ``use:Vertex'' parameter. 
Diagonals within each cell (dotted lines) are automatically generated. 
If ``use:Cell'' is specified instead, or no ``use:'' value is specified, then the game is played on the cells and the following board is generated.
Games can also be played on the edges, or any combination of these three graph elements.

\begin{minipage}{0.5\textwidth}
\begin{boxex}
\begin{ludii}
(board 
   (graph
      vertices:{{0 0}{0 1}
         {1 1}{1 0}{0.5 1.5}}
      edges:{{0 1}{1 2}{2 3}
         {3 0}{1 4}{4 2}}
   )
   use:Cell
)
\end{ludii} 
\end{boxex}
\end{minipage}
\begin{minipage}{0.5\textwidth}
\begin{center}
\includegraphics[width=0.5\linewidth]{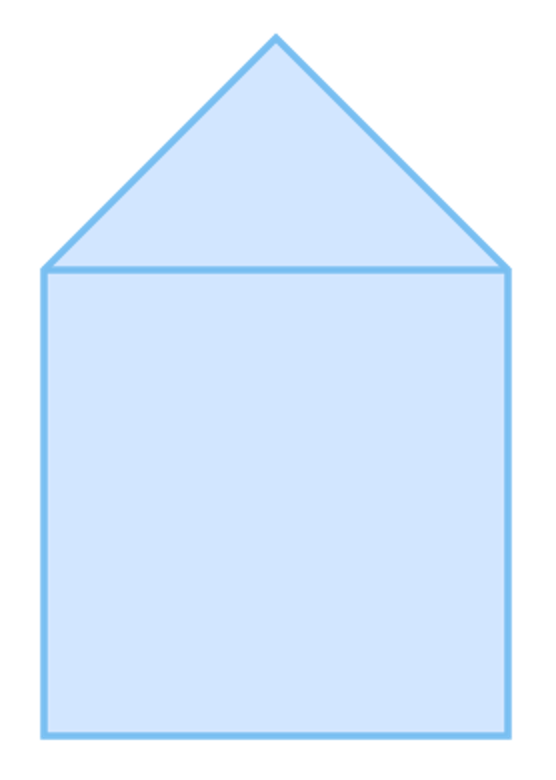}
\end{center}
\end{minipage}

As it can become tedious to specify large numbers of exact vertex positions and edge pairs for more complex boards, a range of graph {\it generators} and {\it operators} are provided, as follows.

\phantom{}

\noindent
{\bf Graph Generators}: 
These generate a base graph according to a specified basis and/or shape. 
The following bases are supported.

\renewcommand{\arraystretch}{1.3}
\arrayrulecolor{white}
\rowcolors{2}{gray!25}{gray!10}
\begin{longtable}{@{}p{.25\textwidth} | p{.65\textwidth}@{}}
\rowcolor{gray!50}
\textbf{Graph Bases} & \textbf{Description} \\
(brick ...) & Brickwork tiling. \\
(celtic ...)  & Celtic knotwork tiling. \\
(hex ...) & Regular hexagonal basis. \\
(mesh ...) & Arbitrary 2D mesh. \\
(morris ...) & Morris board designs. \\
(quadhex ...) & Six-sided quadrilateral mesh, as used for Three-Player Chess. \\
(square ...) & Regular square tiling. \\
(tiling $\langle TilingType \rangle$ ...) & Semi-regular tiling. \\
(tri ...) & Regular triangular tiling. \\
\end{longtable}
\rowcolors{0}{}{}

The following semi-regular tilings are supported.

\renewcommand{\arraystretch}{1.3}
\arrayrulecolor{white}
\rowcolors{2}{gray!25}{gray!10}
\begin{longtable}{@{}p{.2\textwidth} | p{.7\textwidth}@{}}
\rowcolor{gray!50}
\textbf{Semi-Regular Tilings} & \textbf{Description} \\
T488 & Semi-regular tiling made up of octagons with squares in the interstitial gaps. \\
T3464 & Rhombitrihexahedral tiling (e.g. Kensington). \\
T3636 & Semi-regular tiling made up of triangles and hexagons. \\
T4612 & Semi-regular tiling made up of squares, hexagons and dodecagons. \\
T31212 & Semi-regular tiling made up of triangles and dodecagons. \\
T33336 & Semi-regular tiling made up of triangles around hexagons. \\
T33344 & Semi-regular tiling made up of alternating rows of squares and triangles. \\
T33434 & Semi-regular tiling made up of squares and pairs of triangles. \\
T333333\_3343 & Tiling 3.3.3.3.3.3,3.3.4.3.4. \\
\end{longtable}
\rowcolors{0}{}{}

For example, the following ludeme defines a rhombitrihexahedral (T3464) graph with two major units per side, as used in the Kensington board. 

\begin{minipage}{0.5\textwidth}
\begin{boxex}
\begin{ludii}
(graph (tiling T3464 2))
\end{ludii} 
\end{boxex}
\end{minipage}
\begin{minipage}{0.5\textwidth}
\begin{center}
\includegraphics[width=0.75\linewidth]{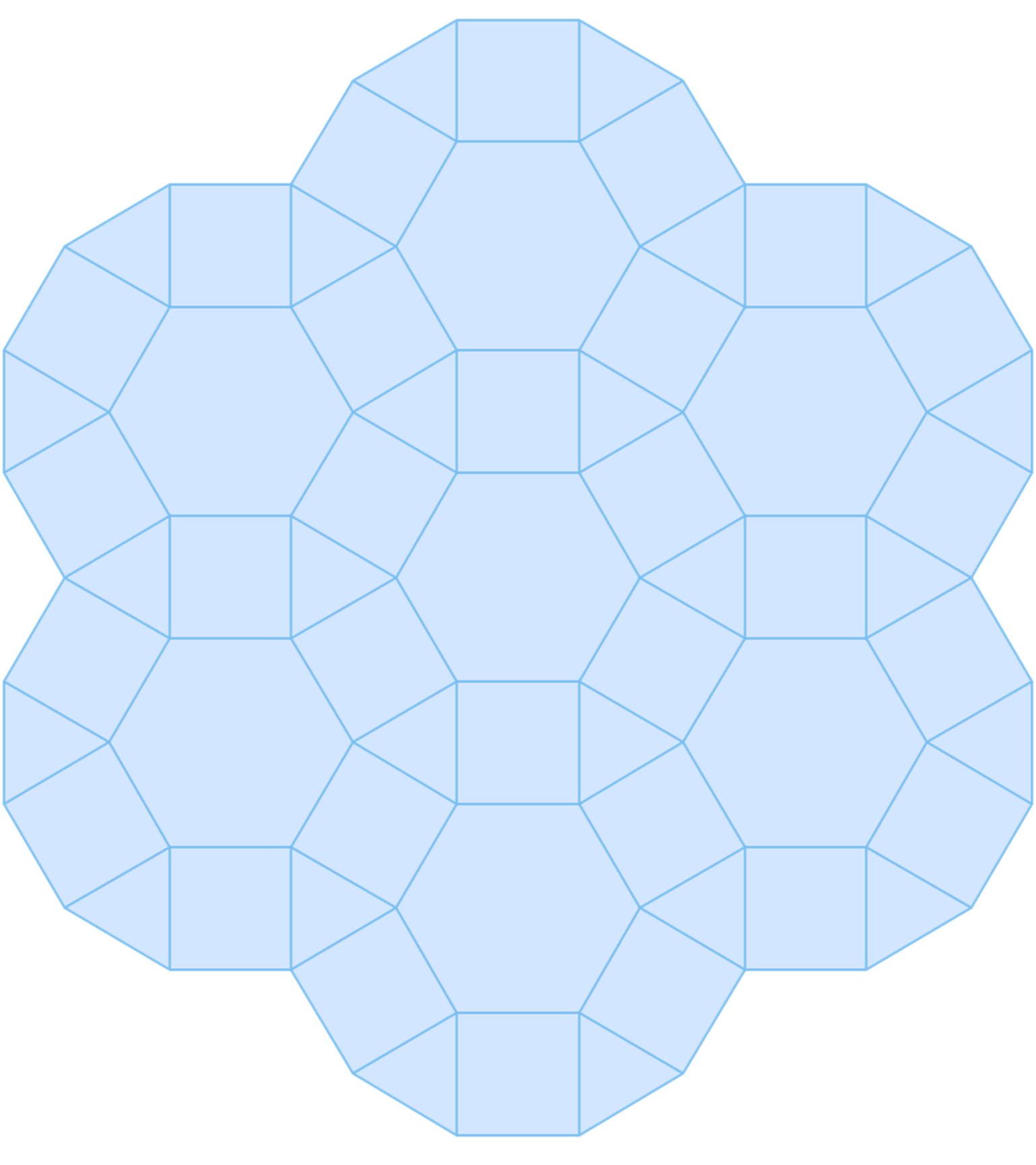}
\end{center}
\end{minipage}

The following board shapes are supported. 
Note that some shapes are assumed as defaults for certain bases, e.g. the T3434 tiling assumes a hexagonal shape unless otherwise specified, as shown in the above figure.

\renewcommand{\arraystretch}{1.3}
\arrayrulecolor{white}
\rowcolors{2}{gray!25}{gray!10}
\begin{longtable}{@{}p{.2\textwidth} | p{.7\textwidth}@{}}
\rowcolor{gray!50}
\textbf{Graph Shapes} & \textbf{Description} \\
(circle ...) & Circular boards. \\
(rectangle ...)  & Rectangular boards. \\
(repeat ...) & Boards composed of repeated user-defined shapes. \\
(shape ...) & Special shapes including Star shaped boards. \\
(spiral ...) & Spiral boards. \\
(wedge ...) & Wedge-shapes sub-board, e.g. for adding external ``arms'' to Alquerque-style boards. \\
\end{longtable}
\rowcolors{0}{}{}

\phantom{}

\noindent
{\bf Graph Operators}: 
These modify an existing graph in specific ways. 
The following operators are provided.

\renewcommand{\arraystretch}{1.3}
\arrayrulecolor{white}
\rowcolors{2}{gray!25}{gray!10}
\begin{longtable}{@{}p{.23\textwidth} | p{.67\textwidth}@{}}
\rowcolor{gray!50}
\textbf{Graph Operators} & \textbf{Description} \\
(add ...) & Add vertices, edges or faces to a graph. \\
(clip ...) & Clip elements from a graph. \\
(complete ...) & Join all vertex pairs with an edge. \\
(dual ...) & Replace the graph with its weak dual. \\
(hole ...) & Remove internal elements of a graph. \\
(intersect ...) & Find the intersection of two graphs. \\
(keep ...) & Keep some elements from a graph and discard the rest. \\
(layers ...) & Create a 3D graph with layers in the z direction. \\
(makeFaces ...) & Generate all non-overlapping faces. \\
(merge ...) & Merge two graphs, keeping all elements of both except where coincident elements are merged. \\
(recoordinate ...) & Regenerate the coordinate labels for the graph elements. \\
(remove ...) & Remove specified elements from a graph. \\
(renumber ...) & Renumber the graph elements (from bottom left rightwards and upwards). \\
(rotate ...) & Rotate the vertices of a graph a specified number of degrees. \\
(scale ...) & Scale the vertex positions of a graph in x, y and/or z directions. \\
(shift ...) & Translate the vertices of a graph in x, y and/or z directions. \\
(skew ...) & Skew a graph by a specified amount. \\
(splitCrossings ...) & Split edge crossings into multiple edges that share a vertex at the crossing point. \\
(subdivide ...) & Subdivide certain faces within a graph. \\
(trim ...) & Trim orphan elements from the graph exterior. \\
(union ...) & Union of two graphs, keeping all elements of both. \\
\end{longtable}
\rowcolors{0}{}{}

Graph operators allow complex board topologies to be easily constructed from simple descriptions. 
For example, the rhombitrihexahedral board shown above can have cells with six or more sides subdivided... 

\begin{minipage}{0.45\textwidth}
\begin{boxex}
\begin{ludii}
(graph 
  (subdivide     
    (tiling T3464 2)
    min:6
  )
)
\end{ludii} 
\end{boxex}
\end{minipage}
\begin{minipage}{0.55\textwidth}
\begin{center}
\includegraphics[width=0.9\linewidth]{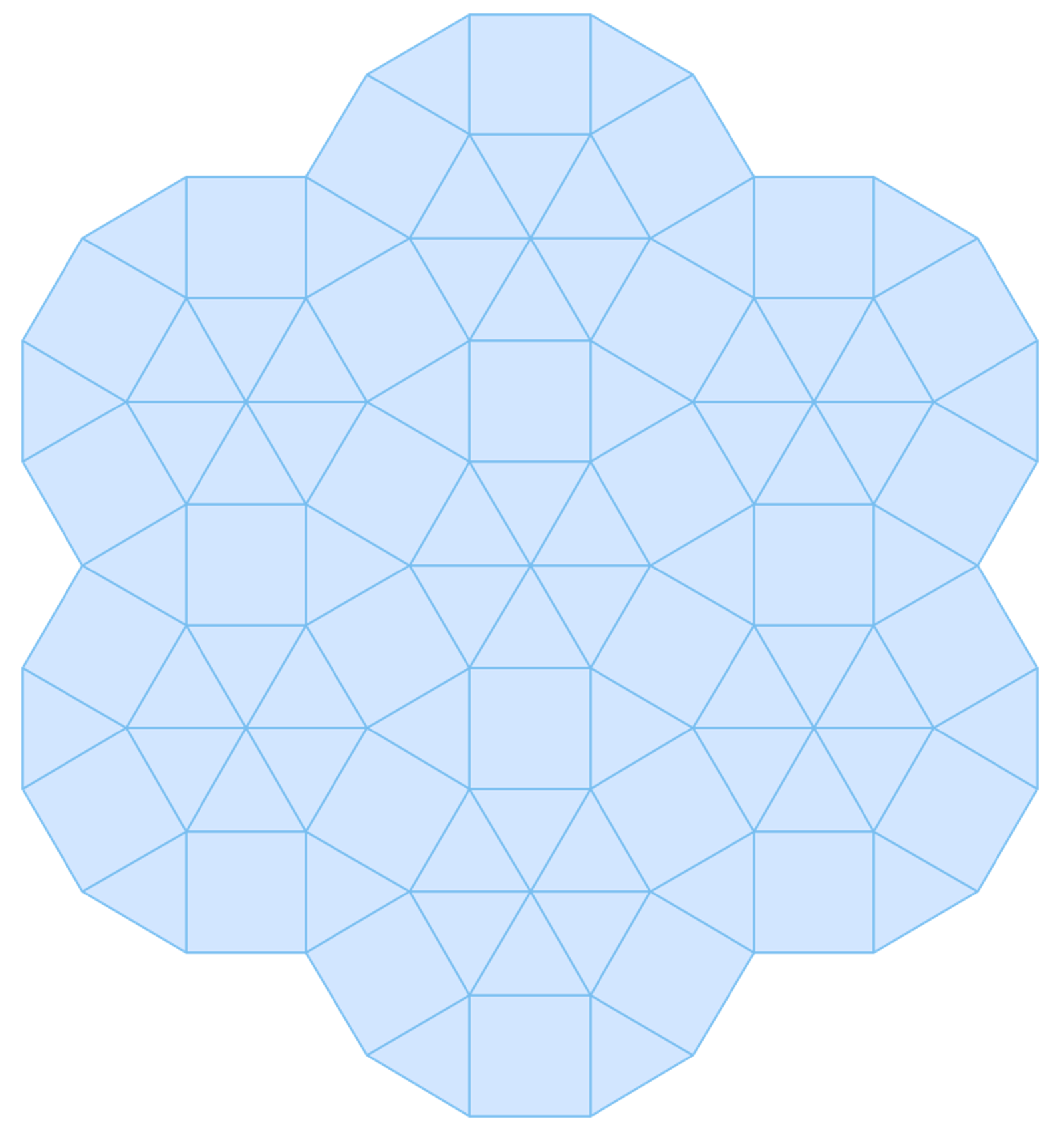}
\end{center}
\end{minipage}

...then be replaced by the weak dual\footnote{The {\it weak dual} of a graph is the graph induced by creating a vertex at the centroid of each face and connecting each pair of vertices generated by adjacent faces with an edge.} of the resulting graph...

\begin{minipage}{0.45\textwidth}
\begin{boxex}
\begin{ludii}
(graph 
  (dual 
    (subdivide     
      (tiling T3464 2)
      min:6
    )
  )
)
\end{ludii} 
\end{boxex}
\end{minipage}
\begin{minipage}{0.55\textwidth}
\begin{center}
\includegraphics[width=0.9\linewidth]{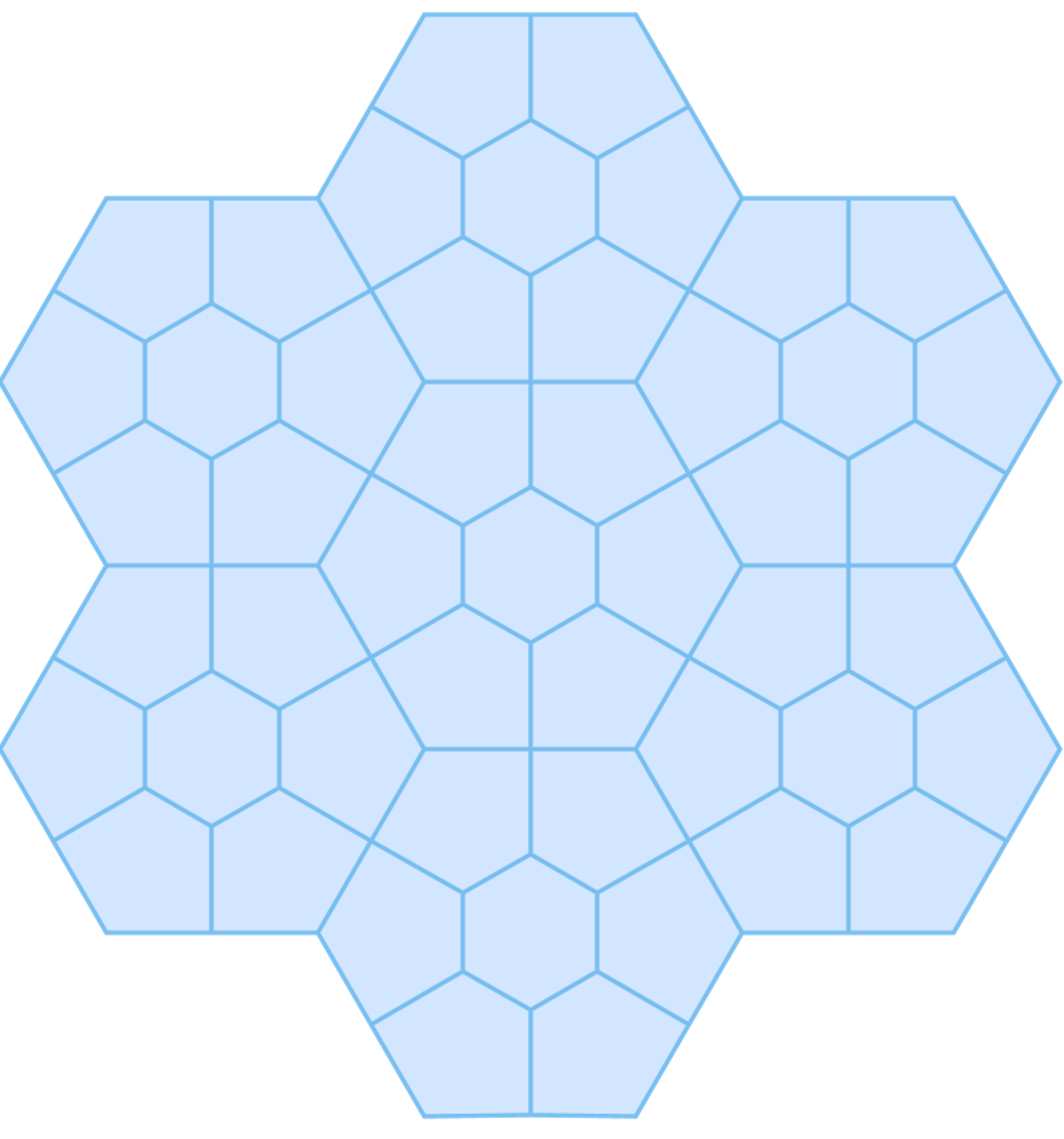}
\end{center}
\end{minipage}

...then have all cells subdivided...

\begin{minipage}{0.45\textwidth}
\begin{boxex}
\begin{ludii}
(graph 
  (subdivide
    (dual 
      (subdivide     
        (tiling T3464 2)
        min:6
      )
    ) 
  )
)
\end{ludii} 
\end{boxex}
\end{minipage}
\begin{minipage}{0.55\textwidth}
\begin{center}
\includegraphics[width=0.9\linewidth]{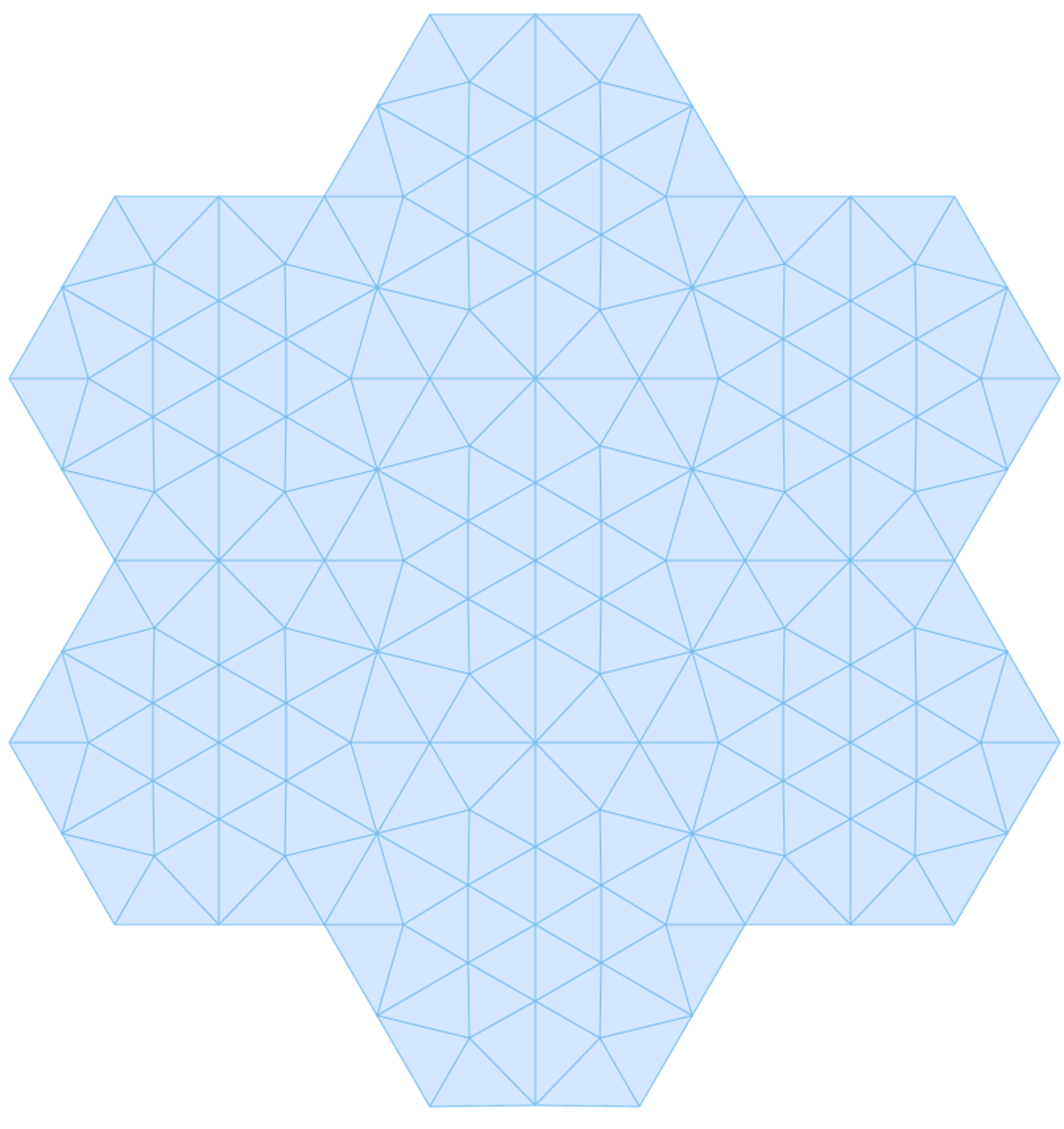}
\end{center}
\end{minipage}

...and again be replaced by the weak dual of the resulting graph. 
The resulting board is distinct and visually interesting, and not previously used for any game that we are aware of.

\begin{minipage}{0.45\textwidth}
\begin{boxex}
\begin{ludii}
(graph 
  (dual 
    (subdivide
      (dual 
        (subdivide     
          (tiling T3464 2)
          min:6
        )
      ) 
    )
  )
)
\end{ludii} 
\end{boxex}
\end{minipage}
\begin{minipage}{0.55\textwidth}
\begin{center}
\includegraphics[width=0.9\linewidth]{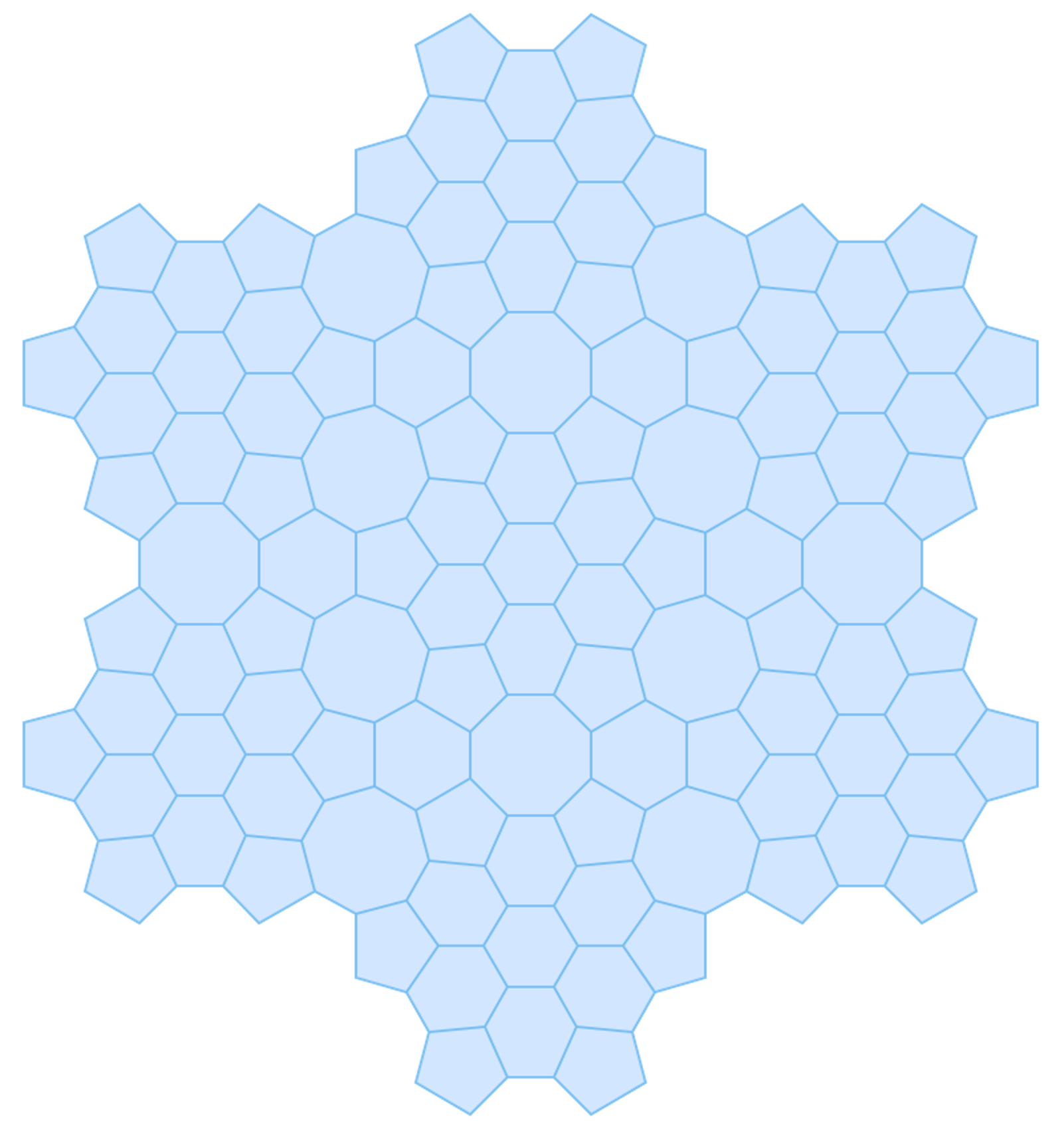}
\end{center}
\end{minipage}

When the game board is generated from a graph produced by a GraphFunction, the following relations are automatically calculated for each site:

\begin{itemize}
    \item {\it Steps}: Step to an adjacent element in each direction (if any). 
    \item {\it Radials}: Contiguous steps to adjacent elements in each direction that minimise change in direction with each step (if any). 
    Radials are used for slide moves, line detection, line-of-sight tests, etc.
\end{itemize}

More details on using GraphFunctions will be provided in the upcoming paper ``General Board Geometry'' (in preparation). 
For the exact syntax of particular GraphFunctions, please see the Ludii Language Reference~\cite{LLR:2020}.

\subsubsection{DimFunction And FloatFunction}

Any Dimension (e.g. the size of a square) used in entry of all the classes defining the graph functions are DimFunctions classes. 
Basically, they are used to allow any common mathematical operators between integers such as: Addition, Subtraction, Division, Product, Absolute value, Power, Maximum and Minimum.

Any Float in entry of all the classes defining the graph functions are FloatFunctions.
They are also the same operators of the DimFunction but also other such as: Cosine, Sine, Tangent, Exponential, Square Root, Logarithm, Logarithm 10.

\subsubsection{Direction System}\label{Section:DirectionSystem}

Ludii supports until 16 compass directions following the intercardinal directions\footnote{Intercardinal directions: \href{https://en.wikipedia.org/wiki/Cardinal_direction}{Wikipedia}}:

\begin{center}
\includegraphics[scale=0.25]{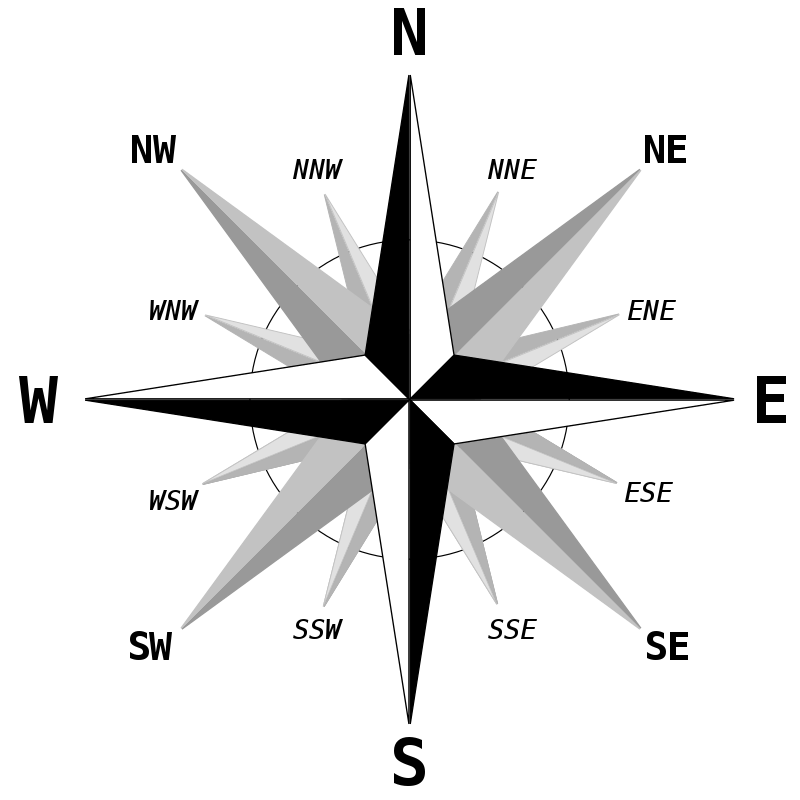}
\end{center}


For the containers involving rotational directions defined with a circular ludeme (circle) or a spiral ludeme (spiral), Ludii adds to the compass directions, the four rotational ones: Outer (Out), Inner (In), Clockwise (CW) and Counter Clockwise (CCW). Finally for the 3D games, Ludii handles all the compass directions for the same layer and the 8 most important compass directions for the downwards and upwards directions: D, DN, DNE, DE, DSE, DS, DSW, DW, DNW and U, UN, UNE, UE, USE, US, USW, UW, UNW.

All these directions are defined in different enumeration types called respectively CompassDirection, RotationalDirection and SpatialDirection, and they are all a DirectionFacing object, used to define the direction faced by a piece if necessary. All these facing directions have methods to define the opposite, left and right direction of each of them in order to handle any potential rotation of a piece during the game.

\begin{center}
\begin{tikzpicture}

\begin{umlpackage}{game}
\begin{umlpackage}{directions}
\umlinterface[rectangle split parts=1, x = 1]{DirectionFacing}{)}{}
\umlclass[rectangle split parts=1, x = 5, y = 2]{CompassDirection}{}{}
\umlclass[rectangle split parts=1, x = 5, y = 0]{RotationalDirection}{}{}
\umlclass[rectangle split parts=1, x = 5, y = -2]{SpatialDirection}{}{}

\umlinterface[rectangle split parts=1, x = -3]{Direction}{}{}
\umlclass[rectangle split parts=1, x = -5, y = -2]{RelativeDirection}{}{}
\umlclass[rectangle split parts=1, x = -1, y = -2]{AbsoluteDirection}{}{}

\end{umlpackage}

\umlinterface[rectangle split parts=2, x = -1, y = 4]{DirectionFunction}{TopologyElement element}{}
\end{umlpackage}

 \umlinherit{DirectionFunction}{Direction}
 \umlinherit{RelativeDirection}{Direction}
 \umlinherit{AbsoluteDirection}{Direction}
 \umlinherit{CompassDirection}{DirectionFacing}
 \umlinherit{RotationalDirection}{DirectionFacing}
 \umlinherit{SpatialDirection}{DirectionFacing}
 \umlunicompo{DirectionFunction}{DirectionFacing}

\end{tikzpicture}
\end{center}

In order to handle any possible direction in a board game, Ludii provides a set of direction functions called DirectionFunction able to return the expecting set of directions according to the context. A direction function can be Absolute in returning constantly the same set of directions of a topology element (playable site) or can be relative and possibly depends on the facing direction of the component placed on this site.

The AbsoluteDirection enumeration type contains equivalents for each possible facing direction and the following ones:
\renewcommand{\arraystretch}{1.3}
\arrayrulecolor{white}
\rowcolors{2}{gray!25}{gray!10}
\begin{longtable}{@{}p{.25\textwidth} | p{.65\textwidth}@{}}
\rowcolor{gray!50}
\textbf{Absolute direction} & \textbf{Description} \\
All & The set of all supported directions of the site. \\
Angled & All angled directions (NW, NNW, WNW, WSW, SE, SW, SSE, SSW, NNE, ESE, ENE, NE). \\
Adjacent & All directions between two sites sharing at least one vertex. \\
Axial & All axial directions (N, S, E, W). \\
Orthogonal & All directions between two sites sharing at least one edge. \\
Diagonal & All directions between two sites in line through a vertex. \\
OffDiagonal & All directions between two cells sharing a vertex but not an edge which are not diagonal. \\
SameLayer & All directions between two sites in the same layer. \\
Upward & All directions upward to a site (U, UN, UE, US, UW, UNW, UNE, USW, USE). \\
Downard & All directions downard to a site (D, DN, DE, DS, DW, DNW, DNE, DSW, DSE). \\
Rotational & All rotational directions (CW, CCW, IN, OUT). \\
\end{longtable}
\rowcolors{0}{}{}

Here are the different step moves of a piece for the 5 common absolute directions for the 3 common tilings: Square, Hexagonal and Triangle:

\renewcommand{\arraystretch}{1.3}
\arrayrulecolor{white}
\begin{longtable}{c|c|c|c}
\rowcolor{gray!50}
\textbf{Absolute Direction} & \textbf{Square} & \textbf{Hexagonal} & \textbf{Triangle}\\
 All &  \raisebox{-.5\height}{\includegraphics[width=0.20\linewidth]{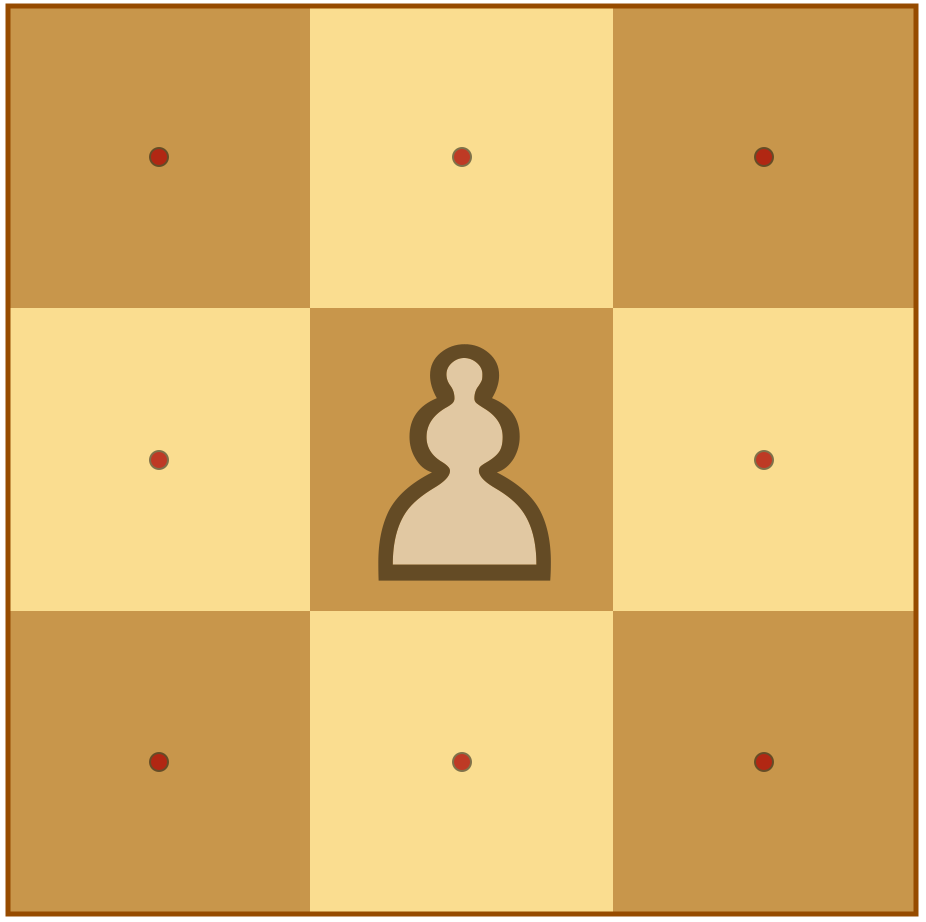}} &  \raisebox{-.5\height}{\includegraphics[width=0.20\linewidth]{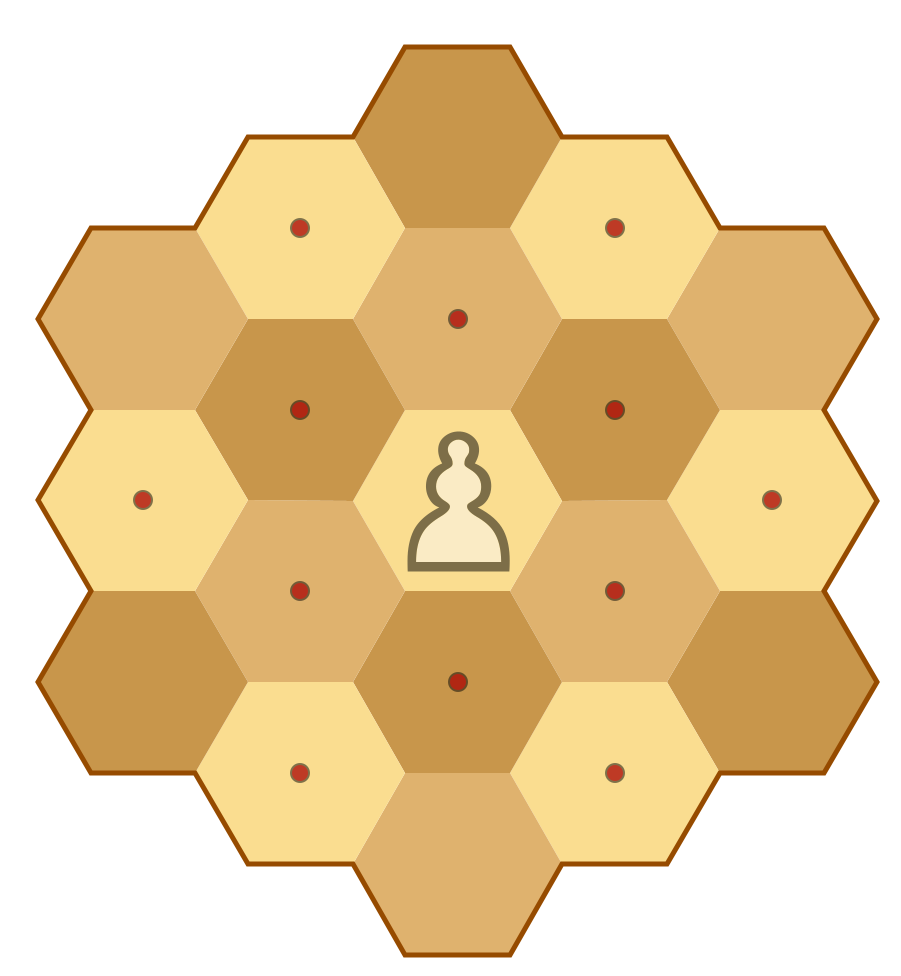}} &  \raisebox{-.5\height}{\includegraphics[width=0.23\linewidth]{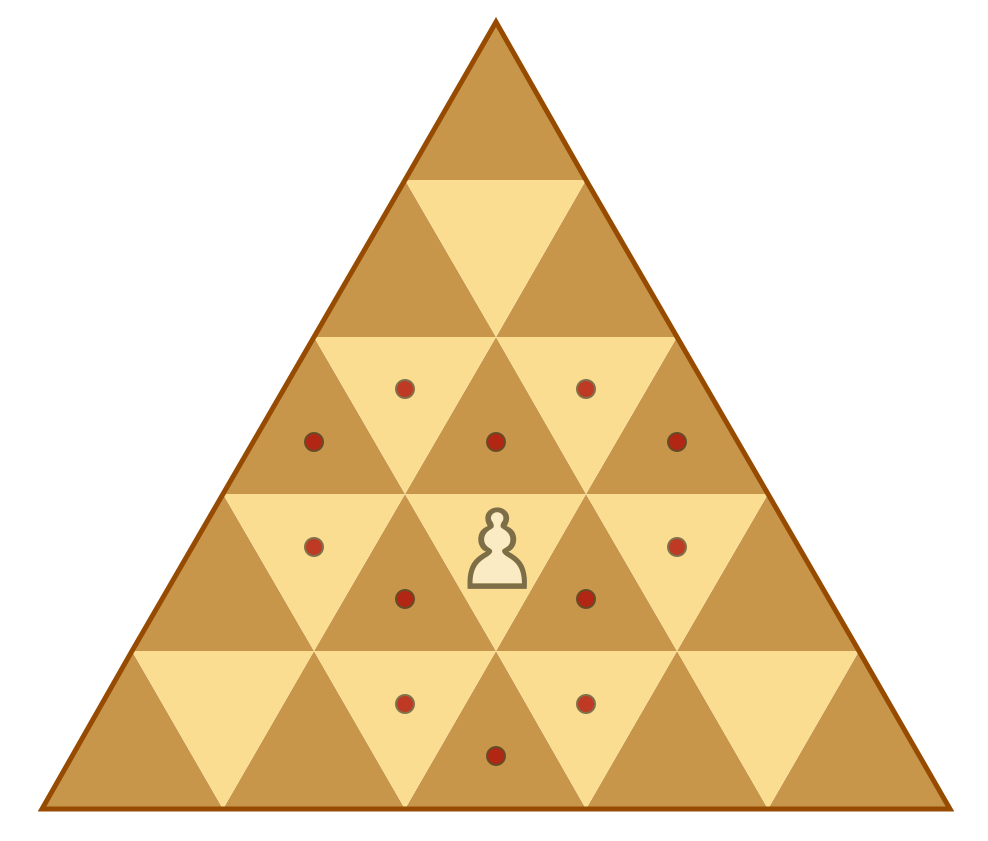}}\\ 
  Adjacent &  \raisebox{-.5\height}{\includegraphics[width=0.20\linewidth]{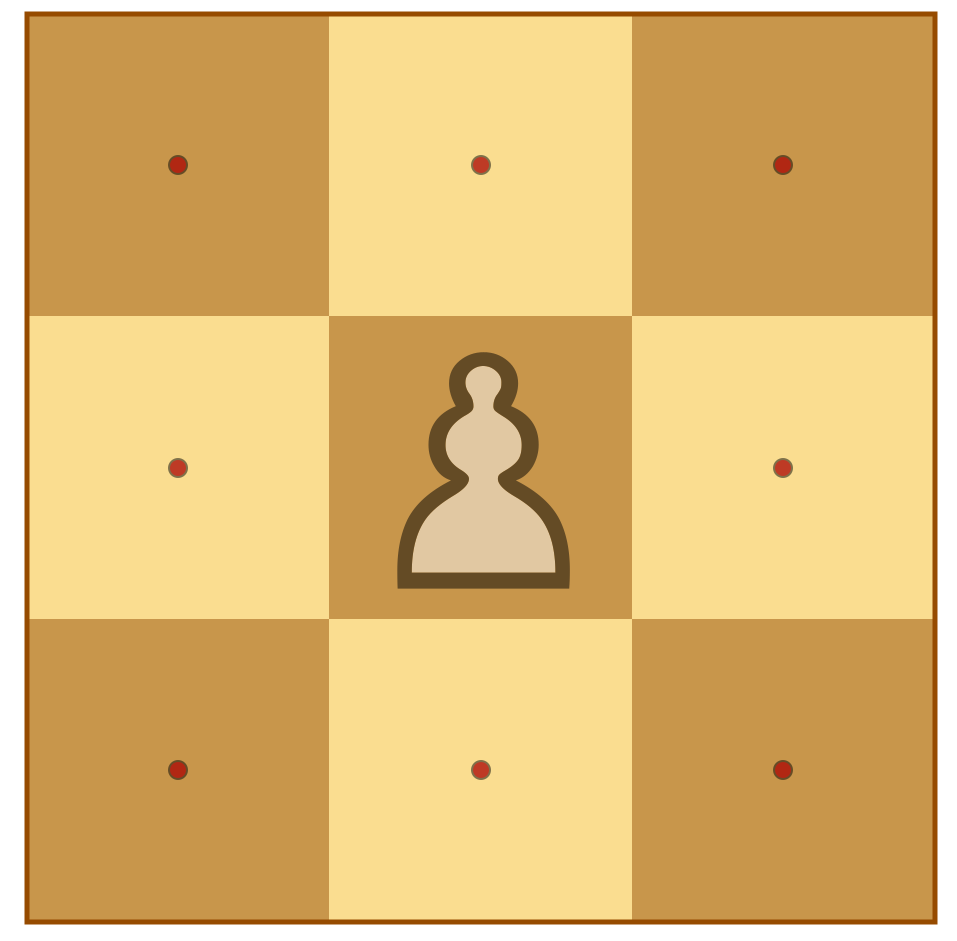}} &  \raisebox{-.5\height}{\includegraphics[width=0.20\linewidth]{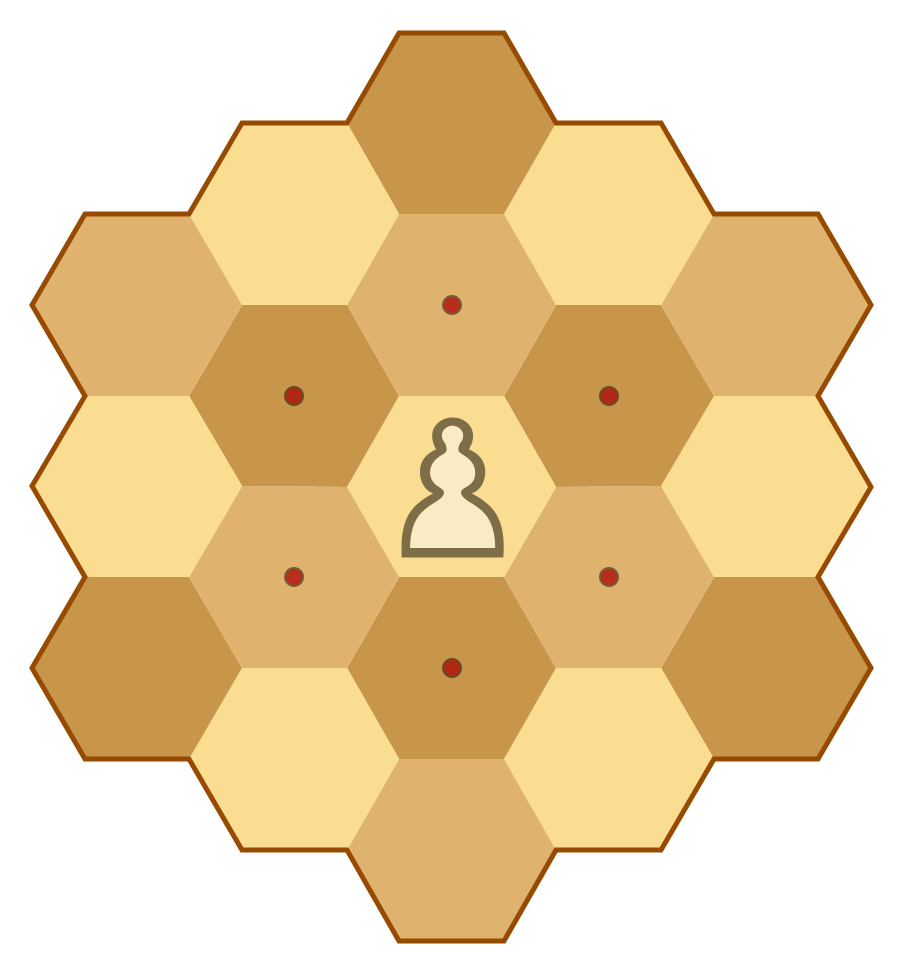}} &  \raisebox{-.5\height}{\includegraphics[width=0.23\linewidth]{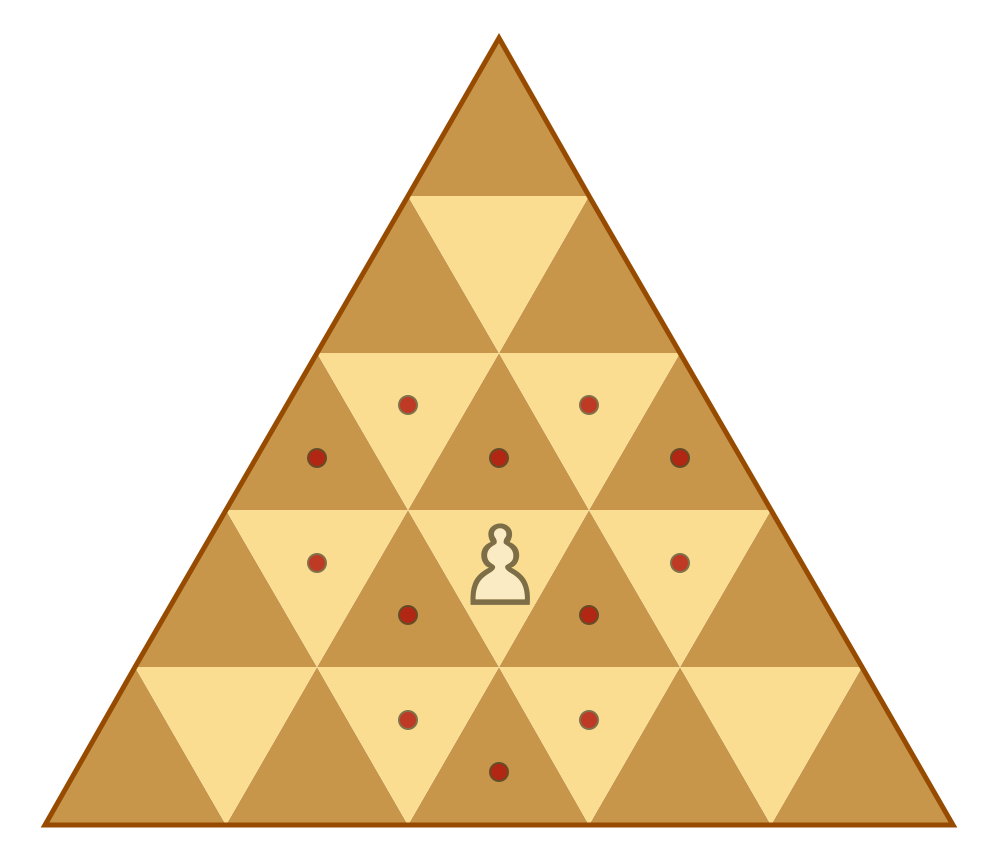}}\\ 
    Orthogonal &  \raisebox{-.5\height}{\includegraphics[width=0.20\linewidth]{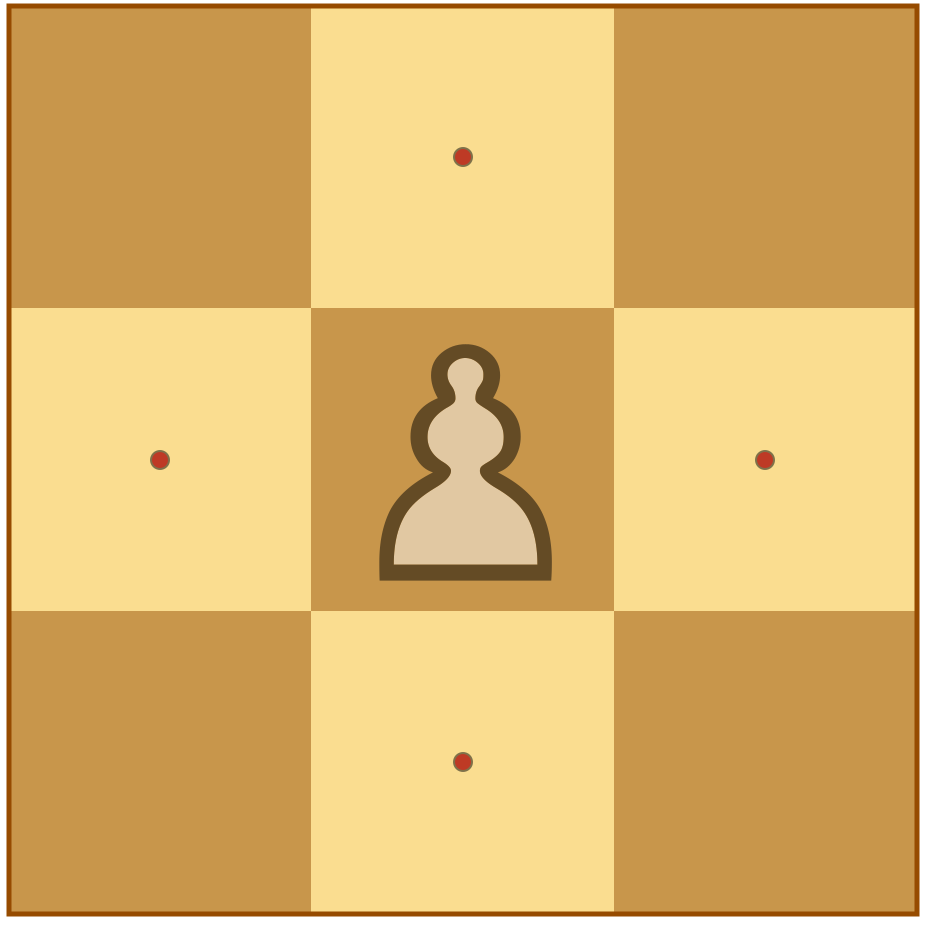}} &  \raisebox{-.5\height}{\includegraphics[width=0.20\linewidth]{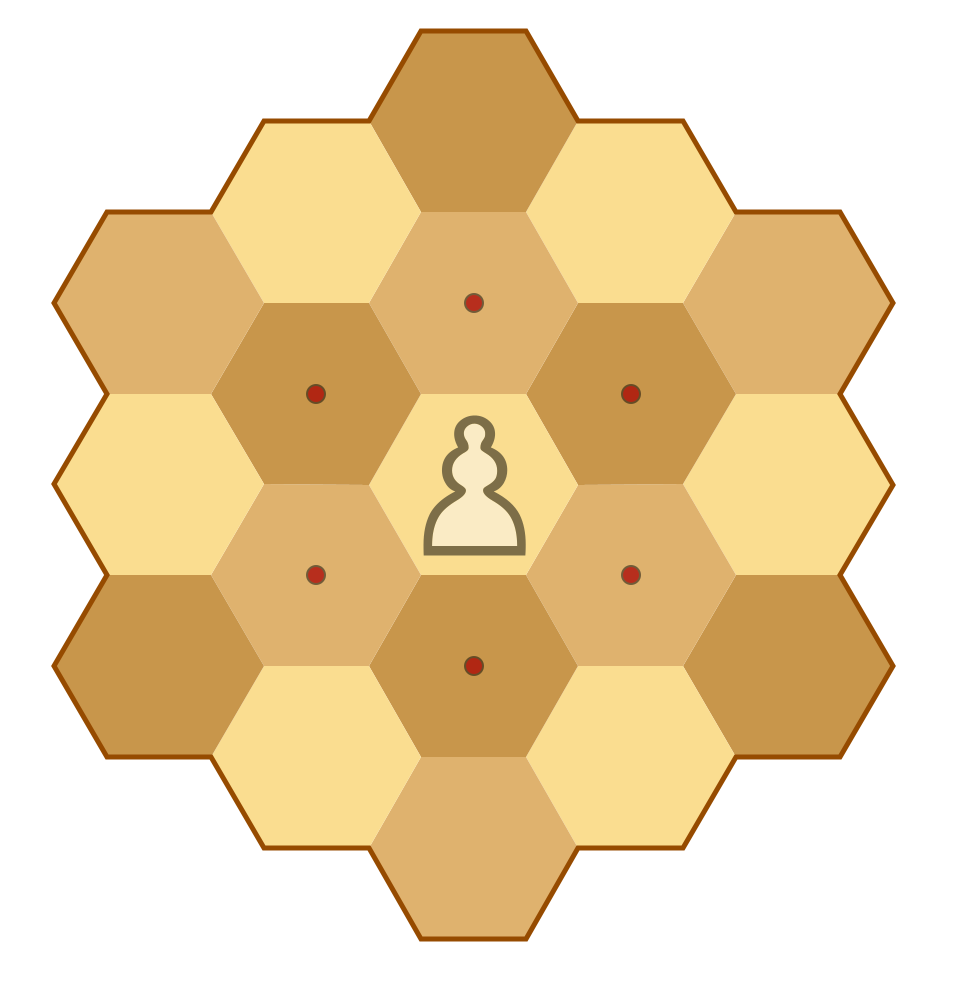}} &  \raisebox{-.5\height}{\includegraphics[width=0.23\linewidth]{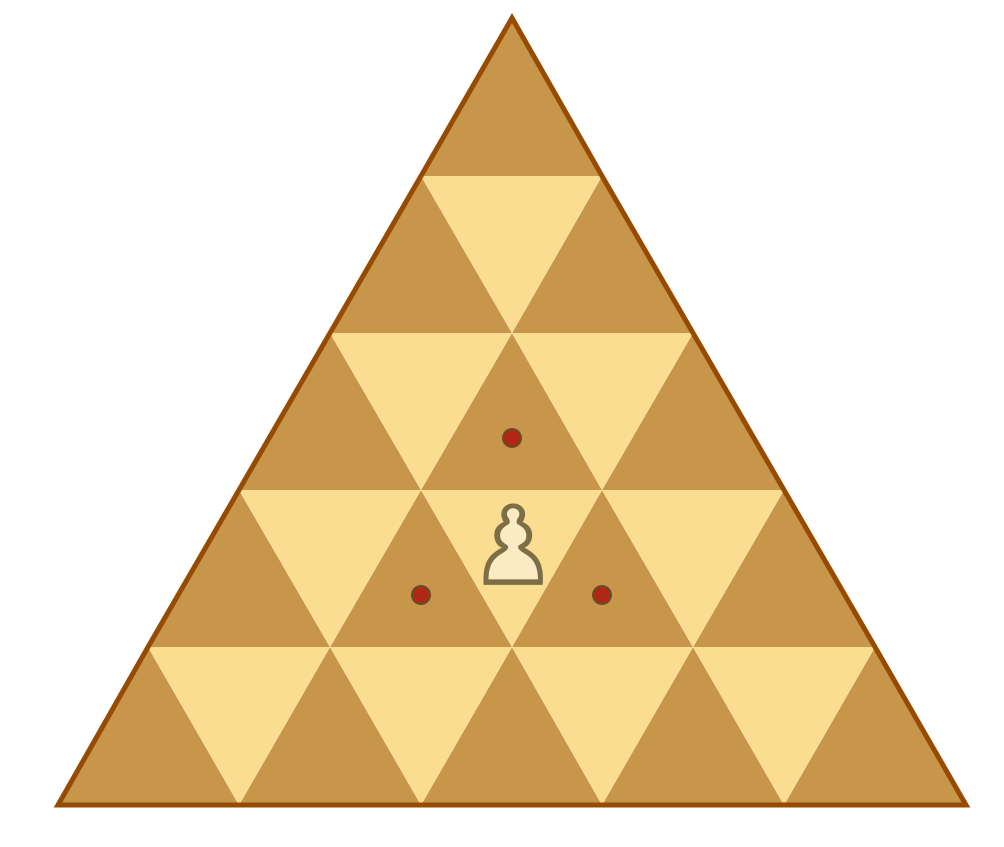}}\\ 
        Diagonal &  \raisebox{-.5\height}{\includegraphics[width=0.20\linewidth]{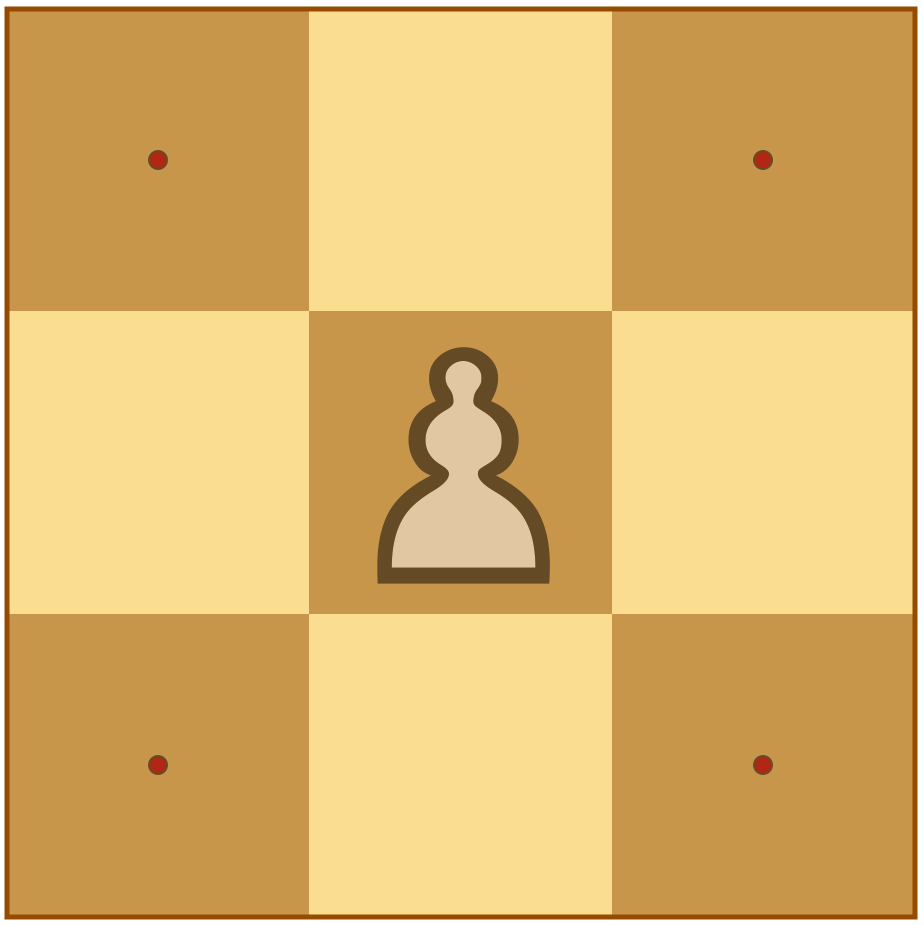}} &  \raisebox{-.5\height}{\includegraphics[width=0.20\linewidth]{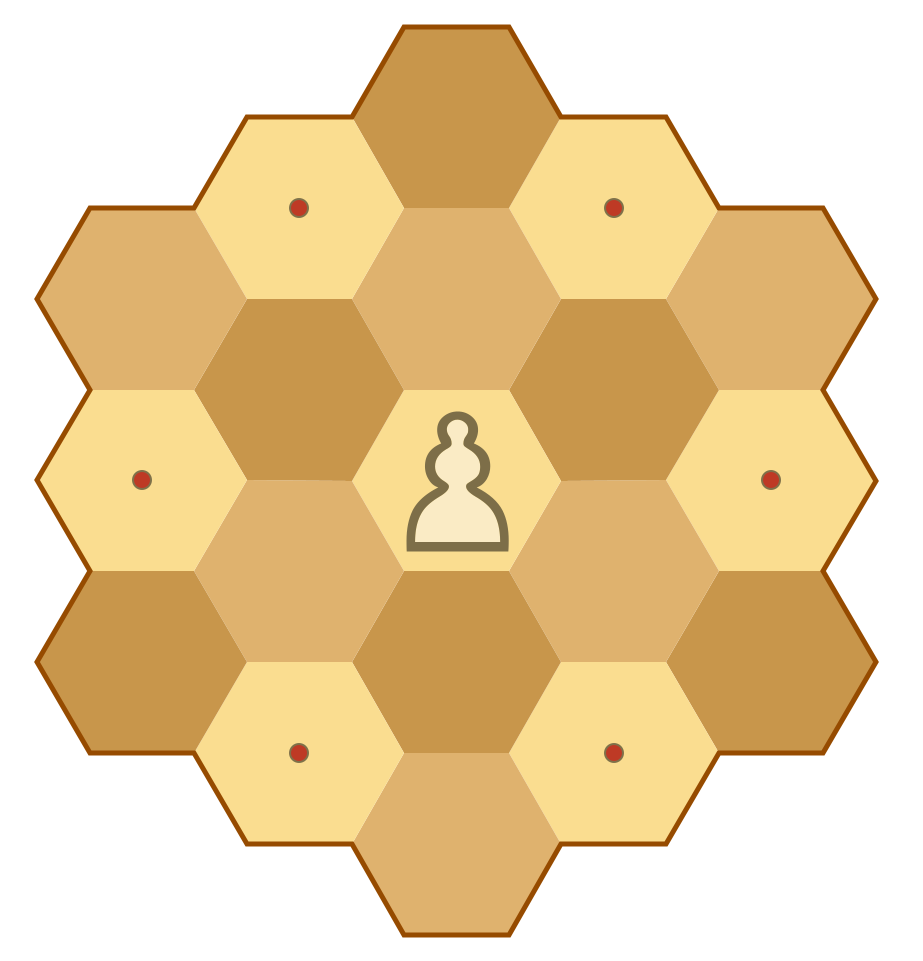}} &  \raisebox{-.5\height}{\includegraphics[width=0.23\linewidth]{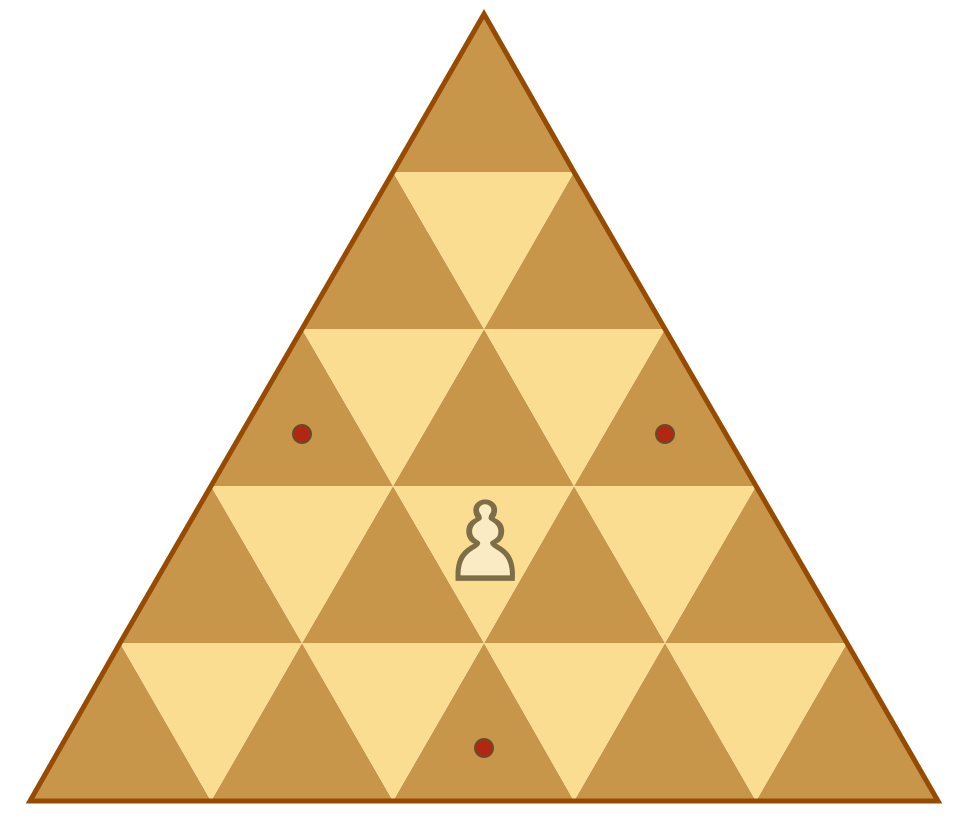}}\\ 
                OffDiagonal &  \raisebox{-.5\height}{\includegraphics[width=0.20\linewidth]{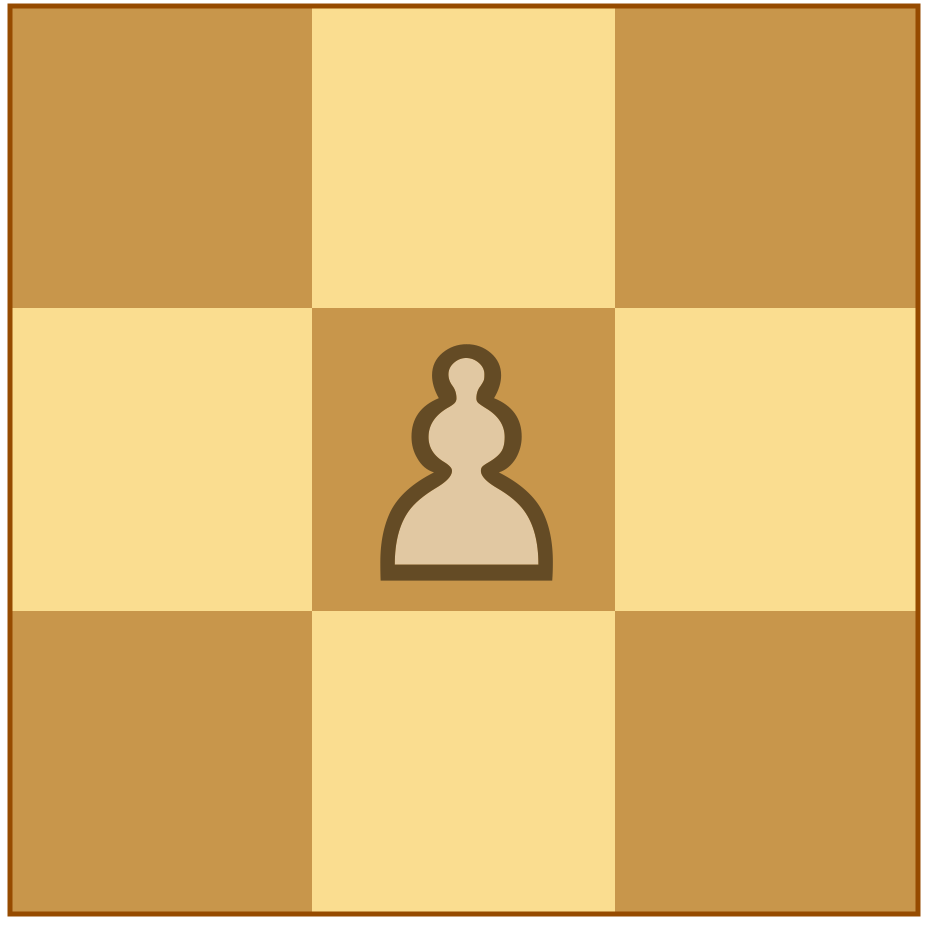}} &  \raisebox{-.5\height}{\includegraphics[width=0.20\linewidth]{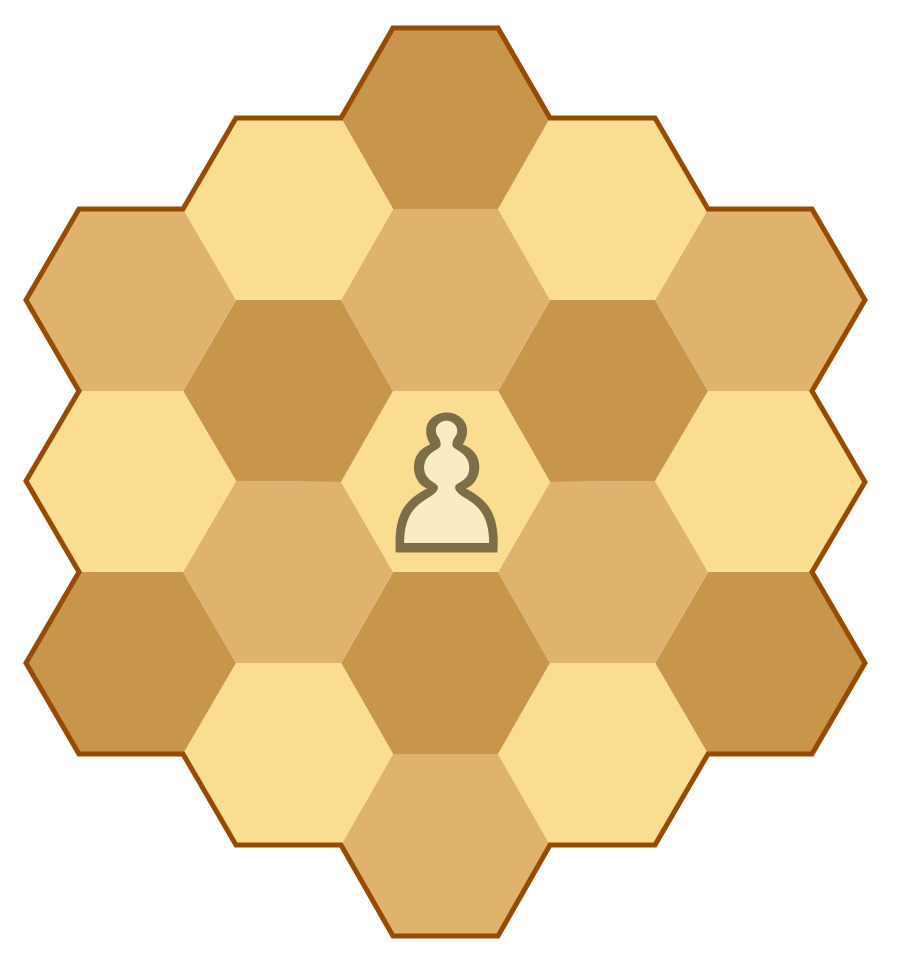}} &  \raisebox{-.5\height}{\includegraphics[width=0.23\linewidth]{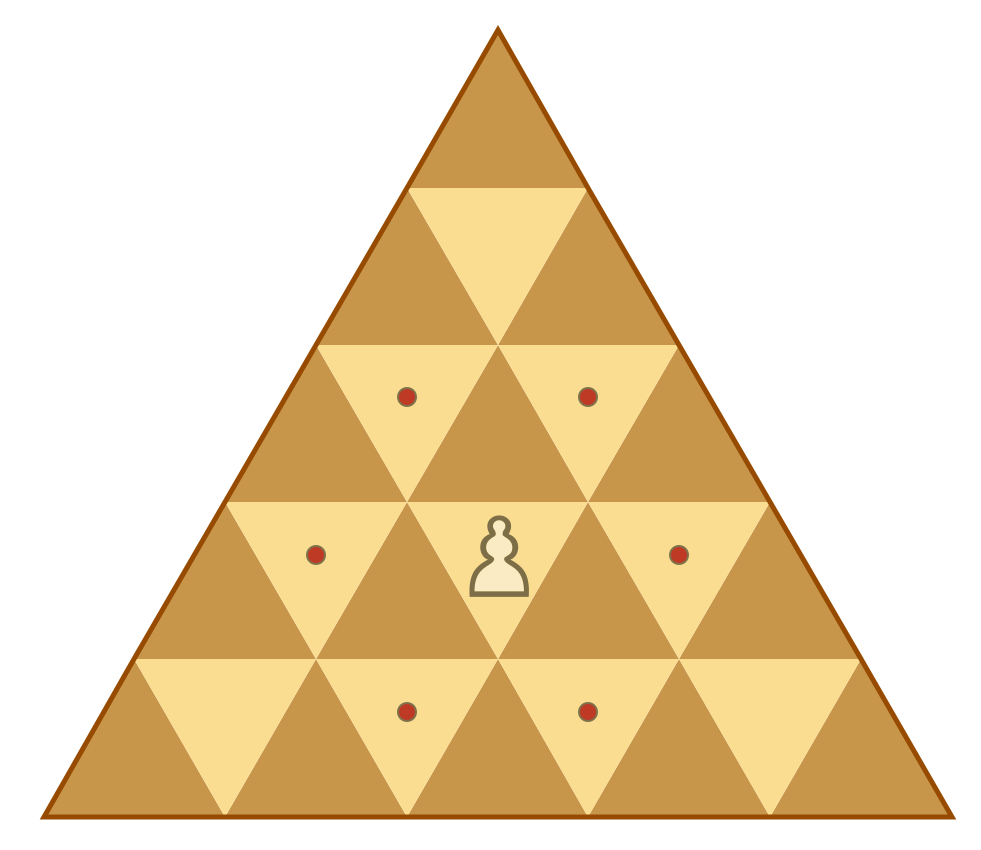}}\\ 
\rowcolor{gray!50}
 \multicolumn{4}{r}{} \\
\end{longtable}

\newpage

The RelativeDirection enumeration type are the following ones:
\renewcommand{\arraystretch}{1.3}
\arrayrulecolor{white}
\rowcolors{2}{gray!25}{gray!10}
\begin{longtable}{@{}p{.25\textwidth} | p{.65\textwidth}@{}}
\rowcolor{gray!50}
\textbf{Relative direction} & \textbf{Description} \\
Forward & The facing direction. \\
Backward & The opposite of the facing direction. \\
Rightward & The most rightward direction between the facing direction and its opposite. \\
Leftward & The most leftward direction between the facing direction and its opposite. \\
Forwards & All forward directions between the leftward and the rightward direction. \\
Backwards & All backward directions between the rightward and the leftward direction. \\
Rightwards & All rightward directions between the forward and the backward direction. \\
Leftwards & All leftward directions between the backward and the forward direction. \\
FL & The direction at the left of the forward direction. \\
FLL & The second direction at the left of the forward direction. \\
FLLL & The third direction at the left of the forward direction. \\
BL & The direction at the left of the backward direction. \\
BLL & The second direction at the left of the backward direction. \\
BLLL & The third direction at the left of the backward direction. \\
FR & The direction at the right of the forward direction. \\
FRR & The second direction at the right of the forward direction. \\
FRRR & The third direction at the right of the forward direction. \\
BR & The direction at the right of the backward direction. \\
BRR & The second direction at the right of the backward direction. \\
BRRR & The third direction at the right of the backward direction. \\
SameDirection & The same direction of the previous move. \\
OppositeDirection & The opposite direction of the previous move. \\
\end{longtable}
\rowcolors{0}{}{}

Here are the different step moves of a piece facing to the north for the common relative directions for the 3 common tilings: Square, Hexagonal and Triangle:

\renewcommand{\arraystretch}{1.3}
\arrayrulecolor{white}
\begin{longtable}{c|c|c|c}
\rowcolor{gray!50}
\textbf{Relative Direction} & \textbf{Square} & \textbf{Hexagonal} & \textbf{Triangle}\\
 Forward &  \raisebox{-.5\height}{\includegraphics[width=0.20\linewidth]{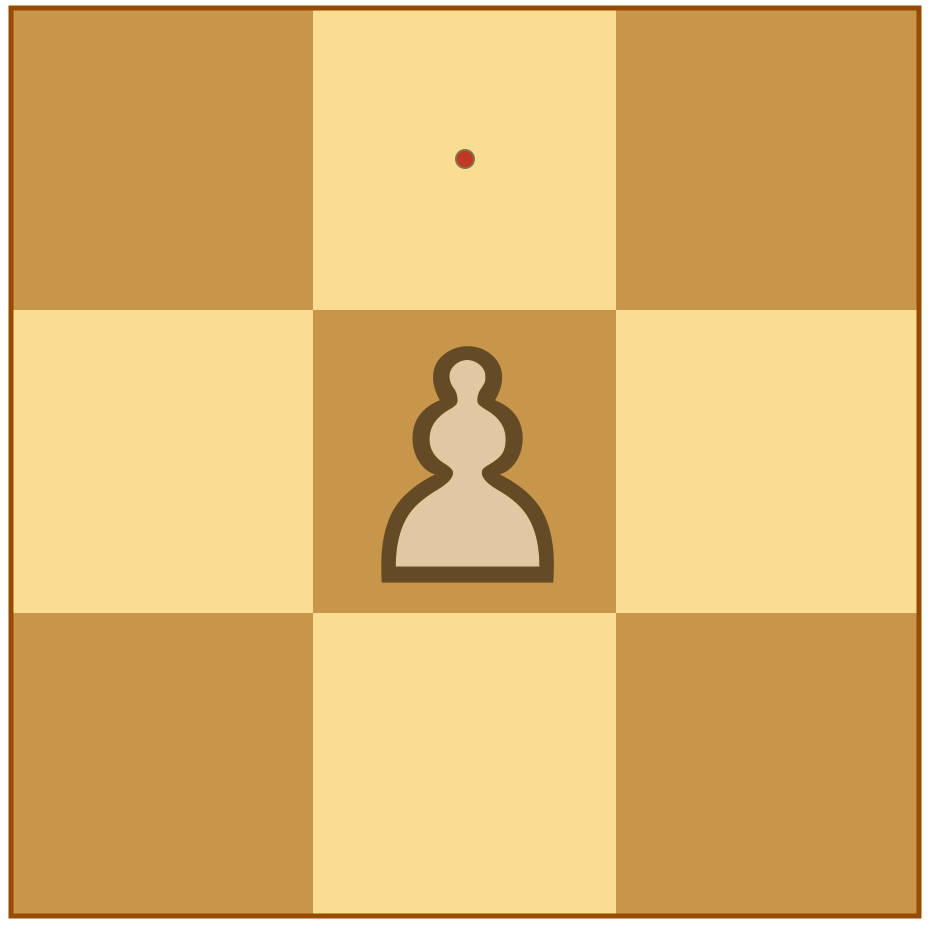}} &  \raisebox{-.5\height}{\includegraphics[width=0.20\linewidth]{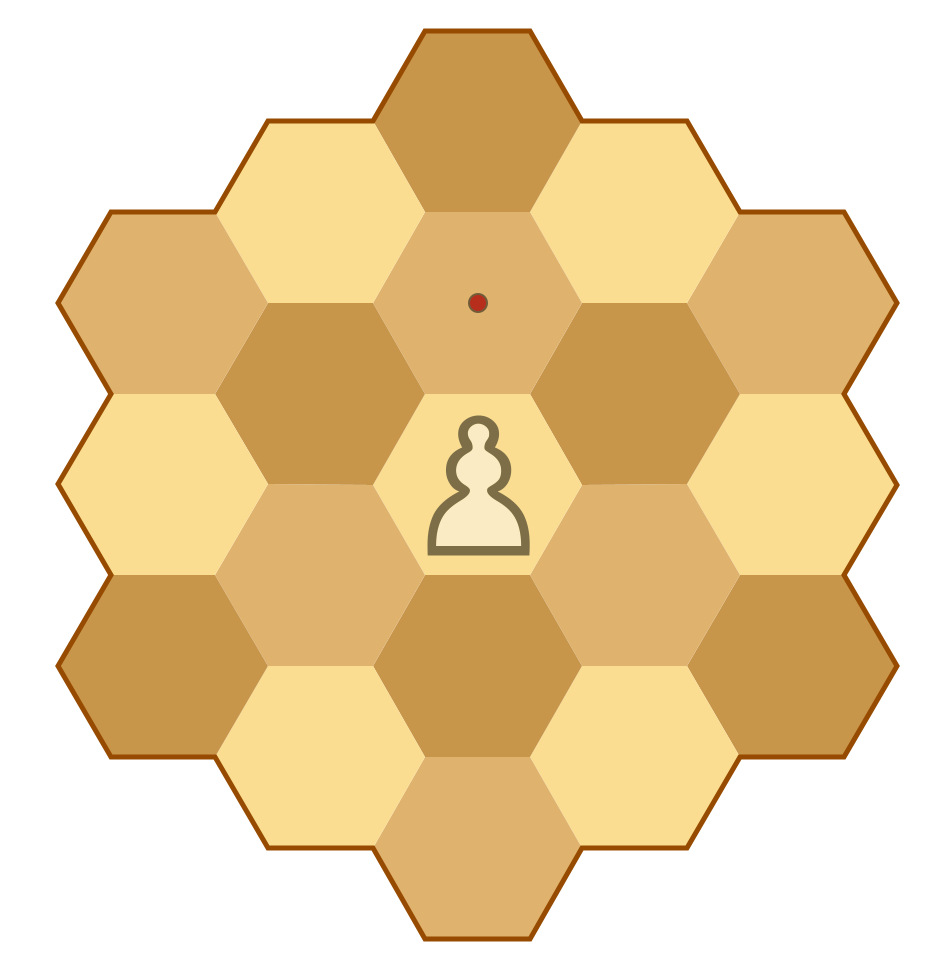}} &  \raisebox{-.5\height}{\includegraphics[width=0.23\linewidth]{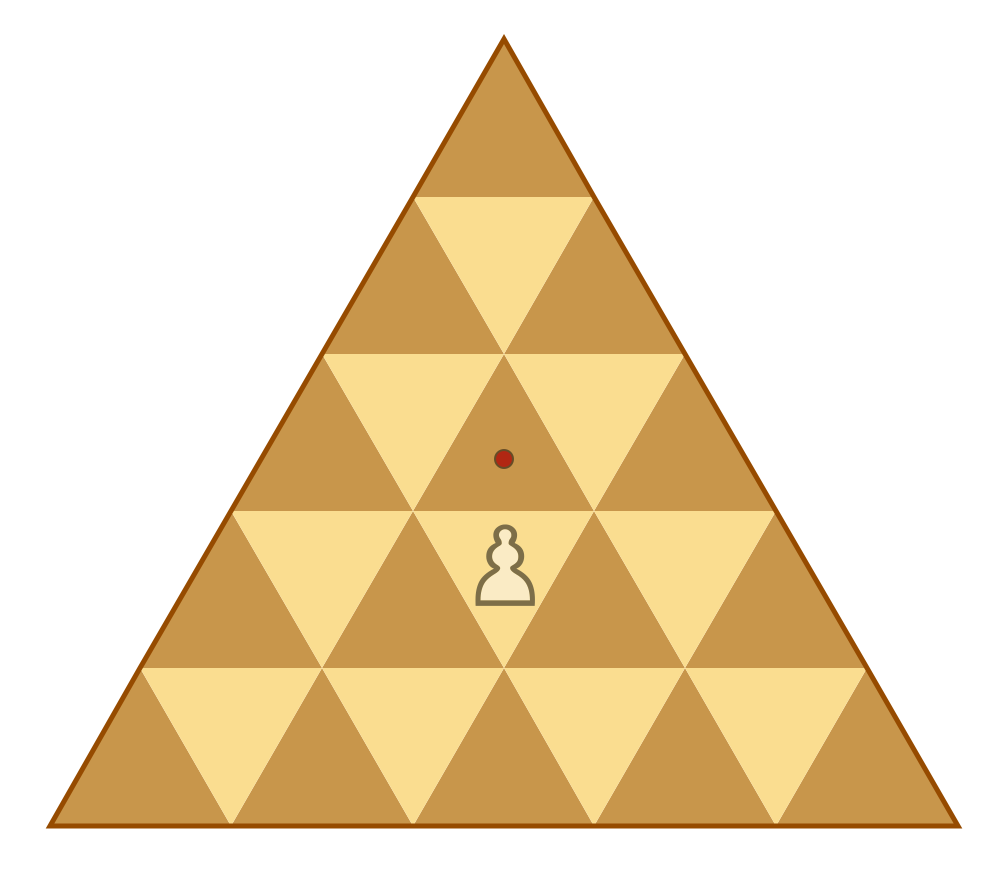}}\\ 
 Backward &  \raisebox{-.5\height}{\includegraphics[width=0.20\linewidth]{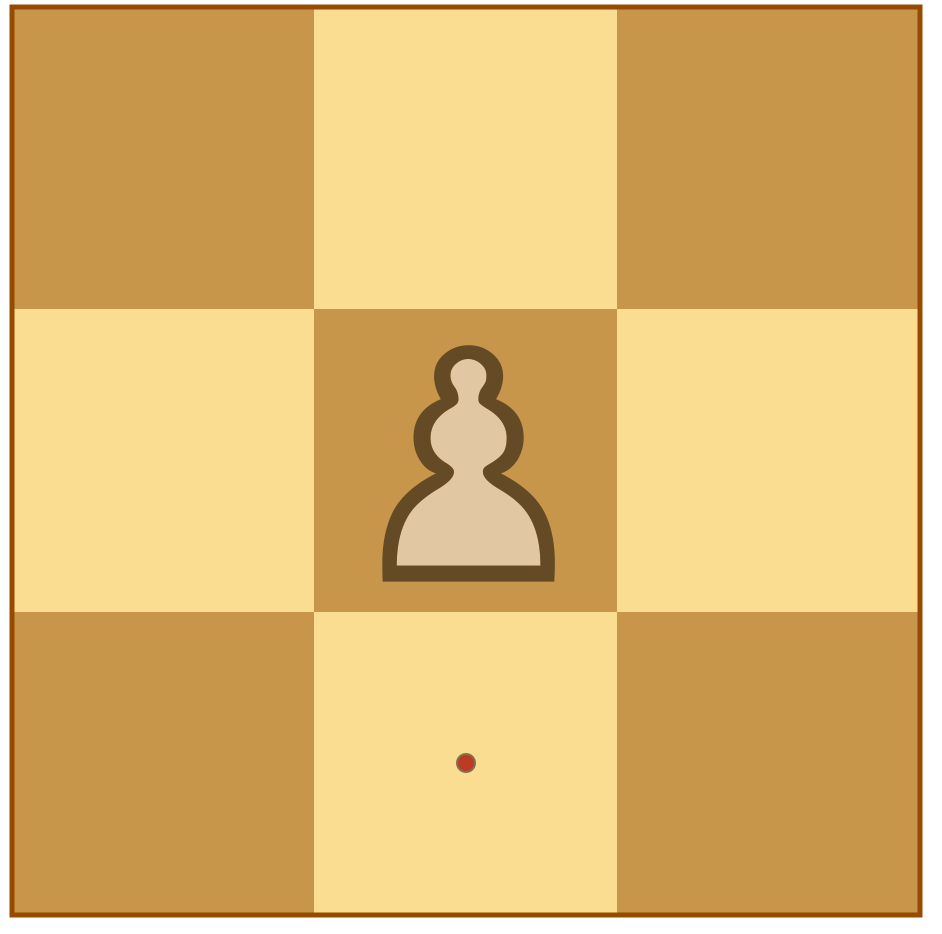}} &  \raisebox{-.5\height}{\includegraphics[width=0.20\linewidth]{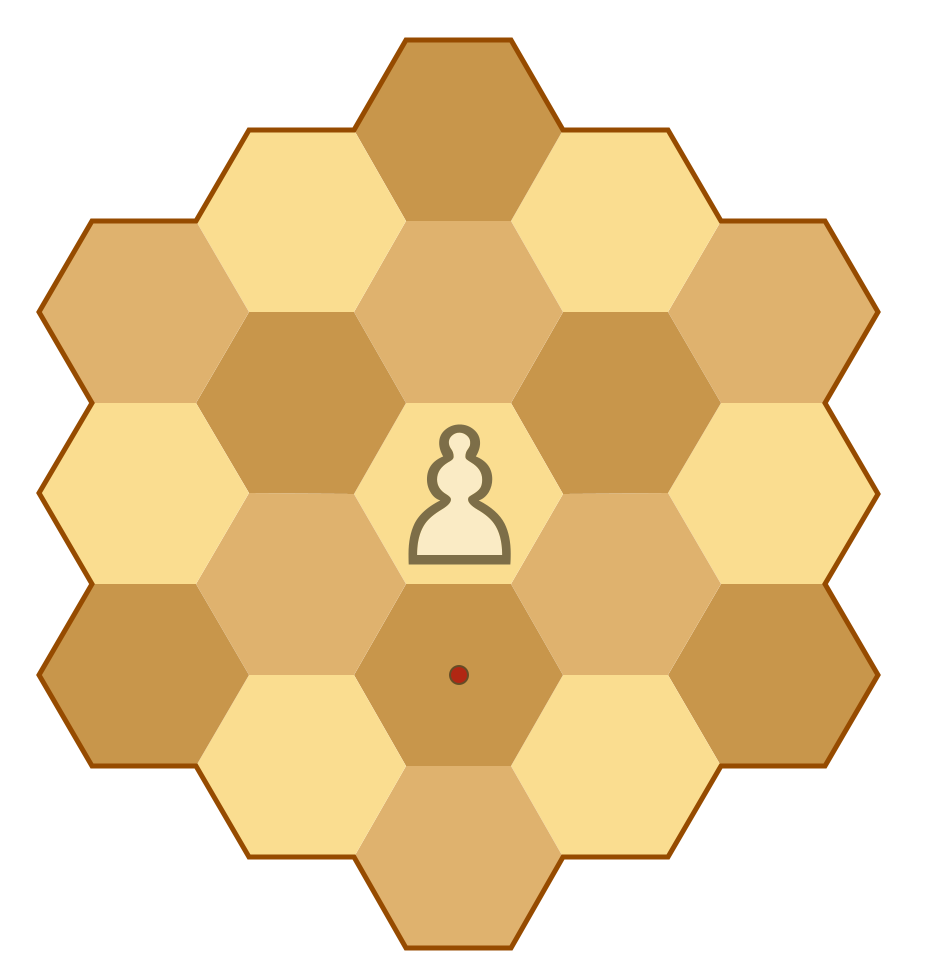}} &  \raisebox{-.5\height}{\includegraphics[width=0.23\linewidth]{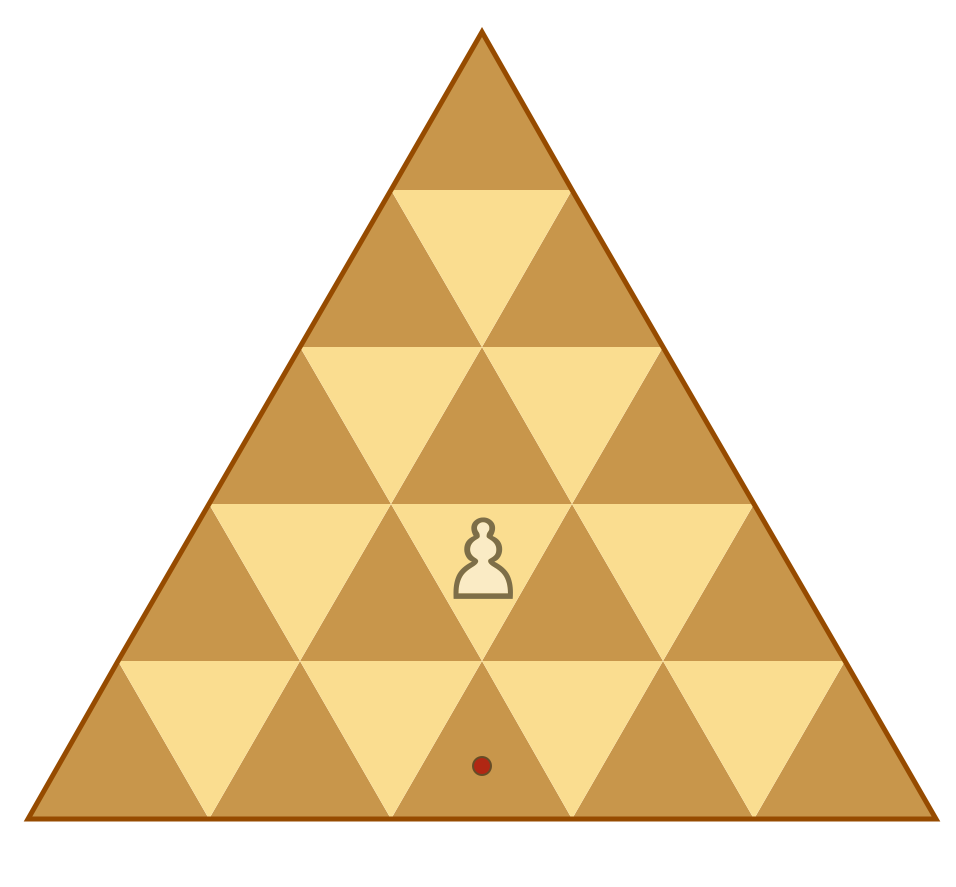}}\\ 
 Leftward &  \raisebox{-.5\height}{\includegraphics[width=0.20\linewidth]{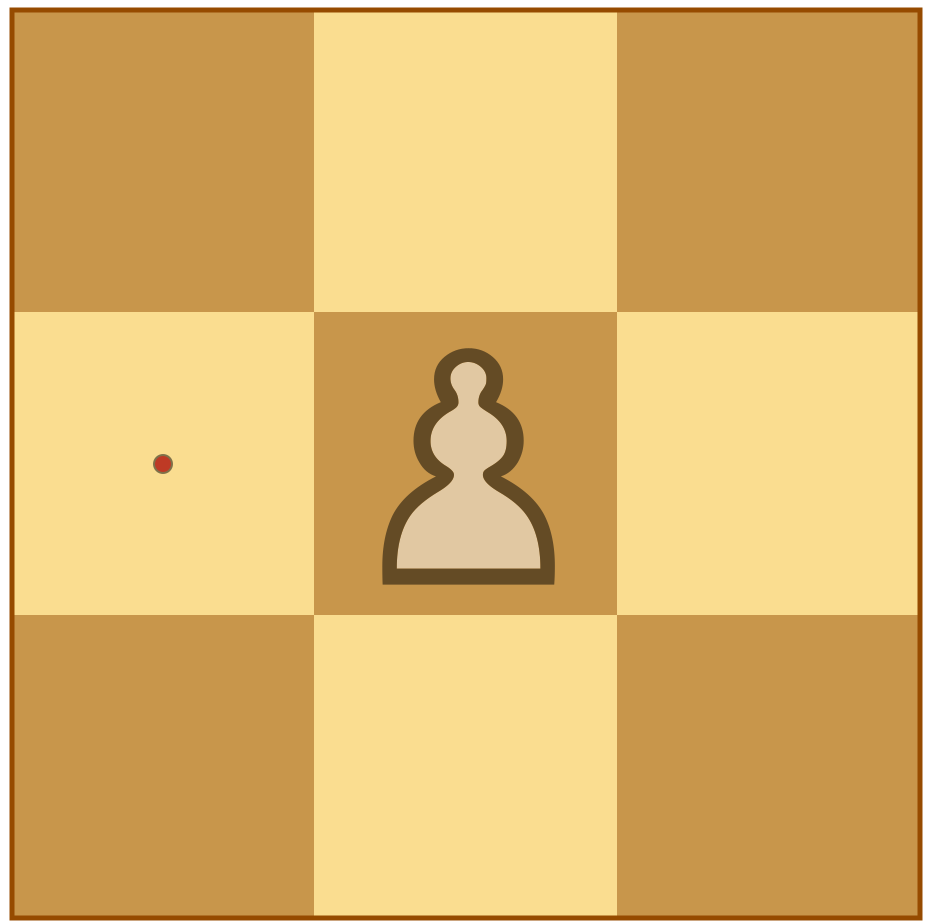}} &  \raisebox{-.5\height}{\includegraphics[width=0.20\linewidth]{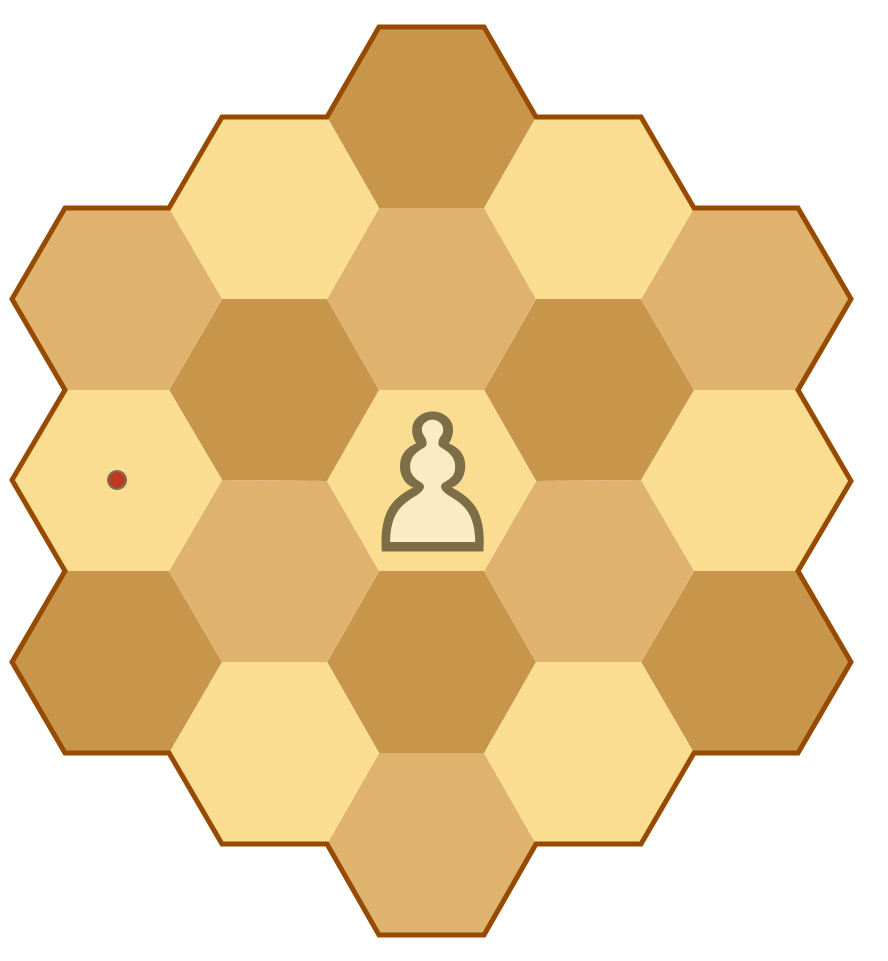}} &  \raisebox{-.5\height}{\includegraphics[width=0.23\linewidth]{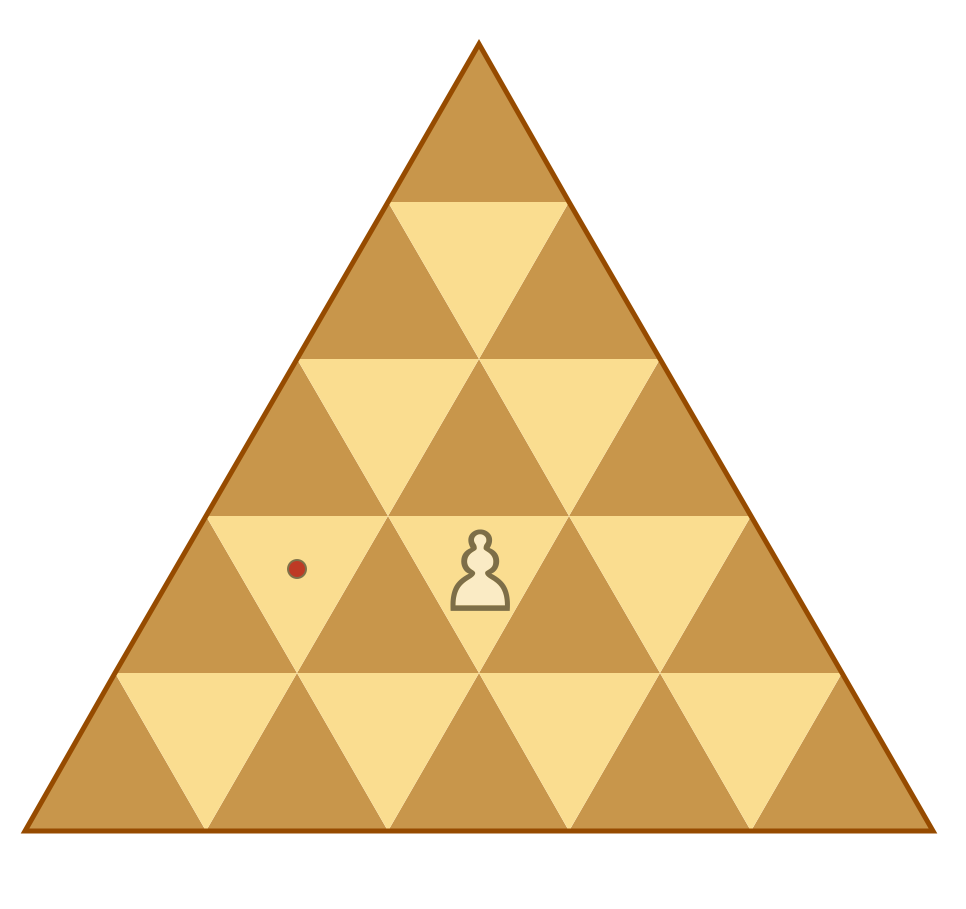}}\\ 
 Rightward &  \raisebox{-.5\height}{\includegraphics[width=0.20\linewidth]{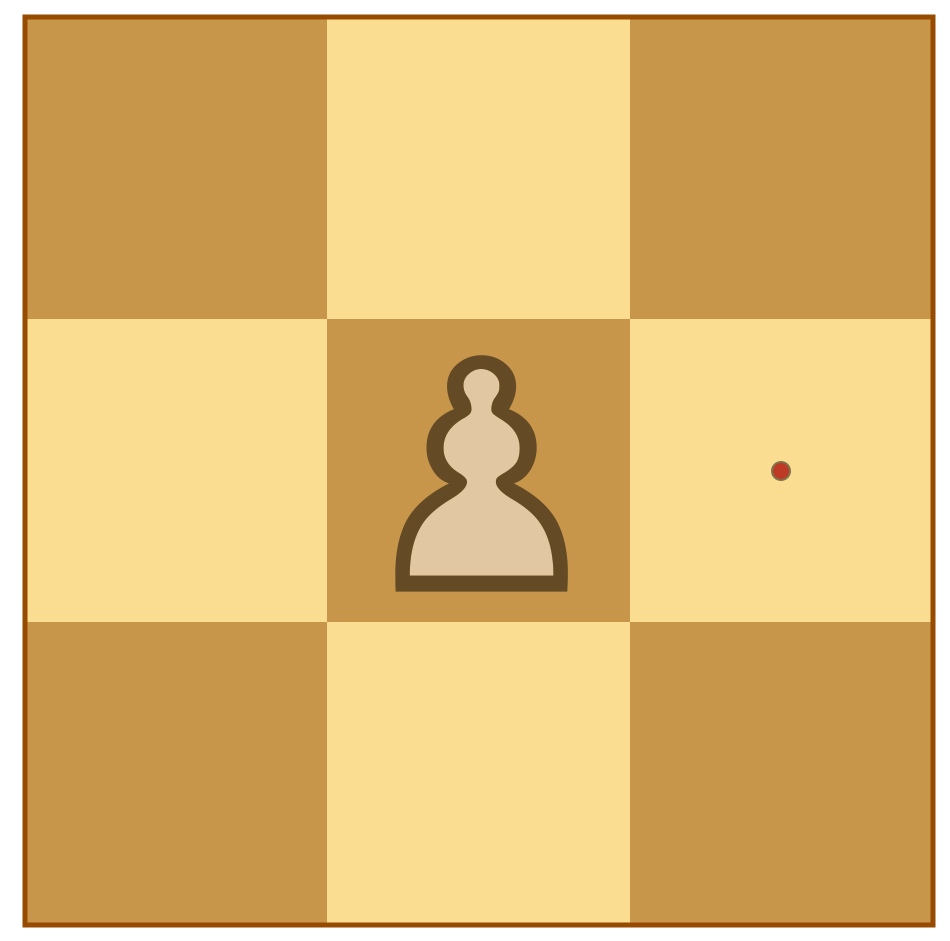}} &  \raisebox{-.5\height}{\includegraphics[width=0.20\linewidth]{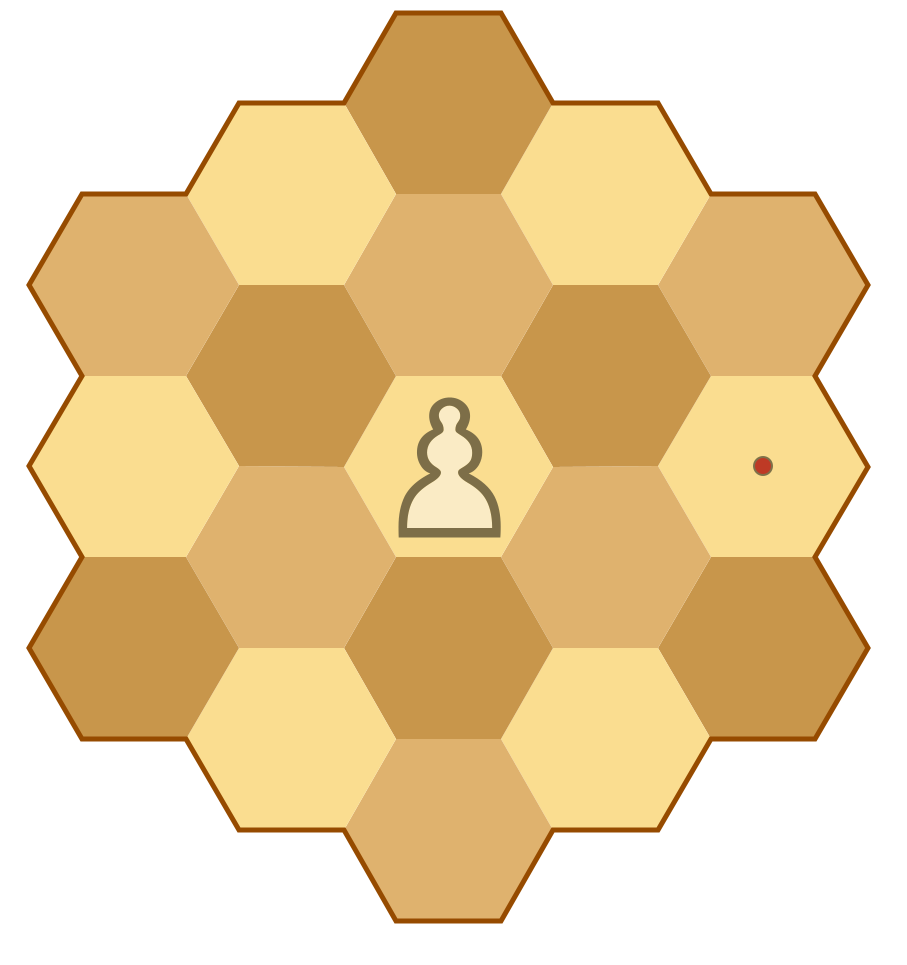}} &  \raisebox{-.5\height}{\includegraphics[width=0.23\linewidth]{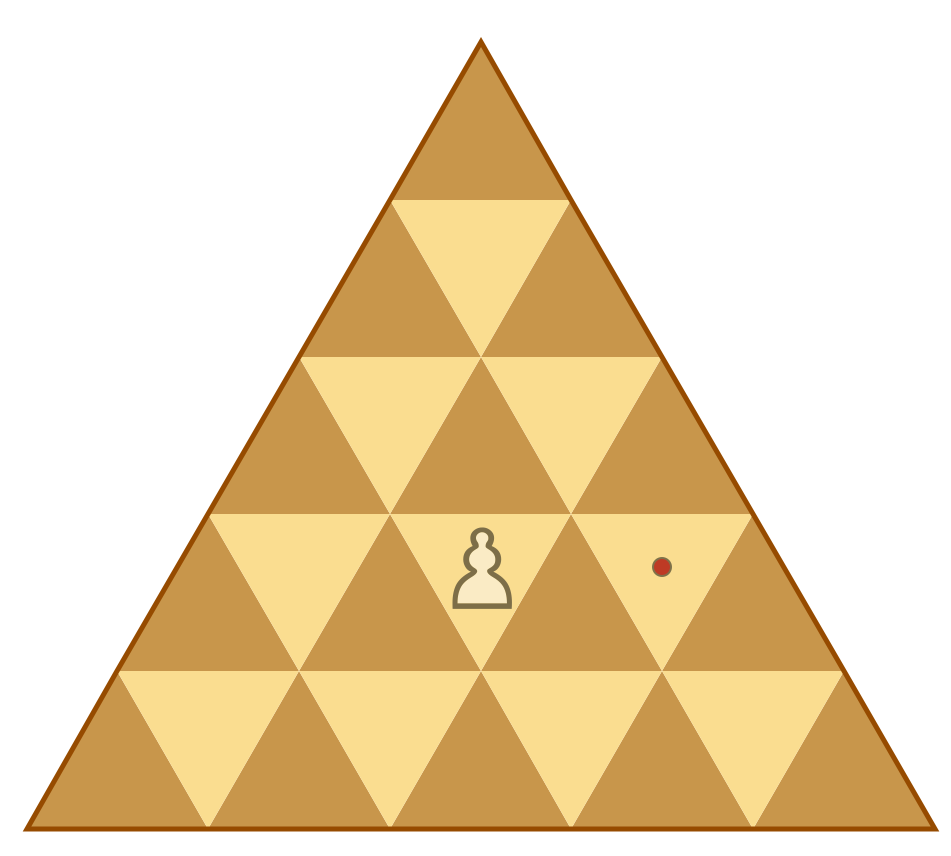}}\\ 
  Forwards &  \raisebox{-.5\height}{\includegraphics[width=0.20\linewidth]{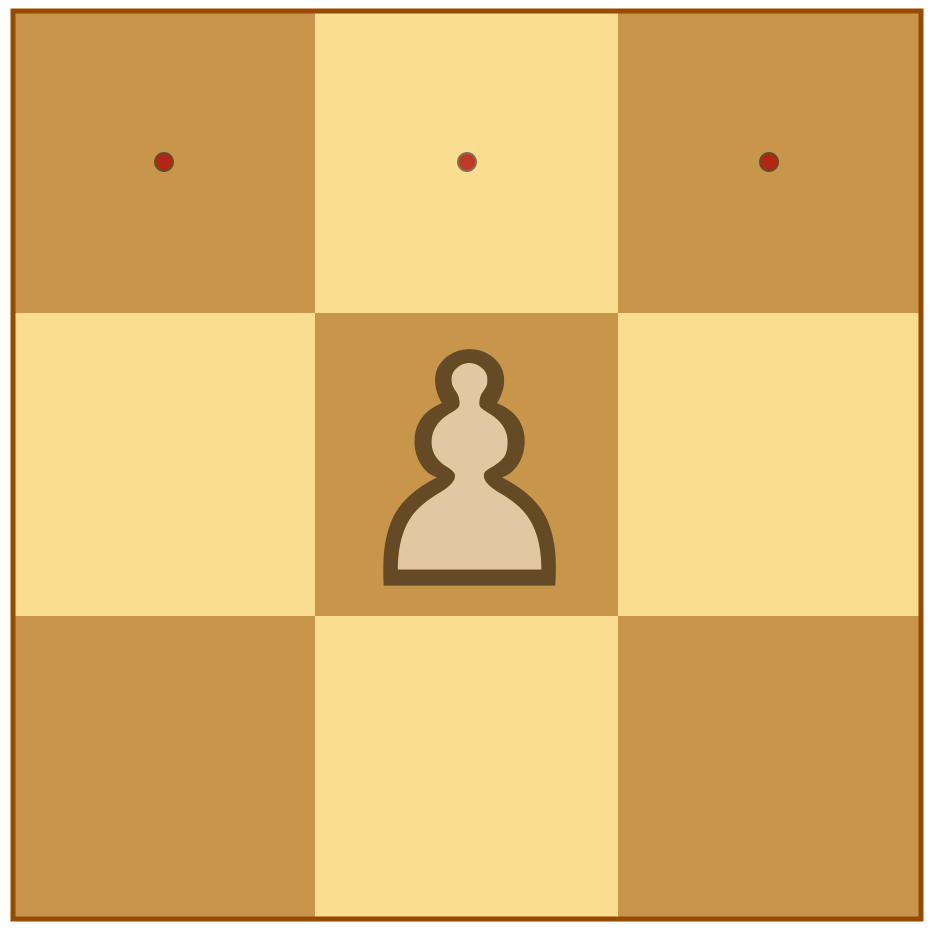}} &  \raisebox{-.5\height}{\includegraphics[width=0.20\linewidth]{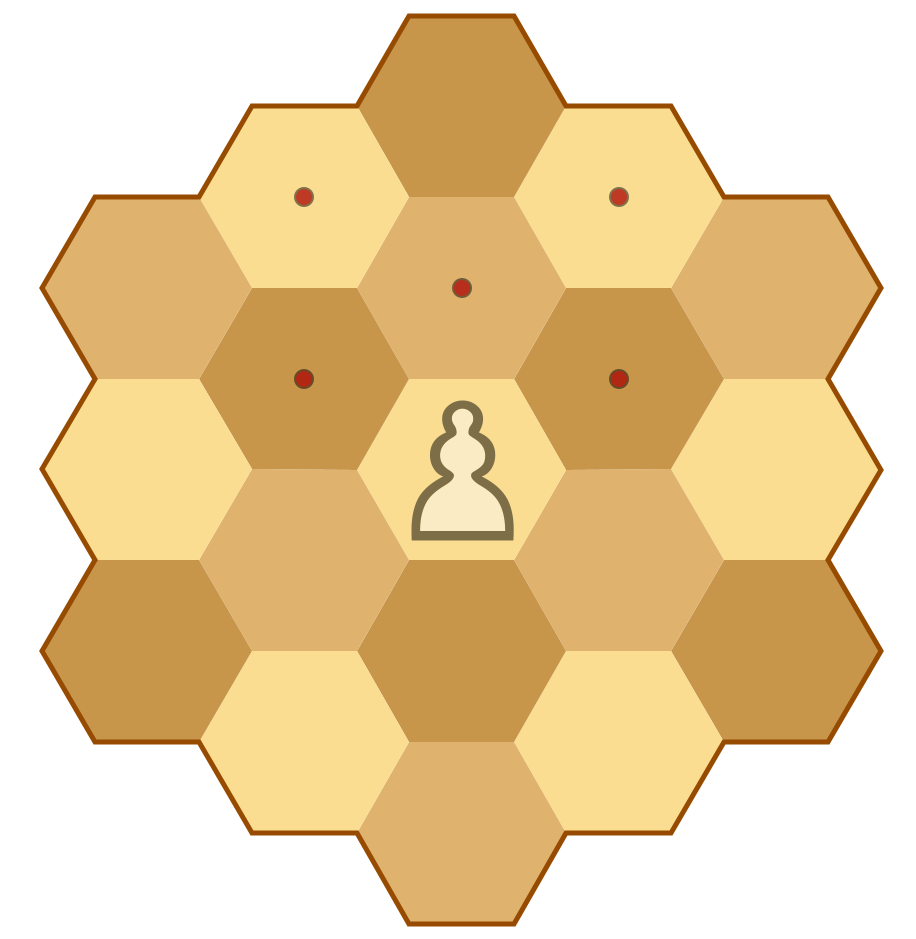}} &  \raisebox{-.5\height}{\includegraphics[width=0.23\linewidth]{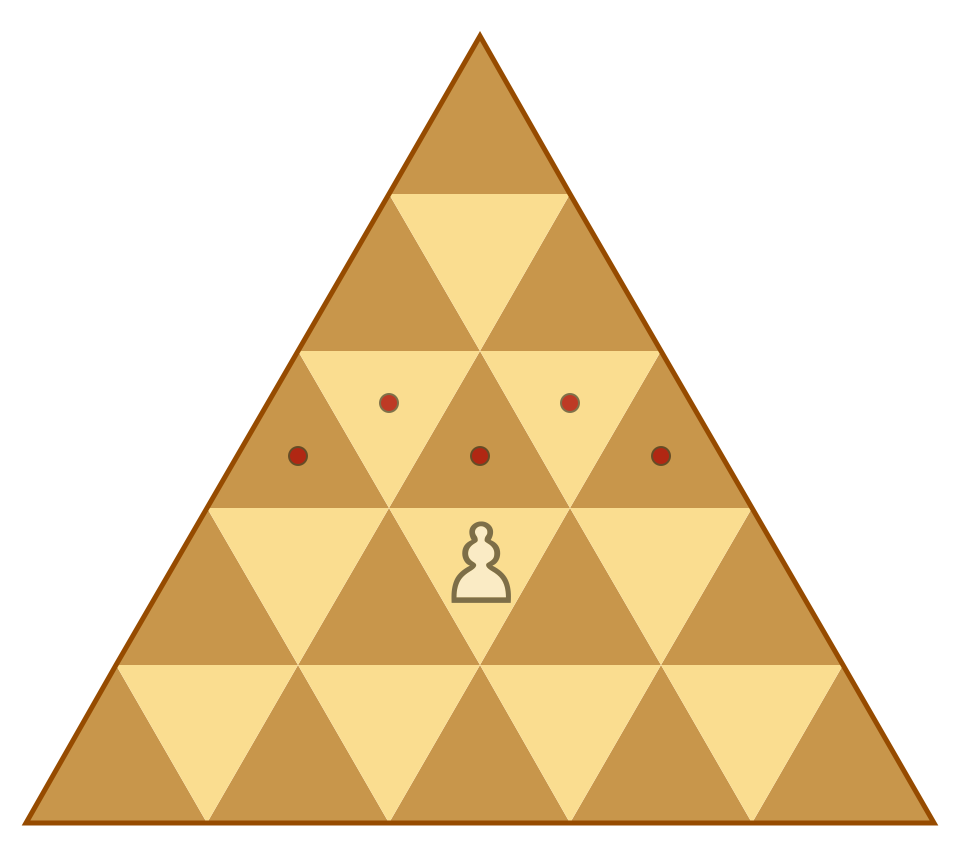}}\\ 
 Backwards &  \raisebox{-.5\height}{\includegraphics[width=0.20\linewidth]{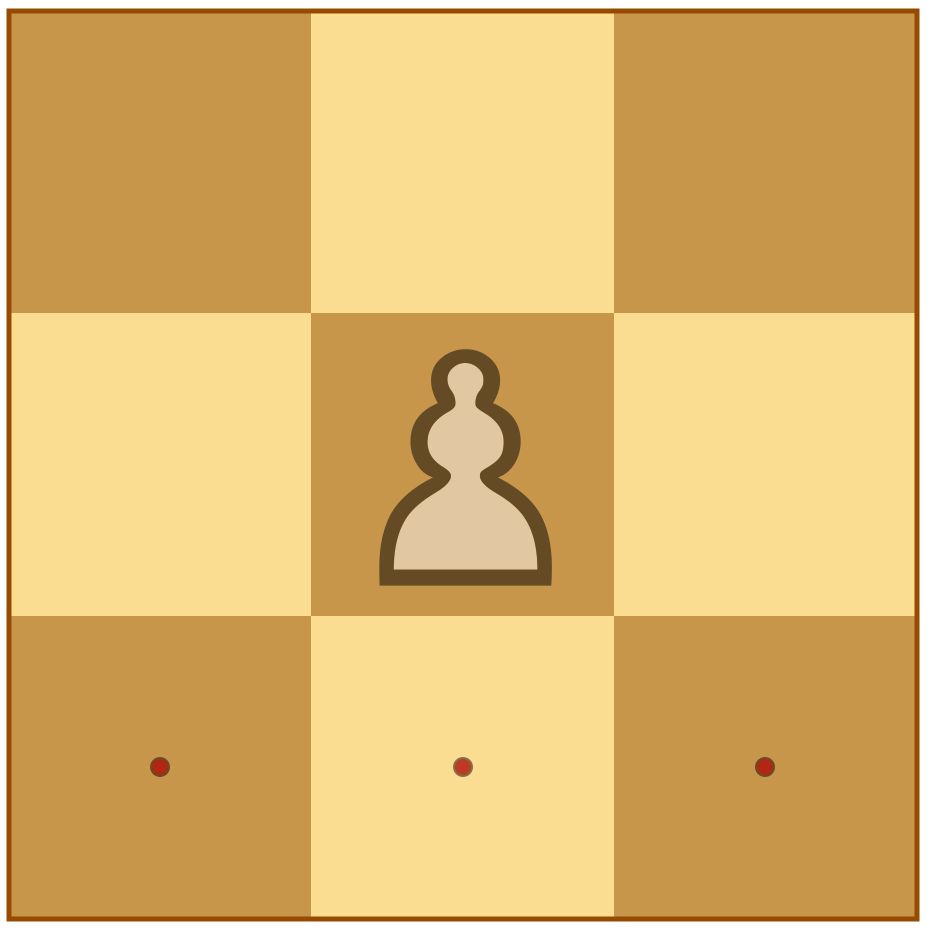}} &  \raisebox{-.5\height}{\includegraphics[width=0.20\linewidth]{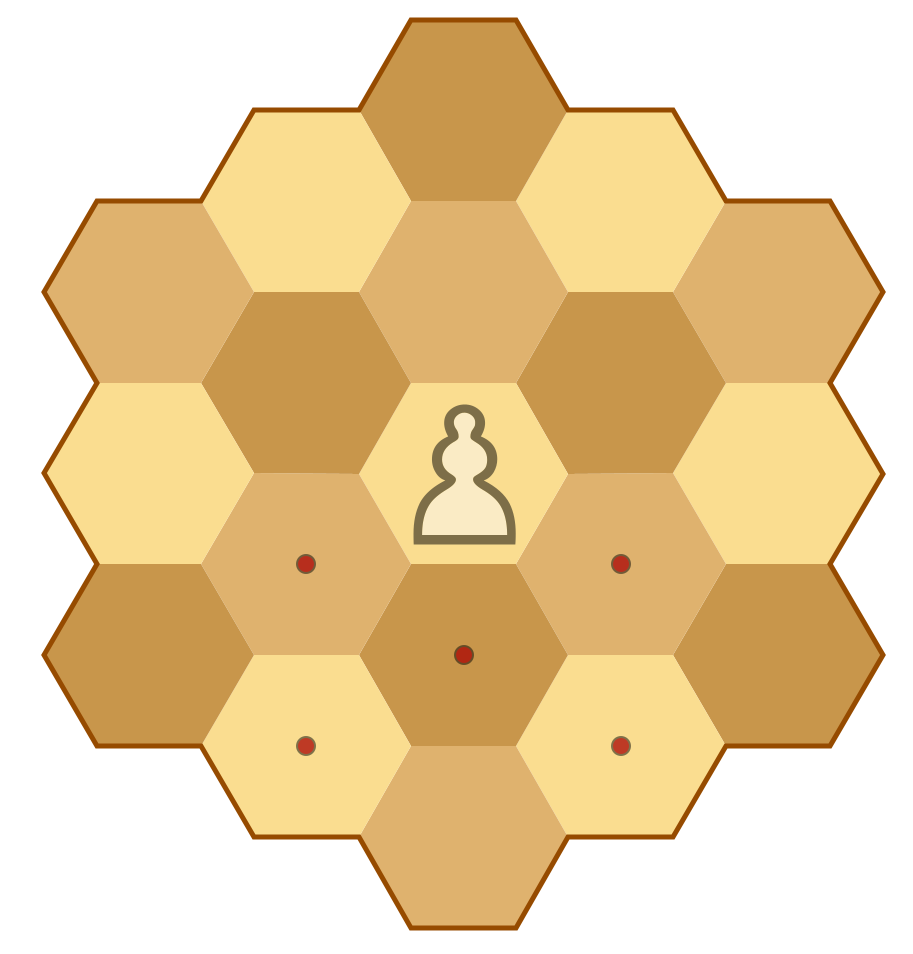}} &  \raisebox{-.5\height}{\includegraphics[width=0.23\linewidth]{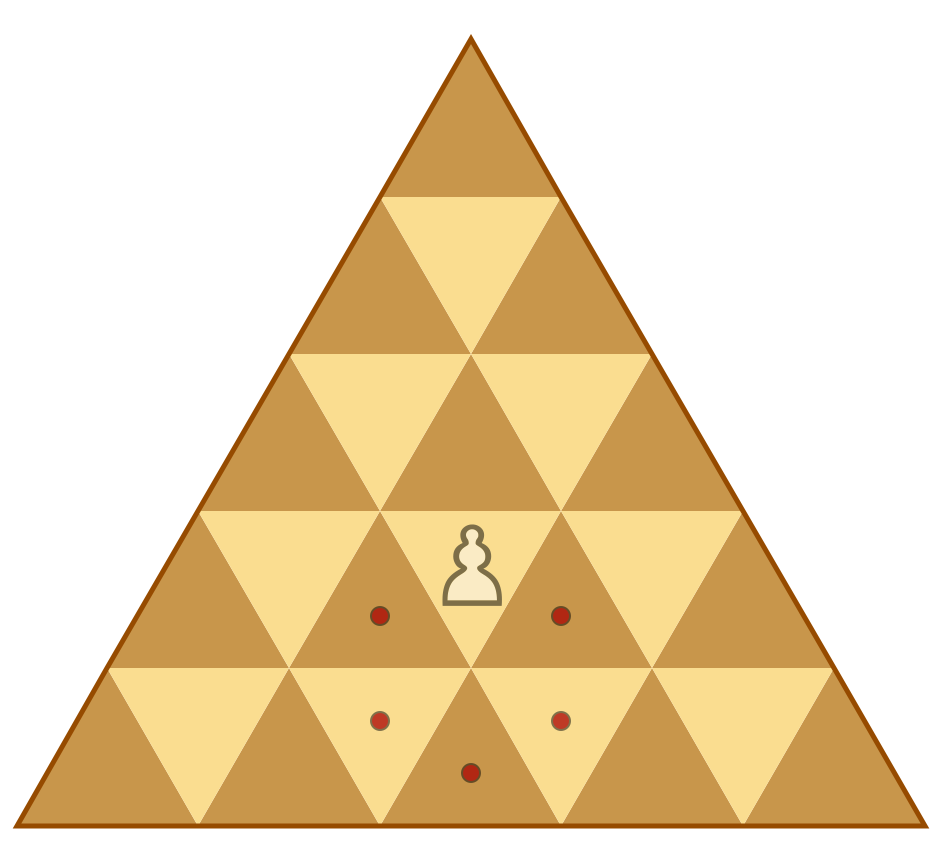}}\\ 
 Leftwards &  \raisebox{-.5\height}{\includegraphics[width=0.20\linewidth]{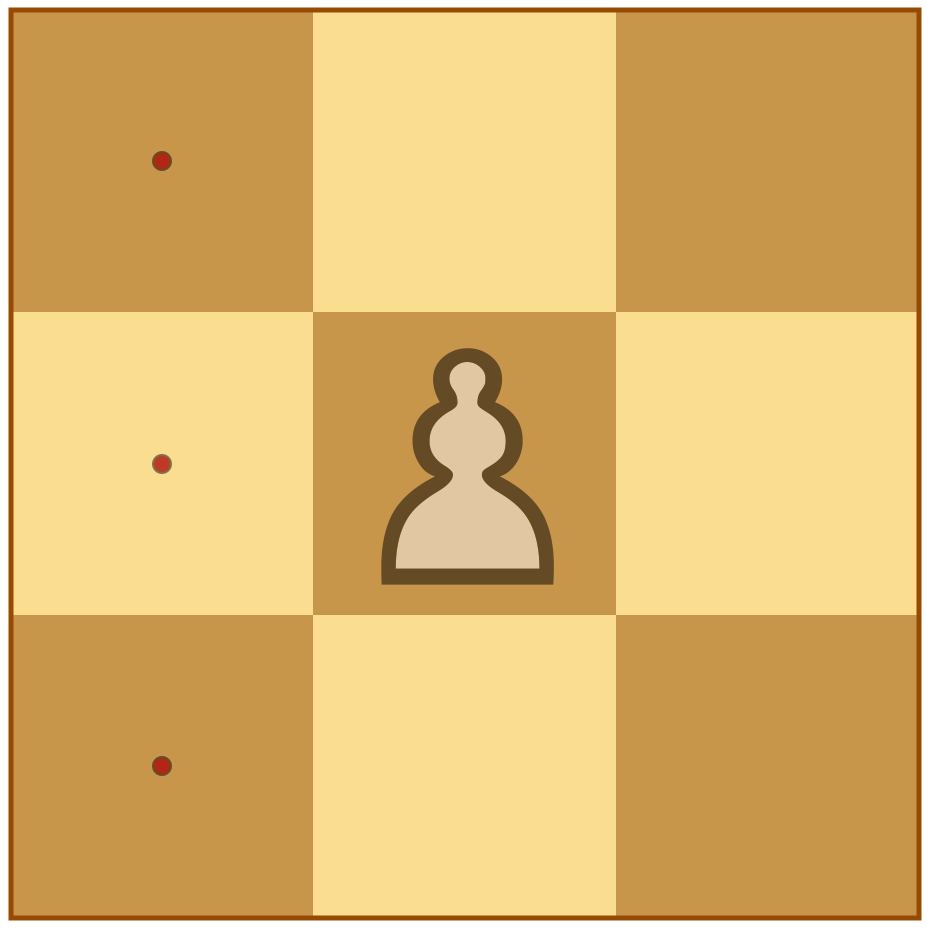}} &  \raisebox{-.5\height}{\includegraphics[width=0.20\linewidth]{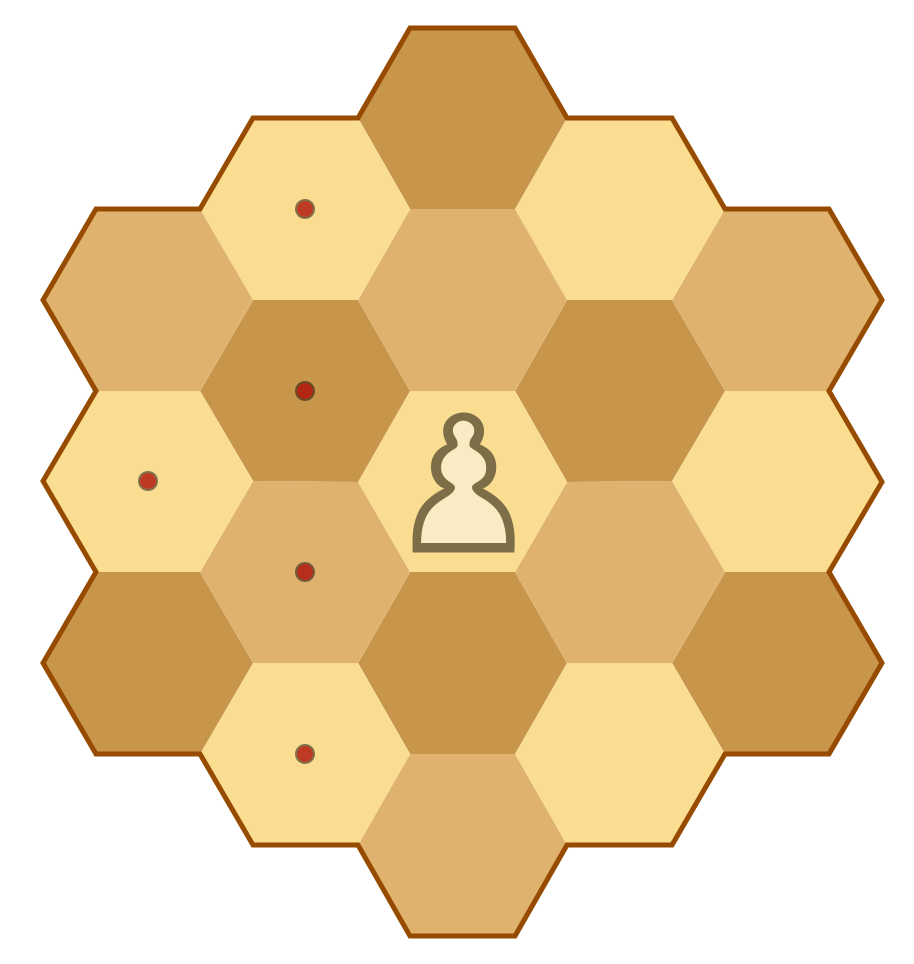}} &  \raisebox{-.5\height}{\includegraphics[width=0.23\linewidth]{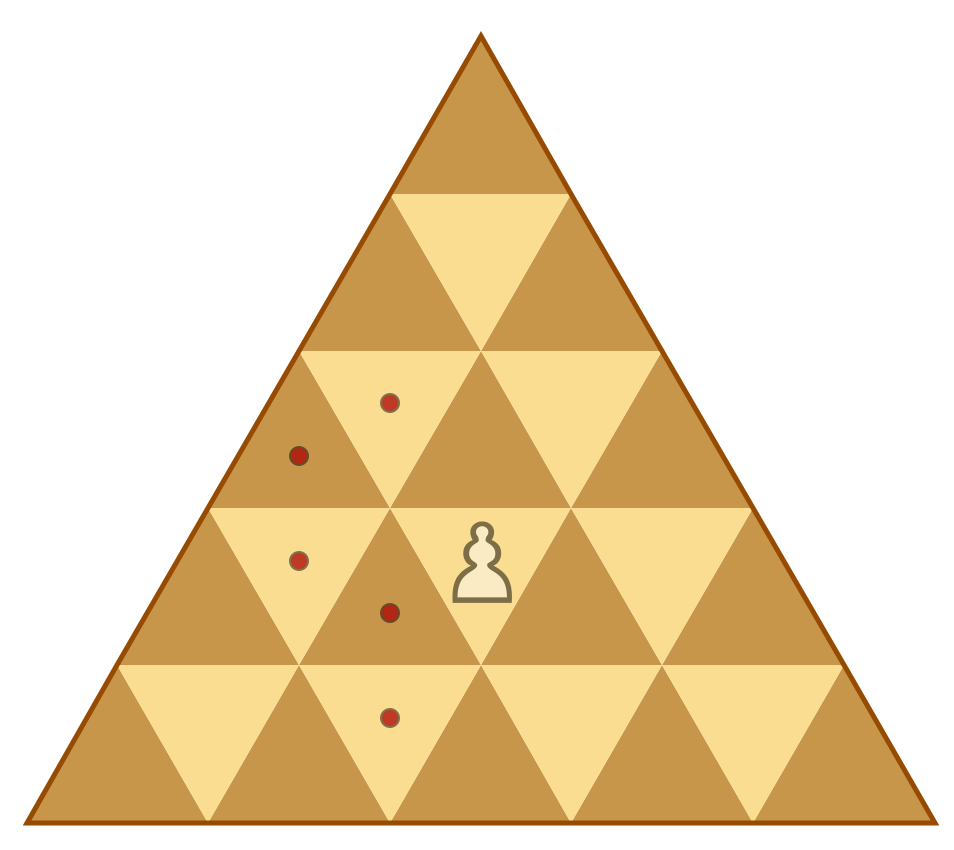}}\\ 
 Rightwards &  \raisebox{-.5\height}{\includegraphics[width=0.20\linewidth]{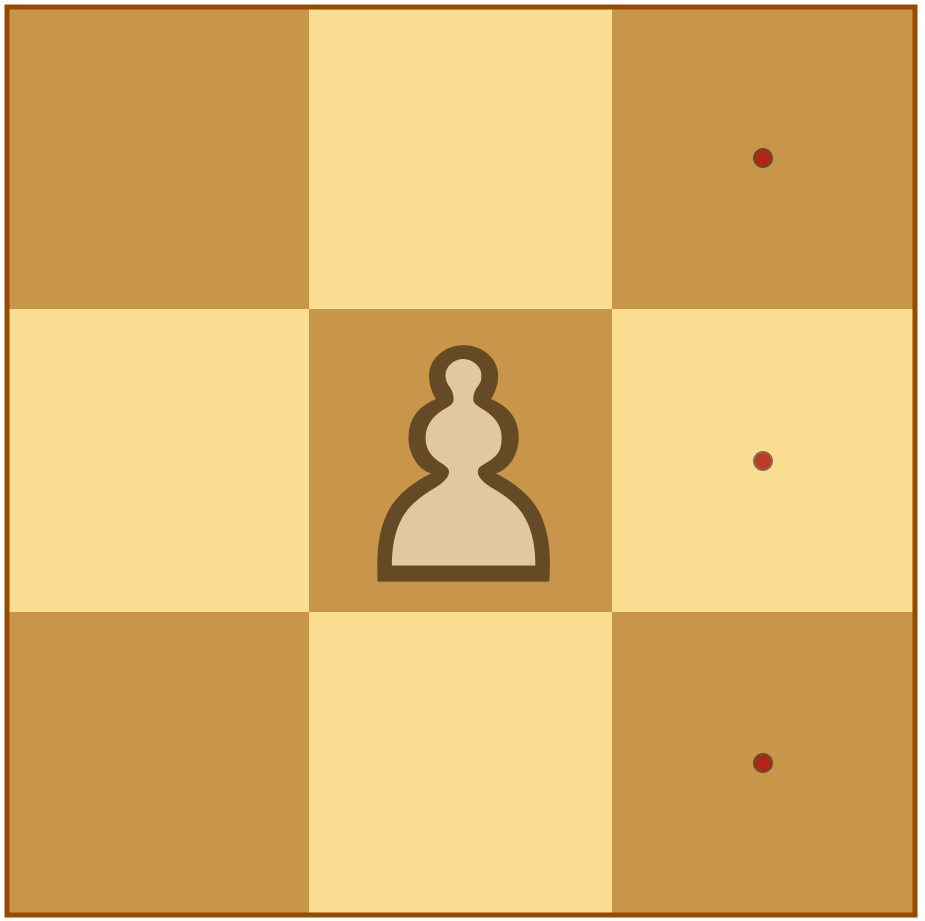}} &  \raisebox{-.5\height}{\includegraphics[width=0.20\linewidth]{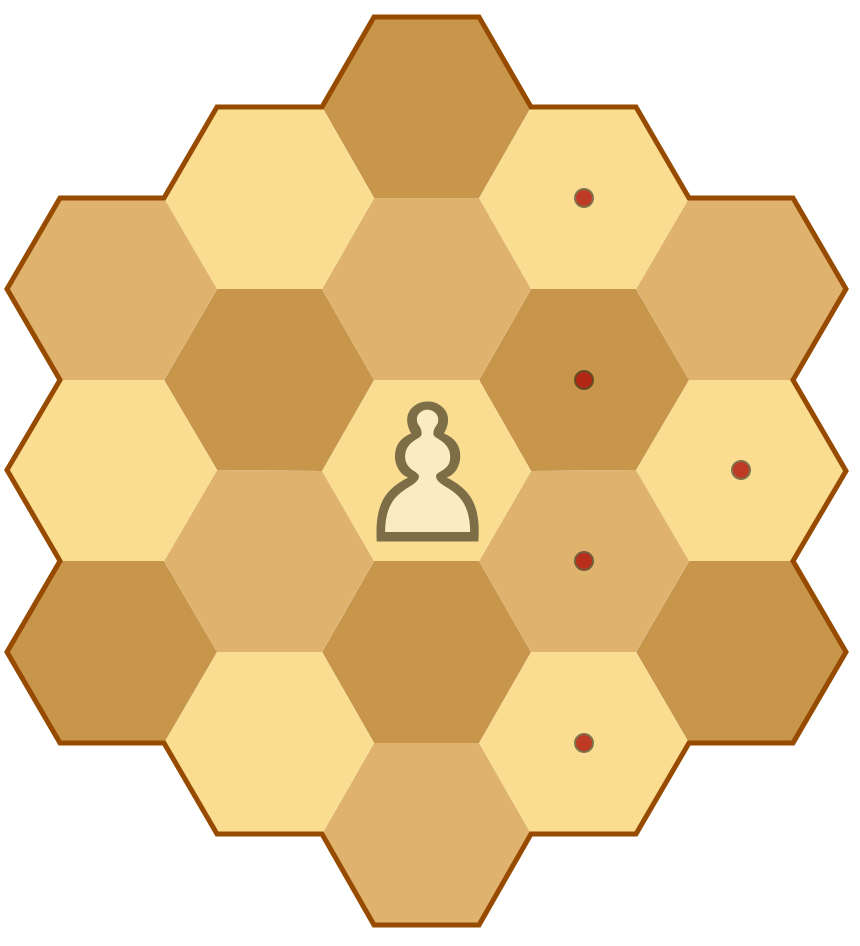}} &  \raisebox{-.5\height}{\includegraphics[width=0.23\linewidth]{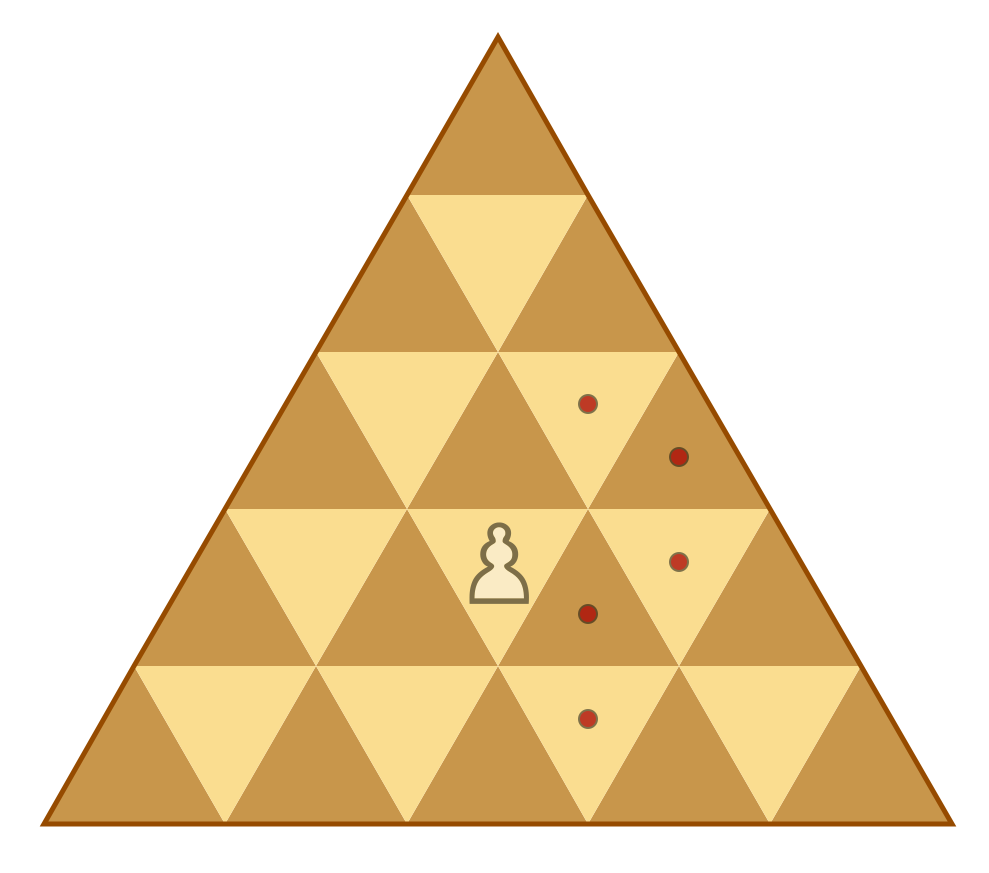}}\\ 
  FR &  \raisebox{-.5\height}{\includegraphics[width=0.20\linewidth]{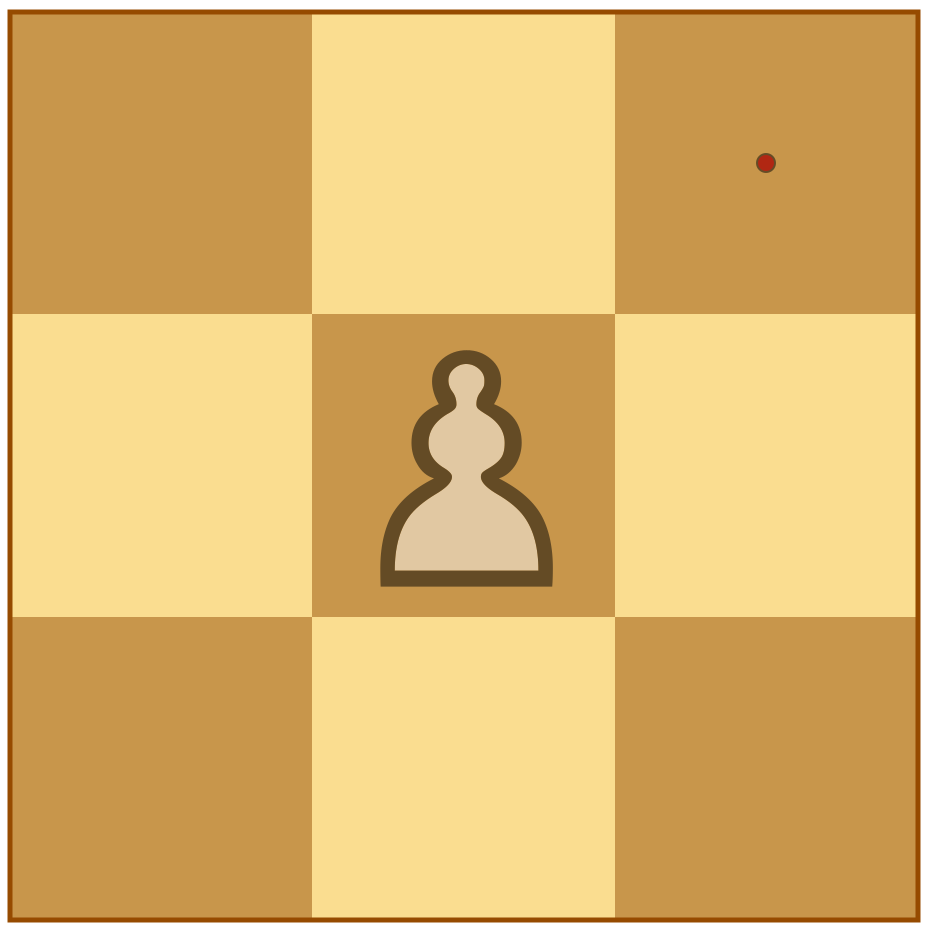}} &  \raisebox{-.5\height}{\includegraphics[width=0.20\linewidth]{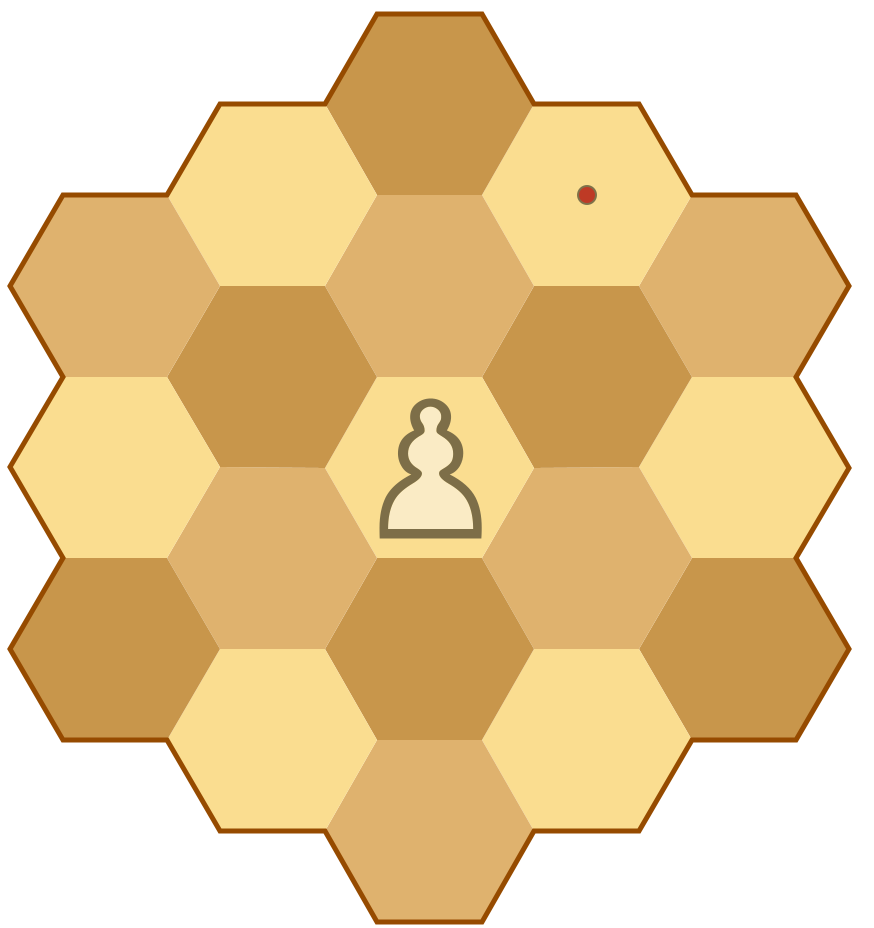}} &  \raisebox{-.5\height}{\includegraphics[width=0.23\linewidth]{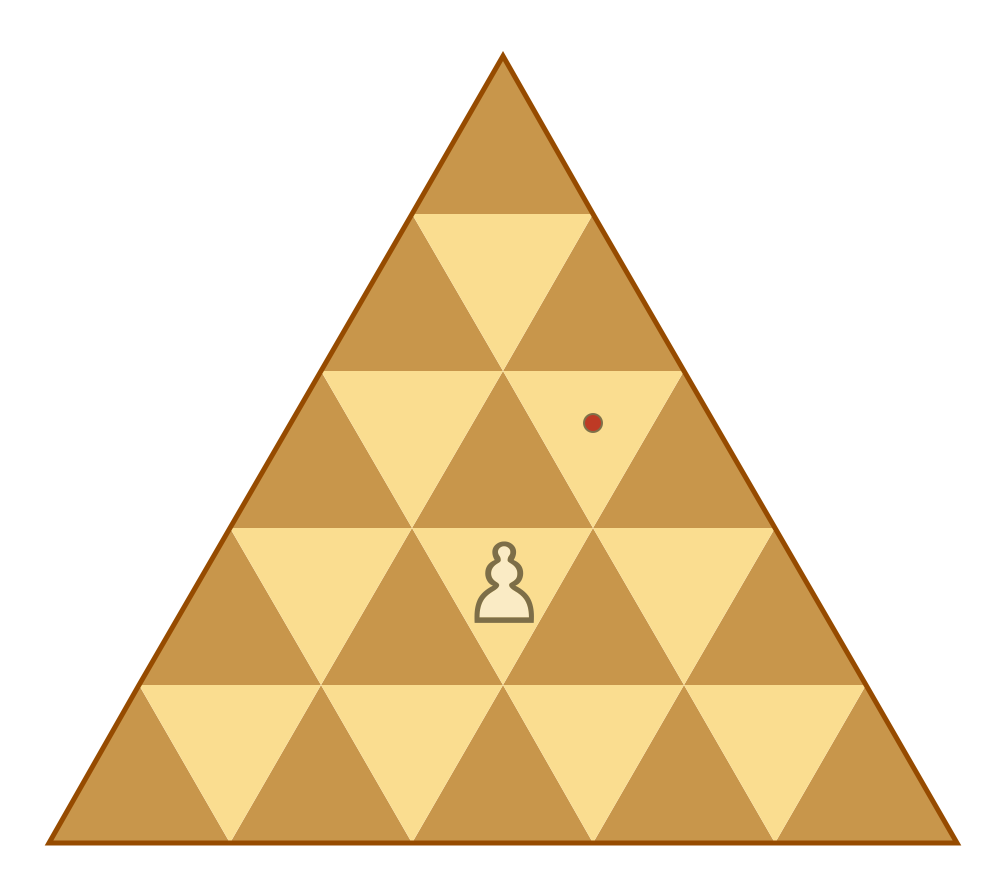}}\\ 
   FL &  \raisebox{-.5\height}{\includegraphics[width=0.20\linewidth]{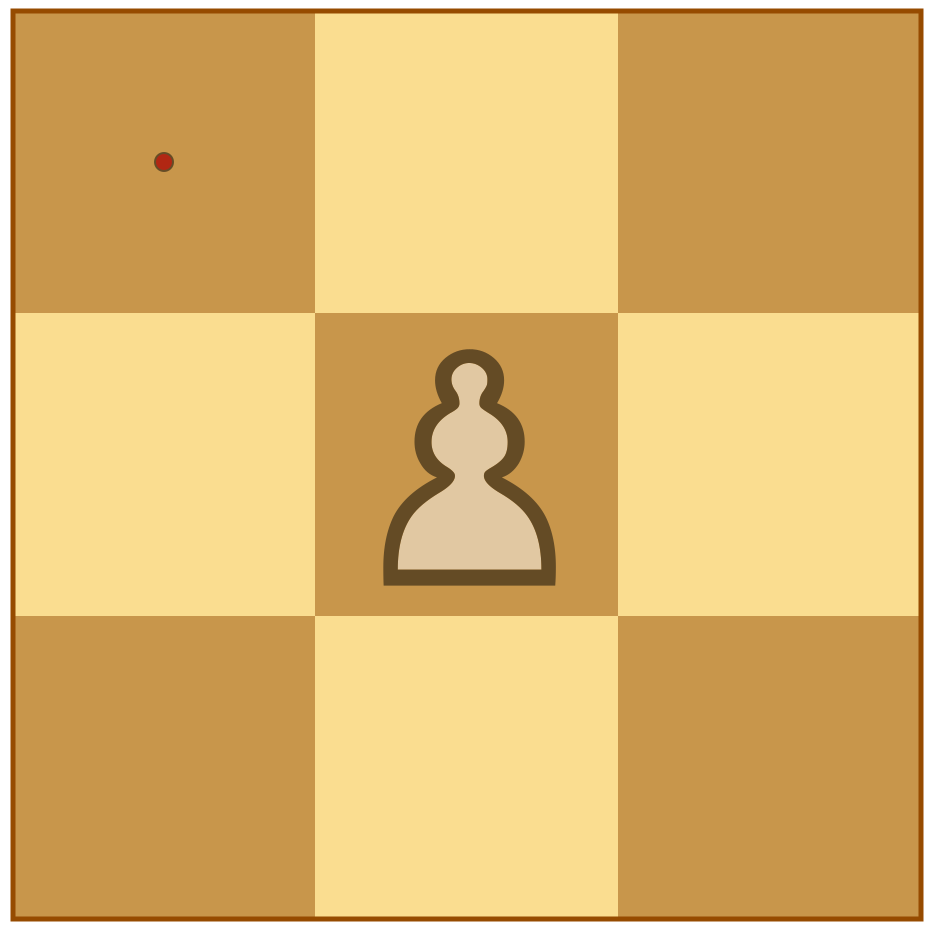}} &  \raisebox{-.5\height}{\includegraphics[width=0.20\linewidth]{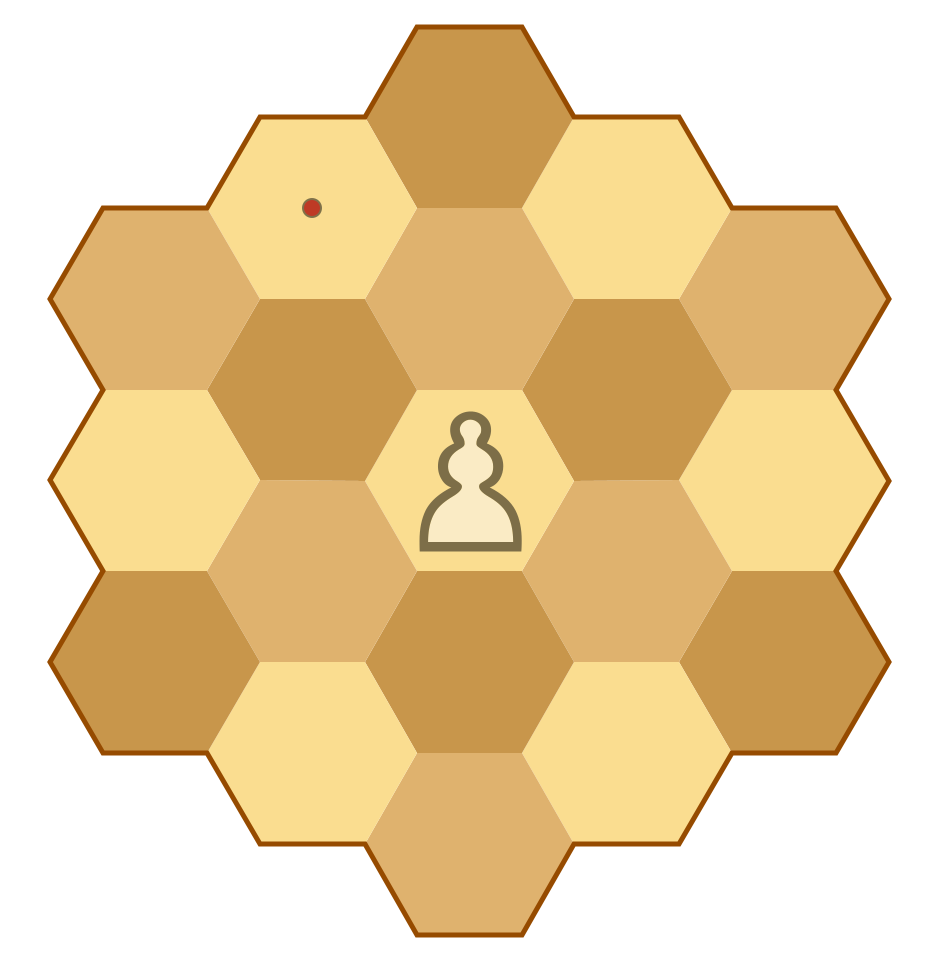}} &  \raisebox{-.5\height}{\includegraphics[width=0.23\linewidth]{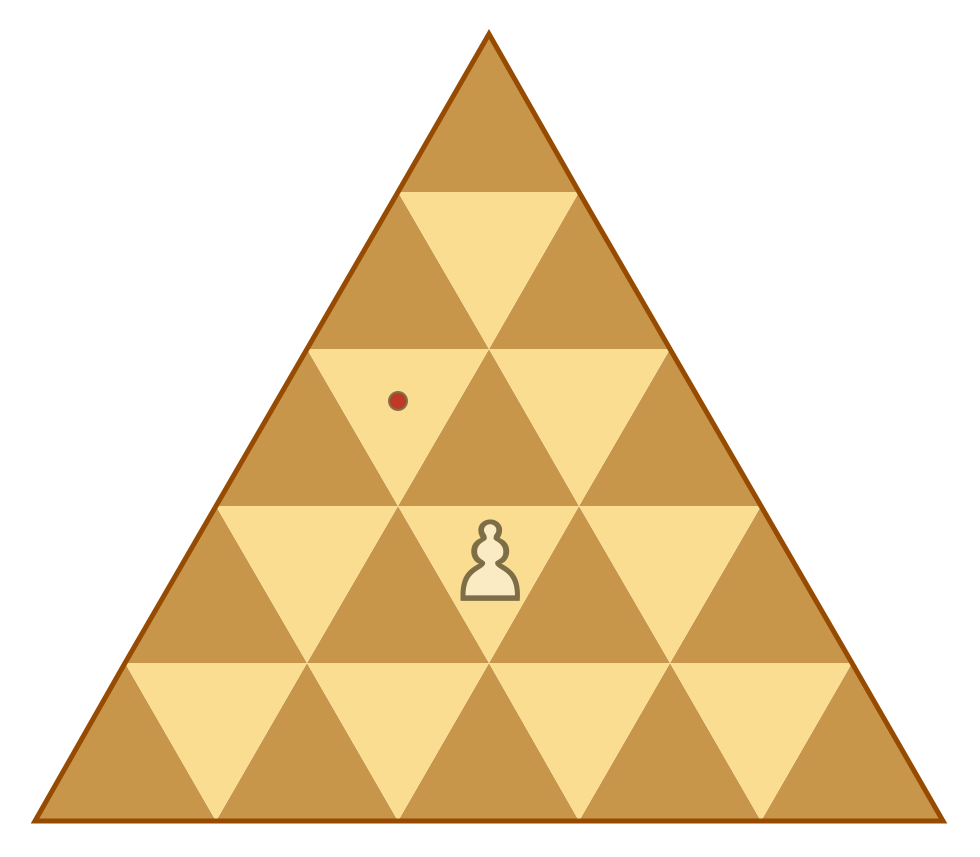}}\\ 
     BR &  \raisebox{-.5\height}{\includegraphics[width=0.20\linewidth]{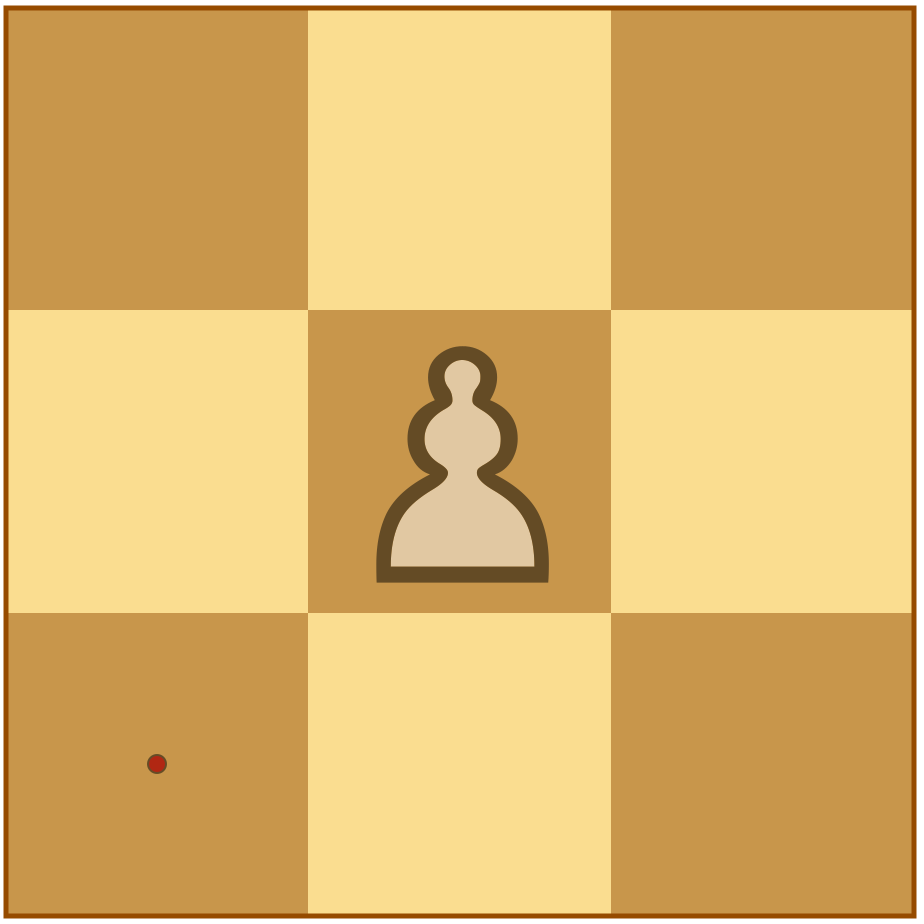}} &  \raisebox{-.5\height}{\includegraphics[width=0.20\linewidth]{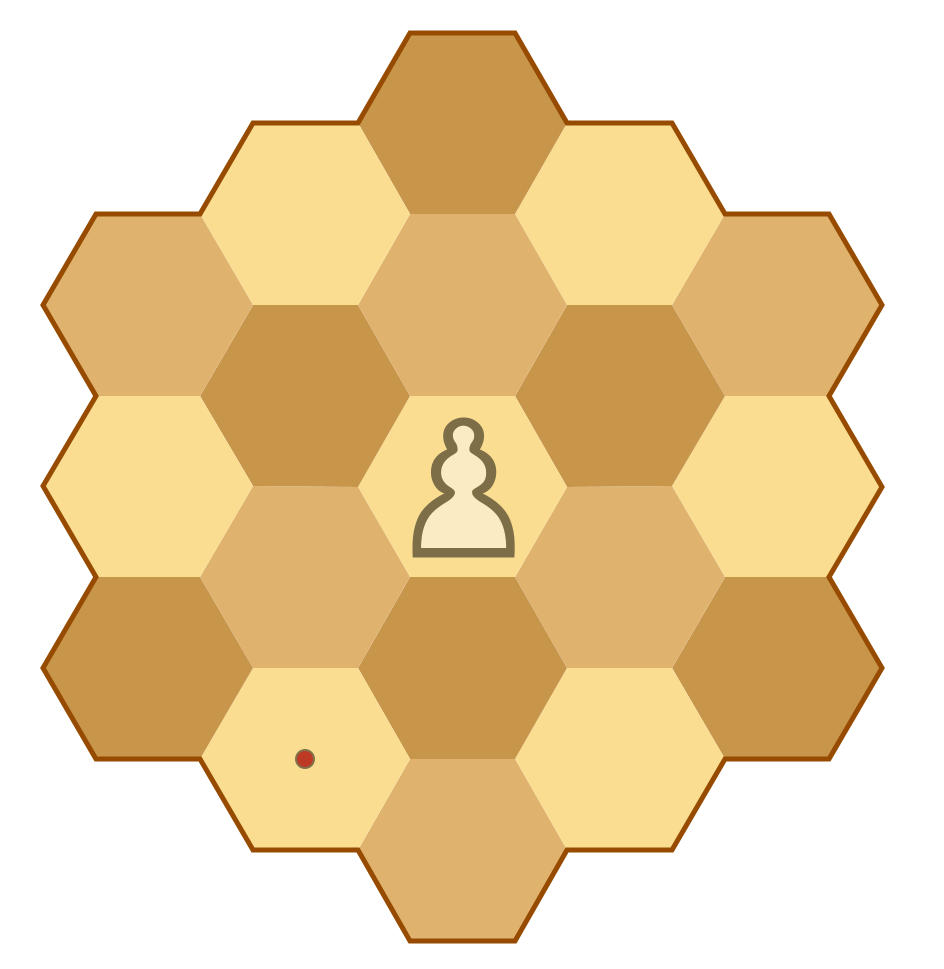}} &  \raisebox{-.5\height}{\includegraphics[width=0.23\linewidth]{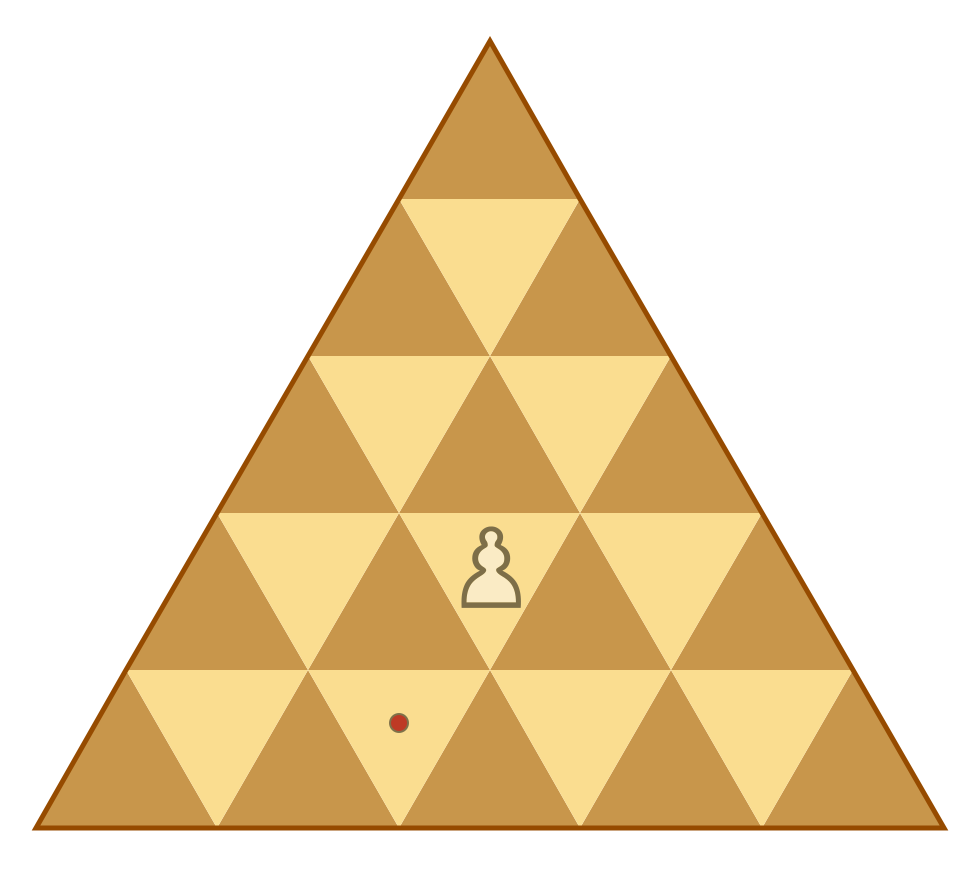}}\\ 
   BL &  \raisebox{-.5\height}{\includegraphics[width=0.20\linewidth]{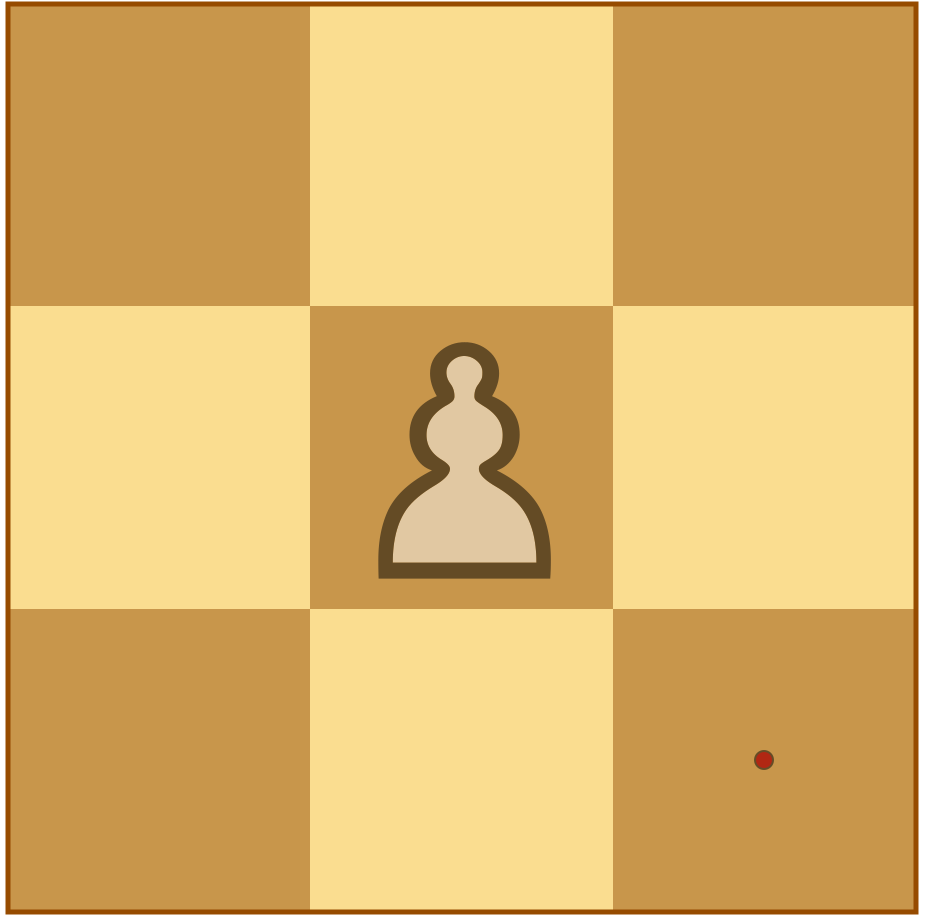}} &  \raisebox{-.5\height}{\includegraphics[width=0.20\linewidth]{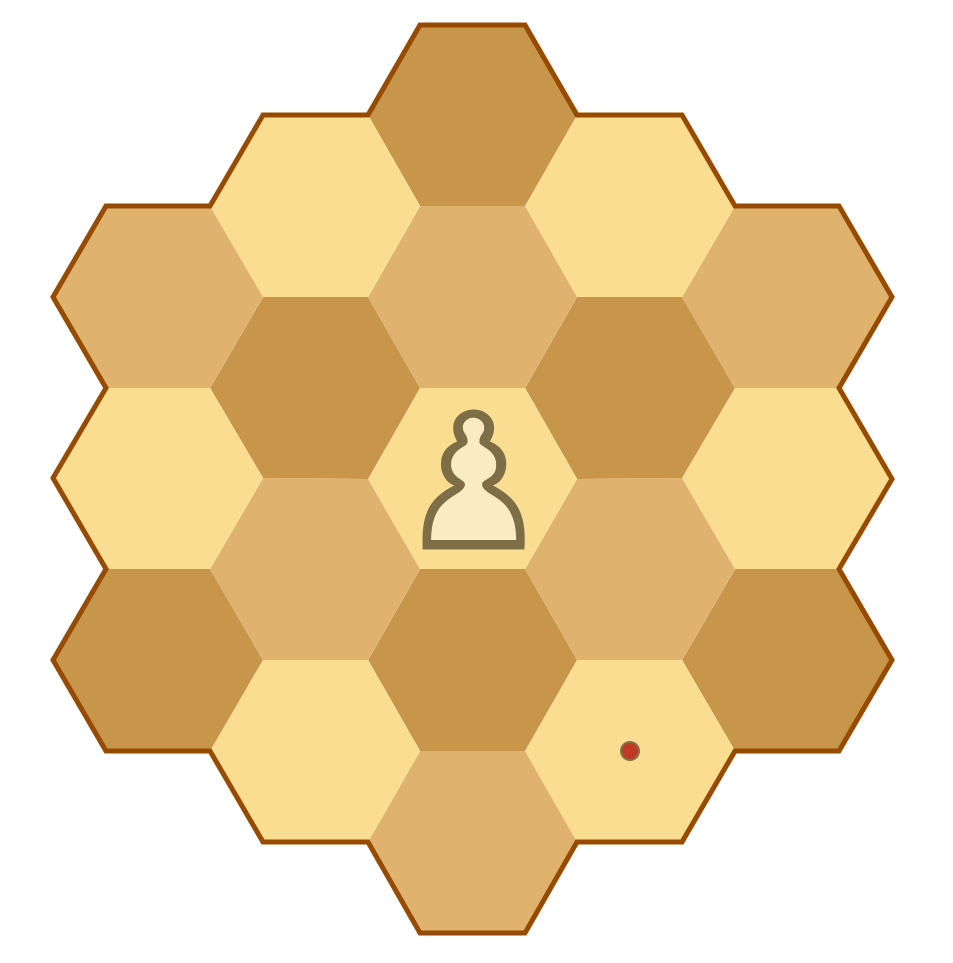}} &  \raisebox{-.5\height}{\includegraphics[width=0.23\linewidth]{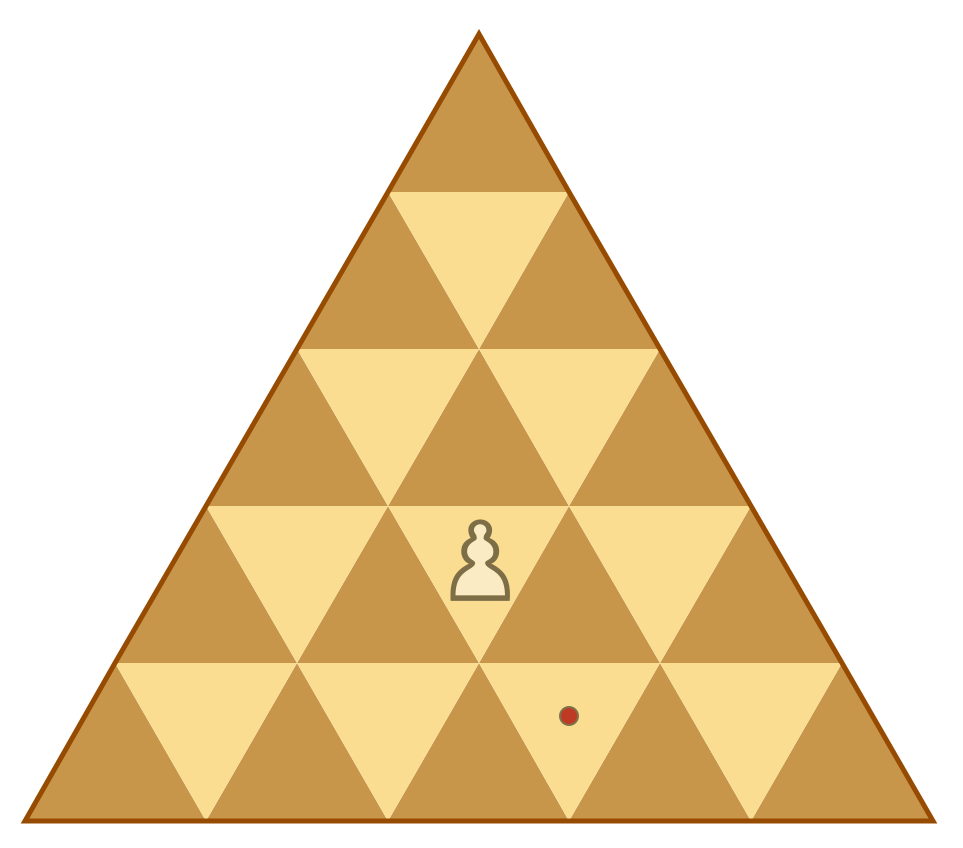}}\\ 
\rowcolor{gray!50}
 \multicolumn{4}{r}{} \\
\end{longtable}

\subsection{Board}

The main board of the game is defined by the ludeme (board ...). 
Firstly, the board evaluates the GraphFunction defining the graph of the board and converts it into the Topology object by associating each pre-generation related to each graph element. Then, according to the flags of the game, more pre-generations are realised, such as the computation of the minimal distance between two graph elements of the same type.

The parameter "use" of the board ludeme is used to define the default graph element used in all the ludemes involved in the game if not specified. If that parameter is not defined, the default site type is cells. Moreover, the dimensions of a board also depend on that parameter. For example, a Chess board has a dimension of 8 because it is played on the cells but a Go board has a dimension on 19 because it is played on the vertices (even if it has 18 cells by row and column). The Board class is also the only link between the logic part of Ludii and the Graphical User Interface in setting the default design of the board according to the default site if not modified by the metadata of the game. 

\subsubsection{Track}
For games involving tracks such as race games, the board builds them. Two ways exist to define a track. One consists of simply listing all the sites from the start to the end. The other one consists of given the starting site of the track then describing in a string the direction and distance of each step of the track. Here is an example of the description of the board of the game "20 Squares"\footnote{20 Squares: \href{https://boardgamegeek.com/boardgame/23211/20-squares}{BGG}} with tracks defined using the second method:

\begin{boxex}
\begin{ludii}
(board 
    (merge 
        (rectangle 3 4) 
        (shift 0 1 (rectangle 1 12))
    )
    {    
    (track "Track1" "3,W,N1,E,End" P1 directed:true)
    (track "Track2" "12,W,S1,E,End" P2 directed:true)
    }
)
\end{ludii} 
\end{boxex}

\begin{center}
\includegraphics[width=\linewidth]{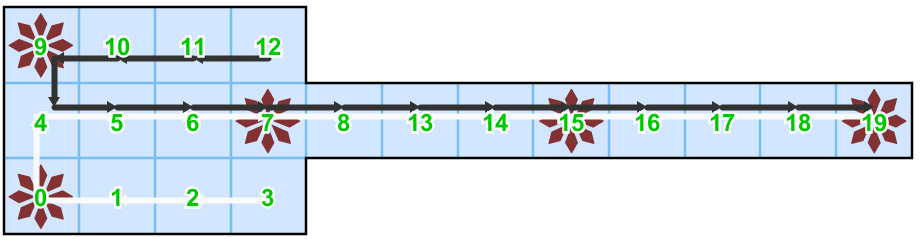}
\end{center}

"Track"1 is defined by "3,W,N1,E,End". The board builds that track by beginning it with Cell 3, then, because of the next step "W", all the sites following the radial starting from Cell 3 to the West (in this order: Cells 2, 1 and 0). Then from the last site reached, because of "N1" the next cell is the cell(s) in the radial starting from Cell 0 to the north at distance 1 max (Cell 4). The Same idea applies for the next step describing by "E" in following the radial starting from Cell 4 to the East (in this order: Cells 5, 6, 7, 8, 13, 14, 15, 16, 17, 18, 19). To finish, the last step is "End" meaning that the track ends at an imaginary site after the last site of the track. This "End" feature is useful for escape race games in which a piece can escape the board by moving one more site after the end of the track. Following that method, the track ordering by the cells, is $\{3$ $2$ $1$ $0$ $4$ $5$ $6$ $7$ $8$ $13$ $14$ $15$ $16$ $17$ $18$ $19$ $End\}$.

A track can be owned by a player in order to have access to all the tracks used only by a specific player (e.g. "Track1" is owned by P1 (white) and "Track2" is owned by P2 (black) in 20 Squares). Two other properties can be defined for a track:
\begin{itemize}
    \item \textbf{directed}: The track is ordered and it is not possible to move backward in it.
    \item \textbf{loop}:  The last site of the track is following by the starting site.
\end{itemize}

The compiler takes into account two special keywords to make the description more concise, which are:
\begin{itemize}
    \item End: An imaginary site one step after the end of the track (its hidden value is -2).
    \item OFF: A value returned by any IntFunction an expected site does not exist (its hidden value is -1). The value UNDEFINED can also be used for that.
\end{itemize}
Moreover, it is important to note that it is also possible to describe a track beginning in a container other than the board, such as the hand of a player, in listing the expecting sites.

\subsubsection{Special cases}

It is possible to define the board ludeme with ranges of values for each graph element. These data are dedicated to deduction puzzles for defining the range of possible values for each variable and will adapt the state of the game as explained in Section \ref{section:DeductionPuzzle}. 

The ludeme Boardless is a special type of "board" extending the Board class and defined only by some regular tiling: Square, Hexagonal or Triangle. In its current implementation a fake board of size 41 following the specified tiling will be generated. However, we plan to improve this in the future by building the board only when the sites are necessary in the following states of the game.

In order to improve the description of specific boards, Ludii proposes two specific ludemes: 
\begin{itemize}
    \item \textbf{MancalaBoard}: To define any mancala board by giving only the number of rows and columns and the type of stores used: None, Outer, Inner.
    \item \textbf{SurakartaBoard}: To define a board similar to the board of Surakarta for any GraphFunction in specifing the number of loops and the starting row to start the loops from.
\end{itemize}

\begin{minipage}{0.5\textwidth}
\includegraphics[width=\linewidth]{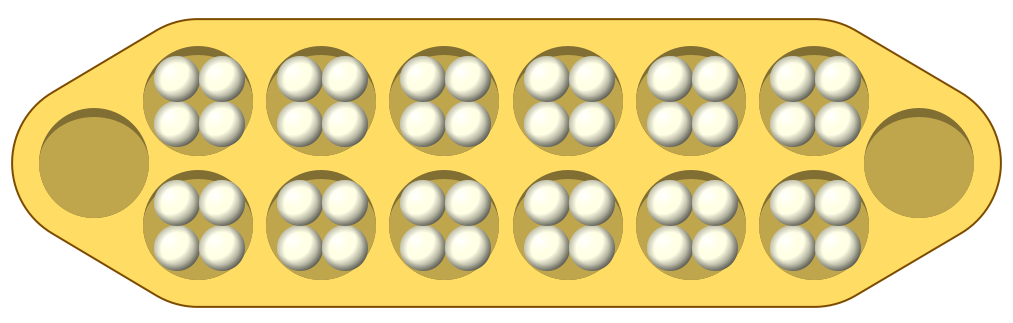}
\end{minipage}
\begin{minipage}{0.5\textwidth}
\includegraphics[width=\linewidth]{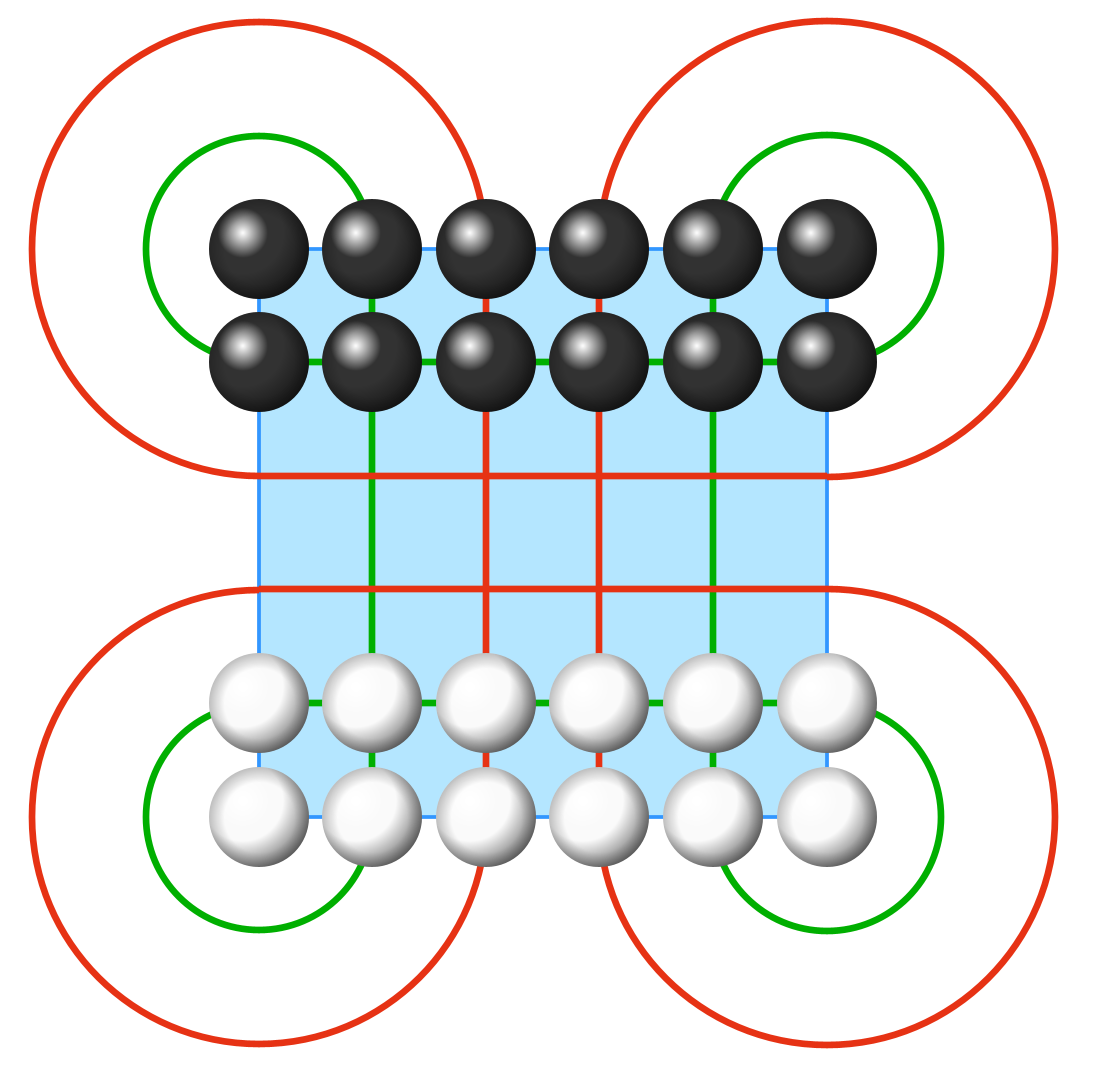}
\end{minipage}

\subsection{Hand, Dice and Deck}\label{Section:HandDiceDeck}

Ludii uses 3 other kinds of container ludemes.\\

The hand of a player defined by the (hand ...) ludeme corresponds simply to a row of square cells defined by the size of each hand. Each hand is owned by a specific player or it is shared between all the players. That's possible to define a hand of the same size for each player in using the ludeme (hand Each size:3), for example.\\

The dice ludeme is defined with the number of dice, by the number of faces of each die (the default is 6), and the value of each face of each die (if not defined, from 1 to the number of faces). 
The dice container is a hand including a site for each die. If not owned, the hand is shared among all players. However, when this ludeme is used, a different component called "DieX" (with X the number of dice created so far) is created for each die. \\

The Deck ludeme is a hand of a single site with a stack of cards modelling a deck of cards. If not owned, that hand is shared between each player. The deck can be customised by just defining the number of cards per suit (default 13) and the number of suits in the deck (default 4) or in defining the deck card by card thanks to the ludeme Card. The ludeme Card defined a card by its rank and its value and potentially its trump rank and trump value. However the cards are still in implementation and that container is susceptible to be modified.


\section{Components}

The components model the different pieces, tiles, dice, dominoes or cards of a game. Each of them are described by different ludemes.

\begin{center}
\begin{tikzpicture}

\begin{umlpackage}{equipment}
\begin{umlpackage}{component}
\umlinterface[rectangle split parts=1]{Component}{}{}
\umlclass[rectangle split parts=1, x = -2, y = 2]{Piece}{}{}
\umlclass[rectangle split parts=2, x = 2, y = 2]{Tile}{Path[] paths}{}
\umlclass[rectangle split parts=1, x = 2, y = -2]{Card}{}{}
\umlclass[rectangle split parts=1, x = -3]{Die}{}{}
\umlclass[rectangle split parts=1, x = -2, y = -2]{Domino}{}{}

\end{umlpackage}
\end{umlpackage}

\umlinherit{Piece}{Component}
\umlinherit{Card}{Component}
\umlinherit[geometry=|-]{Die}{Component}
\umlinherit{Domino}{Component}
\umlinherit{Tile}{Component}

\end{tikzpicture}
\end{center}

The most used component is described by the Piece class.

\begin{minipage}{0.5\textwidth}
\begin{boxex}
\begin{ludii}

(piece "Ball" Each)
\end{ludii} 
\end{boxex}
\end{minipage}
\begin{minipage}{0.5\textwidth}
\begin{center}
\includegraphics[width=0.5\linewidth]{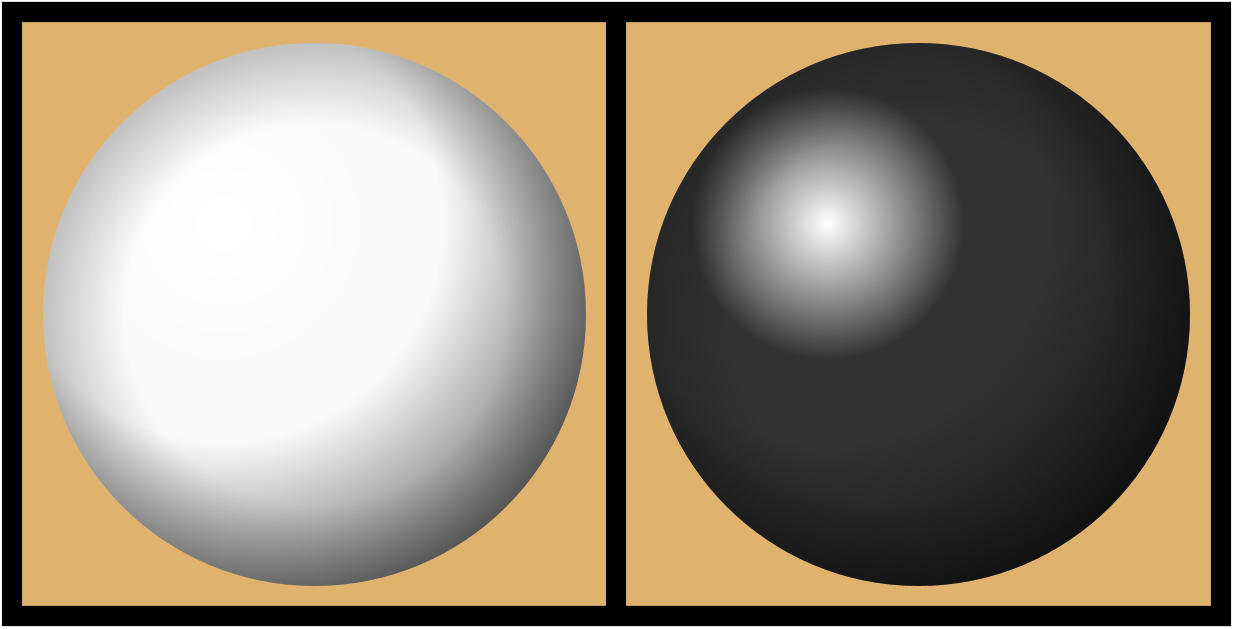}
\end{center}
\end{minipage}

This ludeme generates a component for each player called "Ball". However the name of the component is always its name and the index of the player. For example, for a 2-player game, this ludeme generates a component named "Ball1" for player 1 and a component named "Ball2" for player 2. The parameter "Each" models the owner of the components, if this parameter would be "Neutral" meaning that component is owned by no player, the name of the component would be "Ball0" and if this parameter would be "Shared", meaning this component is owned by all the players, its name would be "Ball" (with no number).

A piece component has many other parameters to specify its properties. As described in Section \ref{Section:DirectionSystem}, a facing direction can be set for the piece in order to use the relative directions for the movement of the pieces. Each piece can also be associated with a Moves ludeme called "generator", which will be used to generate some legal moves starting from each site where such a piece is currently placed. More information about how to model moves will be given in Section \ref{Section:Play}. 

If a piece can be flipped, the flip values of each face of the piece have to be given (e.g. for Reversi\footnote{Reversi: \href{https://en.wikipedia.org/wiki/Reversi}{Wikipedia}}). Each flip value corresponds to the required state value of the site where is placed this piece to be flipped to the other. 

\begin{minipage}{0.5\textwidth}
\begin{boxex}
\begin{ludii}
(piece "Disc" Neutral 
    (flips 1 2)
)
\end{ludii} 
\end{boxex}
\end{minipage}
\begin{minipage}{0.5\textwidth}
\begin{center}
\includegraphics[width=0.5\linewidth]{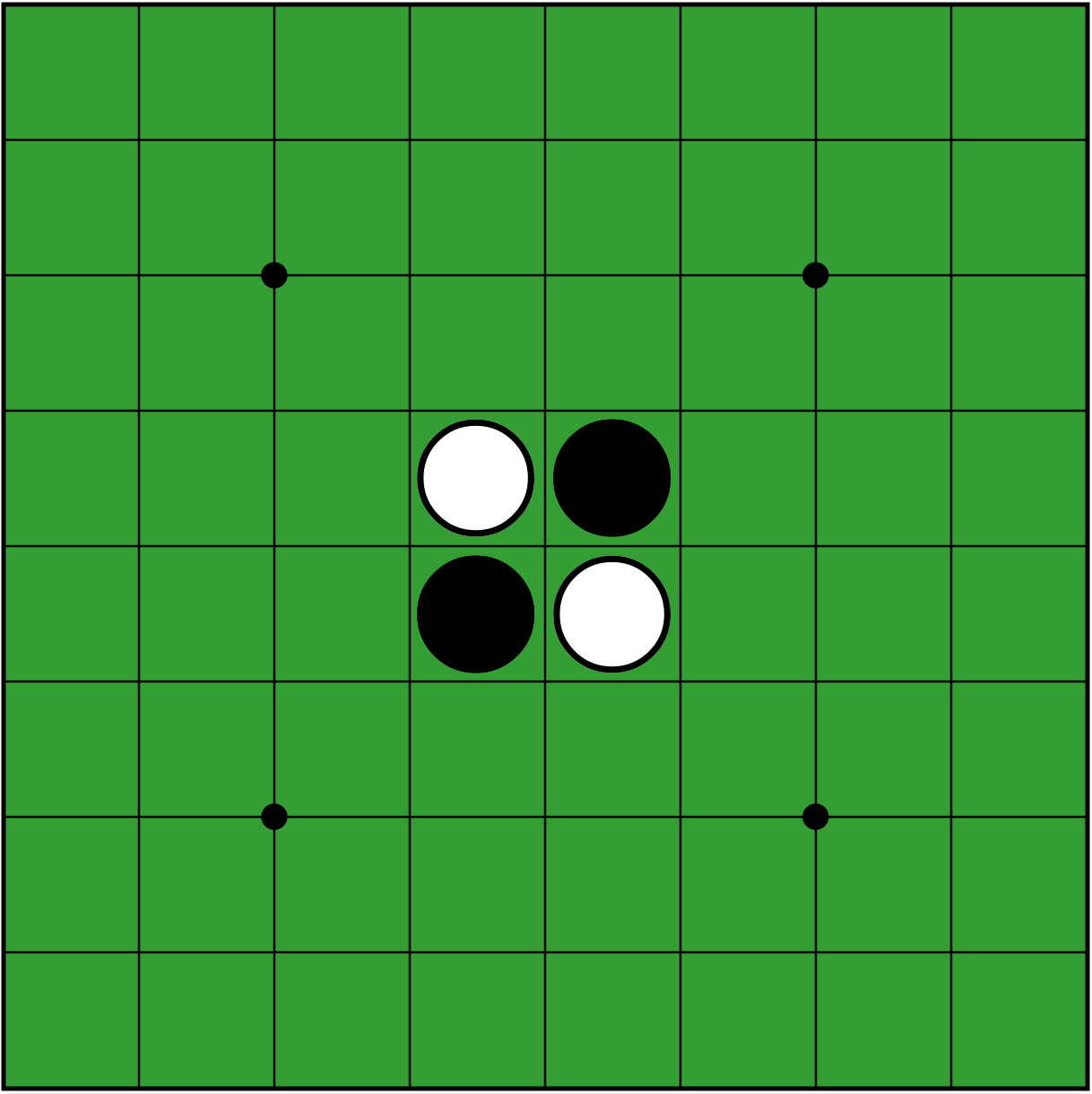}
\end{center}
\end{minipage}

The Tile ludeme is used to specify a component placed only on the cells and shaped like the cells of the board. Some extra parameters can be specified on such components. Firstly, the number of sides of the tile, if not specified, the number of sides is the number of edges of the Cell 0 of the board. Some tiles can have an internal path inside the tile to link some sides to others. Here is an example of the tiles of the game Trax\footnote{Trax: \href{https://boardgamegeek.com/boardgame/748/trax}{Board Game Geek}}:

\begin{minipage}{0.5\textwidth}
\begin{boxex}
\begin{ludii}
(tile "TileX" numSides:4
    { 
    (path from:0 to:2 colour:1)
    (path from:1 to:3 colour:2)
    }
) 
(tile "TileCurved" numSides:4
    { 
    (path from:0 to:1 colour:1)
    (path from:2 to:3 colour:2)
    }
)
\end{ludii} 
\end{boxex}
\end{minipage}
\begin{minipage}{0.5\textwidth}
\begin{center}
\includegraphics[width=0.8\linewidth]{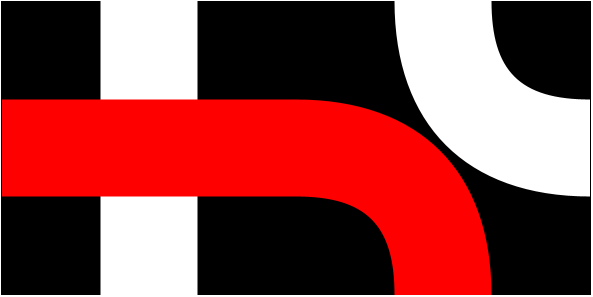}
\end{center}
\end{minipage}

The left piece has a path of a colour 1 between the top and the bottom side and a path of colour 2 between the left side and the right side. The value used in the parameter from: and to: are the number of right orthogonal rotations from the top side to get the expected side. For such a square tile, 0 means the top side, 1 the right side, 2 the bottom side and 3 the left side.

Some tiles can be large enough to be placed on many sites simultaneously. Here is an example of L Game\footnote{L Game: \href{https://en.wikipedia.org/wiki/L_game}{Wikipedia}}, a game with large tiles shaped like the letter L.

\begin{minipage}{0.5\textwidth}
\begin{boxex}
\begin{ludii}
(tile "L" Each 
    { 
    {L F R F F} 
    {R F L F F} 
    }
)
\end{ludii} 
\end{boxex}
\end{minipage}
\begin{minipage}{0.5\textwidth}
\begin{center}
\includegraphics[width=0.45\linewidth]{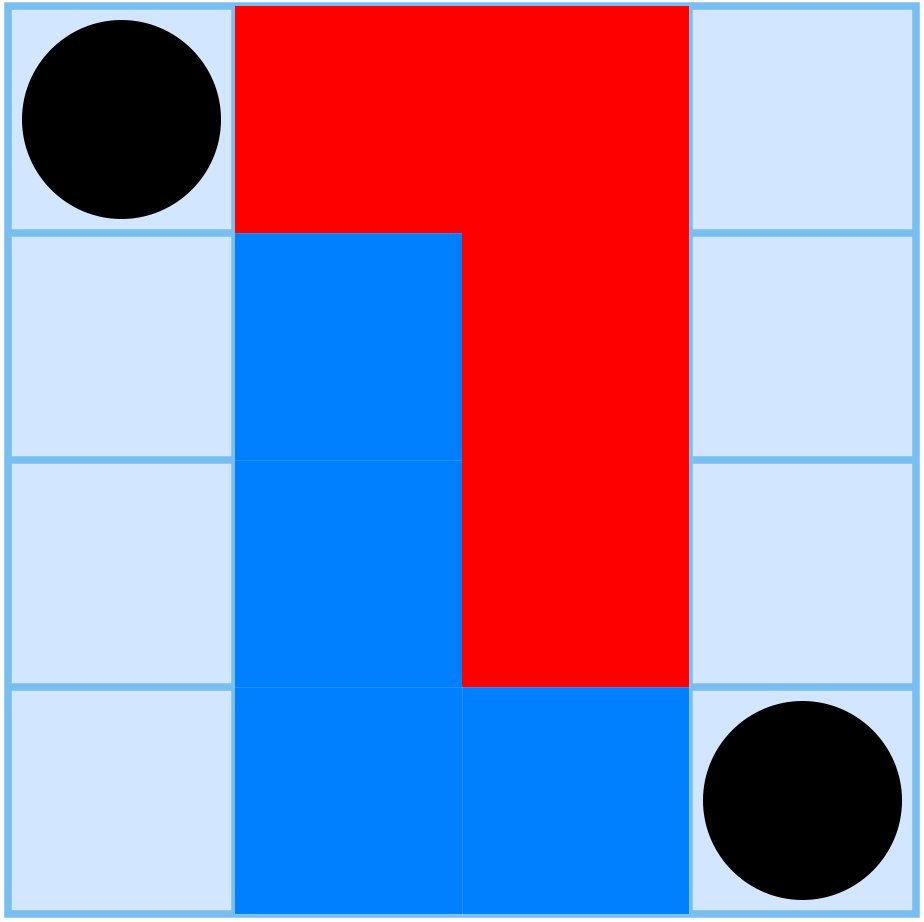}
\end{center}
\end{minipage}

A large tile is defined according to a walk of orthogonal steps. That mechanism is based on Turtle graphics (\href{https://en.wikipedia.org/wiki/Turtle_graphics}{Wikipedia}). In Ludii, a walk starts facing the first orthogonal direction of the board (for a square board, the north), and only 3 parameters describe the walk:
\begin{itemize}
    \item \textbf{F}: Make a forward step in the current facing direction.
    \item \textbf{L}: Modify the facing direction to the left orthogonal direction of the current facing direction.
    \item \textbf{R}: Modify the facing direction to the right orthogonal direction of the current facing direction.
\end{itemize}

Please note that a walk cannot leave the playing area and return.

For the case of a large tile, Ludii generates all possible sets of sites from a root site for each orthogonal direction and for each list of steps defined in the large tile. For the case of this large tile shaped like the letter L used in a square board, for each site, 8 sets of sites are generated (2 sets of steps and 4 possible orthogonal directions from each site). The state value of the root site is used to select the current set of sites used by the large tile, in the case of the L-shaped large tile, the state value is between 0 and 7. This value corresponds to the number of right rotations from the first orthogonal direction of the cells modulo the number of sets of steps used to define the shape of the large tile.  For example, in the picture, the blue large tile rooted from the bottom right site has a state value of 0 and the large tile rooted from the top left site has a state value of 2.

A domino piece is also defined as a large rectangular piece. However, in Ludii a domino uses 8 square sites to be placed (4 sites per side). The reason is because in many dominoes games, it is possible to place a domino in the middle of the long side such as in the next picture. The value of each side of the domino is defined in the Domino ludeme, the first value is allocated to the first 4 sites used by the domino and the second value to the 4 last sites.

\begin{minipage}{0.5\textwidth}
\begin{boxex}
\begin{ludii}
(domino "Domino45" Shared 
    value:4 value2:5
)
\end{ludii} 
\end{boxex}
\end{minipage}
\begin{minipage}{0.5\textwidth}
\begin{center}
\includegraphics[width=0.5\linewidth]{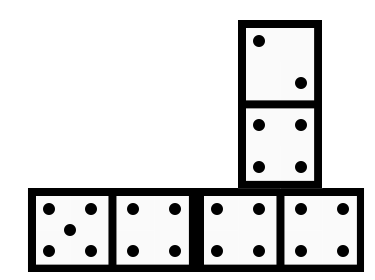}
\end{center}
\end{minipage}

As described in Section \ref{Section:HandDiceDeck}, Ludii can generate a set of dice. Each of the dice is a Die component. A die ludeme has a name, an owner and possibily a facing direction and a value such as the piece and tile ludemes. However this component has also a number of faces and different values associated to each face. These data are obtained by using the container Dice as described in the previous section. During the game, at a specific state, when Ludii needs to obtain the value of a die, it uses the state value of the site where the die is placed. For example, a common die with 6 faces starting from the value 1 to the value 6 has: the state 0 corresponding to the value 1, the state 1 to the value 2, \ldots, the state 5 to the value 6.
        
\begin{minipage}{0.5\textwidth}
\begin{boxex}
\begin{ludii}
(dice num:2)
(dice d:2 from:0 num:4)
\end{ludii} 
\end{boxex}
\end{minipage}
\begin{minipage}{0.5\textwidth}
\begin{center}
\includegraphics[width=0.4\linewidth]{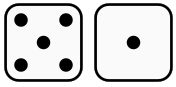}
\includegraphics[width=0.6\linewidth]{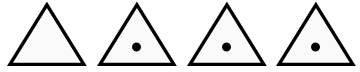}
\end{center}
\end{minipage}

However, to avoid looking at the state value of the sites occupied by the dice at each turn to obtain their values, many shortcuts are implemented in the State class of Ludii (see Section \ref{Section:OtherStateData}). They provide an efficient way to obtain this information when necessary by using different functions ludemes, which will be described in Chapter \ref{Chapter:Functions}. 

Finally, the last type of component is the Card component. As described in the Section \ref{Section:HandDiceDeck}, Ludii generates a set of cards using to the ludeme Deck. Each card is defined by its rank and its value and potentially its trump rank and trump value. However, the cards are still in implementation and this component is susceptible to be modified.

\begin{center}
\includegraphics[width=0.8\linewidth]{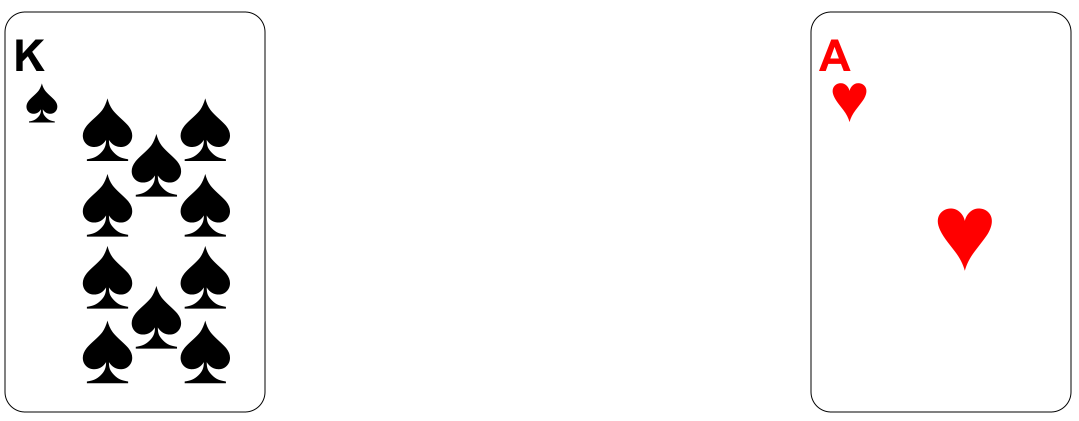}
\end{center}


\section{Other Items}

As described in the beginning of this chapter, Ludii contains different other equipment ludemes. This short section describes each of them briefly.\\

The Dominoes ludeme defines a set of dominoes starting from the Domino with the value 0 for each side to the domino with a specific value for each side. The default maximum value is 6 and, for computational reasons, Ludii limits the value of a side of a domino to 16.\\

The Hints ludeme defines a set of hints used to solve deduction puzzles. This ludeme is described by an array of Hint ludemes, each of which corresponds to a set of sites and a hint value. These data can be used differently according to the constraints defined in the deduction puzzle. As an example, the puzzle Kakuro\footnote{Kakuro: \href{https://en.wikipedia.org/wiki/Kakuro}{Wikipedia}} at its initial state where only the hints are visible, the corresponding expected sum of the values appear in the left or the bottom of the site where they are placed. Here is the lud description of the rules of the game:

\begin{minipage}{0.5\textwidth}
\begin{boxex}
\begin{ludii}
(rules 
    (play
        (satisfy {
            (is Sum (hint))
            (all Different)
        })
    )
    (end 
         (if (is Solved) 
             (result P1 Win)
         )
    )
)
\end{ludii} 
\end{boxex}
\end{minipage}
\begin{minipage}{0.5\textwidth}
\begin{center}
\includegraphics[width=0.8\linewidth]{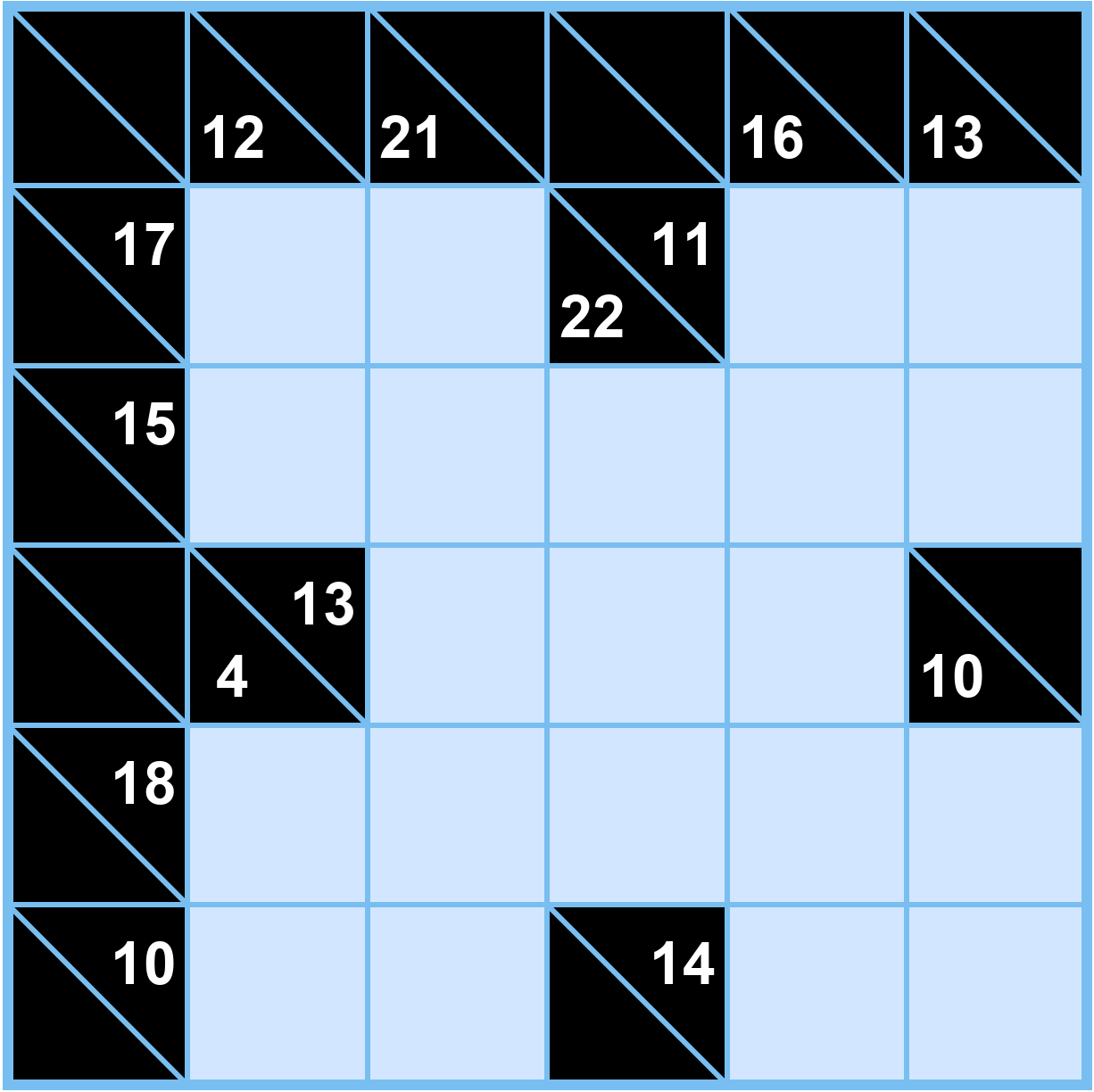}
\end{center}
\end{minipage}

The map equipment is used to create a correspondence between a key and a value. A map is defined by a name and a list of Pair ludemes. The name is optional, if not described, the default name is "MapX" with X the index of the map in the equipment. A pair ludeme is defined by a key and a value. Whatever the entries, the map ludeme converts them to integers. The entries can be coordinates expressed by a string, RoleType of players or directly using IntFunctions (see Section \ref{Section:IntFunction}) or integers. Here is an example of a map used frequently in race games to obtain the site where the pieces can enter into the board for each player.

\begin{boxex}
\begin{ludii}
(map "EntrySite" 
    {
    (pair P1 "A3")
    (pair P2 "A1")
    }
)
\end{ludii} 
\end{boxex}

The last type of equipment is the Regions ludeme. This ludeme defines specific sets of sites used by the game. A region can be defined by a name and owned by a player. The default name is "RegionX" with X being the index of the owner of the region, and if no owner exists, the first available index. The set of sites belonging to the regions can be defined by a list of RegionFunctions (see Section \ref{Section:RegionFunction}) or directly by lists of integers. Here is the description of the home region of each player in English Draughts\footnote{English Draughts: \href{https://en.wikipedia.org/wiki/English_draughts}{Wikipedia}}.

\begin{boxex}
\begin{ludii}
(regions "Home" P1 (sites Bottom))
(regions "Home" P2 (sites Top))
\end{ludii} 
\end{boxex}


\chapter{\textcolor{gray!95}{Rules}}\label{Chapter:Rules}

In Ludii, the rules of a game are divided in 4 types:
\begin{itemize}
    \item \textbf{Meta}: The meta rules of the game applied at each turn whatever the state.
    \item \textbf{Start}: The starting rules defining the initial state. 
    \item \textbf{Play}: The playing rules defining the legal moves of the mover in the current state.
    \item \textbf{End}:  The ending rules defining the conditions to satisfy to end the game and the consequent outcomes.
\end{itemize}

\begin{center}
\begin{tikzpicture}

\begin{umlpackage}{game}
\begin{umlpackage}{rules}
\umlclass[rectangle split parts=1]{Rules}{}{}
\umlclass[rectangle split parts=1, x= -3, y=-1]{Meta}{}{}
\umlclass[rectangle split parts=1, x= -1, y=-2]{Start}{}{}
\umlclass[rectangle split parts=1, x= 1, y=-2]{Play}{}{}
\umlclass[rectangle split parts=1, x= 3, y=-1]{End}{}{}
\end{umlpackage}
\end{umlpackage}

\umlunicompo{Rules}{Meta}
\umlunicompo{Rules}{Start}
\umlunicompo{Rules}{Play}
\umlunicompo{Rules}{End}
\end{tikzpicture}
\end{center}

The following sections detail each type of move.


\section{Meta Rules}

A meta rule is a rule applied at each turn whatever the conditions. These rules are optional. Ludii has some meta rules in its current version:

\renewcommand{\arraystretch}{1.3}
\arrayrulecolor{white}
\rowcolors{2}{gray!25}{gray!10}
\begin{longtable}{@{}p{.28\textwidth} | p{.65\textwidth}@{}}
\rowcolor{gray!50}
\textbf{Meta Rule} & \textbf{Description} \\
(automove) & Apply automatically all the legal moves only applicable to a single site. \\
(gravity ...) & Activates the meta rule to apply a specific gravity. \\
(passEnd ...) & To modify the end of the game in case of all the players pass their turns. \\
(pin ...) & Activates the meta rule to remove from the legal moves the ones which can removea piece which is supported by more than one piece. \\
(swap) & Add the swap move to the legal moves for the first turn of the players different than the first one. \\
(no Repeat ...) & Filter the legal moves reaching a state already reached before. \\
(no Suicide ...) & Filter the legal moves reaching a state leading to a direct loss for the mover. \\
\end{longtable}
\rowcolors{0}{}{}


\section{Start Rules}

The starting rules define a list of movements (list of actions) $A_0$ applied to a state $s_{-1}$ with no piece placed (all the sites are empty in all the containers) and with all the variables set at their default value. After applying $A_0$ to $s_{-1}$, the state created is $s_0$, the initial state of the game. The starting rules are optional, if no starting rules defined, $s_0$ corresponds to $s_{-1}$.

Ludii has different starting rules ludemes. The most used is the Place ludeme. This ludeme places piece(s) (or stack of pieces) on a site (or a set of sites). Simultaneously, that's possible to set the state, the rotation value (based on the facing direction of the piece placed on that site) or the value of the pieces placed. Here are some examples of how it's used:

\begin{boxex}
\begin{ludii}
(place "Commander1" coord:"A1" rotation:1)
(place "Marshal1" 84 value:5)
(place "Pawn1" (sites Row 1))
(place "Rook2" {"A8" "H8"} state:1)
(place Stack "Goat0" (sites {"B2" "D2" "B4"}) counts:{5 5 5})
(place Random {"Ball"} count:29) 
\end{ludii} 
\end{boxex}

The first line places a piece called "Commander" owned by the player 1 on the site corresponding to the coordinate A1 on the board with one rotation to the right according to its facing direction. The second line places a piece called "Marshall" owned by the player 1 on the site 84 and that site will be invisible to the player 2. The third line places a piece called "Pawn" owned by the player 1 on all the sites of the row of index 1. The fourth line places a piece called "Rook" owned by the player 2 on the sites corresponding to the coordinates A8 and H8 with a state site of 1. The fifth line places a stack of 5 pieces called "Goat" owned by no player (neutral pieces) on the sites corresponding to the coordinates B2, D2 and B4. The last line places randomly a component called "Ball" shared by all the players on 29 empty sites.

For all these placement starting rules, Ludii checks if the pieces named in them are defined in the equipment. If they are not, the game will not compile, giving a warning to the game designer of their mistake.

The other common starting rules ludeme is the Set ludeme. Many data can be initiated with this ludeme. Here are some examples

\begin{boxex}
\begin{ludii}
(set Cost 5 Vertex at:10)
(set Phase 1 Cell at:3)
(set Amount 5000)
(set score P1 50)
(set Team 1 {P1 P3})
(set Hidden at:1 to:P1)
(set Hidden What at:5 level:1 to:P2)
\end{ludii} 
\end{boxex}

The first line sets the cost of the vertex 10 to 5. The second line set the phase of the cell 3 to 1. The third line set the amount of all the players to 5000. The fourth line sets the score of the player 1 to 50. The fifth lines puts the player 1 and 3 in the team 1. The sixth line makes invisible the site 1 to the player 1. The last line makes the piece index of site 5 at the level 1 hidden to the player 2.

Some starting rules exist also to iterate through data of the game to create the initial state. They all start with (forEach ...) and are similar to their equivalents in the Moves ludemes. Here are some examples

\begin{boxex}
\begin{ludii}
(forEach Player (set Hidden What at:1 to:Player))
(forEach (players Ally of:(next)) (set Hidden What at:1 to:Player))
(forEach Value min:1 max:5 (set Hidden at:10 level:(value) to:P1))
(forEach Site (sites Top) if:(is Odd (site)) (place "Spy1" (site)))
\end{ludii} 
\end{boxex}

The two last starting rules are the Split ludeme and the Deal ludeme. Both are used for cards or dominoes games. The first ludeme can split a deck of cards in two equals decks, the second one deal a certain number of cards/dominoes to each player. However, both can be modified in a future version of Ludii.


\section{Play Rules}\label{Section:Play}

The playing rules of a game describe the legal moves of the mover in the current state. These legal moves are defined in the Play ludeme thanks to Moves ludemes. 

\begin{center}
\begin{tikzpicture}

\begin{umlpackage}{rules}
\begin{umlpackage}{play}
\umlclass[rectangle split parts=1]{Play}{}{}

\begin{umlpackage}{moves}
\umlclass[rectangle split parts=1, y = -2]{Moves}{}{}
\umlclass[rectangle split parts=1, x = 3, y = -6, alias=ludemeMove]{Move}{}{}
\umlclass[rectangle split parts=1, x = -2, y = -4]{NonDecision}{}{}
\umlclass[rectangle split parts=1, x = -4, y = -6]{Effect}{}{}
\umlclass[rectangle split parts=1, x = 0, y = -6]{Operator}{}{}
\umlclass[rectangle split parts=1, x = 3, y = -4]{Decision}{}{}

\end{umlpackage}
\end{umlpackage}
\end{umlpackage}

\begin{umlpackage}[x = 7,y = -2]{util}
\umlclass[rectangle split parts=1, alias=objectMove]{Move}{}{}
\end{umlpackage}

\umlinherit{Moves}{Decision}
\umlinherit{Moves}{NonDecision}
\umlinherit{NonDecision}{Effect}
\umlinherit{NonDecision}{Operator}
\umlinherit{Decision}{ludemeMove}
\umlunicompo{Play}{Moves}
\umlunicompo{Moves}{objectMove}

\end{tikzpicture}
\end{center}

Each time a new state is reached, the (play ...) ludeme evaluates the Moves object stored in it in calling the method Moves eval(Context context). 
This method returns a Moves object containing a list of util.Move objects (see Section \ref{Section:Move}) modelling the list of all the legal moves and an object called Then.
The Then object corresponds to the consequences of each legal move. This object contains its own Moves object evaluated and applied after applying the move decided by the mover.

The Moves object can be a decision or not as shown in the UML diagram above. A decision move involves a decision taken by a player. A non decision move is an effect applied because of a decision (see Section \ref{Section:effect}) or is an operator (see Section \ref{Section:operator}).

\subsection{Decision Moves}

Ludii has only one single decision move called Move. This ludeme modifies internally all the Move objects evaluated thanks to the eval method. Inside each Move object only one action can be a decision. If the Move object is a decision, the data about the first decision action in the list of actions becomes the data of the move. These data are used by Ludii and the user interface to identify a specific move in the list of legal moves when a decision is taken. 

For example, the move $m$ described by the list of actions $L$ $=$ $\langle$ $ActionRemove(20),$ $ActionMove(10,20),$ $ActionSetScore(1,10)$ $\rangle$ moves a piece from the site 10 to the site 20, removes what is on the site 20 before the move and sets the score of the player 1 to 10. The decision action is the action $ActionMove(10,20)$. When the user clicks on the site 10 then the site 20, the move $m$ is selected and the list $L$ of actions is applied to modify the state $s_i$ to reach the state $s_{i+1}$.

Here is the list of the possible decision moves, followed by a short description:
\renewcommand{\arraystretch}{1.3}
\arrayrulecolor{white}
\rowcolors{2}{gray!25}{gray!10}
\begin{longtable}{@{}p{.28\textwidth} | p{.65\textwidth}@{}}
\rowcolor{gray!50}
\textbf{Decision Move} & \textbf{Description} \\
(move ...) & To decide to move from a site to another. \\
(move Add ...) & To decide to add (a) piece(s) on a site or a set of sites. \\
(move Claim ...) & To decide to claim a site or a set of sites. \\
(move Bet ...) & To decide to bet. \\
(move Hop ...) & To decide to hop. \\
(move Leap ...) & To decide to leap. \\
(move Pass ...) & To decide to pass. \\
(move PlayCard ...) & To decide to play a card. \\
(move Promote ...) & To decide to promote a piece. \\
(move Propose ...) & To decide to propose a subject. \\
(move Remove ...) & Used to remove (a) piece(s) placed on a site or a set of sites. \\
(move Set Rotation ...) & To decide to modify the rotation state of a piece. \\
(move Set NextPlayer ...) & To decide to modify the next player to play. \\
(move Set TrumpSuit ...) & To decide to set the trump suit in a card game. \\
(move Select ...) & To decide to select sites. \\
(move Shoot ...) & To decide to shoot. \\
(move Slide ...) & To decide to slide. \\
(move Step ...) & To decide to step. \\
(move Swap Pieces ...) & To decide to swap two pieces placed on two different sites. \\
(move Swap Players ...) & To decide to swap two players. \\
(move Vote ...) & To decide to vote for a subject. \\
\end{longtable}
\rowcolors{0}{}{}

\subsection{Non Decision Moves}\label{Section:nonDecisionMoves}

The non decision moves ludemes are composed of all the other moves ludemes. Most of them are effects applied because of a decision and a few are operators used to combine or filter moves.

\subsection{Moves Operator}\label{Section:operator}

The Moves operators are used commonly to describe the list of legal moves in generating them in combining many moves or in filtering moves according to some constraints or conditions.

This section briefly describes the logic behind each operator:
\renewcommand{\arraystretch}{1.3}
\arrayrulecolor{white}
\rowcolors{2}{gray!25}{gray!10}
\begin{longtable}{@{}p{.28\textwidth} | p{.65\textwidth}@{}}
\rowcolor{gray!50}
\textbf{Operator} & \textbf{Description} \\
(allCombinations ...) & Generates all combinations in making the cross product between two lists of Move objects. \\
(and ...) & Combines lists of moves in one list. This ludeme is commonly used in a Then ludeme to apply many consequences. \\
(append ...) & Appends the list of moves to each move in the list. \\
(forEach Die ...) & Generating a list of moves in iterating the values of each die (and optionally with combinations of dice). \\
(forEach Direction ...) & Generating a list of moves in iterating each site reached by each possible direction from a site. \\
(forEach Group ...) & Generating a list of moves in iterating each group of pieces. \\
(forEach Piece ...) & Generating a list of moves according to the generators of moves implemented in each component owned by a specific player. \\
(forEach Player ...) & Generating a list of moves in iterating through the indices of the players with the ludeme (player). \\
(forEach Site ...) & Generating a list of moves in iterating each site in a region. \\
(forEach Team ...) & Generating a list of moves in iterating through the teams with the ludeme (team). \\
(forEach Value ...) & Generating a list of moves in iterating between two integers or on an IntArrayFunction. \\
(if ...) & Returning a list of moves if a condition is verified else returning (potentially) another list of moves. \\
(or ...) & Combines lists of moves in one list. \\
\end{longtable}
\rowcolors{0}{}{}

The iterating operators using a (forEach ...) loop are referring to the iterated element by different ludemes. 

To iterate through dice, the ludeme (forEach Die ...) uses the ludeme (pips) to get the value (current number of pips) of each die. 
In the following example, each piece owned by the mover can move to another site in following a track. The number of steps in the track is defined by the number of pips in each die.
\begin{boxex}
\begin{ludii}
(forEach Die 
    (move
        (from (sites Occupied by:Mover))
        (to (trackSite Move steps:(pips)))
    )
)    
\end{ludii} 
\end{boxex}

To iterate through directions, the ludeme (forEach Direction ...) modifies the value of the ludeme (to) to each site reached for each direction from a site.
In the following example, a piece can move to any empty diagonal adjacent site from its current position.
\begin{boxex}
\begin{ludii}
(forEach Direction 
    Diagonal
    (to
        if:(is Empty (to))
        (apply 
            (move (from) (to))
        )
    )
)    
\end{ludii} 
\end{boxex}

To iterate through groups of pieces, the ludeme (forEach Group ...) uses the ludeme (sites) to get the region (set of sites) of each group of pieces.
In the following example, a score of 1 is added to the mover for each group of pieces bigger than one piece owned by the mover.
\begin{boxex}
\begin{ludii}
(forEach Group
    (if (< 1 (count Sites in:(sites)))
        (addScore Mover 1)
    )
)
\end{ludii} 
\end{boxex}

To iterate through the pieces, the ludeme (forEach Piece ...) uses the ludeme (from) to iterate each position of the pieces in using the moves defined in the generator associated to the component. 
In the following example, the pieces named "Ball" owned by the mover are allowed to step to all the empty adjacent sites.
\begin{boxex}
\begin{ludii}
(equipment {
    ...
    (piece "Ball" Each 
        (move
            Step 
            (from) 
            (to if:(is Empty (to)))
        )
    )
})

(play (forEach Piece))
\end{ludii} 
\end{boxex}

To iterate through a region, the ludeme (forEach Site ...) uses the ludeme (site) to iterate each site in the region.
In the following example, all the states of the sites occupied by a piece owned by the mover are set to 1.
\begin{boxex}
\begin{ludii}
(forEach Site
    (sites Occupied by:Mover)
    (set State at:(site) 1)
)
\end{ludii} 
\end{boxex}
To iterate through teams, the ludeme (forEach Team ...) uses the ludeme (team) to iterate each team.
In the following example, the piece index on the site 1 is hidden to each player on a team.
\begin{boxex}
\begin{ludii}
(forEach Team
    (forEach (team)
        (set Hidden What at:1 of:Player)
    )
)
\end{ludii} 
\end{boxex}

To iterate over a range of integers, the ludeme (forEach Value ...) uses the ludeme (value).
In the following example, each piece owned by the mover can move to another site in following a track. The number of steps in the track is between 1 and 6.
\begin{boxex}
\begin{ludii}
(forEach Value min:1 max:6
    (move
        (from (sites Occupied by:Mover))
        (to (trackSite Move steps:(value)))
    )
)
\end{ludii} 
\end{boxex}

\subsection{Effect Moves}\label{Section:effect}

The effects are moves corresponding to a list of actions to apply as an effect of a decision. These effects can be applied before or after the decision according to their use.
All the effects moves ludemes encapsulated in a consequence by the ludeme Then (see Section \ref{Section:Play}) will be applied after the move itself and will refer to the previous state, commonly in using (last From) and (last To) to obtain the site from where began the move to the site where the piece has moved. \\

For example, in the following example, a piece can step from its current site to any adjacent site occupied by an enemy piece. After that move is applied, the consequence \texttt{(remove (last To))} is applied, removing the piece just moved. Assuming the move was from the site 5 to the site 10, the list $L$ of actions to apply is $L$ $=$ $\langle$ $ActionMove(5,10),$ $ActionRemove(10)$ $\rangle$.

\begin{boxex}
\begin{ludii}
(move 
    Step 
    (from) 
    (to
        if:(is Enemy (who at:(to)))
    )
    (then
        (remove (last To))
    )
)
\end{ludii} 
\end{boxex}

The other effects which are different than a consequence will be applied before the decision move. For example, in the following example, a piece can step from its current site to any adjacent site occupied by an enemy piece. But before to move, the enemy piece on the site to move is removed because of the ludeme \texttt{(apply (remove (to)))}. Assumed the move was from the site 5 to the site 10, the list $L$ of actions to apply is $L$ $=$ $\langle$ $ActionRemove(10),$ $ActionMove(5,10)$ $\rangle$.

\begin{boxex}
\begin{ludii}
(move
    Step 
    (from) 
    (to 
        if:(is Enemy (who at:(to)))
        (apply 
            (remove (to))
        )
    )
 )
\end{ludii} 
\end{boxex}

It is important to note that the moves encapsulated inside a (then ...) ludeme are not referring to the current state but to a state modified by the application of the move corresponding to the decision of the mover. That's why, it is needed to use the ludemes (last To) and (last From) inside that ludeme to refer to the 'from' and 'to' sites of the decision just taken and not any variable set during the computation of the moves such as (from), (to), (between), (site) or any iterators. These different iterators will be already reset to their default value when the computation of the moves inside the (then ..) ludeme start.

In the following example, all the occupied sites by a piece owned by the mover can be selected. When a decision is taken to select a site, the piece on that site is flipped. We can see here that the site to flip is referred with (last To) in the (then ...) move contrary to (site) outside of it.

\begin{boxex}
\begin{ludii}
(forEach Site (sites Occupied by:Mover)
    (move Select (from (site))
        (then (flip (last To)))
    )
)
\end{ludii} 
\end{boxex}

\subsection{Requirement Moves}

The requirement moves define criteria that must be satisfied for a move to be legal. These are typically applied to lists of generated moves to filter out those that do not meet the specified criteria. 

This section briefly describes the logic behind each requirement move:
\renewcommand{\arraystretch}{1.3}
\arrayrulecolor{white}
\rowcolors{2}{gray!25}{gray!10}
\begin{longtable}{@{}p{.28\textwidth} | p{.65\textwidth}@{}}
\rowcolor{gray!50}
\textbf{Operator} & \textbf{Description} \\
(avoidStoredState ...) &  Filters the legal moves to avoid to reach a specific state. \\
(do ...) &  Applies a sequence of moves in a specified order. Can also filter some moves according to conditions on the state reached after applying that move. \\
(firstMoveOnTrack ...) &  Returns the first legal move on the track. \\
(max Distance ...) &  Filters the moves to keep only the moves allowing the maximum distance on a track in a turn. \\
(max Captures ...) &  Filters a list of moves to keep only the moves doing the maximum possible number of removing actions (captures). \\
(max Moves ...) &  Filters a list of legal moves to keep only the moves allowing the maximum number of moves in a turn. \\
(priority ...) &  Returns the first list of moves with a non-empty set of moves. \\
(while ...) &  Applies a move while a condition is true. \\
\end{longtable}
\rowcolors{0}{}{}

The ludeme (avoidStoredState ...) is twinned with the ludeme (remember State). When (remember State) is called, the current state is stored and when (avoidStoredState ...) encapsulates a Moves ludeme, all moves reaching the state stored in applying it is removed from the list of moves computed.

The ludeme (do ...) can be used in two manners. 
The first one is to apply a set of Moves before to apply another one, then if the moves corresponding to the next part is not empty, the corresponding moves are returned. In the following example, all the dice are rolled before to compute the moves of each piece.
\begin{boxex}
\begin{ludii}
(do (roll)
    next:(forEach Piece)
)
\end{ludii} 
\end{boxex}

The second manner is to compute a set of moves, then to filter them according to some conditions on the state reached. On the following example, all the moves of each piece are computed and all the ones reaching a state where all the pieces of the next player are removed from the list of moves returning by the ludeme (do ...).
\begin{boxex}
\begin{ludii}
(do 
    (forEach Piece)
    ifAfterwards:(!= 0 (count Sites occupied by:Next))
)
\end{ludii} 
\end{boxex}

Move requirements such as (do ...) can be quite powerful when used correctly, but care must be taken as they can have a high performance overhead.

\subsection{Set Moves}

The Set Moves ludeme sets some aspect of the game state:
\renewcommand{\arraystretch}{1.3}
\arrayrulecolor{white}
\rowcolors{2}{gray!25}{gray!10}
\begin{longtable}{@{}p{.28\textwidth} | p{.65\textwidth}@{}}
\rowcolor{gray!50}
\textbf{Operator} & \textbf{Description} \\
(set Count ...) &  Sets the count of a site (the number of pieces of the same type on that site). \\
(set Counter ...) &  Sets the counter of the game. \\
(set Hidden ...) &  Makes the site invisible to a player. \\
(set Hidden Count ...) &  Makes the number of pieces on a site hidden to a player. \\
(set Hidden Rotation ...) &  Makes the piece rotation on a site hidden to a player. \\
(set Hidden State ...) &  Makes the state site hidden to a player. \\
(set Hidden Value ...) &  Makes the piece value on a site hidden to a player. \\
(set Hidden What ...) &  Makes the piece index on a site hidden to a player. \\
(set Hidden Who ...) &  Makes the piece owner on a site hidden to a player. \\
(set NextPlayer ...) &  Sets the player to play in the next turn. \\
(set Pending ...) &  Sets the state in pending with specific value(s). \\
(set Pot ...) &  Sets the pot. \\
(set Rotation ...) &  Sets the rotation state of a piece. \\
(set Score ...) &  Sets the score of a player. \\
(set State ...) &  Sets the state of a site. \\
(set Team ...) &  Sets a team of players. \\
(set TrumpSuit ...) &  Sets the trump suit of the card game. \\
(set Value ...) &  Sets the value associated to a player. \\
(set Var ...) &  Sets a variable Var associated to the state. \\
\end{longtable}
\rowcolors{0}{}{}

\subsection{Take Moves}

The (take ...) ludeme is a generic ludeme for any move related to take:
\renewcommand{\arraystretch}{1.3}
\arrayrulecolor{white}
\rowcolors{2}{gray!25}{gray!10}
\begin{longtable}{@{}p{.28\textwidth} | p{.65\textwidth}@{}}
\rowcolor{gray!50}
\textbf{Operator} & \textbf{Description} \\
(take Control ...) &  Replaces all the pieces of a player by the equivalent pieces to another player. \\
(take Domino ...) &  Takes a domino from those remaining in the bag. \\
\end{longtable}
\rowcolors{0}{}{}

Concerning the (take Control ...) ludeme, it is important to define in the equipment the same pieces for all the players for it to fully work. The pieces to take control without equivalent are removed.

\subsection{State Moves}

The State Moves ludemes are ludemes modifing the state similarly to the Set Moves.

\renewcommand{\arraystretch}{1.3}
\arrayrulecolor{white}
\rowcolors{2}{gray!25}{gray!10}
\begin{longtable}{@{}p{.28\textwidth} | p{.65\textwidth}@{}}
\rowcolor{gray!50}
\textbf{Operator} & \textbf{Description} \\
(addScore ...) &  Add a value to the current score of a player. \\
(forget Value ...) &  Remove one or more value(s) kept in memory by previous state. \\
(moveAgain) &  Set the next player to the mover for him to play again in the next state. \\
(remember State) &  Store the current state in order to avoid to go back to it with the ludeme (avoidStoredState ...). \\
(remember Value ...) &  Store a value to keep in memory for future states. \\
(swap Players ...) &  Swap two players (in switching their name). \\
(swap Pieces ...) &  Swap two pieces placed on two different sites. \\
\end{longtable}
\rowcolors{0}{}{}

\subsection{Others}

\renewcommand{\arraystretch}{1.3}
\arrayrulecolor{white}
\rowcolors{2}{gray!25}{gray!10}
\begin{longtable}{@{}p{.28\textwidth} | p{.65\textwidth}@{}}
\rowcolor{gray!50}
\textbf{Operator} & \textbf{Description} \\
(add ...) &  To add (a) piece(s) on a site. \\
(apply ...) &  To apply an effect if a condition is satisfied. \\
(attract ...) &  To attract all the pieces as close as possible to a site in each direction. \\
(bet ...) & To bet an amount. \\
(claim ...) & To claim a site by adding a piece of the colour of the owner. \\
(custodial ...) & To apply an effect to all flanked sites. \\
(deal ...) & To deal dominoes or cards to the players. \\
(directionCapture ...) & To capture all the pieces in the same or opposite direction of the last move. \\
(enclose ...) & To apply an effect to all the pieces enclosed between group(s) of pieces containing the last piece moved or placed. \\
(flip ...) & To flip a piece from its current state to the corresponding flipped state defined in the piece. \\
(fromTo ...) & To move a piece from a site to another site. \\
(hop ...) & To jump from a site to another in following a radial. \\
(intervene ...) & To apply an effect to pieces flanking (a) site(s). \\
(leap ...) & To move to a site resulting from a turtle graphic walk. \\
(note ...) & To notify a player with a message. \\
(pass ...) & To pass the turn. \\
(playCard ...) & To play a card from a fan to the board. \\
(promote ...) & To promote a piece to another. \\
(propose ...) & To propose a subject to vote. \\
(push ...) & To push pieces in a direction. \\
(remove ...) & To remove (a) piece(s) placed on a site or a set of sites. \\
(roll ...) & To roll dice. \\
(satisfy ...) & To return the legal moves of a deduction puzzle. \\
(select ...) & To select one or two sites (from and to). \\
(shoot ...) & To add a piece to any site in the line of sight of a piece. \\
(slide ...) & To move a piece to any site in its line of sight. \\
(sow ...) & To sow pieces to many sites in following a track. \\
(step ...) & To step a piece from a site to another connected one. \\
(surround ...) & To apply an effect to all the sites surrounded. \\
(then ...) & To encapsulate effects in a consequence moves. \\
(trigger ...) & To trigger an event. \\
(vote ...) & To vote for a subject. \\
\end{longtable}
\rowcolors{0}{}{}

In the following part, more examples and explanations are providing for the most common moves with many possible uses. The blue/red dots corresponds to the legal moves according to these descriptions. 


\subsubsection{Add}

The ludeme {\it Add} can be used to add one or many pieces on different sites and according to different conditions. It is generally used as a decision move, an effect or a consequence.
The main parameters of this ludeme are:
\begin{itemize}
    \item (piece ...) describes the piece to add (Default: (piece (mover))).
    \item (to ...) describes the site or region to place add the piece.
    \item count: specifies the number of pieces to add (Default: 1).
    \item stack: specifies if the piece is added on top of a stack or not.
\end{itemize}
Here are some examples of how it's used in a 3x3 square board using the cells by default.

\renewcommand{\arraystretch}{1.3}
\arrayrulecolor{white}
\begin{longtable}{@{}M{.6\textwidth} | M{.4\textwidth}@{}}
\rowcolor{gray!50}
\textbf{Description} & \textbf{Result}\\
A ball can be added to the empty cells. 
 \begin{boxexnonbreakable}
\begin{ludii}
(move Add 
    (to 
        (sites Empty)
    )
)
\end{ludii} 
\end{boxexnonbreakable} 
&  
\begin{center}
\includegraphics[width=0.60\linewidth]{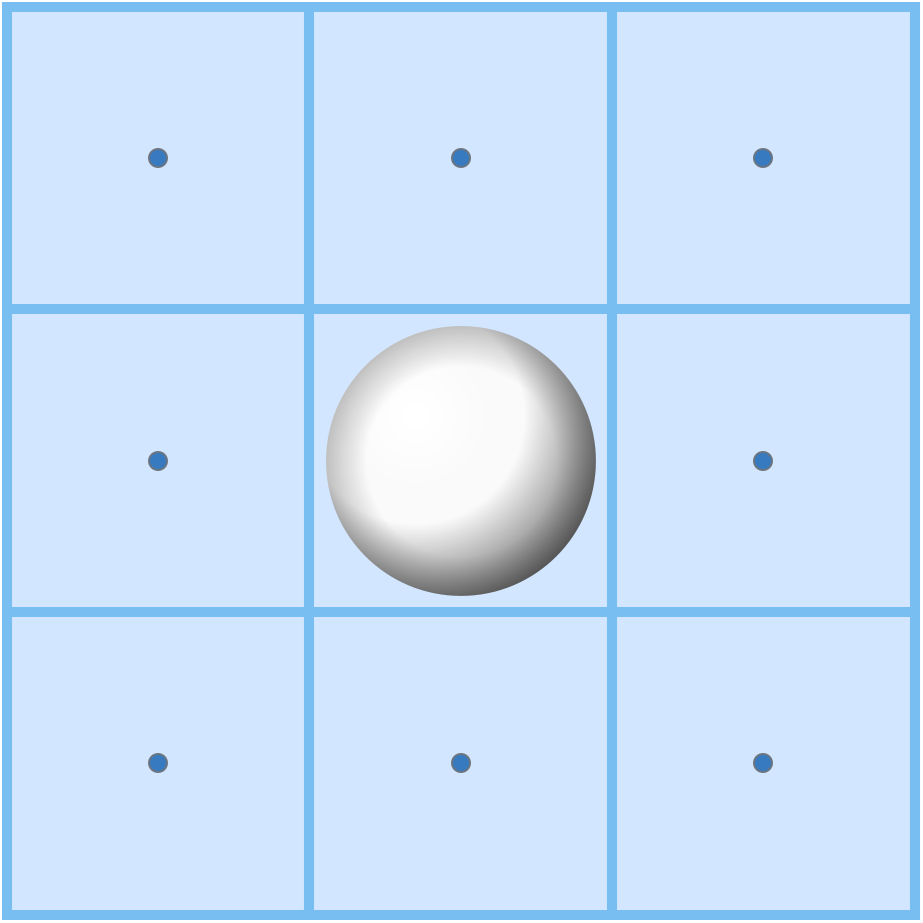}
\end{center} \\ 

Two balls can be added to the empty cells.
 \begin{boxexnonbreakable}
\begin{ludii}
(move Add 
    (to (sites Empty)) 
    count:2 stack:true
)
\end{ludii} 
\end{boxexnonbreakable} 
&  
\begin{center}
\includegraphics[width=0.60\linewidth]{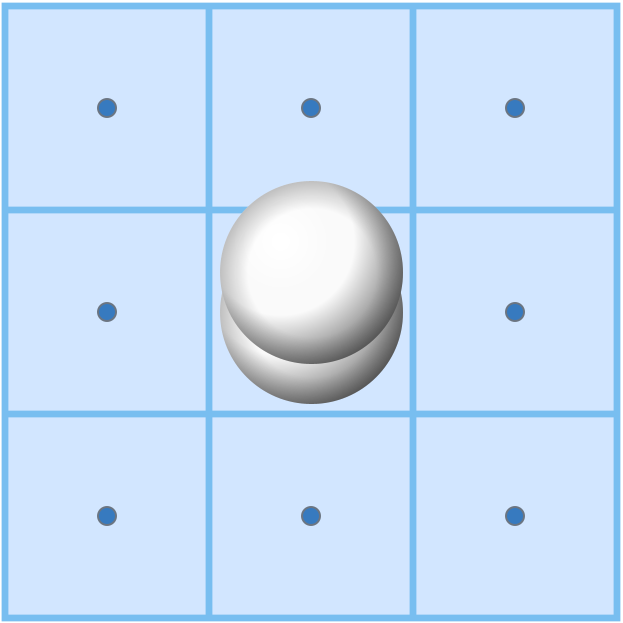}
\end{center} \\ 

A king can be added to the empty vertices.
 \begin{boxexnonbreakable}
\begin{ludii}
(move Add 
    (piece (id "King" Mover))
    (to Vertex 
        (sites Empty Vertex)
    ) 
)
\end{ludii} 
\end{boxexnonbreakable} 
&  
\begin{center}
\includegraphics[width=0.60\linewidth]{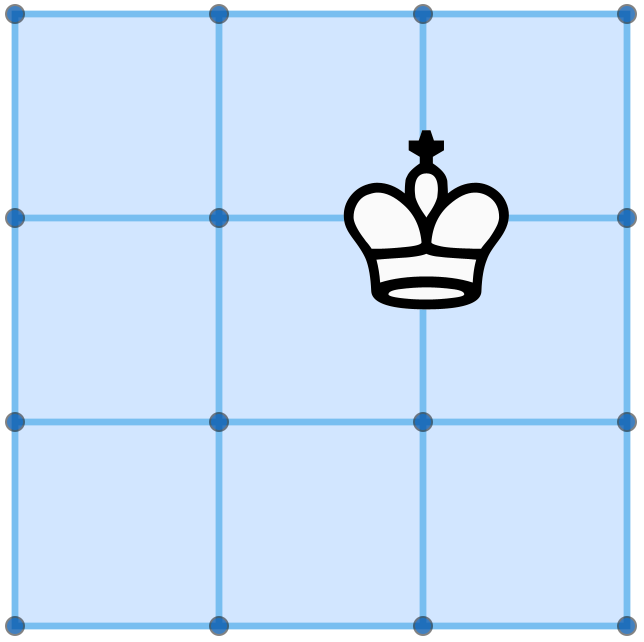}
\end{center} \\ 

A ball can be added to the empty edges.
 \begin{boxexnonbreakable}
\begin{ludii}
(move Add 
    (to Edge 
        (sites Empty Edge)
    ) 
)
\end{ludii} 
\end{boxexnonbreakable} 
&  
\begin{center}
\includegraphics[width=0.60\linewidth]{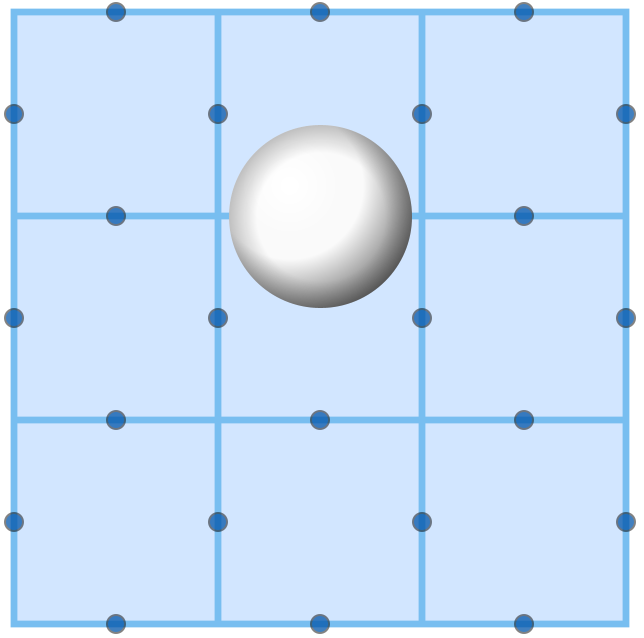}
\end{center} \\ 

A disc can be added only on enemy stack of size 1.
 \begin{boxexnonbreakable}
\begin{ludii}
(move Add 
    (to 
        (sites Occupied by:Next)
        if:(= 1 (size Stack at:(to)))
    )
    stack:true
)
\end{ludii} 
\end{boxexnonbreakable}  
&  
\begin{center}
\includegraphics[width=0.60\linewidth]{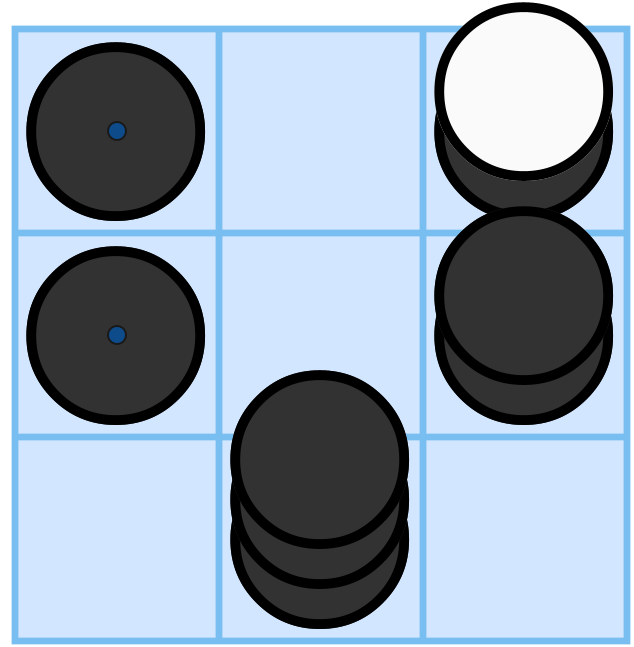}
\end{center} \\ 
\end{longtable}


\subsubsection{Attract}

The ludeme {\it Attract} used to attract all the pieces in different direction from a site (e.g. Feed the Ducks\footnote{Feed the Ducks: \href{https://www.nestorgames.com/rulebooks/FEEDTHEDUCKS_EN.pdf}{nestorgames.com}}) is generally used as a consequence after moving or placing a piece.
The main parameters of this ludeme are:
\begin{itemize}
    \item (from ...) describes the attracting site (default: (from (last To))).
    \item The direction to attract (default: Adjacent).
\end{itemize}

Here are some examples of how it's use in a 8x8 square board using the cells by default:
\renewcommand{\arraystretch}{1.3}
\arrayrulecolor{white}
\begin{longtable}{@{}M{.54\textwidth} | M{.23\textwidth}@{} | M{.23\textwidth}@{}}
\rowcolor{gray!50}
\textbf{Description} & \textbf{Before} & \textbf{After}\\

Attract pieces in all directions.

 \begin{boxex}
\begin{ludii}
(move 
    Add 
    (to (sites Empty)) 
    (then 
        (attract (from (last To)))
    )
)
\end{ludii} 
\end{boxex} 
&  
\begin{center}
\raisebox{-.50\height}{\includegraphics[width=0.9\linewidth]{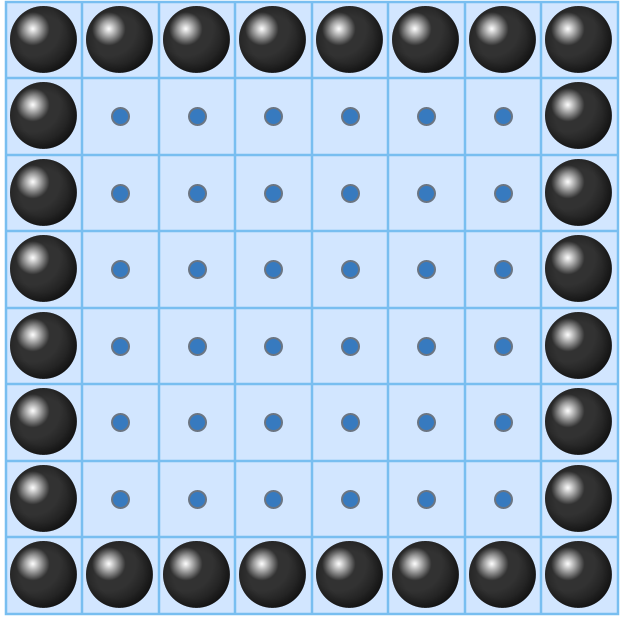}}
\end{center} 
&  
\begin{center}
\raisebox{-.50\height}{\includegraphics[width=0.9\linewidth]{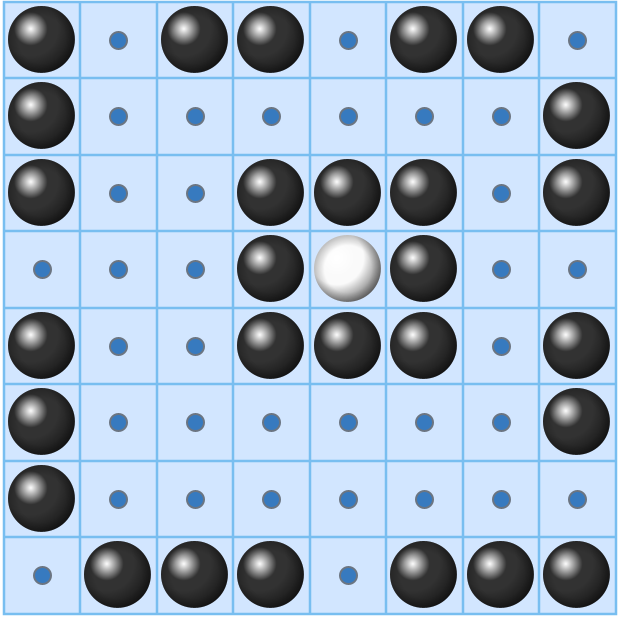}}
\end{center} \\ 

Attract pieces in orthogonal directions.

 \begin{boxex}
\begin{ludii}
(move 
    Add 
    (to (sites Empty)) 
    (then 
        (attract 
            (from (last To)) 
            Orthogonal
        )
    )
)
\end{ludii} 
\end{boxex} 
&  
\begin{center}
\raisebox{-.50\height}{\includegraphics[width=0.9\linewidth]{figs/AttractAllBefore.PNG}}
\end{center} 
&  
\begin{center}
\raisebox{-.50\height}{\includegraphics[width=0.9\linewidth]{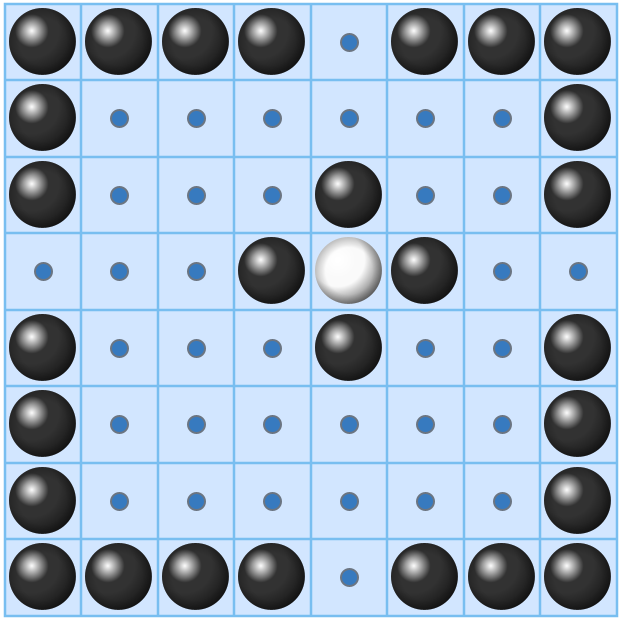}}
\end{center} \\ 

\end{longtable}


\subsubsection{Custodial}

The ludeme {\it Custodial} is generally used as a consequence after moving or placing a piece to capture flanked pieces (e.g. Hnefatfl\footnote{Talf games: \href{https://en.wikipedia.org/wiki/Tafl_games}{Wikipedia}}).
The main parameters of that ludeme are:
\begin{itemize}
    \item (from ...) describes the site possibly causing the custodial effect (default: (from (last To))).
    \item (between ...) describes the data about the flanked site(s). (between) ludeme is used to iterate each potential flanked site. The condition is described with the parameter "if:" and the effect to apply described inside the ludeme (apply ...). It is also possible to define the maximum number of pieces to flank in each direction. 
    \item (to ...) describes the flanking sites. The (to) ludeme is used to iterate each potential flanking site. The condition on that site is described with the parameter "if:". 
\end{itemize}

Here are some examples of how it's used in a hex board of size 3 using the cells by default:
\renewcommand{\arraystretch}{1.3}
\arrayrulecolor{white}
\begin{longtable}{@{}M{.75\textwidth} | M{.25\textwidth}@{}}
\rowcolor{gray!50}
\textbf{Description} & \textbf{Before/After}\\

After stepping a piece, each flanked enemy is removed.

 \begin{boxex}
\begin{ludii}
(move Step 
    (to if:(is Empty (to)))
    (then 
        (custodial 
            (from (last To))
            (between 
                (max 1)
                if:(is Enemy (who at:(between)))
                (apply (remove (between)))
            )
            (to if:(is Friend (who at:(to))))
        )
    )
)
\end{ludii} 
\end{boxex} 
&  
\begin{center}
\raisebox{-.50\height}{\includegraphics[width=0.8\linewidth]{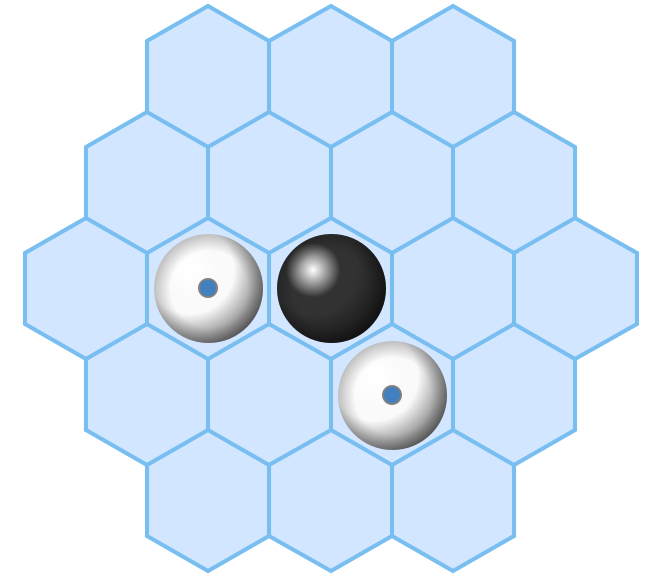}}
\end{center}
\begin{center}
\raisebox{-.50\height}{\includegraphics[width=0.8\linewidth]{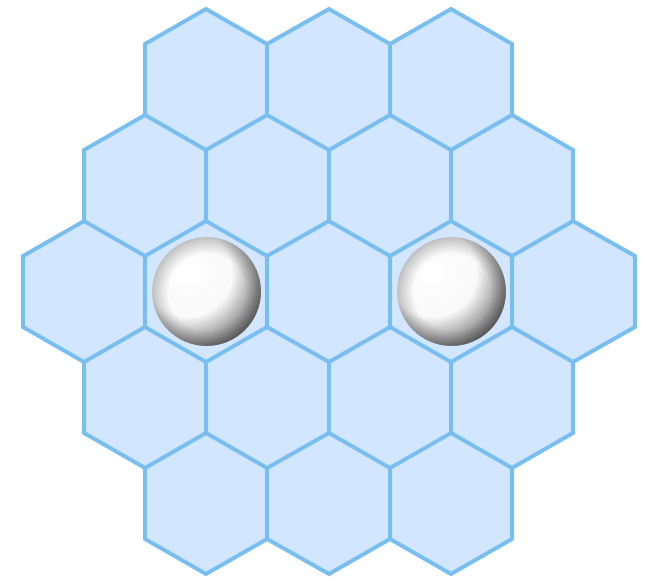}}
\end{center} \\ 

After stepping a piece, any line of flanked enemy pieces is removed.

 \begin{boxex}
\begin{ludii}
(move Step 
    (to if:(is Empty (to)))
    (then 
        (custodial 
            (from (last To))
            (between 
                if:(is Enemy (who at:(between)))
                (apply (remove (between)))
            )
            (to if:(is Friend (who at:(to))))
        )
    )
)
\end{ludii} 
\end{boxex} 
&  
\begin{center}
\raisebox{-.50\height}{\includegraphics[width=0.8\linewidth]{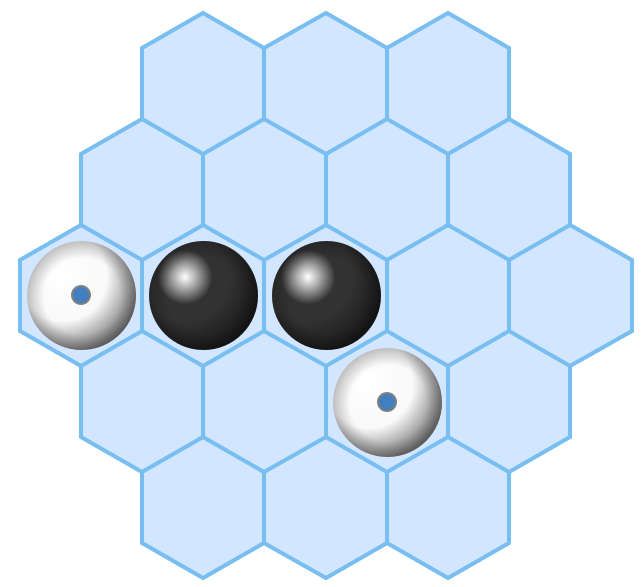}}
\end{center}
\begin{center}
\raisebox{-.50\height}{\includegraphics[width=0.8\linewidth]{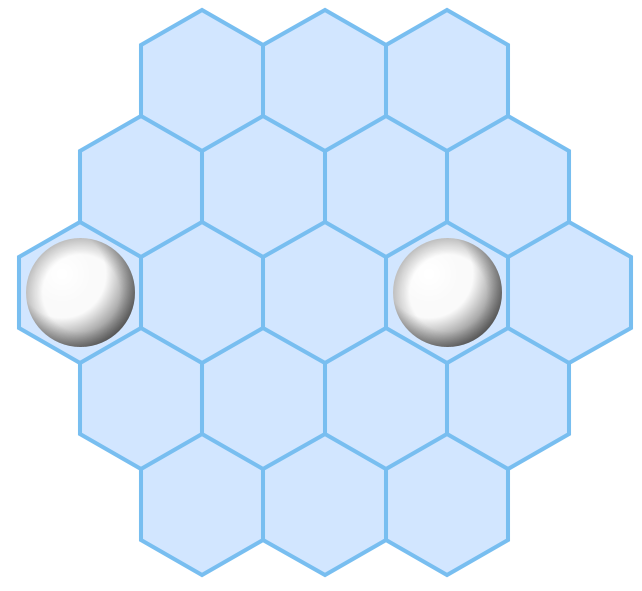}}
\end{center} \\ 

\end{longtable}


\subsubsection{DirectionCapture}

The ludeme {\it DirectionCapture} is generally used as a consequence after moving or placing a piece (e.g. Fanorona\footnote{Fanorona: \href{https://en.wikipedia.org/wiki/Fanorona}{Fanorona}}) to capture all the pieces in the same or opposite direction of the move.
The main parameters of that ludeme are:
\begin{itemize}
    \item (from ...) describes the site starting the capture (default: (from (last To))). In case of a capture in the opposite direction that site would have to be (from (last From)).
    \item (to ...) described the data of the sites. The (to) ludeme is used to iterate each site in the direction, when one site does not satisfy the condition the ludeme stops to check the next sites in that direction. The condition is described with the parameter "if:" and the effect to apply described inside the ludeme (apply ...). 
\end{itemize}

Here are some examples of how it's used in a hex board of size 3 using the cells by default:
\renewcommand{\arraystretch}{1.3}
\arrayrulecolor{white}
\begin{longtable}{@{}M{.75\textwidth} | M{.25\textwidth}@{}}
\rowcolor{gray!50}
\textbf{Description} & \textbf{Before/After}\\

After stepping a piece, all the enemy pieces in the same direction of the move are removed.

 \begin{boxex}
\begin{ludii}
(move Step 
    (to if:(is Empty (to)))
    (then 
        (directionCapture 
            (from (last To)) 
            (to 
                if:(is Enemy (who at:(to))) 
                (apply (remove (to)))
            )
        )
    )
)
\end{ludii} 
\end{boxex} 
&  
\begin{center}
\raisebox{-.50\height}{\includegraphics[width=0.8\linewidth]{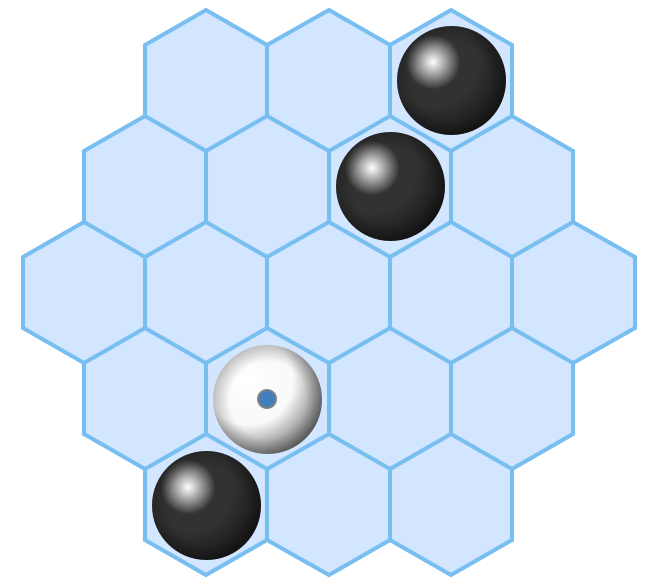}}
\end{center}
\begin{center}
\raisebox{-.50\height}{\includegraphics[width=0.8\linewidth]{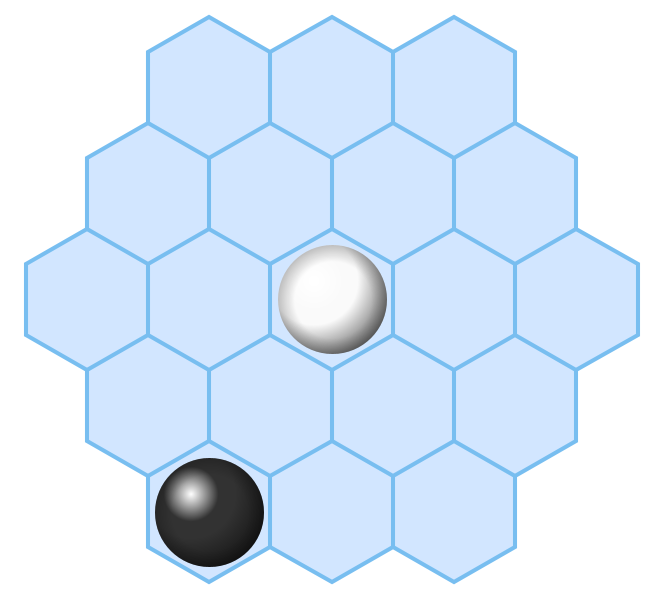}}
\end{center} \\ 

After stepping a piece, all the enemy pieces in the opposite direction of the move are removed.

 \begin{boxex}
\begin{ludii}
(move Step 
    (to if:(is Empty (to)))
    (then 
        (directionCapture 
            (from (last From)) 
            (to 
                if:(is Enemy (who at:(to))) 
                (apply (remove (to)))
            )
            opposite:true
        )
    )
)
\end{ludii} 
\end{boxex} 
&  
\begin{center}
\raisebox{-.50\height}{\includegraphics[width=0.8\linewidth]{figs/CaptureDirectionBefore.PNG}}
\end{center}
\begin{center}
\raisebox{-.50\height}{\includegraphics[width=0.8\linewidth]{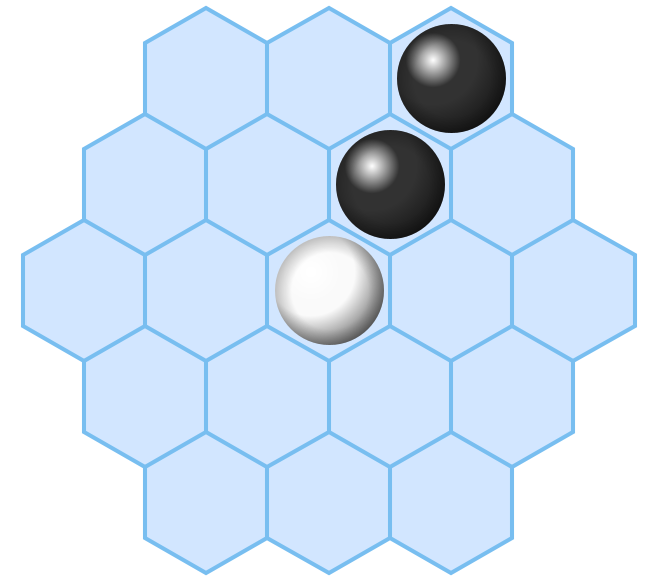}}
\end{center} \\ 

\end{longtable}


\subsubsection{Enclose}
The ludeme {\it Enclose} is generally used as a consequence after moving or placing a piece (e.g. Go\footnote{Go: \href{https://en.wikipedia.org/wiki/Go_(game)}{Wikipedia}}) to capture all the pieces enclosed because of the last move.
The main parameters of that ludeme are:
\begin{itemize}
    \item (from ...) describes the site possibly causing the enclose effect (default: (from (last To))).
    \item The directions used to enclose pieces (default: Adjacent).
    \item (between ...) describes the data on the enclosed sites. The (between) ludeme is used to iterate each potential enclosed site. The condition is described with the parameter "if:" and the effect to apply described inside the ludeme (apply ...).
\end{itemize}

Here are some examples of how it's used in a 9x9 go board using the vertices by default:
\renewcommand{\arraystretch}{1.3}
\arrayrulecolor{white}
\begin{longtable}{@{}M{.75\textwidth} | M{.25\textwidth}@{}}
\rowcolor{gray!50}
\textbf{Description} & \textbf{Before/After}\\

After placing a piece, the enemy pieces enclosed orthogonally are removed.

 \begin{boxex}
\begin{ludii}
(move Add
    (to (sites Empty))
    (then 
        (enclose 
            (from (last To)) 
            Orthogonal 
            (between 
                if:(is Enemy (who at:(between))) 
                (apply (remove (between))) 
            )
        )
    )
)
\end{ludii} 
\end{boxex} 
&  
\begin{center}
\raisebox{-.50\height}{\includegraphics[width=0.8\linewidth]{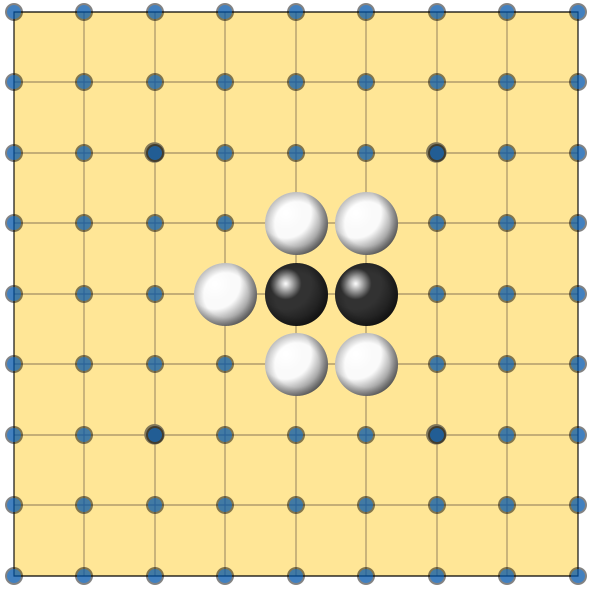}}
\end{center}
\begin{center}
\raisebox{-.50\height}{\includegraphics[width=0.8\linewidth]{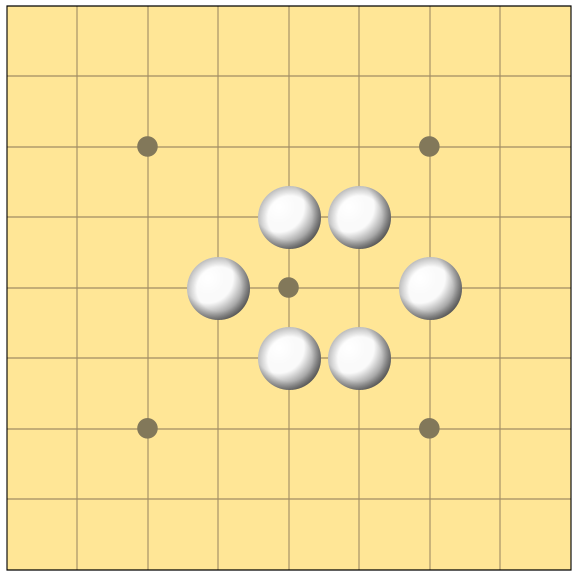}}
\end{center} \\ 

After placing a piece, the enemy pieces enclosed adjacently are removed.

 \begin{boxex}
\begin{ludii}
(move Add
    (to (sites Empty))
    (then 
        (enclose 
            (from (last To)) 
            (between 
                if:(is Enemy (who at:(between))) 
                (apply (remove (between))) 
            )
        )
    )
)
\end{ludii} 
\end{boxex} 
&  
\begin{center}
\raisebox{-.50\height}{\includegraphics[width=0.8\linewidth]{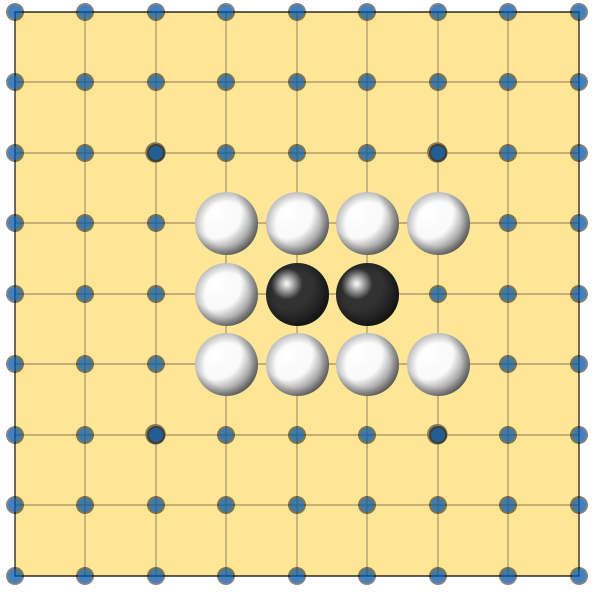}}
\end{center}
\begin{center}
\raisebox{-.50\height}{\includegraphics[width=0.8\linewidth]{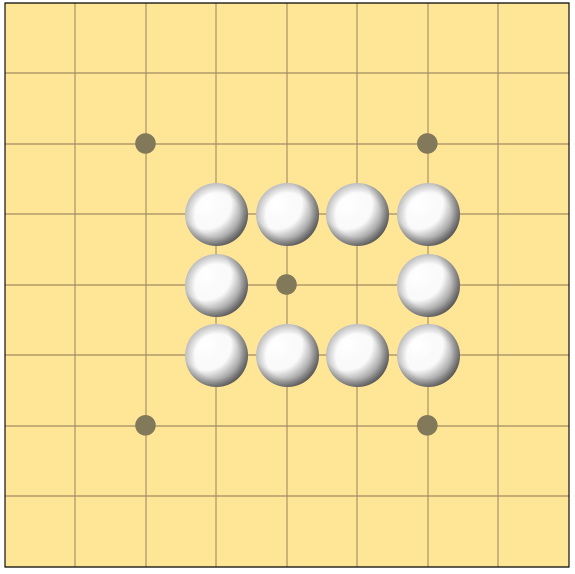}}
\end{center} \\ 

\end{longtable}


\subsubsection{Hop}
The ludeme {\it Hop} is used to jump over site(s) (e.g. Draughts\footnote{Draughts: \href{https://en.wikipedia.org/wiki/Draughts}{Wikipedia}}) and potentially to capture the pieces placed on the sites jumped. It is generally used in the generator of moves of the pieces.
The main parameters of that ludeme are:
\begin{itemize}
    \item (from ...) describes the origin site of the move (default: (from)).
    \item The directions allowed to jump (default: Adjacent).
    \item (between ...) describes the data on the jumped sites. The (between) ludeme is used to iterate each potential jumped site. The condition is described with the parameter "if:" and the effect to apply described inside the ludeme (apply ...). The maximum number of sites before to jump is described with the parameter "before:", the range of sites to jump is described with a RangeFunction and the maximum number of sites to stop the move after the jumped sites is described with the parameter "after:".
    \item (to ...) describes the data on the sites to move. The (to) ludeme is used to iterate each potential target site. The condition is described with the parameter "if:" and the effect to apply described inside the ludeme (apply ...).
\end{itemize}

Here are some examples of how it's used in a 7x7 square board using the cells by default:
\renewcommand{\arraystretch}{1.3}
\arrayrulecolor{white}
\begin{longtable}{@{}M{.75\textwidth} | M{.25\textwidth}@{}}
\rowcolor{gray!50}
\textbf{Description} & \textbf{Before/After}\\

A piece can jump an enemy piece to an empty site to capture it.

 \begin{boxex}
\begin{ludii}
(move Hop 
    (between 
        if:(is Enemy (who at:(between))) 
        (apply (remove (between)))
    )
    (to if:(is Empty (to)))
)
\end{ludii} 
\end{boxex} 
&  
\begin{center}
\raisebox{-.50\height}{\includegraphics[width=0.8\linewidth]{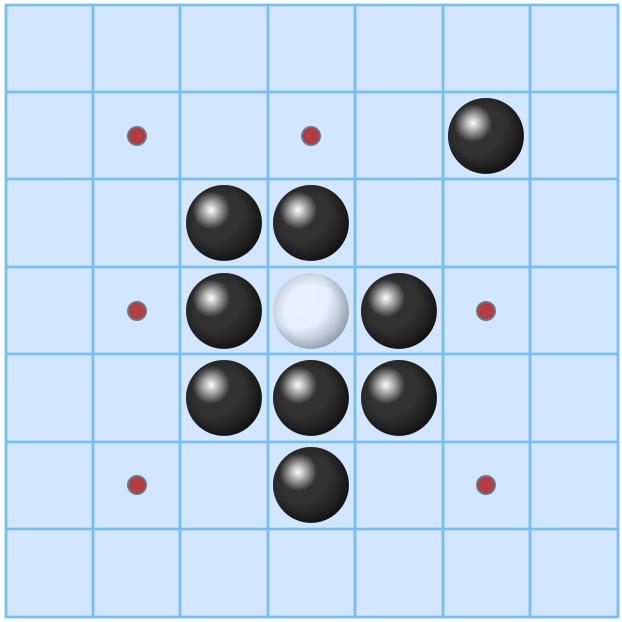}}
\end{center}
\begin{center}
\raisebox{-.50\height}{\includegraphics[width=0.8\linewidth]{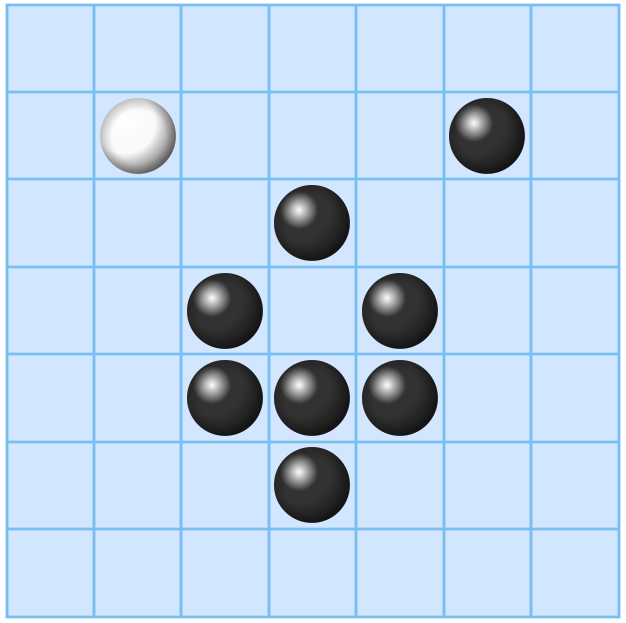}}
\end{center} \\ 

A piece can jump a diagonal enemy piece to an empty site to capture it.

 \begin{boxex}
\begin{ludii}
(move Hop 
    Diagonal
    (between 
        if:(is Enemy (who at:(between))) 
        (apply (remove (between)))
    )
    (to if:(is Empty (to)))
)
\end{ludii} 
\end{boxex} 
&  
\begin{center}
\raisebox{-.50\height}{\includegraphics[width=0.8\linewidth]{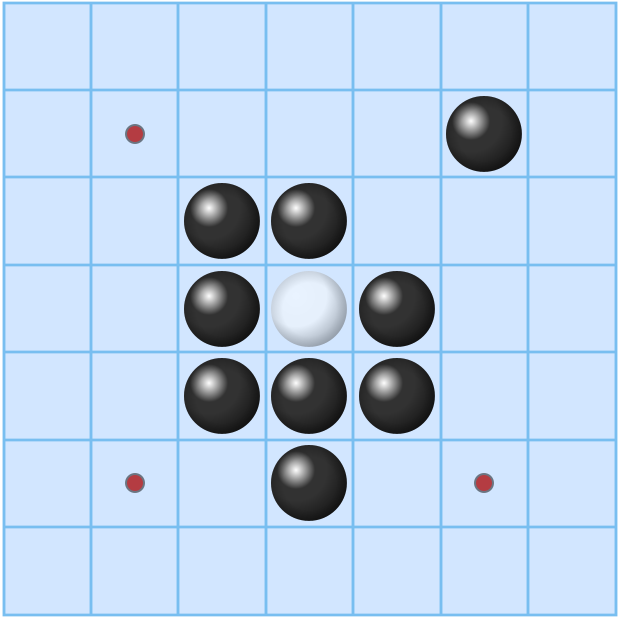}}
\end{center}
\begin{center}
\raisebox{-.50\height}{\includegraphics[width=0.8\linewidth]{figs/CaptureHopEmptyAfter.PNG}}
\end{center} \\ 

A piece can jump an orthogonal enemy piece to an empty site to capture it.

 \begin{boxex}
\begin{ludii}
(move Hop 
    Orthogonal
    (between 
        if:(is Enemy (who at:(between))) 
        (apply (remove (between)))
    )
    (to if:(is Empty (to)))
)
\end{ludii} 
\end{boxex} 
&  
\begin{center}
\raisebox{-.50\height}{\includegraphics[width=0.8\linewidth]{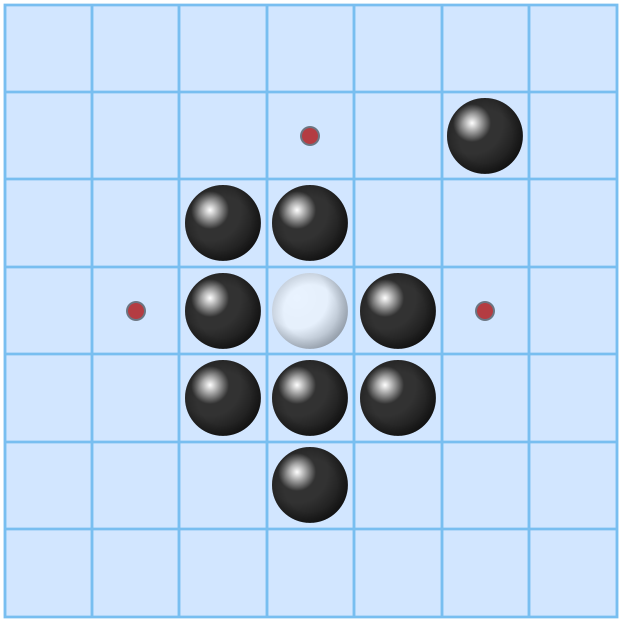}}
\end{center}
\begin{center}
\raisebox{-.50\height}{\includegraphics[width=0.8\linewidth]{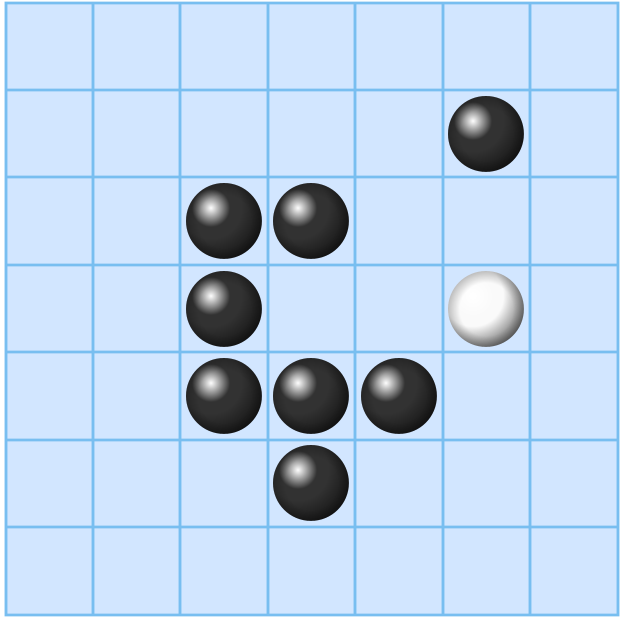}}
\end{center} \\ 

A piece can jump an orthogonal enemy piece to a site occupied by an enemy to capture both enemy pieces.

 \begin{boxex}
\begin{ludii}
(move Hop 
    Orthogonal
    (between 
        if:(is Enemy (who at:(between))) 
        (apply (remove (between)))
    )
    (to 
        if:(is Enemy (who at:(to))) 
        (apply (remove (to)))
    )
)
\end{ludii} 
\end{boxex} 
&  
\begin{center}
\raisebox{-.50\height}{\includegraphics[width=0.8\linewidth]{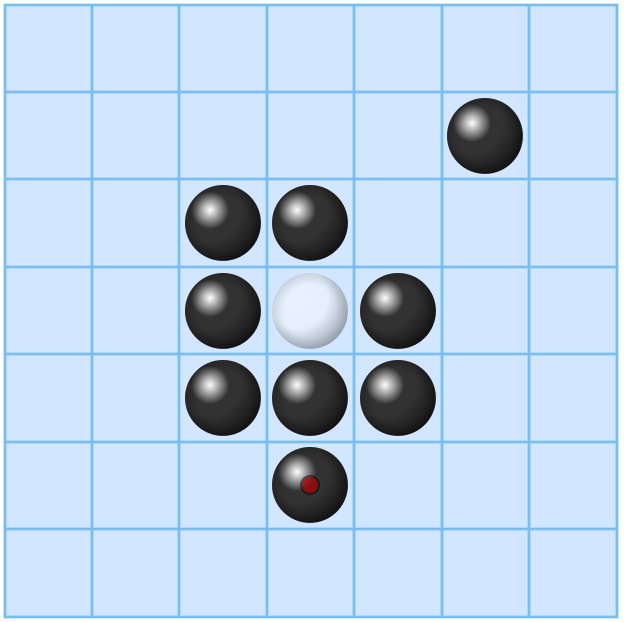}}
\end{center}
\begin{center}
\raisebox{-.50\height}{\includegraphics[width=0.8\linewidth]{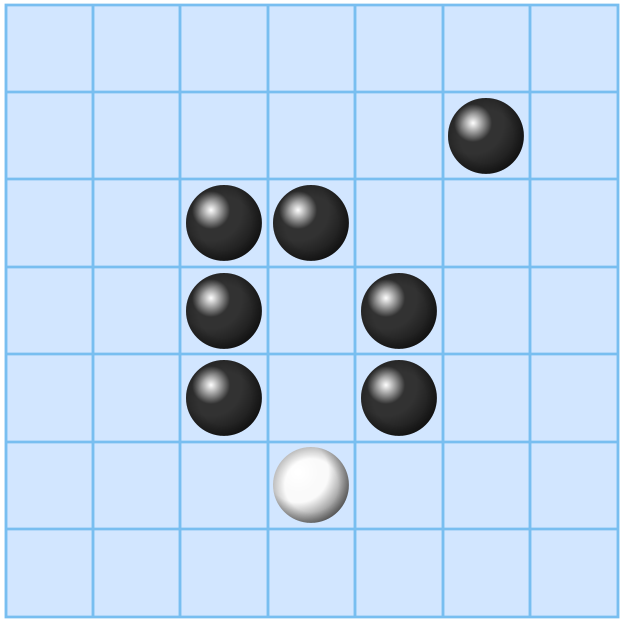}}
\end{center} \\ 

A piece can jump one or two orthogonal enemy pieces to an empty site in capturing the jumped pieces.

 \begin{boxex}
\begin{ludii}
(move Hop 
    Orthogonal
    (between 
        (range 1 2)
        if:(is Enemy (who at:(between))) 
        (apply (remove (between)))
    )
    (to if:(is Empty (to)))
)
\end{ludii} 
\end{boxex} 
&  
\begin{center}
\raisebox{-.50\height}{\includegraphics[width=0.8\linewidth]{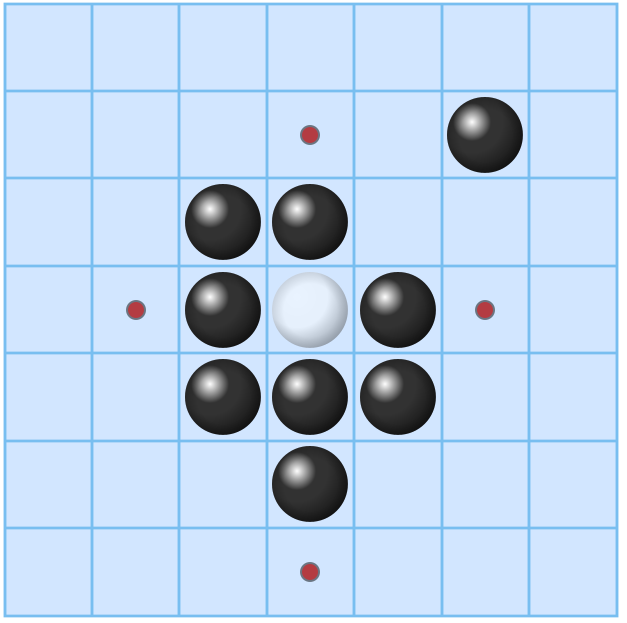}}
\end{center}
\begin{center}
\raisebox{-.50\height}{\includegraphics[width=0.8\linewidth]{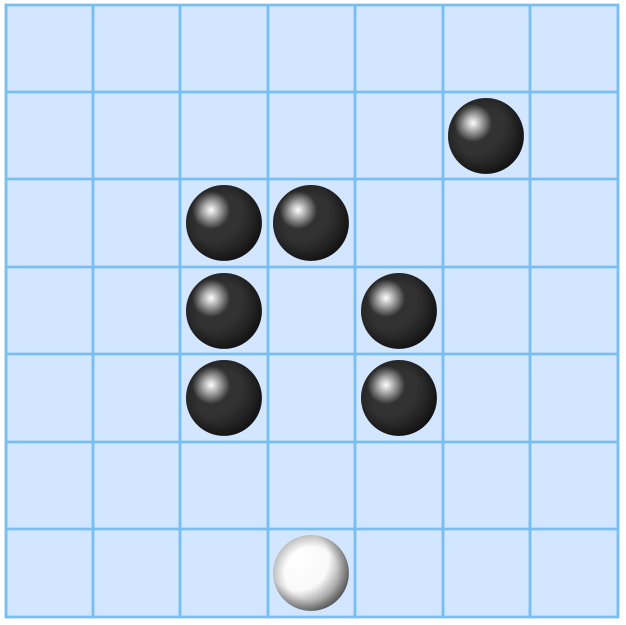}}
\end{center} \\ 

A piece can jump an orthogonal enemy piece to an empty site at a maximum distance of 2 after the jumped sites in capturing the jumped piece.

 \begin{boxex}
\begin{ludii}
(move Hop 
    Orthogonal
    (between 
        after:2
        if:(is Enemy (who at:(between))) 
        (apply (remove (between)))
    )
    (to if:(is Empty (to)))
)
\end{ludii} 
\end{boxex} 
&  
\begin{center}
\raisebox{-.50\height}{\includegraphics[width=0.8\linewidth]{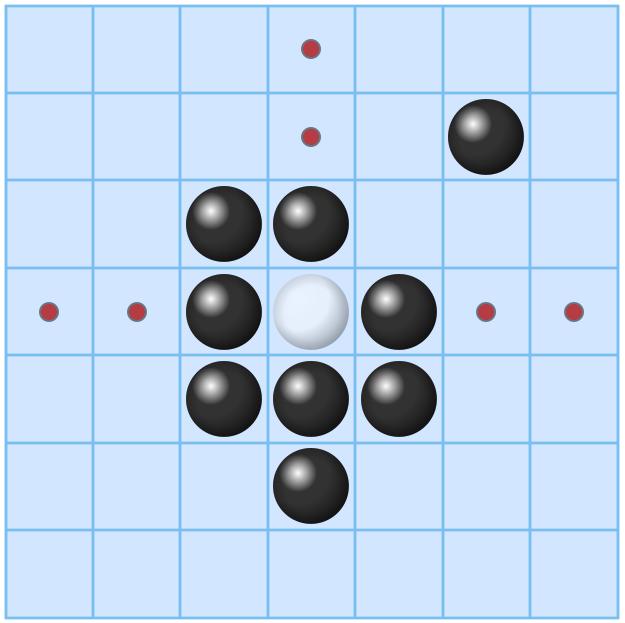}}
\end{center}
\begin{center}
\raisebox{-.50\height}{\includegraphics[width=0.8\linewidth]{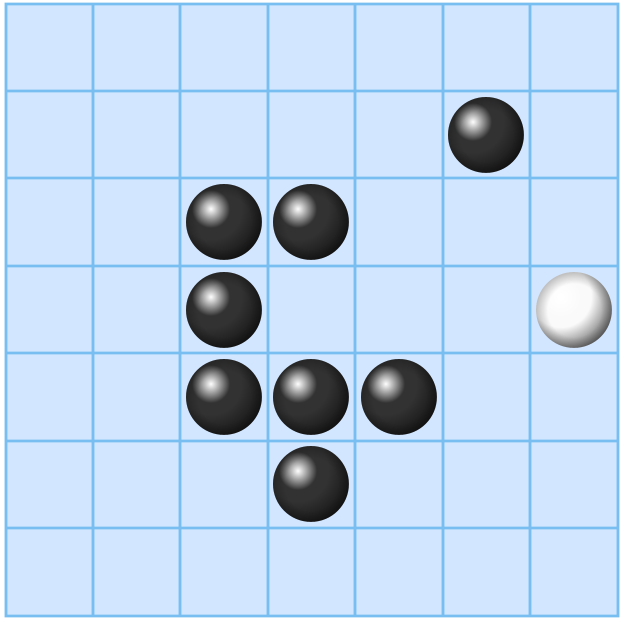}}
\end{center} \\ 

A piece can jump a diagonal enemy piece to an empty site at a maximum distance of 2 before the jumped sites in capturing the jumped pieces.

 \begin{boxex}
\begin{ludii}
(move Hop 
    Diagonal
    (between 
        before:2
        if:(is Enemy (who at:(between))) 
        (apply (remove (between)))
    )
    (to if:(is Empty (to)))
)
\end{ludii} 
\end{boxex} 
&  
\begin{center}
\raisebox{-.50\height}{\includegraphics[width=0.8\linewidth]{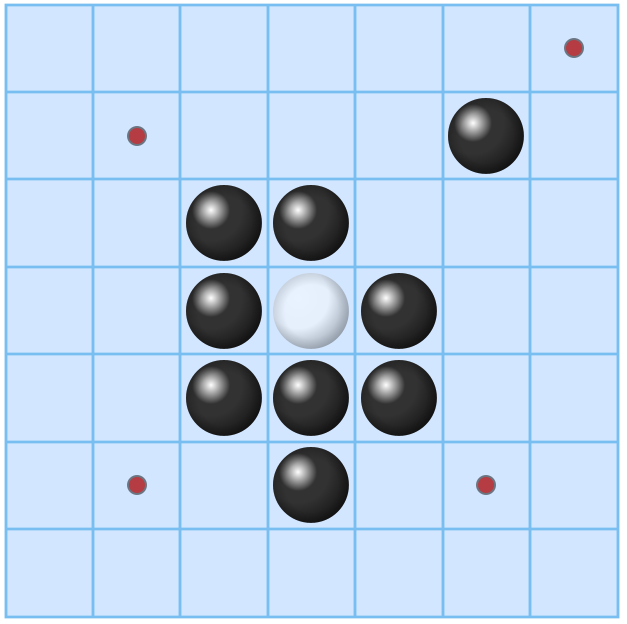}}
\end{center}
\begin{center}
\raisebox{-.50\height}{\includegraphics[width=0.8\linewidth]{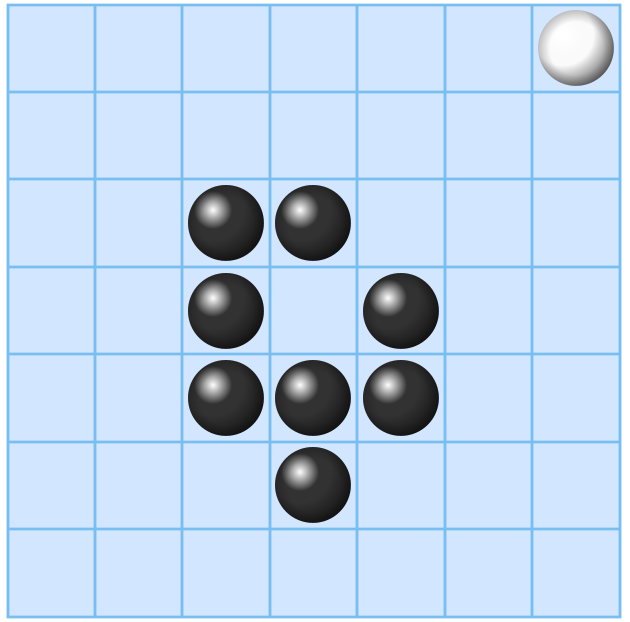}}
\end{center} \\ 

\end{longtable}


\subsubsection{Intervene}

The ludeme {\it Intervene} is generally used as a consequence after moving or placing a piece to capture pieces flanking the target site (e.g. Mak-yek\footnote{Mak-yek: \href{https://en.wikipedia.org/wiki/Mak-yek}{Wikipedia}}).
The main parameters of that ludeme are:
\begin{itemize}
    \item (from ...) describes the possible flanking site (default: (from (last To))).
    \item (to ...) describes the data on the flanking sites. The (to) ludeme is used to iterate each potential flanking site. The condition is described with the parameter "if:" and the effect to apply described inside the ludeme (apply ...).
\end{itemize}

Here is a example of how it's used in a hex board of size 3 using the cells by default:
\renewcommand{\arraystretch}{1.3}
\arrayrulecolor{white}
\begin{longtable}{@{}M{.75\textwidth} | M{.25\textwidth}@{}}
\rowcolor{gray!50}
\textbf{Description} & \textbf{Before/After}\\

After stepping a piece, the enemy pieces flanking the moved piece are removed.

 \begin{boxexnonbreakable}
\begin{ludii}
(move Step 
    (to if:(is Empty (to)))
    (then 
        (intervene 
            (from (last To))
            (to 
                if:(is Enemy (who at:(to)))
                (apply (remove (to)))
            )
        )
    )
)
\end{ludii} 
\end{boxexnonbreakable} 
&  
\begin{center}
\raisebox{-.50\height}{\includegraphics[width=0.8\linewidth]{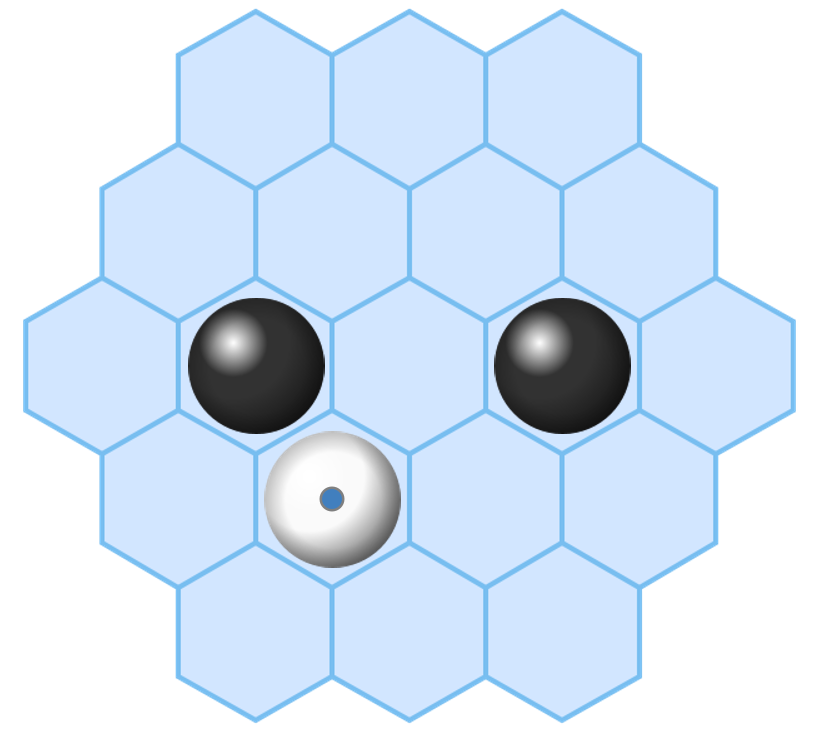}}
\end{center}
\begin{center}
\raisebox{-.50\height}{\includegraphics[width=0.8\linewidth]{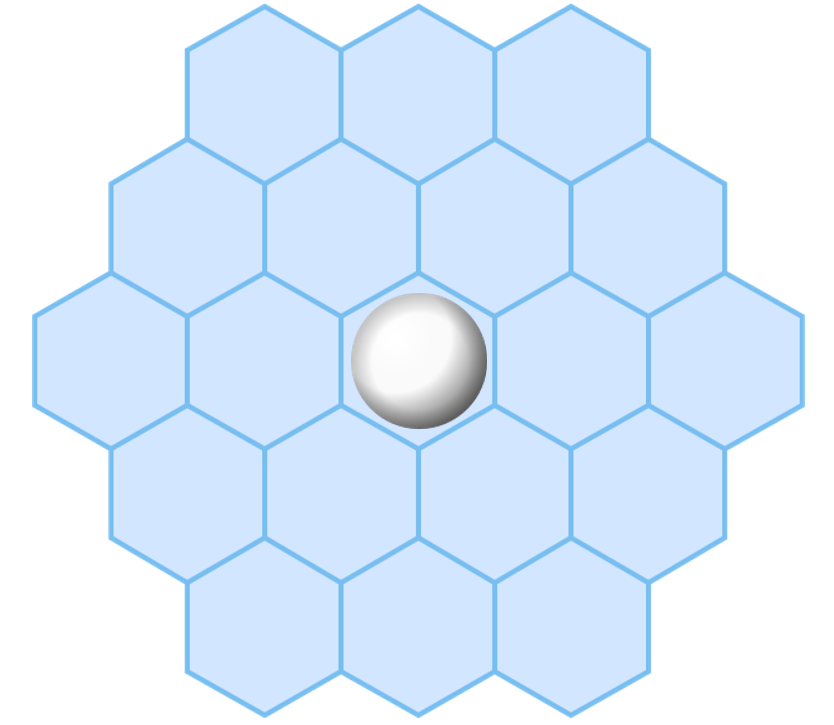}}
\end{center} \\ 
\end{longtable}


\subsubsection{Leap}

The ludeme {\it Leap} is used to jump sites by following a Turtle graphics walk (see \href{https://en.wikipedia.org/wiki/Turtle_graphics}{Wikipedia}) like the knight in Chess\footnote{Chess: \href{https://en.wikipedia.org/wiki/Chess}{Wikipedia}} to move or capture pieces by replacement. It is generally used in the generator of moves of the pieces.
The main parameters of that ludeme are:
\begin{itemize}
    \item (from ...) describes the origin site of the move (default: (from)).
    \item The walk(s) to follow described between braces $\{$ and $\}$.
    \item (to ...) describes the date on the sites to move. The (to) ludeme is used to iterate each potential site to move. The condition is described with the parameter "if:" and the effect to apply described inside the ludeme (apply ...).
\end{itemize}

By default, all the orthogonal rotations of the walk will be used to compute the legal moves. However, that's possible to filter them in keeping only the forwards moves (according to the facing direction of the piece) thanks to the parameter "forward:true" and that's also possible to compute the legals moves with not all the rotations but in using only the primary direction (the north for a square board) with the parameter "rotations:false".

Here are some examples of how it's used in a square board of size 7 using the cells by default:

\renewcommand{\arraystretch}{1.3}
\arrayrulecolor{white}
\begin{longtable}{@{}M{.75\textwidth} | M{.25\textwidth}@{}}
\rowcolor{gray!50}
\textbf{Description} & \textbf{Before/After}\\

A knight piece can leap as a chess knight to move to an empty site or to capture an enemy piece.

 \begin{boxexnonbreakable}
\begin{ludii}
(move Leap 
    { {F F R F} {F F L F} }
    (to 
        if:(not (is Friend (who at:(to))))
        (apply 
            (if (is Enemy (who at:(to)))
                (remove (to) )
            )
        ) 
    ) 
)
\end{ludii}
\end{boxexnonbreakable} 
&  
\begin{center}
\raisebox{-.50\height}{\includegraphics[width=0.8\linewidth]{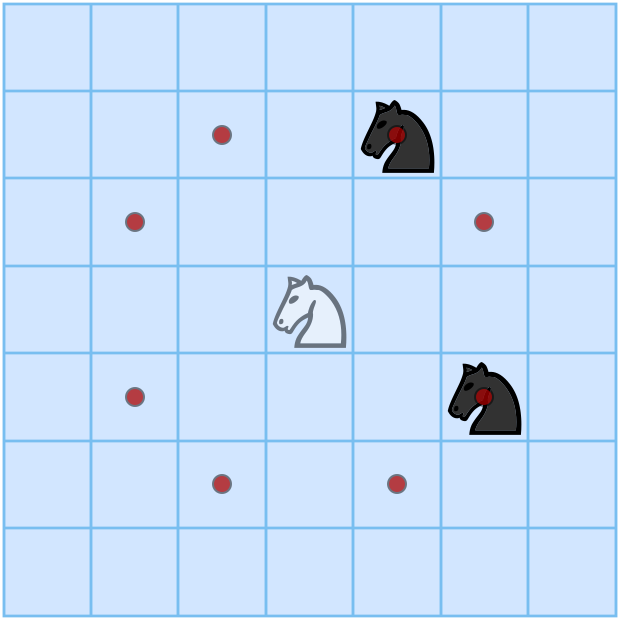}}
\end{center}
\begin{center}
\raisebox{-.50\height}{\includegraphics[width=0.8\linewidth]{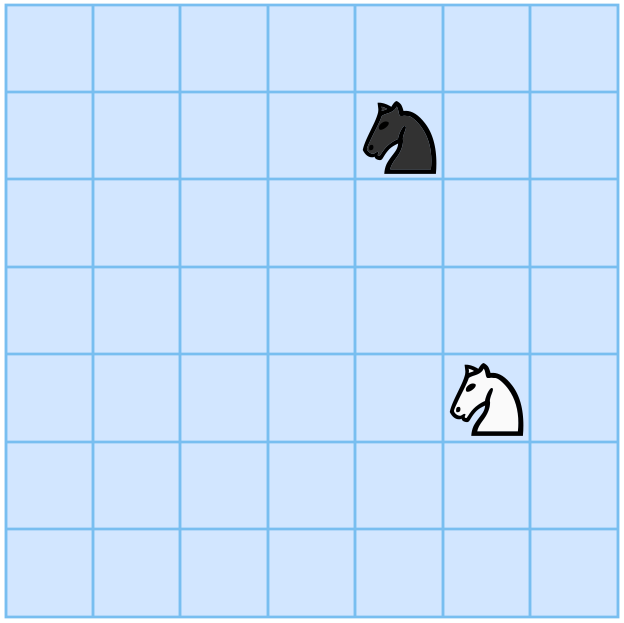}}
\end{center} \\ 

A knight piece facing to the north can leap as a large L only in the forward directions to move to an empty site or to capture an enemy piece.

 \begin{boxexnonbreakable}
\begin{ludii}
(move Leap 
    { {F F F R F} {F F F L F} }
    forward:true
    (to 
        if:(not (is Friend (who at:(to)))) 
        (apply 
            (if (is Enemy (who at:(to)))
                (remove (to) )
            )
        ) 
    ) 
)
\end{ludii} 
\end{boxexnonbreakable} 
&  
\begin{center}
\raisebox{-.50\height}{\includegraphics[width=0.8\linewidth]{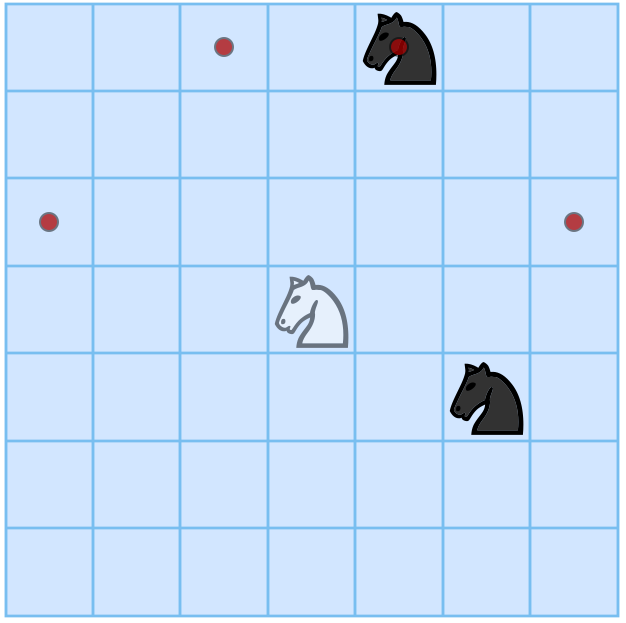}}
\end{center}
\begin{center}
\raisebox{-.50\height}{\includegraphics[width=0.8\linewidth]{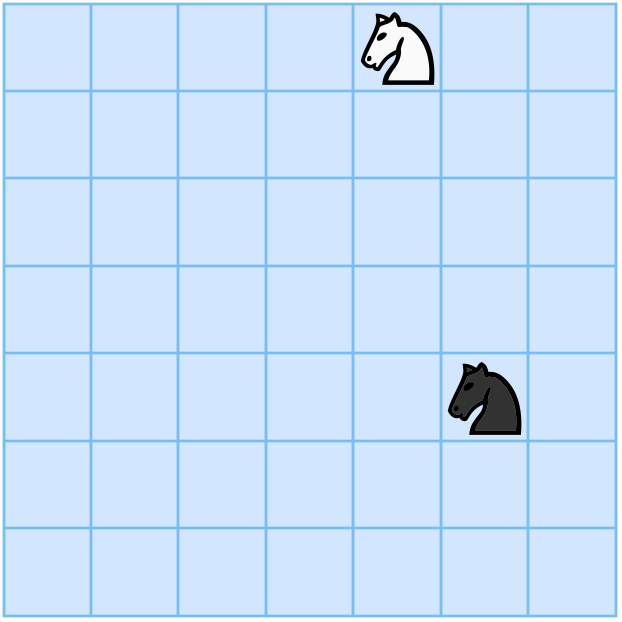}}
\end{center} \\ 

A knight piece can leap with a specific walk (not all the rotations) to move to an empty site or to capture an enemy piece.

 \begin{boxexnonbreakable}
\begin{ludii}
(move Leap 
    { {F F F R F} }
    rotations:false
    (to 
        if:(not (is Friend (who at:(to))))
        (apply 
            (if (is Enemy (who at:(to)))
                (remove (to) )
            )
        ) 
    ) 
)
\end{ludii} 
\end{boxexnonbreakable} 
&  
\begin{center}
\raisebox{-.50\height}{\includegraphics[width=0.8\linewidth]{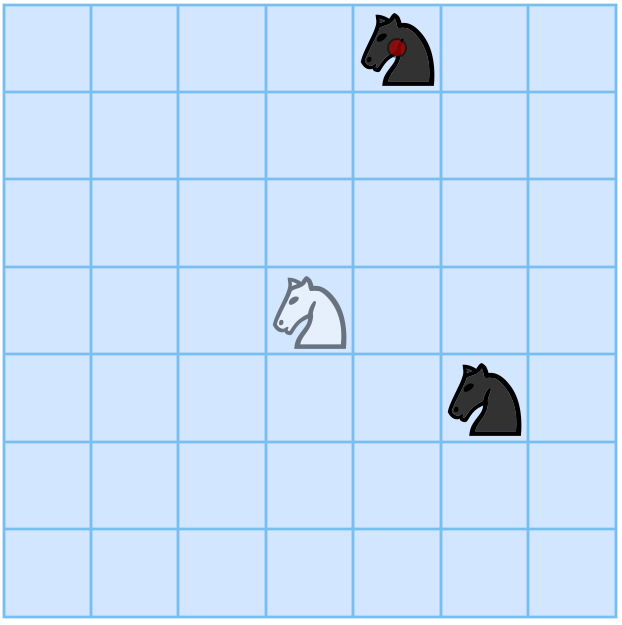}}
\end{center}
\begin{center}
\raisebox{-.50\height}{\includegraphics[width=0.8\linewidth]{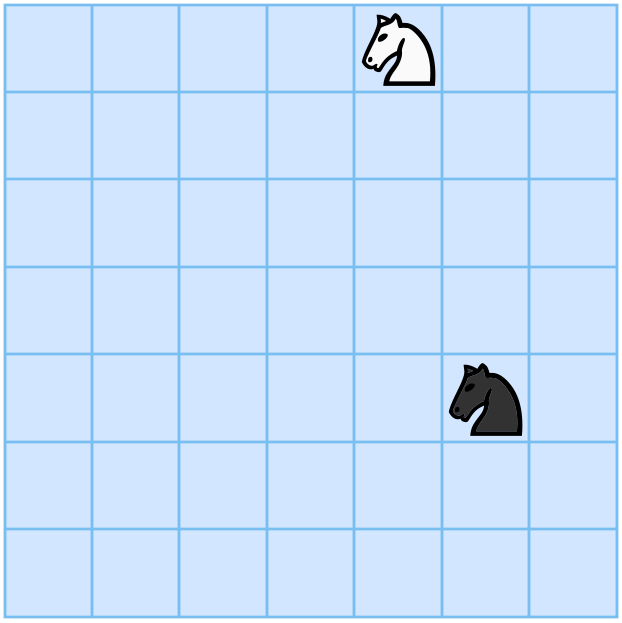}}
\end{center} \\ 
\end{longtable}


\subsubsection{Promote}

The ludeme {\it Promote} uses to convert a piece to another can be used as a consequence after moving or placing a piece or as a decision move (e.g. Chess\footnote{Chess: \href{https://en.wikipedia.org/wiki/Chess}{Wikipedia}}).
The main parameters of that ludeme are:
\begin{itemize}
    \item The site where is the piece to promote.
    \item The list of possible promotion piece types described in a (piece ...) ludeme.
    \item The owner of these piece types.
\end{itemize}

Here is a example of how it's used in a square board of size 3 using the cells by default:
\renewcommand{\arraystretch}{1.3}
\arrayrulecolor{white}
\begin{longtable}{@{}M{.75\textwidth} | M{.25\textwidth}@{}}
\rowcolor{gray!50}
\textbf{Description} & \textbf{Before/After}\\

A Pawn piece can promote to a queen, a knight, a bishop or a rook of the current player.

 \begin{boxex}
\begin{ludii}
(move Promote 
    (where (id "Pawn" Mover)) 
    (piece {"Queen" "Knight" "Bishop" "Rook"}) 
    Mover
)
\end{ludii} 
\end{boxex} 
&  
\begin{center}
\raisebox{-.50\height}{\includegraphics[width=0.8\linewidth]{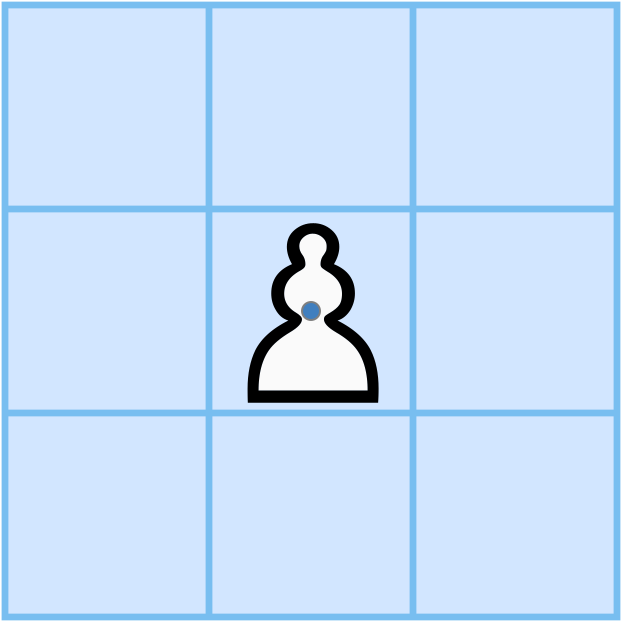}}
\end{center}
\begin{center}
\raisebox{-.50\height}{\includegraphics[width=0.8\linewidth]{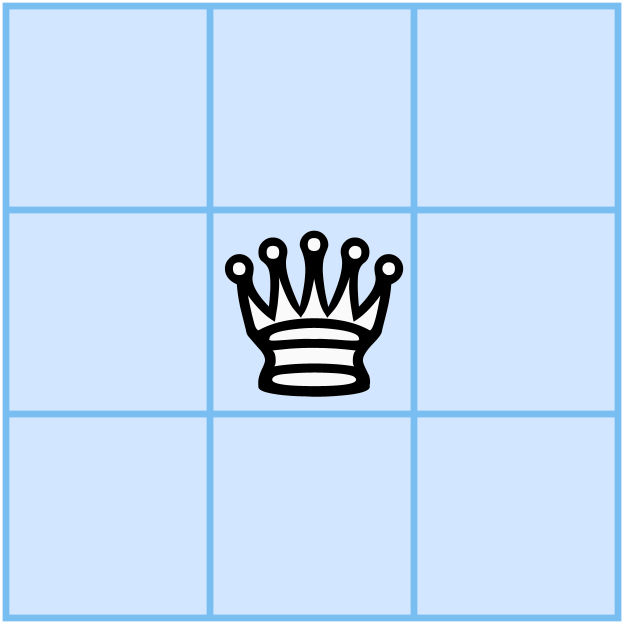}}
\end{center} \\ 
\end{longtable}


\subsubsection{Push}

The ludeme {\it Push} is generally used as a consequence to push a set of pieces in a specific direction.
The main parameters of that ludeme are:
\begin{itemize}
    \item (from ...) describes the origin site of the move (default (last To)).
    \item The direction to push.
\end{itemize}

Here is a example of how it's used in a square board of size 5 using the cells by default:
\renewcommand{\arraystretch}{1.3}
\arrayrulecolor{white}
\begin{longtable}{@{}M{.75\textwidth} | M{.25\textwidth}@{}}
\rowcolor{gray!50}
\textbf{Description} & \textbf{Before/After}\\

A King piece is selected and pushes all the pieces one step to the north.

 \begin{boxex}
\begin{ludii}
(move
    Select
    (from 
        (sites Occupied by:Mover 
            component:"King"
        )
    )
    (then 
        (push (from (last To)) N)
    )
)
\end{ludii} 
\end{boxex} 
&  
\begin{center}
\raisebox{-.50\height}{\includegraphics[width=0.8\linewidth]{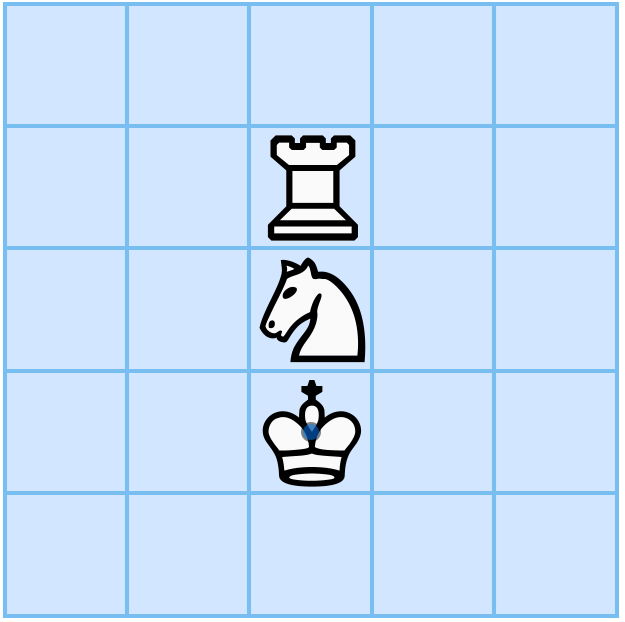}}
\end{center}
\begin{center}
\raisebox{-.50\height}{\includegraphics[width=0.8\linewidth]{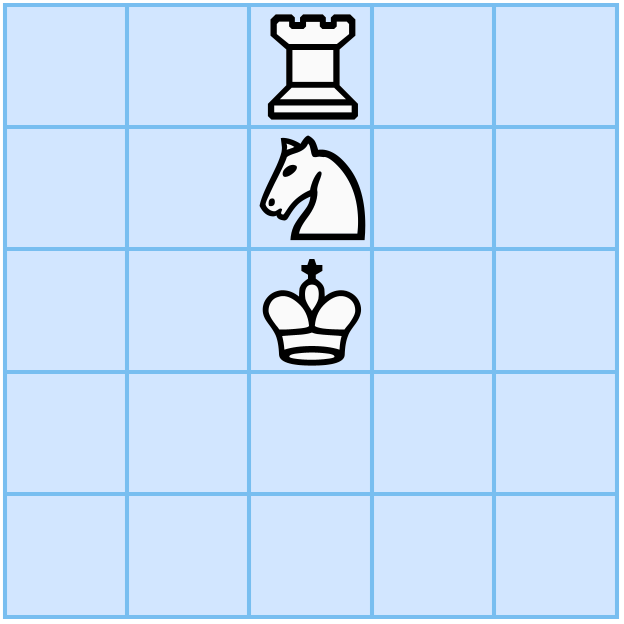}}
\end{center} \\ 
\end{longtable}


\subsubsection{Remove}

The ludeme {\it Remove} used to remove one or many piece(s) from the board can be used as an effect or a consequence like in different examples in that document, but can also be used as a decision move.

The main parameters of that ludeme are:
\begin{itemize}
    \item The site(s) to remove the piece(s) on. 
    \item count: describes the number of pieces to remove.
\end{itemize}

Here is a example of how it's used in a square board of size 5 using the cells by default:
\renewcommand{\arraystretch}{1.3}
\arrayrulecolor{white}
\begin{longtable}{@{}M{.75\textwidth} | M{.25\textwidth}@{}}
\rowcolor{gray!50}
\textbf{Description} & \textbf{Before/After}\\

Any piece owned by the current player can be removed from the board.

 \begin{boxex}
\begin{ludii}
(move Remove
    (sites Occupied by:Mover)
)
\end{ludii} 
\end{boxex} 
&  
\begin{center}
\raisebox{-.50\height}{\includegraphics[width=0.8\linewidth]{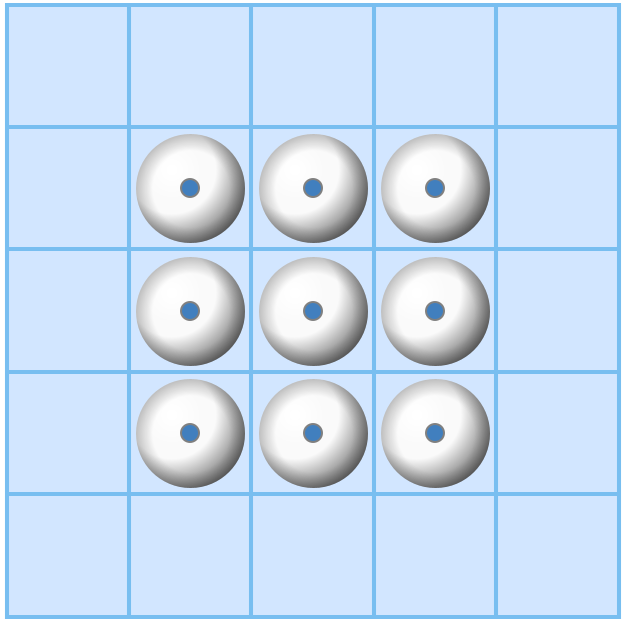}}
\end{center}
\begin{center}
\raisebox{-.50\height}{\includegraphics[width=0.8\linewidth]{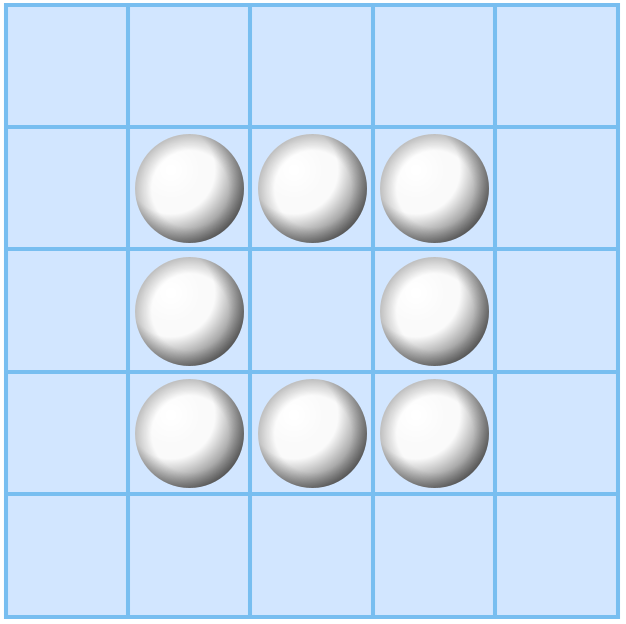}}
\end{center} \\ 
\end{longtable}


\subsubsection{Shoot}

The ludeme {\it Shoot} is used to place a piece in any site in the line of sight of another piece (e.g. Amazons\footnote{Amazons: \href{https://en.wikipedia.org/wiki/Game_of_the_Amazons}{Wikipedia}}). It is generally used has a decision.
The main parameters of that ludeme are:
\begin{itemize}
    \item (piece ...) describes the piece to shoot.
    \item (from ...) describes the origin site of the move (default (last To)).
    \item The directions allowed to shoot (default: Adjacent).
    \item (to ...) describes the date of the site to shoot. The (to) ludeme is used to iterate each potential target site. The condition is described with the parameter "if:". In Each direction, the sites are checked, if the condition is false, the piece can not be shoot in that site and farther in the same direction (default (to if:(is Empty (to)))). 
\end{itemize}

Here are some examples of how it's used in a square board of size 5 using the cells by default:
\renewcommand{\arraystretch}{1.3}
\arrayrulecolor{white}
\begin{longtable}{@{}M{.75\textwidth} | M{.25\textwidth}@{}}
\rowcolor{gray!50}
\textbf{Description} & \textbf{Before/After}\\

A ball is shot from the piece placed in the centre to any sites in any adjacent direction.

 \begin{boxex}
\begin{ludii}
(move Shoot 
    (piece "Ball")
    (from (centrePoint))
)
\end{ludii} 
\end{boxex} 
&  
\begin{center}
\raisebox{-.50\height}{\includegraphics[width=0.8\linewidth]{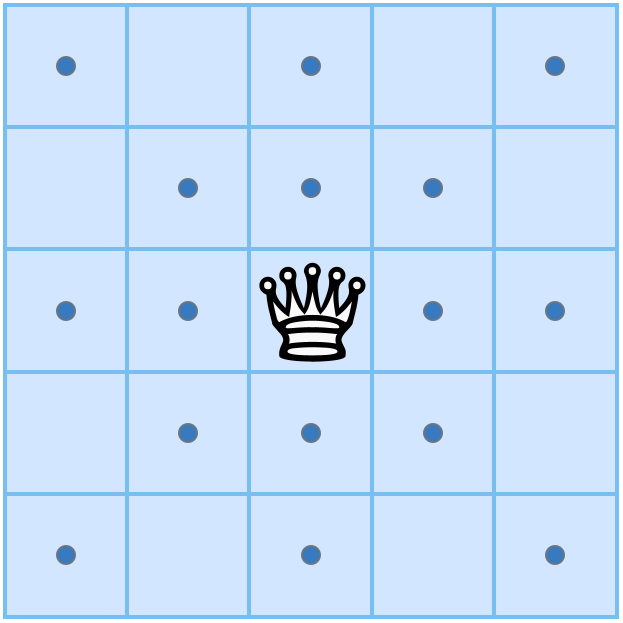}}
\end{center}
\begin{center}
\raisebox{-.50\height}{\includegraphics[width=0.8\linewidth]{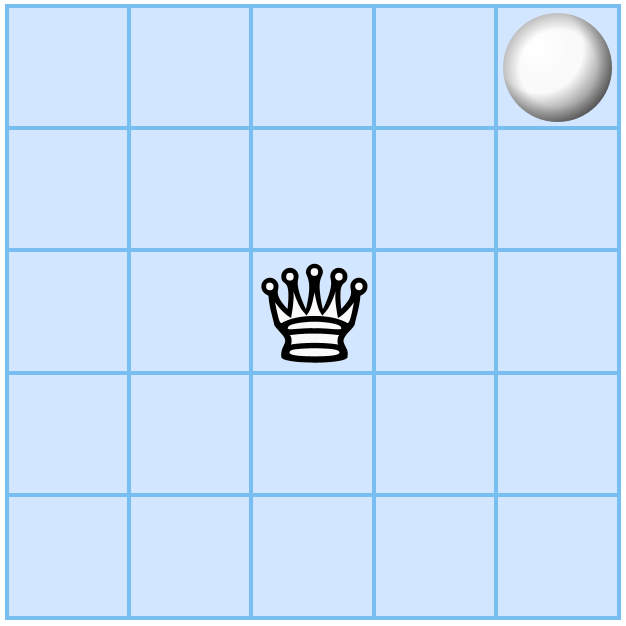}}
\end{center} \\ 

A ball is shot from the piece placed in the centre to any sites in any orthogonal direction.

 \begin{boxex}
\begin{ludii}
(move Shoot 
    (piece "Ball")
    (from (centrePoint))
    Orthogonal
    (to if:(is Empty (to)))
)
\end{ludii} 
\end{boxex} 
&  
\begin{center}
\raisebox{-.50\height}{\includegraphics[width=0.8\linewidth]{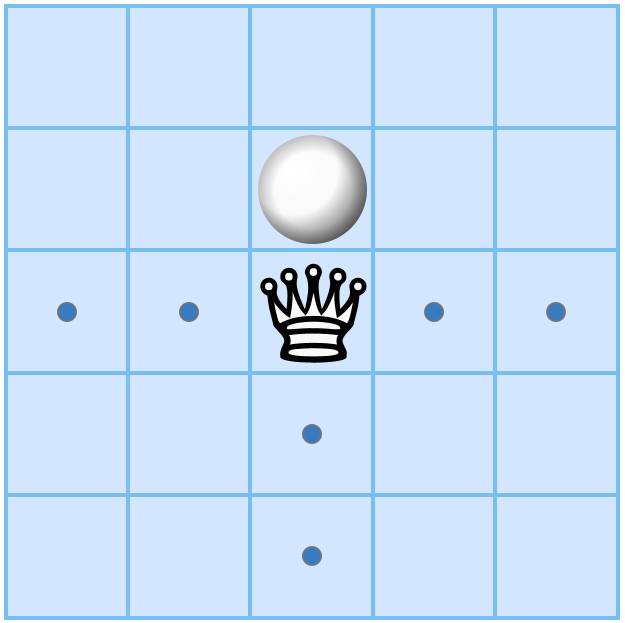}}
\end{center}
\begin{center}
\raisebox{-.50\height}{\includegraphics[width=0.8\linewidth]{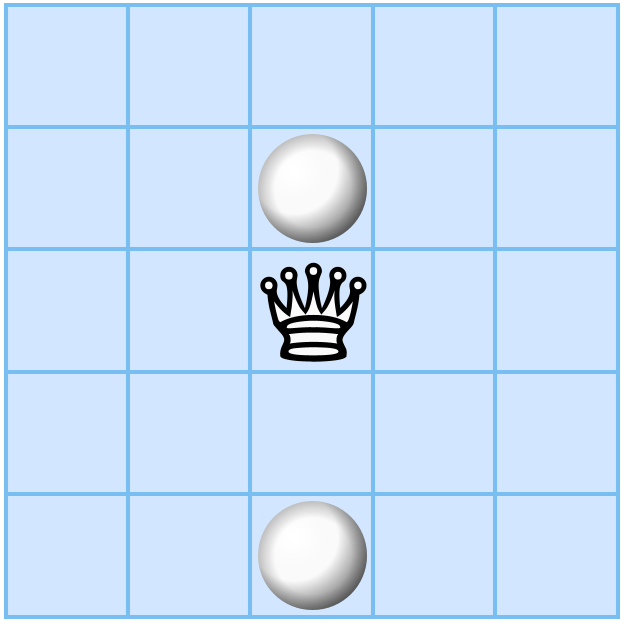}}
\end{center} \\ 
\end{longtable}


\subsubsection{Slide}

The ludeme {\it Slide} is used to move a piece to any site in different directions and potentially to capture a piece by replacement (e.g. Chess\footnote{Chess: \href{https://en.wikipedia.org/wiki/Chess}{Wikipedia}}). We generally use it in the generator of moves of a component.

The main parameters of that ludeme are:
\begin{itemize}
    \item (from ...) describes the origin site of the move (default (from)).
    \item The directions to slide (default: Adjacent).
    \item (between ...) describes the data about the sites between the (from) site and the (to) site. The ludeme (between) is used to iterate these sites. The condition is described with the parameter "if:". In Each direction, the sites are checked, if the condition is false, the piece can not be slide in that site and farther in the same direction. The moving piece can trail another piece in each site, that can be described with the parameter "trail:".
    \item (to ...) describes the data about the sites to stop to slide. The (to) ludeme is used to iterate these sites. The condition is described with the parameter "if:" and the effect to apply described inside the ludeme (apply ...). 
\end{itemize}

Here are some examples of how it's used in a square board of size 5 using the cells by default:
\renewcommand{\arraystretch}{1.3}
\arrayrulecolor{white}
\begin{longtable}{@{}M{.75\textwidth} | M{.25\textwidth}@{}}
\rowcolor{gray!50}
\textbf{Description} & \textbf{Before/After}\\

A queen piece can slide in each empty site in all the directions.

 \begin{boxex}
\begin{ludii}
(move Slide)
\end{ludii} 
\end{boxex} 
&  
\begin{center}
\raisebox{-.50\height}{\includegraphics[width=0.8\linewidth]{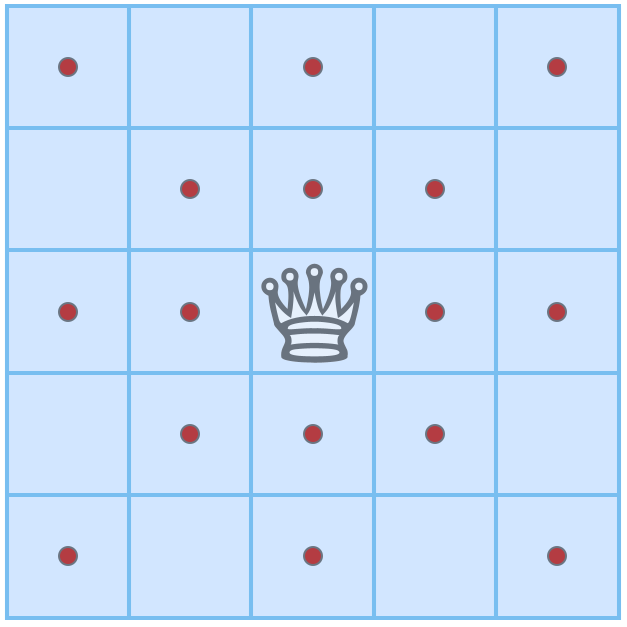}}
\end{center}
\begin{center}
\raisebox{-.50\height}{\includegraphics[width=0.8\linewidth]{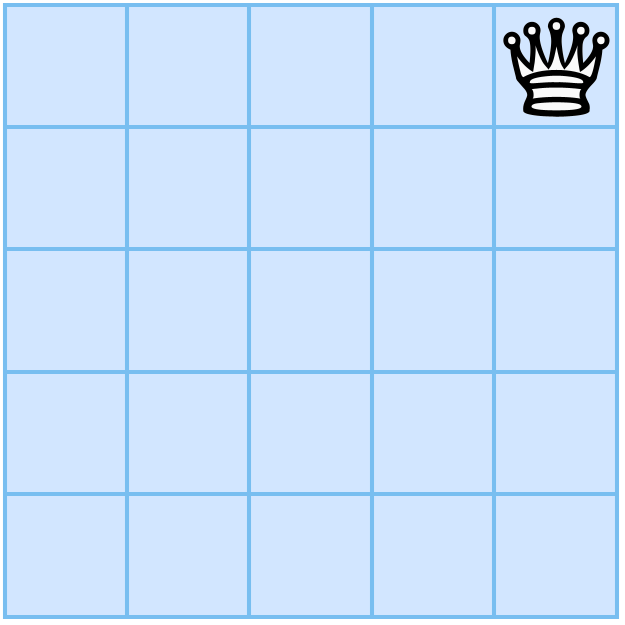}}
\end{center} \\ 

A queen piece can slide in each empty site in all the directions and trail a "Disc" piece.

 \begin{boxex}
\begin{ludii}
(move Slide
    (between
        trail:(id "Disc")
    )
)
\end{ludii} 
\end{boxex} 
&  
\begin{center}
\raisebox{-.50\height}{\includegraphics[width=0.8\linewidth]{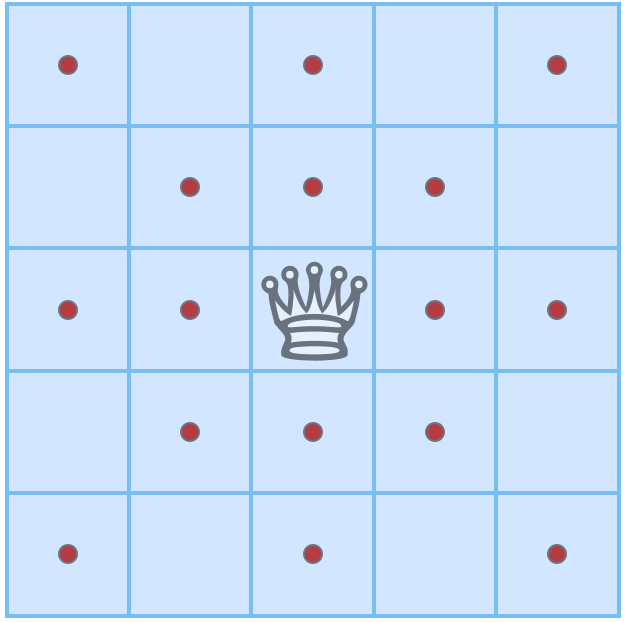}}
\end{center}
\begin{center}
\raisebox{-.50\height}{\includegraphics[width=0.8\linewidth]{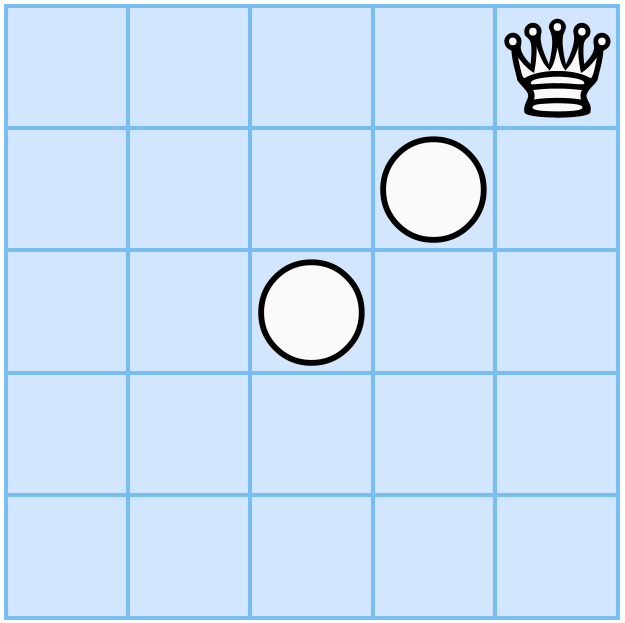}}
\end{center} \\ 

A rook piece can slide in each empty site in all the orthogonal directions and possibly capture an enemy piece.

 \begin{boxex}
\begin{ludii}
(move Slide
    Orthogonal
    (to if:(is Enemy (who at:(to)))
        (apply (remove (to)))
    )
)
\end{ludii} 
\end{boxex} 
&  
\begin{center}
\raisebox{-.50\height}{\includegraphics[width=0.8\linewidth]{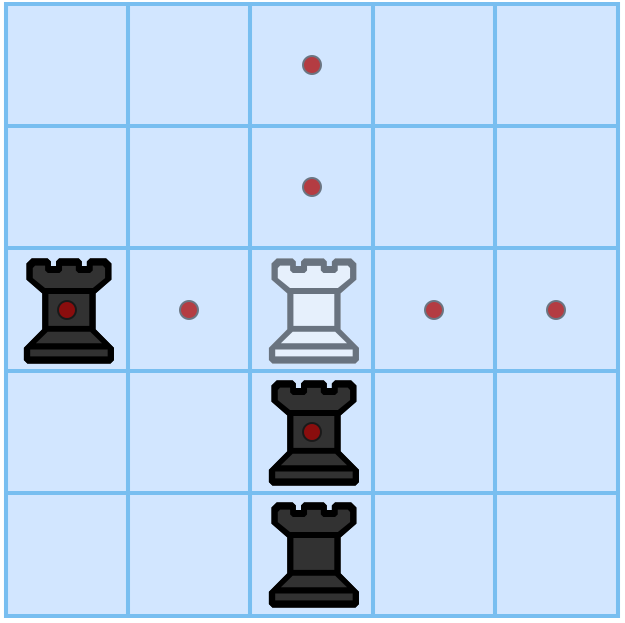}}
\end{center}
\begin{center}
\raisebox{-.50\height}{\includegraphics[width=0.8\linewidth]{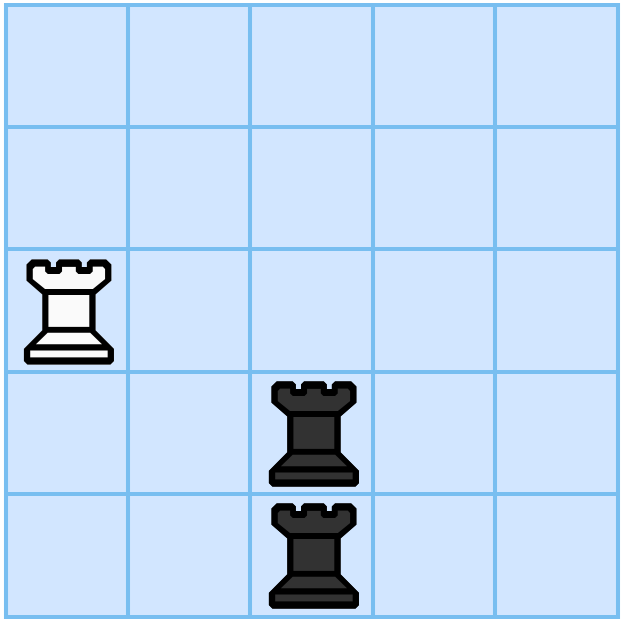}}
\end{center} \\ 

\end{longtable}


\subsubsection{Sow}

The ludeme {\it Sow} is used has a consequence after selected a site to sow a certain number of pieces in following a track (e.g. mancala games\footnote{mancala games: \href{https://en.wikipedia.org/wiki/Mancala}{Wikipedia}}).
The main parameters of that ludeme are:
\begin{itemize}
    \item count: describes the number of pieces to sow (default: (count at:(last To))).
    \item numPerHole: describes the number of pieces to sow in each hole (default: 1).
    \item if: describes the condition on the last site of the sowing to apply an effect on it. That site is described by the ludeme (to).
    \item sowEffect: describes the effect to apply to each hole sowed.
    \item apply: describes the effect to apply to the last site of the sowing. That site is described by the ludeme (to).
    \item includeSelf: if true, the sowing includes the origin site in the sowing (default: true).
    \item origin: if true, a piece is sowing to the origin site before to begin the sowing (default: false).
    \item skipIf: describes a condition to skip to sow to a site. That site is described by the ludeme (to).
    \item backtracking: if true and if the condition to apply the effect in the last site is true, the condition and the effect can also be applied to the previous sites in the sowing until the condition is false (default: false).
    \item forward: if true and if the condition to apply the effect in the last site is true, the condition and the effect can also be applied to the next sites in the sowing until the condition is false (default: false).
\end{itemize}

Here is a example of how it's used in a mancala board of 2 rows and 6 columns with stores:
\renewcommand{\arraystretch}{1.3}
\arrayrulecolor{white}
\begin{longtable}{@{}M{.62\textwidth} | M{.38\textwidth}@{}}
\rowcolor{gray!50}
\textbf{Description} & \textbf{Before/After}\\

Any site in the track with at least a piece can be selected then the pieces in it are sowing in following a counter clockwise direction. If after sowing the site has 4 pieces, the pieces are captured to the site mapped to the mover.

 \begin{boxexnonbreakable}
\begin{ludii}
(equipment {
    (mancalaBoard 2 6
        (track "Track" "1,E,N,W" 
            loop:true
        )
    )
    (piece "Seed" Shared)
    (map {
        (pair P1 FirstSite) 
        (pair P2 LastSite)
    })
})

(move Select 
    (from 
        (sites Track) 
        if:(> (count at:(from)) 0)
    )
    (then 
        (sow
            if:(= (count at:(to)) 4)
            apply:(fromTo 
                (from (to)) 
                (to (mapEntry (mover))) 
                count:(count at:(to))
            ) 
        )
    )
)
\end{ludii} 
\end{boxexnonbreakable} 
&  
\begin{center}
\raisebox{-.50\height}{\includegraphics[width=0.9\linewidth]{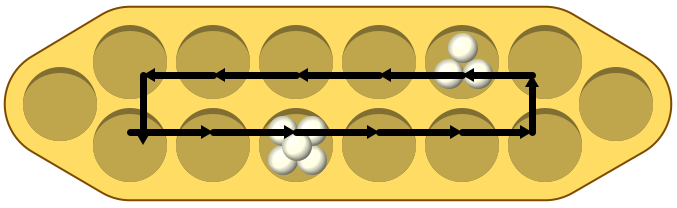}}
\end{center}
~ \newline ~ \newline ~ \newline ~ \newline
\begin{center}
\raisebox{-.50\height}{\includegraphics[width=0.9\linewidth]{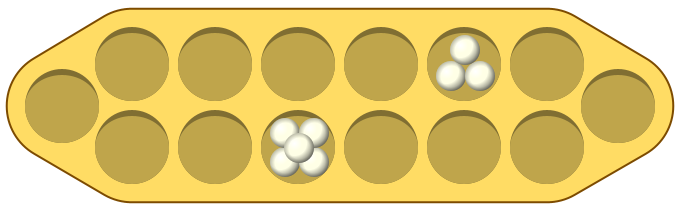}}
\end{center}
~ \newline ~ \newline ~ \newline ~ \newline
\begin{center}
\raisebox{-.50\height}{\includegraphics[width=0.9\linewidth]{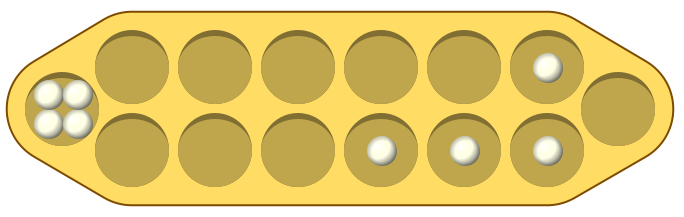}}
\end{center} \\ 
\end{longtable}


\subsubsection{Step}

The ludeme {\it Step} can be used to step a piece from a site to another connected site like the king in Chess\footnote{Chess: \href{https://en.wikipedia.org/wiki/Chess}{Wikipedia}} to potentially capture a piece by replacement. It is generally used in the generator of the pieces.
The main parameters of that ludeme are:
\begin{itemize}
    \item (from ...) describes the origin site (default: (from)).
    \item The directions allowed to step (default: Adjacent).
    \item (to ...) describes the data about the sites to step. The (to) ludeme is used to iterate each site to move. The condition is described with the parameter "if:" and the effect to apply described inside the ludeme (apply ...). 
\end{itemize}

~ \newline

Here are some examples of how it's used in a square board of size 5 using the cells by default:
\renewcommand{\arraystretch}{1.3}
\arrayrulecolor{white}
\begin{longtable}{@{}M{.70\textwidth} | M{.30\textwidth}@{}}
\rowcolor{gray!50}
\textbf{Description} & \textbf{Before/After}\\

A king piece can step to an adjacent empty site or to a site occupied by an enemy to capture it.

\begin{boxex}
\begin{ludii}
(move Step
     (to if:(is Empty (to)))
)
\end{ludii}
\end{boxex} 
&  
\begin{center}
\raisebox{-.50\height}{\includegraphics[width=0.8\linewidth]{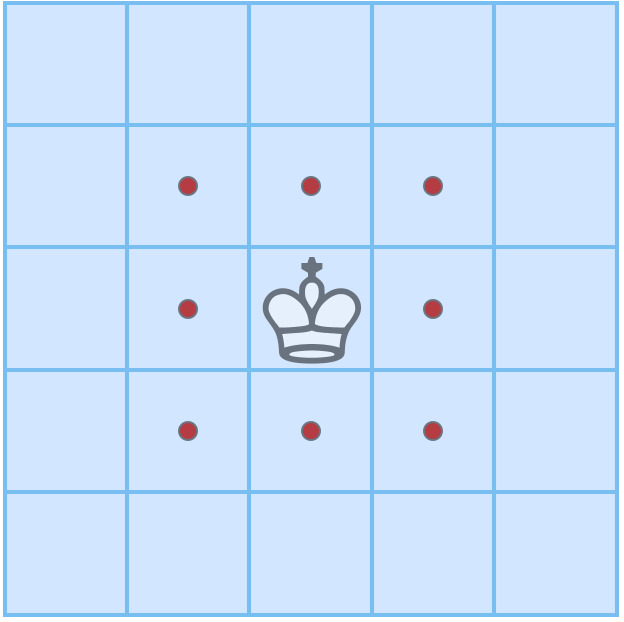}}
\end{center}
\begin{center}
\raisebox{-.50\height}{\includegraphics[width=0.8\linewidth]{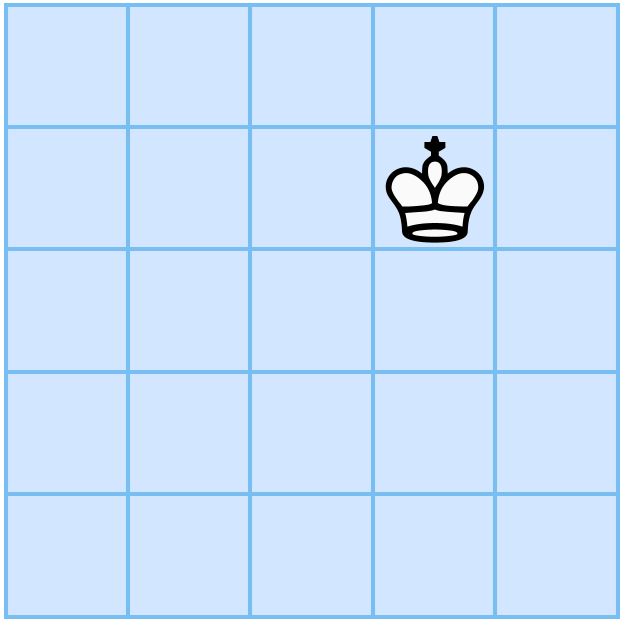}}
\end{center} \\ 

A king piece can step to an orthogonal empty site or to a site occupied by an enemy to capture it.

\begin{boxex}
\begin{ludii}
(move Step
     Orthogonal
     (to 
        if:(not (is Friend (who at:(to))))
        (apply 
            (if (is Enemy (who at:(to)))
                (remove (to))
            )
        )
     )
)
\end{ludii}
\end{boxex} 
&  
\begin{center}
\raisebox{-.50\height}{\includegraphics[width=0.8\linewidth]{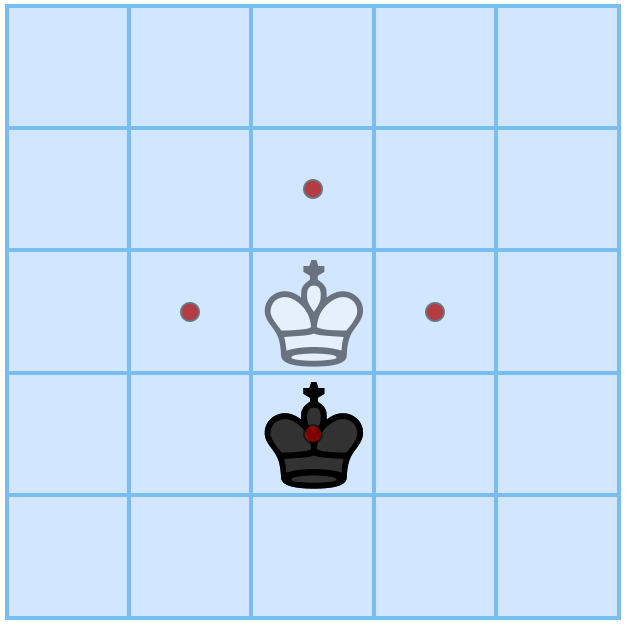}}
\end{center}
\begin{center}
\raisebox{-.50\height}{\includegraphics[width=0.8\linewidth]{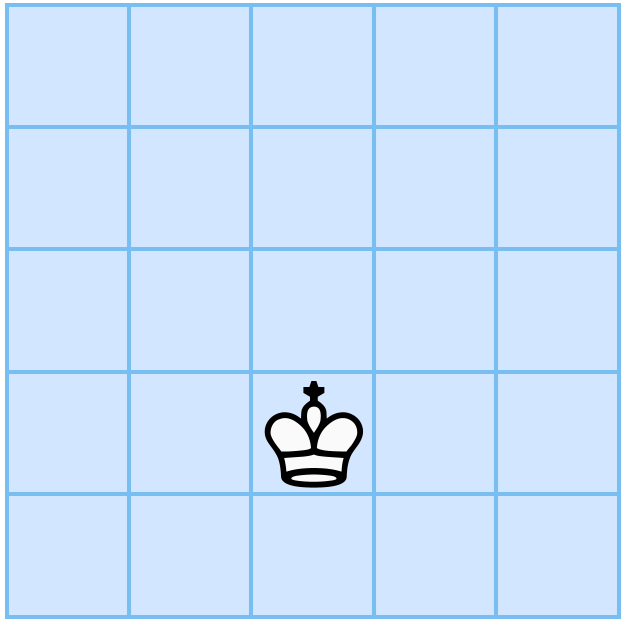}}
\end{center} \\ 
\end{longtable}


\subsubsection{Surround}

The ludeme {\it Surround} is generally used as a consequence to apply an effect on pieces surrounding after placing or moving a piece (e.g. Hnefatfl\footnote{Talf games: \href{https://en.wikipedia.org/wiki/Tafl_games}{Wikipedia}}).
The main parameters of that ludeme are:
\begin{itemize}
    \item (from ...) describes the site possibly causing the surrounding of sites (default: (from)).
    \item The directions used to surround (default: Adjacent).
    \item (between ...) describes the data about the potential surrounded sites. The (between) ludeme is used to iterate them. The condition is described with the parameter "if:" and the effect to apply described inside the ludeme (apply ...). 
    \item (to ...) describes the surrounding sites. The (to) ludeme is used to iterate them. The condition is described with the parameter "if:". 
\end{itemize}

~ \newline

Here is a example of how it's used in a square board of size 5 using the cells by default:
\renewcommand{\arraystretch}{1.3}
\arrayrulecolor{white}
\begin{longtable}{@{}M{.75\textwidth} | M{.25\textwidth}@{}}
\rowcolor{gray!50}
\textbf{Description} & \textbf{Before/After}\\

A thrall piece can step to an empty site, then if any enemy piece orthogonally surrounded by friends pieces is removed.

\begin{boxex}
\begin{ludii}
(move Step
    (to if:(is Empty (to)))
    (then
        (surround 
            (from (last To)) 
            Orthogonal 
            (between 
                if:(is Enemy (who at:(between))) 
                (apply (remove (between)))
            ) 
            (to 
                if:(is Friend (who at:(to)))
            )
        )
    )
)
\end{ludii}
\end{boxex} 
&  
\begin{center}
\raisebox{-.50\height}{\includegraphics[width=0.8\linewidth]{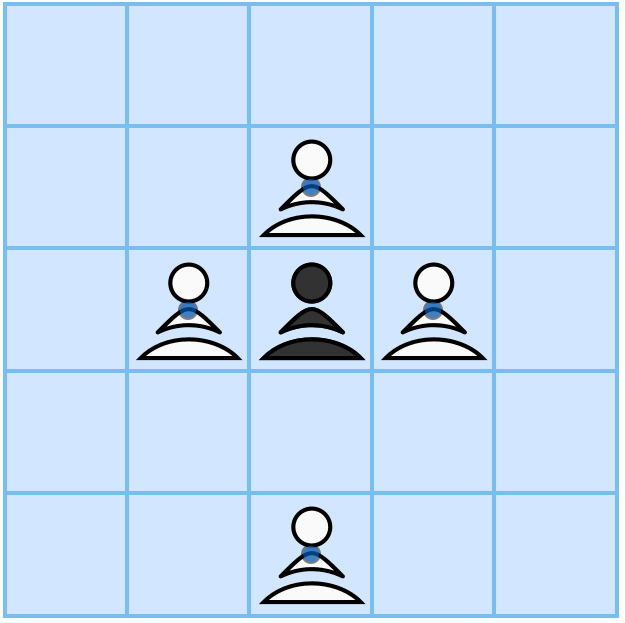}}
\end{center}
\begin{center}
\raisebox{-.50\height}{\includegraphics[width=0.8\linewidth]{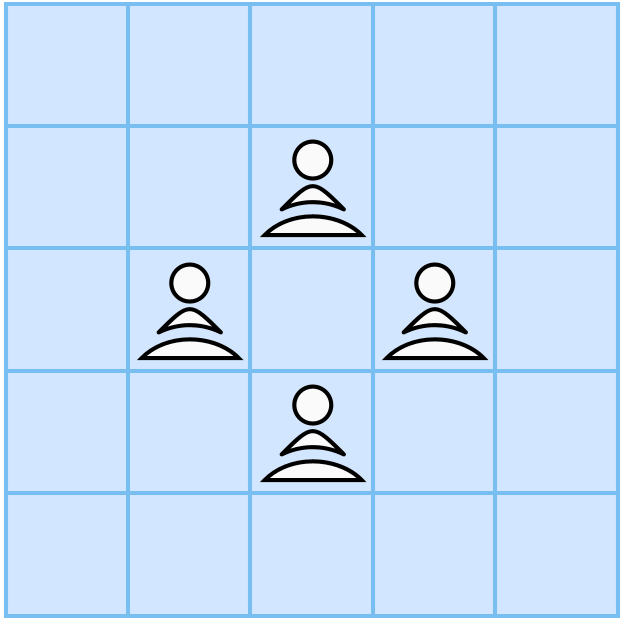}}
\end{center} \\ 

\end{longtable}


\section{End Rules}

The ending rules describe when and how play can terminate for one or more of the players. In Ludii, the ending rules start by condition(s) encapsulated in an (if ...) ludeme, followed by the description of the outcome of the player(s).
In games with 2 players or less, an (end ...) rule ludeme describes the end of the game, but for games with more players, the game can continue if play did not yet terminate for at least 2 of the players.

All the conditions are described with BooleanFunctions (see Section \ref{Chapter:BooleanFunctions}). The conditions can be nested within other (if ...).

Two ways are possible to describe the outcomes of the players:
\begin{itemize}
    \item (result ...): The result ludeme has 2 parameters:
    \begin{itemize}
    \item A roletype describing the player or the team assigned to that ending state.
    \item The ResultType, which can be Win (the player got the best available rank), Loss (the player got the worst available rank), Draw (all the remaining players get the middle of the available ranks). 
    \end{itemize}
    \item (byScore ...): To end the game and order the players according to their scores. Their final scores can be set in that ludeme too.
\end{itemize}

The following example gives the best available rank to the mover if a line of size 3 minimum is detected with the pieces of the mover.
\begin{boxex}
\begin{ludii}
(end 
    (if 
        (is Line 3) 
        (result Mover Win)
    )
)
\end{ludii}
\end{boxex} 

The following example if the player is P1, to get the best available rank it needs to make a line of size 3 minimum else it has to make a line of size 4 minimum.
\begin{boxex}
\begin{ludii}
(end 
    (if (is Mover P1)
        (if 
            (is Line 3) 
            (result P1 Win)
        )
        (if 
            (is Line 4) 
            (result P2 Win)
        )
    )
)
\end{ludii}
\end{boxex} 

The following example ends the game if all the player just passed in ordering them according to their final score. This score is set for each player in being their current score added to the size of their territory.
\begin{boxex}
\begin{ludii}
(end
    (if (all Passed)
        (byScore {
            (score P1 (+ (score P1) (size Territory P1))) 
            (score P2 (+ (score P2) (size Territory P2)))
        })
    )
)
\end{ludii}
\end{boxex} 

It is also possible to specify end results that trigger simultaneously for multiple, but not necessarily all, players. The following example causes all players who become blocked to simultaneously lose as soon as they become blocked, and causes a player to win when they complete a connected chain of pieces between their regions.
\begin{boxex}
\begin{ludii}
(end {
    (forEach NonMover if:(is Blocked Player) (result Player Loss))
    (if (is Connected Mover) (result Mover Win))
})
\end{ludii}
\end{boxex} 

\subsection{Player Rankings}

In every $n$-player game, there are $n$ ranks to be distributed among the players; ranks $1$, $2$, $\dots$, $n$. Low numbers are more desirable than high numbers (i.e., rank $1$ is always the best rank). 

Whenever a player achieves a ``Win'' result, that player is assigned the next best rank that has not already been assigned. For instance, in a $2$-player game where neither player has achieved an outcome yet, this will normally yield rank $1$ for the winner. In a game with $6$ players, for example, a ``Win'' result would yield a rank of $3$ if two other players have already previously obtained ``Win'' results (and hence already claimed ranks $1$ and $2$).

Similarly, whenever a player achieves a ``Loss'' result, that player is assigned the next worst rank that has not already been assigned. In a typical $2$-player game situation, that would simply be rank $2$. In a situation in a $5$-player game, where 2 other players have already previously lost (and hence claimed ranks $5$ and $4$), this would be rank $3$.

A ``Draw'' result always terminates the game, and assigns the average of all remaining, unassigned ranks to all players that are still active. In the simple case of a $2$-player game, a ``Draw'' result means that both players obtain a rank of $1.5$. Such non-integer ranks are unusual to see for humans, but convenient for AI implementations because it leads to a result that is naturally in between the the ``Win'' and ``Loss'' outcomes, and therefore also leads to correctly ordered preferences for different outcomes. In a $6$-player game where one player has already lost, and two players have already won, the remaining unassigned ranks would be $2$, $3$, and $4$. A ``Draw'' result in such a situation would yield a shared rank of $\frac{2 + 3 + 4}{3} = 3$ for each of the three remaining players.

When a ``Win'' (or ``Loss'') result triggers for $n > 1$ different players simultaneously, the average of the next $n$ best (or worst) ranks available is assigned to those $n$ players, and those $n$ ranks are counted as already having been assigned. For example, suppose that two players win simultaneously in a $4$-player game, with all ranks still left unassigned. The best two ranks available are $1$ and $2$, which means that these two players obtain a shared rank of $\frac{1 + 2}{2} = 1.5$. If at a later stage a third player also wins, ranks $1$ and $2$ both count as already having been assigned, which means that this third player achieves a rank of $3$ (and finally the last remaining player defaulting to rank $4$).


\chapter{\textcolor{gray!95}{Functions}}\label{Chapter:Functions}

Except the moves, each ludeme corresponds to a particular type of function.


\section{BooleanFunction} \label{Chapter:BooleanFunctions}

The Boolean functions return true or false. Except the math Boolean functions, all of them are starting by one of these keywords: all, can, is, no, was.

In this section, descriptions of the Boolean functions not specifically dedicated to deduction puzzles appear before the ones designed specifically for them.

\subsection{Math}

The mathematics Boolean functions are all the classical mathematics operators returning a Boolean:

\renewcommand{\arraystretch}{1.3}
\arrayrulecolor{white}
\rowcolors{2}{gray!25}{gray!10}
\begin{longtable}{@{}p{.13\textwidth} | p{.80\textwidth}@{}}
\rowcolor{gray!50}
\textbf{Boolean} & \textbf{Description} \\
(= ...) & Returns true if all the Integer functions return the same value. \\
(!= ...) & Returns true if all the Integer functions return a different value. \\
($<$ ...) & Returns true if the value returning by the first parameter is smaller than the value returning by the second parameter. \\
($<$= ...) & Returns true if the value returning by the first parameter is smaller or equal to the value returning by the second parameter. \\
($>$ ...) & Returns true if the value returning by the first parameter is greater than the value returning by the second parameter. \\
($>$= ...) & Returns true if the value returning by the first parameter is greater or equal to the value returning by the second parameter. \\
(and ...) & Returns true if each function described in the ludeme returns true. \\
(if ...) & Returns the result of a Boolean function among two according to a condition. \\
(not ...) & Returns true if the Boolean ludeme described in the ludeme returns false. \\
(or ...) & Returns true if one of the functions described in the ludeme returns true. \\
(toBool ...) & Returns true if the result of the function in entry is 0. \\
(xor ...) & Returns the Boolean result of a xor operation between two values. \\
\end{longtable}
\rowcolors{0}{}{}

\subsection{All}

The "all" Boolean ludeme returns true if many aspects of a game satisfy a condition:

\renewcommand{\arraystretch}{1.3}
\arrayrulecolor{white}
\rowcolors{2}{gray!25}{gray!10}
\begin{longtable}{@{}p{.18\textwidth} | p{.75\textwidth}@{}}
\rowcolor{gray!50}
\textbf{Boolean} & \textbf{Description} \\
(all DiceEqual) &  Returns True if all the dice have the same values. \\
(all DiceUsed) &  Returns True if all the dice were used by previous moves in the same turn. \\
(all Passed) &  Returns True if all the players pass in their last turn. \\
(all Sites ...) &  Returns True if all the sites of a region satisfy the condition. \\
(all Values ...) &  Returns True if all the values of an array satisfy the condition. \\
\end{longtable}
\rowcolors{0}{}{}

With the ludeme (all Sites ...), each site is the region is iterated by the ludeme (site). 
In the following example, the ludeme (all Sites ...) returns true if all the sites occupied by the next player are not currently in a line of size 3 or more.
\begin{boxex}
\begin{ludii}
(all Sites
    (sites Occupied by:Next) 
    if:(not 
        (is Line 3 through:(site))
    )
)    
\end{ludii} 
\end{boxex}

\subsection{Can}

The "can" Boolean ludeme is composed in the current version of a single ludeme:

\renewcommand{\arraystretch}{1.3}
\arrayrulecolor{white}
\rowcolors{2}{gray!25}{gray!10}
\begin{longtable}{@{}p{.18\textwidth} | p{.75\textwidth}@{}}
\rowcolor{gray!50}
\textbf{Boolean} & \textbf{Description} \\
(can Move ...) &  Returns True if one of the moves described as a parameter of the ludeme is legal. \\
\end{longtable}
\rowcolors{0}{}{}

\subsection{Is}

The "is" ludeme is the most common Boolean ludeme and can be used to check many aspects of a game:

\renewcommand{\arraystretch}{1.3}
\arrayrulecolor{white}
\rowcolors{2}{gray!25}{gray!10}
\begin{longtable}{@{}p{.23\textwidth} | p{.70\textwidth}@{}}
\rowcolor{gray!50}
\textbf{Boolean} & \textbf{Description} \\
(is Active ...) &  Returns True if a specific player did not win or lose. \\
(is AnyDie ...) &  Returns True if any die is equal to a specific value. \\
(is Blocked ...) &  Returns True whether regions cannot possibly be connected by a player. \\
(is CaterpillarTree ...) &  Returns True if the graph is the largest caterpillar tree. \\
(is Connected ...) &  Returns True when regions are connected by a group of pieces. \\
(is Crossing ...) &  Returns True if two edges are crossing. \\
(is Cycle ...) &  Returns True if the game is currently in a cycle of state (a succession of states repeated at least three times). \\
(is Decided ...) &  Returns True if a subject is voted. \\
(is Empty ...) &  Returns True if a site is empty. \\
(is Enemy ...) &  Returns True if a player is an enemy to the mover. \\
(is Even ...) &  Returns True if a value is even. \\
(is Flat ...) &  Returns True if all the pieces have pieces under them or are in the ground (for 3D games). \\
(is Friend ...) &  Returns True if a player is a friend to the mover. \\
(is Full ...) &  Returns True if no site is empty. \\
(is Hidden ...) & Returns True if the site is invisible to a player \\
(is Hidden Count ...) & Returns True if the number of pieces on a site is hidden to a player \\
(is Hidden Rotation ...) & Returns True if piece rotation on a site is hidden to a player \\
(is Hidden State ...) & Returns True if the state site is hidden to a player \\
(is Hidden Value ...) & Returns True if the piece value is hidden to a player \\
(is Hidden What ...) & Returns True if the piece index is hidden to a player \\
(is Hidden Who ...) & Returns True if the piece owner is hidden to a player \\
(is In  ...) &  Returns True if a site belongs to a region. \\
(is LastFrom  ...) &  Returns True if the graph element type of the 'from' location of the last move is the expected one. \\
(is LastTo  ...) &  Returns True if the graph element type of the 'to' location of the last move is the expected one. \\
(is Line ...) &  Returns True if a line of a specific size is detected. \\
(is Loop ...) &  Returns True if a loop of pieces with at least a site in the middle is detected. \\
(is Mover ...) &  Returns True if a player is the mover. \\
(is Next ...) &  Returns True if a player is the next player to play. \\
(is Occupied ...) &  Returns True if a site is not empty. \\
(is Odd ...) &  Returns True if a value is odd. \\
(is Pattern ...) &  Returns True if the sites following a specific pattern satisfied the conditions. \\
(is Path ...) &  Returns True if a path or a cycle of a specific size with pieces owned by a specific player is detected. \\
(is Pending ...) &  Returns True if the state is in pending (due to the use of (set Pending ...)). \\
(is PipsMatch ...) &  Returns True if the pips of all the dominoes match to their neighbours. \\
(is Proposed ...) &  Returns True if a subject is proposed for voting. \\
(is Prev ...) &  Returns True if a player is the previous player who played. \\
(is RegularGraph ...) &  Returns True if the induced graph (according to the parameters of the ludeme) is regular. \\
(is Related ...) &  Returns True if two sites are related to each other (e.g. Orthogonal or Diagonal). \\
(is Repeat ...) &  Returns True if the current state was previously already reached. \\
(is SidesMatch ...) &  Returns True if the terminus of a tile matches with its neighbours (e.g. Trax). \\
(is SpanningTree ...) &  Returns True if the graph is a spanning tree. \\
(is Target ...) &  Returns True if a set of sites is in a specific configurations of pieces. \\
(is Threatened ...) &  Returns True if a location can be reached by an enemy piece. \\
(is Tree ...) &  Returns True if the graph is a tree. \\
(is TreeCenter ...) &  Returns True if the last site selected is the centre of the tree. \\
(is Trigger ...) &  Returns True if a specific event for a specific player was triggered by the ludeme (trigger ...). \\
(is Visited ...) &  Returns True if a site was already used in the previous moves of the same turn by a player. \\
(is Within ...) &  Returns True if a specific piece is currently placed on at least one site of a region. \\
\end{longtable}
\rowcolors{0}{}{}

In the following part, more examples and explanations are providing for some BooleanFunctions related to space constraints.

\subsubsection{is Connected}
The ludeme (is Connected ...) is used to detect a sequence of pieces connected regions (e.g. Hex\footnote{Hex: \href{https://en.wikipedia.org/wiki/Hex}{Wikipedia}}).

The main parameters of that ludeme are:
\begin{itemize}
    \item The number of regions to connect (default: All).
    \item at: describes the site through the connection has to be checked (default: (last To)).
    \item The directions allowed to connect regions.
    \item The disjointed regions to connect (default: the regions associeted with the mover).
    \item The RoleType of the owner of the pieces making the connection.
\end{itemize}

Here is a example of how it's used.
\renewcommand{\arraystretch}{1.3}
\arrayrulecolor{white}
\begin{longtable}{@{}M{.6\textwidth} | M{.4\textwidth}@{}}
\rowcolor{gray!50}
\textbf{Description} & \textbf{Satisfying}\\

A connection of the regions associated with the mover has to be made.

 \begin{boxexnonbreakable}
\begin{ludii}
    (equipment  {
        ...
        (regions P1 { (sites Side NE) (sites Side SW) } )
        (regions P2 { (sites Side NW) (sites Side SE) } )
    })
    
    (is Connected Mover)
\end{ludii} 
\end{boxexnonbreakable} 
&  
\begin{center}
\raisebox{-.50\height}{\includegraphics[width=0.8\linewidth]{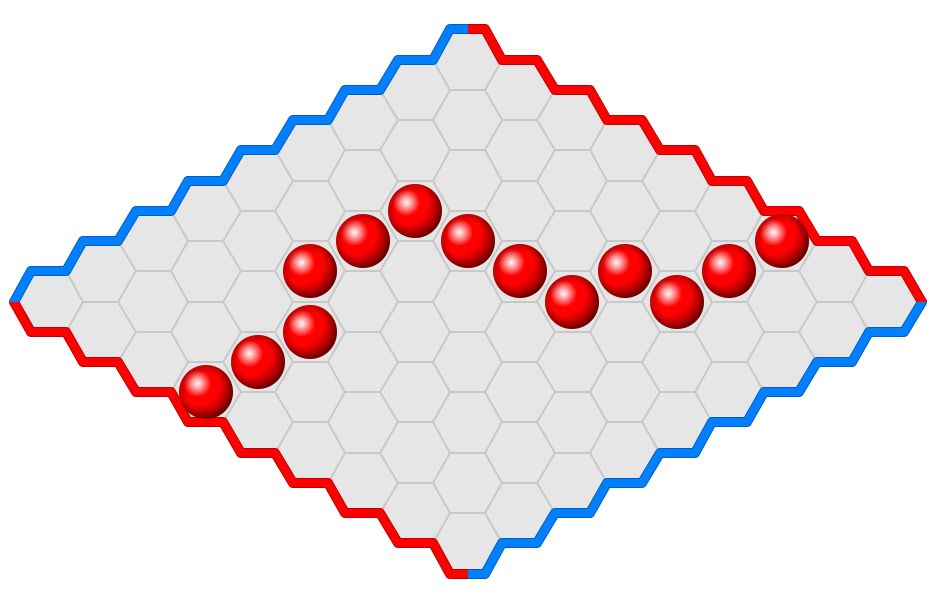}}
\end{center}\\
\end{longtable}

\subsubsection{is Line}
The ludeme (is Line ...) is used to detect a sequence of pieces places in a line pattern (e.g. Gomoku\footnote{Gomoku: \href{https://en.wikipedia.org/wiki/Gomoku}{Wikipedia}}).
The main parameters of that ludeme are:
\begin{itemize}
    \item The length of the line.
    \item The directions allowed to detect a line (default: Adjacent).
    \item through: describes the site through has to pass the line (default: (last To)).
    \item throughAny: describes the region of sites through at least one the line has to pass.
    \item The RoleType of the owner of the pieces making the line.
    \item what: describes the index of the component making the line.
    \item whats: describes the indices of the components making the line.
    \item exact: if true, line has to have exactly the right length, if false the line can be longer.
    \item if: describes the condition on the pieces making the line. The ludeme (to) is used to iterate the potential sites in the line.
    \item byLevel: if true, the line can also be made between different levels of stack of pieces.
\end{itemize}

Here are some examples of how it's used.
\renewcommand{\arraystretch}{1.3}
\arrayrulecolor{white}
\begin{longtable}{@{}M{.6\textwidth} | M{.4\textwidth}@{}}
\rowcolor{gray!50}
\textbf{Description} & \textbf{Satisfying}\\

A line of minimum size 3 has to be made.

 \begin{boxex}
\begin{ludii}
(is Line 3)
\end{ludii} 
\end{boxex} 
&  
\begin{center}
\raisebox{-.50\height}{\includegraphics[width=0.8\linewidth]{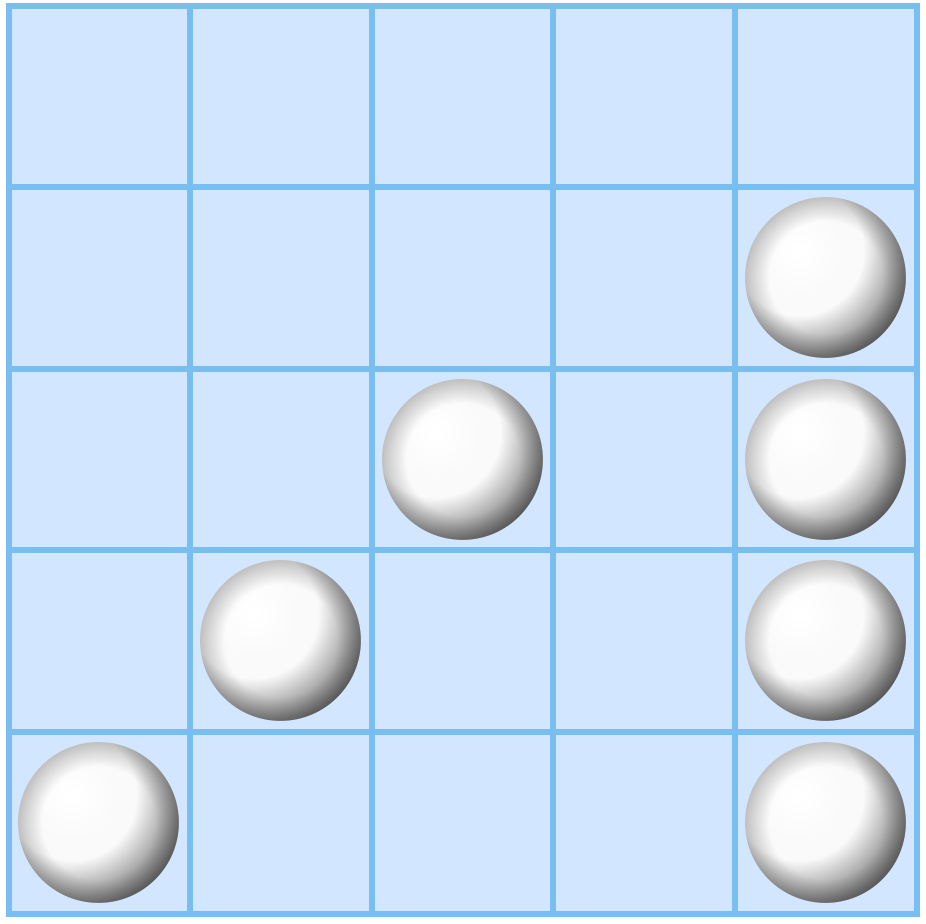}}
\end{center}\\

A line of exactly size 3 has to be made orthogonally.

 \begin{boxex}
\begin{ludii}
(is Line 3 Orthogonal exact:3)
\end{ludii} 
\end{boxex} 
&  
\begin{center}
\raisebox{-.50\height}{\includegraphics[width=0.8\linewidth]{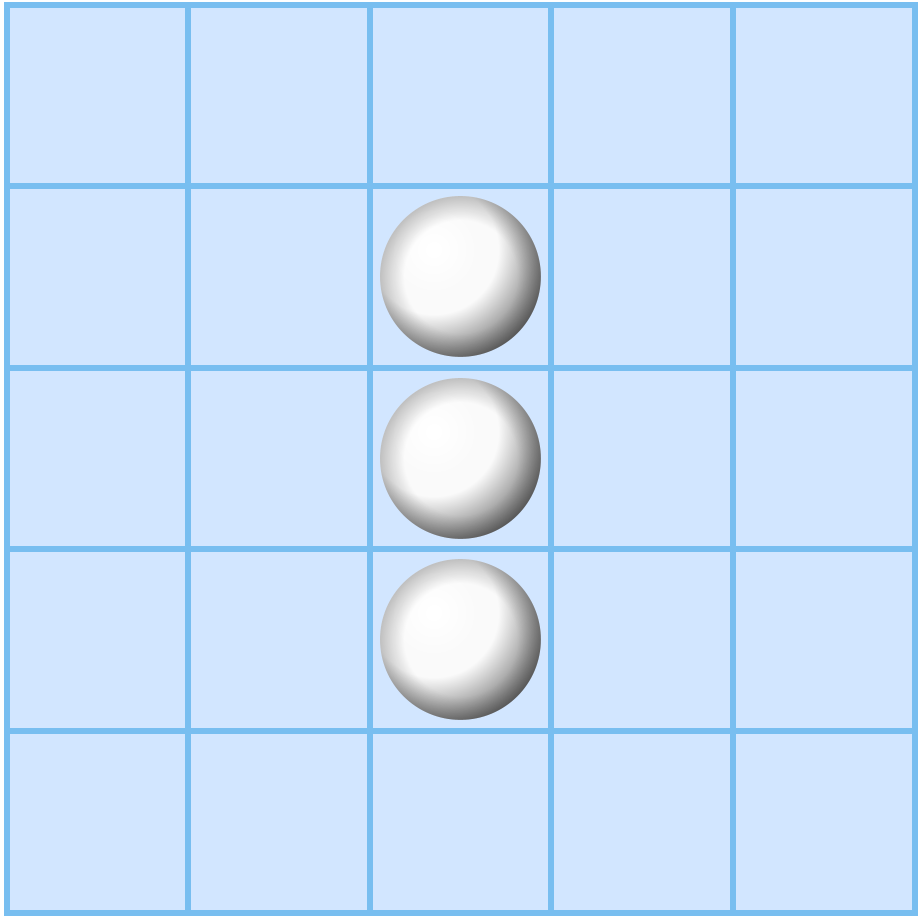}}
\end{center}\\

A line of minimum size 3 has to be made with any pieces owned by P1.

 \begin{boxex}
\begin{ludii}
(is Line 3 P1)
\end{ludii} 
\end{boxex} 
&  
\begin{center}
\raisebox{-.50\height}{\includegraphics[width=0.8\linewidth]{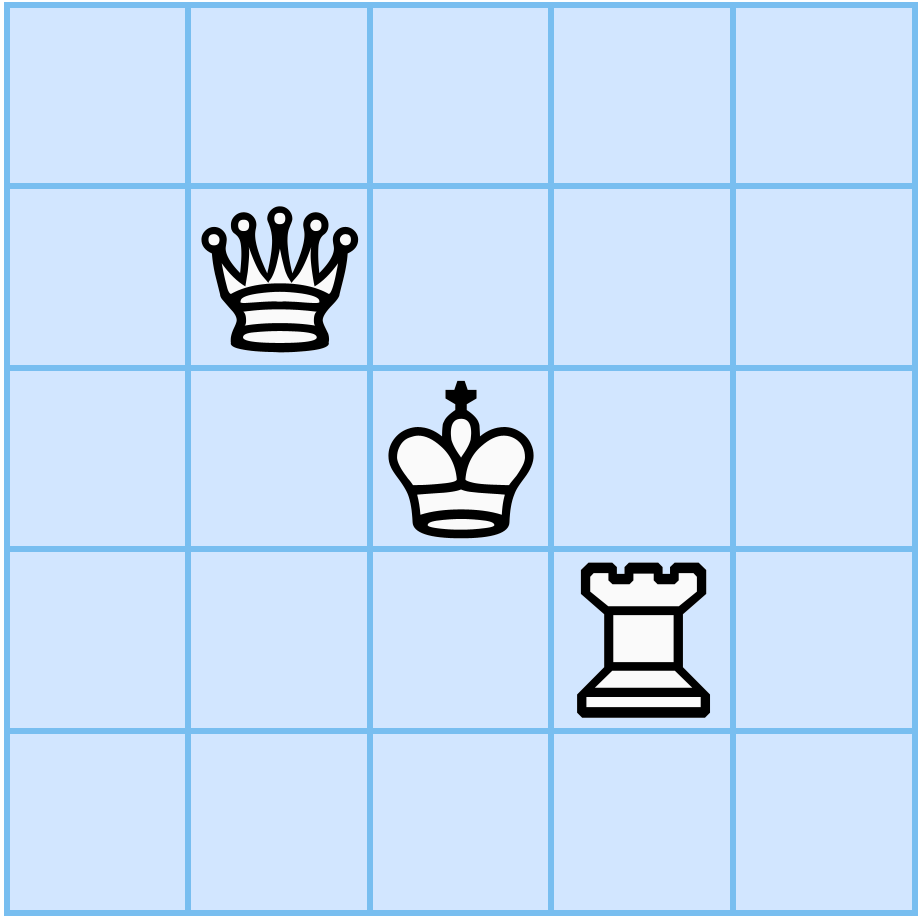}}
\end{center}\\
\end{longtable}

\subsubsection{is Loop}
The ludeme (is Loop ...) is used to detect a sequence of pieces looping thought at least a site (e.g. Havannah\footnote{Havannah: \href{https://en.wikipedia.org/wiki/Havannah}{Wikipedia}}).
The main parameters of that ludeme are:
\begin{itemize}
    \item The RoleType of the owner of the pieces enclosed by the loop (default: Empty sites).
    \item The directions allowed to detect the loop (default: Adjacent).
    \item The indice of the piece used to make the loop (default: (mover)).
    \item The site through has to pass the loop (default: (last To)).
    \item path: if true, the loop can be detected through tiles (e.g. Trax\footnote{Trax:\href{https://en.wikipedia.org/wiki/Trax_(game)}{Wikipedia}}).
\end{itemize}

Here are some examples of how it's used.
\renewcommand{\arraystretch}{1.3}
\arrayrulecolor{white}
\begin{longtable}{@{}M{.6\textwidth} | M{.4\textwidth}@{}}
\rowcolor{gray!50}
\textbf{Description} & \textbf{Satisfying}\\

A loop around empty sites has to be made.

 \begin{boxex}
\begin{ludii}
(is Loop)
\end{ludii} 
\end{boxex} 
&  
\begin{center}
\raisebox{-.50\height}{\includegraphics[width=0.8\linewidth]{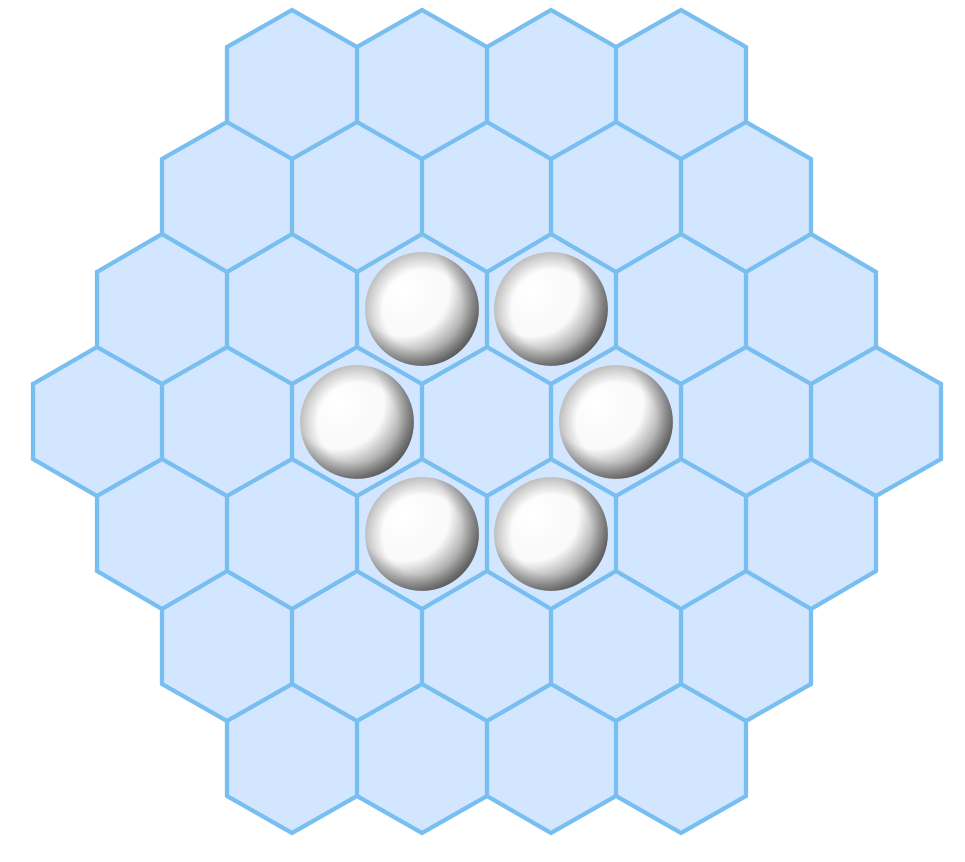}}
\end{center}\\

A loop using path of tiles pieces has to be made.

 \begin{boxex}
\begin{ludii}
(is Loop path:true)
\end{ludii} 
\end{boxex} 
&  
\begin{center}
\raisebox{-.50\height}{\includegraphics[width=0.8\linewidth]{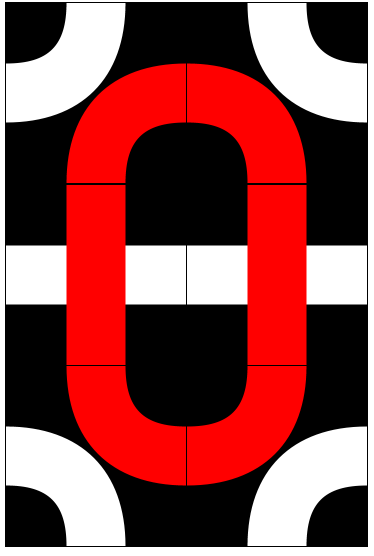}}
\end{center}\\
\end{longtable}

\subsubsection{is Pattern}
The ludeme (is Pattern ...) is used to detect pieces in specific pattern.
The main parameters of that ludeme are:
\begin{itemize}
    \item The walk describing the pattern.
    \item from: describes the site through has to pass the pattern (default (last To)).
    \item what: describes the indice of the piece used to make the pattern (default: (mover)).
    \item whats: describes the indices of the piece(s) used alternatively to make the pattern.
\end{itemize}

Here are some examples of how it's used.
\renewcommand{\arraystretch}{1.3}
\arrayrulecolor{white}
\begin{longtable}{@{}M{.6\textwidth} | M{.4\textwidth}@{}}
\rowcolor{gray!50}
\textbf{Description} & \textbf{Satisfying}\\

A square pattern has to be made.

\begin{boxex}
\begin{ludii}
(is Pattern {F R F R F R F})
\end{ludii} 
\end{boxex} 
&  
\begin{center}
\raisebox{-.50\height}{\includegraphics[width=0.8\linewidth]{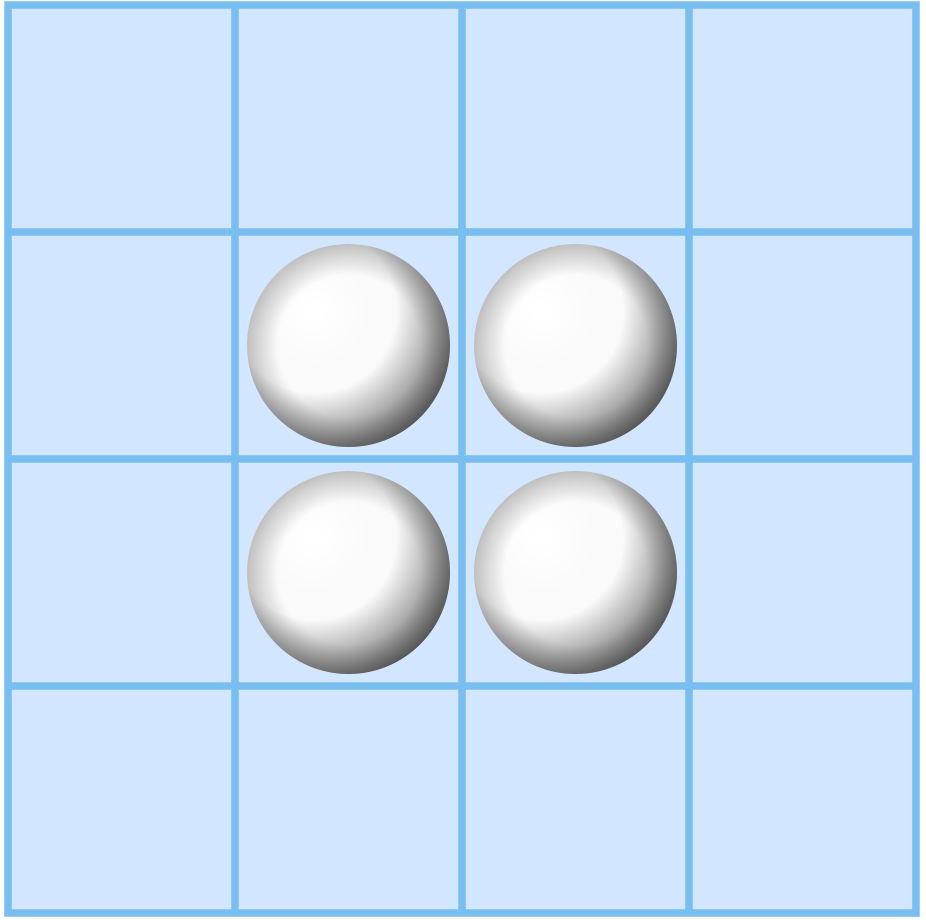}}
\end{center}\\

A square pattern has to be made with alternately a piece of the mover and a piece of the next player.

 \begin{boxex}
\begin{ludii}
(is Pattern {F R F R F R F} 
    whats:{(mover) (next)}
)
\end{ludii} 
\end{boxex} 
&  
\begin{center}
\raisebox{-.50\height}{\includegraphics[width=0.8\linewidth]{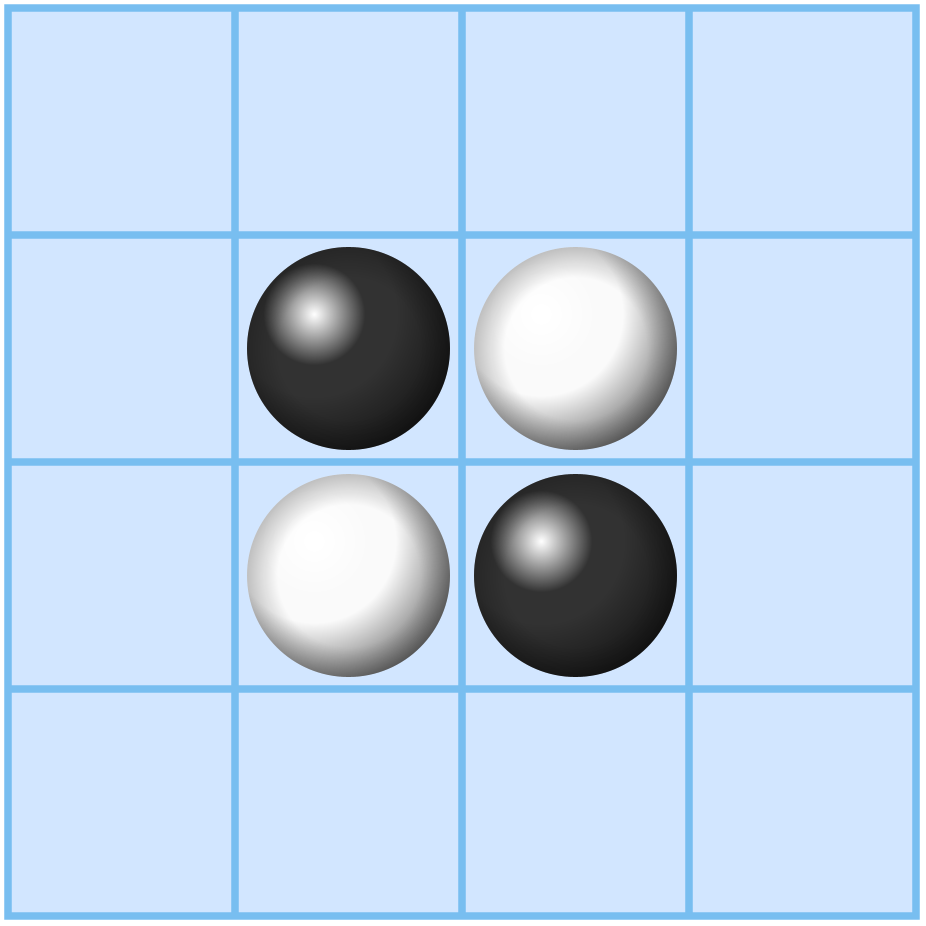}}
\end{center}\\
\end{longtable}

\subsection{No}

The "no" Boolean ludeme is composed in the current version of a single ludeme:

\renewcommand{\arraystretch}{1.3}
\arrayrulecolor{white}
\rowcolors{2}{gray!25}{gray!10}
\begin{longtable}{@{}p{.18\textwidth} | p{.75\textwidth}@{}}
\rowcolor{gray!50}
\textbf{Boolean} & \textbf{Description} \\
(no Moves ...) &  Returns True if a specific player has no legal moves in this state. \\
(no Pieces ...) &  Returns True if a specific piece type (or all piece types) is not placed. \\
\end{longtable}
\rowcolors{0}{}{}

\subsection{Was}

The "Was" Boolean ludeme is composed in the current version of a single ludeme and refers to a previous state.

\renewcommand{\arraystretch}{1.3}
\arrayrulecolor{white}
\rowcolors{2}{gray!25}{gray!10}
\begin{longtable}{@{}p{.18\textwidth} | p{.75\textwidth}@{}}
\rowcolor{gray!50}
\textbf{Boolean} & \textbf{Description} \\
(was Pass) &  Returns True if the last move was a pass. \\
\end{longtable}
\rowcolors{0}{}{}

\subsection{Deduction Puzzle}

Some Boolean functions are designed specically for deductions puzzles. As described in the Section \ref{section:DeductionPuzzle}, the deduction puzzles have a special state in Ludii and consequently using any of these Boolean functions in the description of a game will convert the game to a such state.

\renewcommand{\arraystretch}{1.3}
\arrayrulecolor{white}
\rowcolors{2}{gray!25}{gray!10}
\begin{longtable}{@{}p{.18\textwidth} | p{.75\textwidth}@{}}
\rowcolor{gray!50}
\textbf{Boolean} & \textbf{Description} \\
(all Different) &  Returns True if all variables are set and have different values. \\
(is Count ...) &  Returns True if the number of values set to variables and equal to a specific value is equal to a specific number. \\
(is Unique ...) &  Returns True if a value is set only one time to variables. \\
(is Sum ...) &  Returns True if the sum of all the values set to variables is equal to a value. \\
(is Solved ...) &  Returns True if all the variables are set and all the constraints of the puzzle are satisfied. \\
\end{longtable}
\rowcolors{0}{}{}


\section{IntFunction}\label{Section:IntFunction}

The Int functions return an integer value. For each of these functions, if nothing is specified in the documentation related to the function, the value UNDEFINED (-1) is returning if the result can not be computed.

\subsection{Math}

The mathematics Int functions are all the classical mathematics operators returning an integer:

\renewcommand{\arraystretch}{1.3}
\arrayrulecolor{white}
\rowcolors{2}{gray!25}{gray!10}
\begin{longtable}{@{}p{.13\textwidth} | p{.80\textwidth}@{}}
\rowcolor{gray!50}
\textbf{IntFunction} & \textbf{Description} \\
(+ ...) & Returns the sum of all the Int functions in parameters of that ludeme. \\
(- ...) & Returns the result of the subtraction of an Int function by another. \\
(* ...) & Returns the product of all the Int functions in parameters of that ludeme. \\
($\backslash$ ...) & Returns the (integer) division of an Int function by another. \\
($\%$ ...) & Returns the modulo of an Int function by another. \\
($\wedge$ ...) & Returns the result of an Int function to the power of the result of another Int function. \\
(abs ...) & Returns the absolute value of the result of an Int function. \\
(if ...) & Returns the result of an Int function among two according to a condition. \\
(max ...) & Returns the maximal value returning by a set of Int functions. \\
(min ...) & Returns the minimal value returning by a set of Int functions. \\
(toInt ...) & Converts a Float function or a Boolean Function to an Int function.. \\
\end{longtable}
\rowcolors{0}{}{}

\subsection{Board}

The board Int functions return an integer related to the board:

\renewcommand{\arraystretch}{1.3}
\arrayrulecolor{white}
\rowcolors{2}{gray!25}{gray!10}
\begin{longtable}{@{}p{.18\textwidth} | p{.75\textwidth}@{}}
\rowcolor{gray!50}
\textbf{IntFunction} & \textbf{Description} \\
(ahead ...) & Returns the index of the site at a specific distance in a specific direction from a site. \\
(centrePoint ...) & Returns the index of the site in the centre of the board. If the centre is many sites, the minimum index between them. \\
(column of:...) & Returns the index of the column of a site. \\
(coord ...) & Returns the index of the site corresponding to a coordinate on the board. \\
(cost ...) & Returns the cost/weight of a graph element. \\
(handSite ...) & Returns the index of a site in the hand of a player. \\
(id ...) & Returns the index of a player or of a component. \\
(layer of:...) & Returns the index of the layer of a site. \\
(mapEntry ...) & Returns the integer value mapping to the entry. \\
(phase of:...) & Returns the phase index of a site. \\
(regionSite ...) & Returns the index of a specific site in a region. \\
(row of:...) & Returns the index of the row of a site. \\
(where ...) & Returns the minimal index of a site where is placed a specific piece. \\
\end{longtable}
\rowcolors{0}{}{}

\subsection{Card}

The card Int functions return an integer related to card aspect. However, these ludemes are still in progress and can change in the future:

\renewcommand{\arraystretch}{1.3}
\arrayrulecolor{white}
\rowcolors{2}{gray!25}{gray!10}
\begin{longtable}{@{}p{.23\textwidth} | p{.70\textwidth}@{}}
\rowcolor{gray!50}
\textbf{IntFunction} & \textbf{Description} \\
(card Rank ...) & Returns the index of the rank of a card. \\
(card Suit ...) & Returns the index of the suit of a card. \\
(card TrumpRank ...) & Returns the index of the trump rank of a card. \\
(card TrumpSuit ...) & Returns the index of the suit, which is the trump. \\
(card TrumpValue ...) & Returns the trump value of a card. \\
\end{longtable}
\rowcolors{0}{}{}

\subsection{Count}

The count Int ludeme is used to count a property within the game such as the number of players, components, sites, turns, groups, etc. 

Here is the list of all of them:
\renewcommand{\arraystretch}{1.3}
\arrayrulecolor{white}
\rowcolors{2}{gray!25}{gray!10}
\begin{longtable}{@{}p{.26\textwidth} | p{.67\textwidth}@{}}
\rowcolor{gray!50}
\textbf{IntFunction} & \textbf{Description} \\
(count ...) & Returns the number of pieces in a site (if they are not in a stack). \\
(count Active) & Returns the number of active players. \\
(count Adjacent ...) & Returns the number of adjacent sites of a site. \\
(count Cells) & Returns the number of cells of the board. \\
(count Columns) & Returns the number of columns of the board. \\
(count Diagonal ...) & Returns the number of diagonal sites of a site. \\
(count Edges) & Returns the number of edges of the board. \\
(count Groups) & Returns the total number of groups of pieces owned by a player. \\
(count Liberties) & Returns the number of liberties (empty sites) of a group of pieces starting from a site. \\
(count Moves) & Returns the number of moves done so far. \\
(count MovesThisTurn) & Returns the number of moves done so far in this turn. \\
(count Neighbours ...) & Returns the number of neighbours sites of a site. \\
(count Off ...) & Returns the number of off-diagonal sites of a site. \\
(count Orthogonal ...) & Returns the number of orthogonal sites of a site. \\
(count Pieces ...) & Returns the number of pieces of a player. \\
(count Pips) & Returns the total number of pips of all the dice. \\
(count Phases) & Returns the number of phases of the game. \\
(count Players) & Returns the number of players. \\
(count Rows) & Returns the number of rows of the board. \\
(count Sites in:...) & Returns the number of sites in a region function. \\
(count Stack ...) & Returns the number of pieces in stack(s) according to conditions. \\
(count Steps ...) & Returns the distance between two sites. The step between two sites can be defined with a specific relation type or with a Step moves ludeme. \\
(count StepsOnTrack ...) & Returns the number of steps between two sites in following a track. \\
(count Trials) & Returns the number of trials done so far (for match game). \\
(count Turns) & Returns the number of turns done so far. \\
(count Vertices) & Returns the number of vertices of the board. \\
\end{longtable}
\rowcolors{0}{}{}

\subsection{Dice}

The dice Int ludeme is composed in the current version of a single ludeme:

\renewcommand{\arraystretch}{1.3}
\arrayrulecolor{white}
\rowcolors{2}{gray!25}{gray!10}
\begin{longtable}{@{}p{.23\textwidth} | p{.70\textwidth}@{}}
\rowcolor{gray!50}
\textbf{IntFunction} & \textbf{Description} \\
(face) & Returns the current value of a dice according to the site state where it is placed. \\
\end{longtable}
\rowcolors{0}{}{}

\subsection{Iterator}

The iterator Int functions return an integer used as an iterator.
These iterators can be used in ludemes looping through values such as (forEach ...) or for ludeme checking many sites in each direction such as many moves ludemes.

Here is the list of them:
\renewcommand{\arraystretch}{1.3}
\arrayrulecolor{white}
\rowcolors{2}{gray!25}{gray!10}
\begin{longtable}{@{}p{.23\textwidth} | p{.70\textwidth}@{}}
\rowcolor{gray!50}
\textbf{IntFunction} & \textbf{Description} \\
(between) & Returns the current value of the between iterator. \\
(edge) & Returns the current value of the edge iterator. \\
(from) & Returns the current value of the from iterator. \\
(hint) & Returns the current value of the hint iterator. \\
(level) & Returns the current value of the level iterator. \\
(pips) & Returns the current value of the pips iterator. \\
(player) & Returns the current value of the player iterator. \\
(site) & Returns the current value of the site iterator. \\
(team) & Returns the current value of the team iterator. \\
(to) & Returns the current value of the to iterator. \\
(track) & Returns the current value of the track iterator. \\
(value) & Returns the current value of the value iterator. \\
\end{longtable}
\rowcolors{0}{}{}

\subsection{Last}

The last Int ludeme returns integers information about the last state:

\renewcommand{\arraystretch}{1.3}
\arrayrulecolor{white}
\rowcolors{2}{gray!25}{gray!10}
\begin{longtable}{@{}p{.23\textwidth} | p{.70\textwidth}@{}}
\rowcolor{gray!50}
\textbf{IntFunction} & \textbf{Description} \\
(last From) & Returns the index of the 'from' site of the last move. \\
(last To) & Returns the index of the 'to' site of the last move. \\
\end{longtable}
\rowcolors{0}{}{}

\subsection{Match}

The match Int ludeme is composed in the current version of a single ludeme and returns information about match:

\renewcommand{\arraystretch}{1.3}
\arrayrulecolor{white}
\rowcolors{2}{gray!25}{gray!10}
\begin{longtable}{@{}p{.23\textwidth} | p{.70\textwidth}@{}}
\rowcolor{gray!50}
\textbf{IntFunction} & \textbf{Description} \\
(matchScore ...) & Returns the score of a player in the match. \\
\end{longtable}
\rowcolors{0}{}{}

\subsection{Size}

The size Int ludeme is composed in the current version of a single ludeme and returns information about match:

\renewcommand{\arraystretch}{1.3}
\arrayrulecolor{white}
\rowcolors{2}{gray!25}{gray!10}
\begin{longtable}{@{}p{.23\textwidth} | p{.70\textwidth}@{}}
\rowcolor{gray!50}
\textbf{IntFunction} & \textbf{Description} \\
(size Array ...) & Returns the size of an IntArrayFunction. \\
(size Group ...) & Returns the size of the group of pieces starting from a specific site. \\
(size Large Piece ...) & Returns the number of sites occupied by a large piece. \\
(size Stack ...) & Returns the size of the stack placed on a site. \\
(size Territory ...) & Returns the number of sites enclosed by the pieces of a player. \\
\end{longtable}
\rowcolors{0}{}{}

\subsection{Stacking}

The stacking Int ludeme is composed in the current version of a single ludeme:

\renewcommand{\arraystretch}{1.3}
\arrayrulecolor{white}
\rowcolors{2}{gray!25}{gray!10}
\begin{longtable}{@{}p{.23\textwidth} | p{.70\textwidth}@{}}
\rowcolor{gray!50}
\textbf{IntFunction} & \textbf{Description} \\
(topLevel ...) & Returns the index of the topmost level of a stack placed at a specific site. \\
\end{longtable}
\rowcolors{0}{}{}

\subsection{State}

The state Int ludemes returns an integer value based on the current game state:

\renewcommand{\arraystretch}{1.3}
\arrayrulecolor{white}
\rowcolors{2}{gray!25}{gray!10}
\begin{longtable}{@{}p{.23\textwidth} | p{.70\textwidth}@{}}
\rowcolor{gray!50}
\textbf{IntFunction} & \textbf{Description} \\
(amount ...) & Returns the amount of a player set with the ludeme (set Amount ...). \\
(counter) & Returns the counter used in the state or set with (set Counter ...). \\
(mover) & Returns the index of the player who has to move. \\
(next) & Returns the index of the player who has to move in the next state. \\
(pot) & Returns the value of the pot set with the ludeme (set Pot ...). \\
(pre ...) & Returns the index of the player who has played the previous move. \\
(rotation ...) & Returns the index of the rotation of a piece placed on a site. \\
(score ...) & Returns the score of a player. \\
(state ...) & Returns the state of a piece placed on a site. \\
(var ...) & Returns the value of a variable var set in using the ludeme (set Var ...). \\
(what ...) & Returns the index of the piece placed on a site. \\
(who ...) & Returns the index of the player owning a piece placed on a site. \\
\end{longtable}
\rowcolors{0}{}{}

\subsection{Tile}

The tile Int ludeme is composed in the current version of a single ludeme:

\renewcommand{\arraystretch}{1.3}
\arrayrulecolor{white}
\rowcolors{2}{gray!25}{gray!10}
\begin{longtable}{@{}p{.23\textwidth} | p{.70\textwidth}@{}}
\rowcolor{gray!50}
\textbf{IntFunction} & \textbf{Description} \\
(pathExtent ...) & Returns the maximum extent of a path. \\
\end{longtable}
\rowcolors{0}{}{}

\subsection{TrackSite}

The trackSite Int ludemes return integer value related to a site belonging to a track:

\renewcommand{\arraystretch}{1.3}
\arrayrulecolor{white}
\rowcolors{2}{gray!25}{gray!10}
\begin{longtable}{@{}p{.23\textwidth} | p{.70\textwidth}@{}}
\rowcolor{gray!50}
\textbf{IntFunction} & \textbf{Description} \\
(trackSite Move ...) & Returns the index of the site from a site belonging to the track after a specific number of steps. \\
(trackSite EndTrack ...) & Returns the index of the site at the end of the track. \\
(trackSite FirstSite ...) & Returns the index of the first site verifying a condition. \\
\end{longtable}
\rowcolors{0}{}{}

\subsection{Value}

The value Int ludemes return integer value related to the keyword value:

\renewcommand{\arraystretch}{1.3}
\arrayrulecolor{white}
\rowcolors{2}{gray!25}{gray!10}
\begin{longtable}{@{}p{.23\textwidth} | p{.70\textwidth}@{}}
\rowcolor{gray!50}
\textbf{IntFunction} & \textbf{Description} \\
(value at:...) & Returns the value of a piece at a specific location. \\
(value Player) & Returns the value associated to a player set with (set Value Player ...). \\
(value Pending) & Returns the pending value of the state set with (set Pending ...). \\
(value Random ...) & Returns a random value within a range. \\
\end{longtable}
\rowcolors{0}{}{}

\subsection{Constants}

The Ludii's compiler can convert a few constants to integers:

\renewcommand{\arraystretch}{1.3}
\arrayrulecolor{white}
\rowcolors{2}{gray!25}{gray!10}
\begin{longtable}{@{}c | c@{}}
\rowcolor{gray!50}
\textbf{Constants} & \textbf{Value} \\
Off & -1 \\
End & -2 \\
Undefined & -1 \\
Unused & -4 \\
NoPiece & 0 \\
Infinity & 1000000000 \\
\end{longtable}


\section{IntArrayFunction}\label{Section:IntArrayFunction}

The Int Array functions return an array of integer values. These functions are not commonly used because of the Region functions (see Section \ref{Section:RegionFunction}) which are dedicated to set of sites. However some ludemes refer directly to these functions. The next subsections give a short description of each of them.

\subsection{Math}

The mathematics Int Array functions contains classical mathematics operators returning an array of integers:

\renewcommand{\arraystretch}{1.3}
\arrayrulecolor{white}
\rowcolors{2}{gray!25}{gray!10}
\begin{longtable}{@{}p{.23\textwidth} | p{.70\textwidth}@{}}
\rowcolor{gray!50}
\textbf{IntArrayFunction} & \textbf{Description} \\
(difference ...) & Returns an array containing the integers in the first parameter that are not in the second parameter. \\
(intersection ...) & Returns an array containing the integers which are in each array in entry. \\
(results ...) & Returns an array of all the results of the function in parameter for each site 'from' to each site 'to'. \\
(union ...) & Returns an array containing the integers on each array in entry. \\
\end{longtable}
\rowcolors{0}{}{}

\subsection{Players}

The Players Int Array functions return an array of players indices.

\renewcommand{\arraystretch}{1.3}
\arrayrulecolor{white}
\rowcolors{2}{gray!25}{gray!10}
\begin{longtable}{@{}p{.23\textwidth} | p{.70\textwidth}@{}}
\rowcolor{gray!50}
\textbf{IntArrayFunction} & \textbf{Description} \\
(players Team....) & Returns the indices of all the players in a team. \\
(players ... of:...) & Returns the indices of players related to the player index specified. \\
\end{longtable}
\rowcolors{0}{}{}

\subsection{Sizes}

The Sizes Int Array functions return an array of sizes of regions.

\renewcommand{\arraystretch}{1.3}
\arrayrulecolor{white}
\rowcolors{2}{gray!25}{gray!10}
\begin{longtable}{@{}p{.23\textwidth} | p{.70\textwidth}@{}}
\rowcolor{gray!50}
\textbf{IntArrayFunction} & \textbf{Description} \\
(group ...) & Returns the sizes of the group computing according to the parameters. \\
\end{longtable}
\rowcolors{0}{}{}

\subsection{State}

The State Int Array functions return an array of integers depending on the current state.

\renewcommand{\arraystretch}{1.3}
\arrayrulecolor{white}
\rowcolors{2}{gray!25}{gray!10}
\begin{longtable}{@{}p{.23\textwidth} | p{.70\textwidth}@{}}
\rowcolor{gray!50}
\textbf{IntArrayFunction} & \textbf{Description} \\
(rotations ...) & Converts a list of facing directions to a list of integer directions according to the rotation of the piece placed on a site. \\
\end{longtable}
\rowcolors{0}{}{}

\subsection{Values}

The Values Int Array functions return an array of integers according to the parameter.

\renewcommand{\arraystretch}{1.3}
\arrayrulecolor{white}
\rowcolors{2}{gray!25}{gray!10}
\begin{longtable}{@{}p{.28\textwidth} | p{.65\textwidth}@{}}
\rowcolor{gray!50}
\textbf{IntArrayFunction} & \textbf{Description} \\
(values Remembered ...) & Returns the current list of values remembered in the state because of the use of the move (remember Value ...). \\
\end{longtable}
\rowcolors{0}{}{}


\section{RegionFunction}\label{Section:RegionFunction}

The Region functions return a list of Integer corresponding to a list of indices of sites. An empty list is returned in case of no site in the result of a Region function.

\subsection{Math}

The mathematics Region functions are mathematics operators about set or about set of vertices/faces in a graph:

\renewcommand{\arraystretch}{1.3}
\arrayrulecolor{white}
\rowcolors{2}{gray!25}{gray!10}
\begin{longtable}{@{}p{.18\textwidth} | p{.75\textwidth}@{}}
\rowcolor{gray!50}
\textbf{RegionFunction} & \textbf{Description} \\
(difference ...) & Returns the sites in the first Region function parameter without the sites in the second Region function parameter. \\
(expand ...) & Returns the sites of a region expanding in one or all the directions with a specific distance. \\
(if ...) & Returns the result of a Region function among two according to a condition. \\
(intersection ...) & Returns the sites in common between many regions. \\
(union ...) & Returns all the sites of many regions. \\
\end{longtable}
\rowcolors{0}{}{}

\subsection{Last}

The last Region ludeme returns regions information about the last state:

\renewcommand{\arraystretch}{1.3}
\arrayrulecolor{white}
\rowcolors{2}{gray!25}{gray!10}
\begin{longtable}{@{}p{.23\textwidth} | p{.70\textwidth}@{}}
\rowcolor{gray!50}
\textbf{RegionFunction} & \textbf{Description} \\
(last Between) & Returns the indices of the 'between' sites of the last decision move. \\
\end{longtable}
\rowcolors{0}{}{}

\subsection{ForEach}

The forEach Region ludeme is used to filter a region or to create a region based on different player indices:

\renewcommand{\arraystretch}{1.3}
\arrayrulecolor{white}
\rowcolors{2}{gray!25}{gray!10}
\begin{longtable}{@{}p{.30\textwidth} | p{.63\textwidth}@{}}
\rowcolor{gray!50}
\textbf{RegionFunction} & \textbf{Description} \\
(forEach ... ) &  Returns the sites satisfying a constraint from a given region. \\
(forEach (players ...) ...) &  Returns a region in iterating player indices. \\
(forEach Team ) &  Returns a region in iterating the teams. \\
\end{longtable}
\rowcolors{0}{}{}

With the ludeme (forEach ...), each site in the region is described by the ludeme (site). 
In the following example, the ludeme (forEach ...) returns the sites occupied by the next player that are not currently in a line of size 3 or more.
\begin{boxex}
\begin{ludii}
(forEach
    (sites Occupied by:Next) 
    if:(not 
        (is Line 3 through:(site))
    )
)    
\end{ludii} 
\end{boxex}

With the ludeme (forEach (players ...) ...), each player in the IntArrayFunction is described by the ludeme (player). 
In the following example, the ludeme (forEach (players ...) ...) returns the sites occupied by each player ally with the mover.
\begin{boxex}
\begin{ludii}
(forEach
    (players Ally of:(mover))
    (sites Occupied by:Player) 
)    
\end{ludii} 
\end{boxex}

With the ludeme (forEach Team ...), each team is iterated with the ludeme (team).
In the following example, the ludeme (forEach Team ...) returns the sites occupied by each player of each team.
\begin{boxex}
\begin{ludii}
(forEach Team
    (forEach (team)
        (sites Occupied by:Player) 
    )
)    
\end{ludii} 
\end{boxex}

\subsection{Sites}

The (sites ...) Region ludeme can return many type of regions:

\renewcommand{\arraystretch}{1.3}
\arrayrulecolor{white}
\rowcolors{2}{gray!25}{gray!10}
\begin{longtable}{@{}p{.28\textwidth} | p{.65\textwidth}@{}}
\rowcolor{gray!50}
\textbf{RegionFunction} & \textbf{Description} \\
(sites) & Returns the sites iterated by ludemes iterated though many regions such as (forEach Group ...). \\
(sites {...}) & Returns a set of sites corresponding to a list of indices or of a list of coordinates. \\
(sites Angled) & Returns all the angled edges of the board. \\
(sites Around ...) & Returns the sites around a given region or site according to specified directions and conditions. \\
(sites Axial) & Returns all the axial edges of the board. \\
(sites Between) & Returns all the 'between' sites of a set of moves. \\
(sites Board) & Returns all the sites of the board. \\
(sites Bottom) & Returns all the sites at the bottom of the board. \\
(sites Cell ...) & Returns all the vertices of a cell. \\
(sites Centre) & Returns all the sites in the centre of a board. \\
(sites Column ...) & Returns all the sites of a column. \\
(sites ConcaveCorners ...) & Returns all the concave corners sites of the board. \\
(sites ConvexCorners ...) & Returns all the convex corners sites of the board. \\
(sites Corners ...) & Returns all the corners sites of the board. \\
(sites Direction ...) & Returns the sites in a direction from a site. \\
(sites Distance ...) & Returns the sites at a specific distance from another. \\
(sites Edge ...) & Returns all the vertices of an edge. \\
(sites Empty) & Returns all the empty sites. \\
(sites From) & Returns all the 'from' sites of a set of moves. \\
(sites Hand ...) & Returns all the sites of a hand container. \\
(sites Hidden ...) & Returns all the sites invisible to a player \\
(sites Hidden Count ...) & Returns all the sites with the piece number hidden to a player \\
(sites Hidden Rotation ...) & Returns all the sites with the piece rotation hidden to a player \\
(sites Hidden Value ...) & Returns all the sites with the piece value hidden to a player \\
(sites Hidden State ...) & Returns all the sites with the state hidden to a player \\
(sites Hidden What ...) & Returns all the sites with the piece index hidden to a player \\
(sites Hidden Who ...) & Returns all the sites with the piece owner hidden to a player \\
(sites Hint ...) & Returns all the sites with a hint on them. \\
(sites Horizontal) & Returns all the horizontal edges of the board. \\
(sites Incident ...) & Returns the sites connected to a specific site. \\
(sites Inner) & Returns all the inner sites of the board. \\
(sites LargePiece ...) & Returns all the sites used by a large piece from the root. \\
(sites LastFrom) & Returns all the 'from' sites of the last move. \\
(sites LastTo) & Returns all the 'to' sites of the last move. \\
(sites Layer ...) & Returns all the sites of a layer. \\
(sites Left) & Returns all the sites at the leftmost of the board. \\
(sites LineOfPlay) & Returns all the sites in the line of play of domino games. \\
(sites LineOfSight ...) & Returns all the sites in a specific line of sight from a site. \\
(sites Loop ...) & Returns all the sites making a loop or inside the loop. \\
(sites Occupied ...) & Returns the sites occupied by pieces owned by a given player in a given container. \\
(sites Major) & Returns all the sites used as major generators to generate the graph. \\
(sites Minor) & Returns all the sites used as minor generators to generate the graph. \\
(sites Outer) & Returns all the outer sites of the board. \\
(sites P...) & Returns the sites corresponding to regions defined as owned by a given player in the equipment. \\
(sites Pattern ...) & Returns the sites in a specific pattern. \\
(sites Pending) & Returns all the sites in pending in the current state. \\
(sites Perimeter) & Returns all the perimeter sites of the board. \\
(sites Phase ...) & Returns all the sites in a specific phase. \\
(sites Playable) & Returns all the playable sites of a boardless game. \\
(sites Right) & Returns all the sites at the rightmost of the board. \\
(sites Random ...) & Returns a list of random sites. \\
(sites Row ...) & Returns all the sites of a row. \\
(sites Side ...) & Returns all the sites of a given side of the board. \\
(sites Slash) & Returns all the slash edges of the board. \\
(sites Slosh) & Returns all the slosh edges of the board. \\
(sites Start ...) & Returns the sites occupied by pieces of a given type in the initial state. \\
(sites State ...) & Returns all the sites with a specific state site. \\
(sites To) & Returns all the 'to' sites of a set of moves. \\
(sites ToClear) & Returns all the sites to remove the pieces on when the turn of the player will be over. \\
(sites Top) & Returns all the sites at the topmost of the board. \\
(sites Track ...) & Returns all the sites of a track. \\
(sites Vertical) & Returns all the vertical edges of the board. \\
(sites Walk ...) & Returns all the sites reached in following a walk. \\
(sites Winning ...) & Returns all the winning sites for a player. \\
\end{longtable}
\rowcolors{0}{}{}


\section{DirectionFunctions}\label{Section:DirectionFunction}

The Direction functions return a list of Absolute directions. They are used inside ludemes (generally Moves ludemes) to specify set of directions (e.g. (move Slide ...)).

\renewcommand{\arraystretch}{1.3}
\arrayrulecolor{white}
\rowcolors{2}{gray!25}{gray!10}
\begin{longtable}{@{}p{.34\textwidth} | p{.60\textwidth}@{}}
\rowcolor{gray!50}
\textbf{DirectionFunctions} & \textbf{Description} \\
(difference ...) & Returns the set of absolute directions of the first entry which are not in the second entry. \\
(directions ... from:... to:...) & Returns the direction between 'from' and 'to' if exists. \\
(directions ...) & Converts the corresponding set of absolute directions or relative directions in entry to absolute directions. \\
(directions Random ...) & Returns a random set of directions. \\
(if ...) & Returns a set of directions according to a condition. \\
\end{longtable}
\rowcolors{0}{}{}

\section{RangeFunctions}\label{Section:RangeFunction}

The Range functions are used in some ludemes to specify a minimum and/or a maximum to obtain a range.

\renewcommand{\arraystretch}{1.3}
\arrayrulecolor{white}
\rowcolors{2}{gray!25}{gray!10}
\begin{longtable}{@{}p{.28\textwidth} | p{.65\textwidth}@{}}
\rowcolor{gray!50}
\textbf{RangeFunction} & \textbf{Description} \\
(exact ...) & To define a range of a single value. \\
(min ...) & To define a range with a minimum but no maximum. \\
(max ...) & To define a range with a maximum but no minimum. \\
(range ...) & To define a range with a minimum and a maximum. \\
\end{longtable}
\rowcolors{0}{}{}









\chapter{\textcolor{gray!95}{Options}}\label{Chapter:Options}


For many games there exist alternative rule sets and other variable aspects, such as different board sizes, number of pieces, starting positions, and so on. 
The Ludii game compiler supports an {\it option} mechanism to allow such alternatives for a game to be defined in a single description, to avoid the need to implement each one in its own file. 
Options are defined outside the main {\tt game} ludeme but typically used within it. 
Options are typically declared directly below the {\tt game ...)} ludeme, for clarity.

Options are instantiated at compile time and can be arbitrarily large, including choices between complete game descriptions if desired. 
Multiple options can be specified in combination, to give dozens or even hundreds of variant rule sets for a single game description.

Each set of options is declared with the {\tt option} keyword and constitutes an option {\it category} with a number of option {\it items}. 
Each option item has a name and one or more {\it arguments} followed by a short description.  
The following example shows options for board sizes and for the ending rules:
\begin{boxex}
\begin{ludii}
(option "Board Size" <Board> args:{ <size> } {
    (item "5x5"    <5>   "The game is played on a 5x5 board.")   
    (item "10x10"  <10>  "The game is played on a 10x10 board.")   
    (item "19x19"  <19>  "The game is played on a 19x19 board.")*   
})

(option "End Rules" <Result> args:{ <type>} {
    (item "Standard" <Win>  "The player to reach the end wins.")*   
    (item "Misere"   <Loss> "The player to reach the end loses.")   
})
\end{ludii} 
\end{boxex}

Option items are referenced in the game description by tag-argument pairs {\tt <Tag:arg>}. 
However if the option has only one option, the parameter can be not specified.
For example, this option call in a game description. The first one does not specify the parameter contrary to the second one.
\begin{boxex}
\begin{ludii}
(board (hex Diamond <Board>)) 

(end 
    (if 
        (is Connected Mover) 
        (result Mover <Result:type>)
    )
) 
\end{ludii} 
\end{boxex}

In order to specify the main/default option(s) of a game, an asterisk may optionally be appended to the end of each option item (e.g. option ``19x19" or ``Standard" in the examples).
If no user-selected options are specified when a game is compiled, then the highest priority item within each category becomes the current option for that category. 

The names of the option categories can be anything, however conventionally, we try to use the following ones when that's possible:
\begin{itemize}
\item ``Board":  To modify the kind of board, mainly for the GUI.
\item ``Board Size": To modify the size of the board.
\item ``Challenge": For the different challenge of a deduction puzzle.
\item ``End Rules": To modify the ending rules of the game.
\item ``Play Rules": To modify the playing rules of the game.
\item ``Players": To modify the number of players.
\item ``Start Rules": To modify the starting rules of the game.
\item ``Tiling": To modify the tiling of the board.
\item ``Turn Limit": To modify the turn limit.
\item ``Variant" or ``Version": To modify many rule types in the same time.
\end{itemize}

\section{Rulesets}

In addition to the {\tt option} mechanism, Ludii also supports a {\tt ruleset} mechanism that allows user to declare custom rule sets defined by combinations of options. 

In the following example, two rulesets are defined, one corresponding to a board of size 19x19 with the standard victory rules and another with a board of size 10x10 with the misere ending rules.
\begin{boxex}
\begin{ludii}
(rulesets { 
        (ruleset "Ruleset/Standard" {
            "Board Size/19x19" "End Rules/Standard"
        })
        
        (ruleset "Ruleset/Misere" {
            "Board Size/10x10" "End Rules/Misere"
        })
})
\end{ludii} 
\end{boxex}

The convention to name each {\it ruleset} is to start them by "Ruleset/" then the name of the ruleset.


\chapter{\textcolor{gray!95}{Defines}}\label{Chapter:Defines}

The $\it define$ is a mechanism used to encapsulate a part of a description that can be repeated many times in the description, they can be viewed as a macro.

Defines are useful for:
\begin{itemize}
\item Simplifying game descriptions by wrapping complex ludeme structures into simple labels.
\item Giving meaningful names to useful game concepts.
\item Gathering repeated ludeme structures that are duplicated across multiple games into a single reusable definition. 
\end{itemize}

When a game has to be compiled, Ludii first expands all the defines before compiling the game description. A define has a name and possibly can be parameterised. 
The parameters of a define are described with the symbol \# followed by the indices of the parameters of the define: \#1, \#2, \ldots, \#n for n parameters.

In the following example we define a hop move, which can be parameterised by the condition of the piece to jump, the effect to apply to it and the condition on the site to move to. 
\begin{boxex}
\begin{ludii}
(define "Hop" 
    (move Hop 
        (between 
            if:#1
            #2
        )
        (to if:#3)
    )
)
\end{ludii} 
\end{boxex}

To call a define, the name of the define is used at the beginning of the parenthesis, then each ludeme parameter value is provided such as in the following example of the hop move in the "Ball" piece jumping an enemy piece to capture it.
However, if the middle parameter does not have to be provided the symbol $\sim$ can be used, such as in the example of the hop move in the "Disc" piece, jumping any enemy pieces without removing it.
\begin{boxex}
\begin{ludii}
(piece "Ball" 
    ("Hop" 
        (is Enemy (who at:(between))) 
        (apply (remove (between)))
        (is Empty (to))
    )
) 

(piece "Disc" 
    ("Hop" 
        (is Enemy (who at:(between))) 
        ~ 
        (is Empty (to))
    )
) 
\end{ludii} 
\end{boxex}

As an advice for game designers, the define structure can also be used to help to make sense of some technical descriptions.
For example, the state of a site or the pending values can be used for many purposes and may not be representative of what they really describe.
However in using a define with a descriptive name, this is possible to give them a sense to make a better description.
For example, in the description of Chess, the pending values are used to describe "En Passant" but that is clear only in using the following define 
\begin{boxex}
\begin{ludii}
(define "SetEnPassantLocation"
    (then 
        (set Pending (ahead (last To) Backward))
    )
)
\end{ludii} 
\end{boxex}

Defines can occur anywhere in the game description file even within ludemes but are typically at the top of the file, before the {\tt game} ludeme, to impart some structures. 
Defines can be nested within other defines but not themselves if they have exactly the same parameters called.
The convention is to name each $\it define$ with a single compound word in ``UpperCamelCase'' format.


\section{Ludemeplexes}

Defines constitute {\it ludemeplexes}, which are useful structures of component {\it ludemes}\footnote{Ludemes are ``game memes'' or units of game-related information~\cite{Parlett2016}.}. 
These can be defined either:

\begin{itemize}
    \item globally in their own .def file, in which case they become {\it known defines}, or
    \item locally within any .lud file in which they can be used.
    
\end{itemize}

The current list of known defines is available in the Ludii Language Reference~\cite{LLR:2020}. 


\chapter{\textcolor{gray!95}{Conclusion}}\label{Chapter:Conclusion}

The document will be periodically updated to reflect developments in the game logic of future Ludii releases, and in accordance with feedback from Ludii users. 
Please post any suggestions for improvements to the relevant Ludii Forum (\href{https://ludii.games/forums}{Here}) or email us at \url{ludii.games@gmail.com}.


\chapter{\textcolor{gray!95}{Acknowledgements}}\label{Chapter:Acknowledgements}

This research is part of the European Research Council-funded Digital Ludeme Project (ERC Consolidator Grant \#771292), led by Cameron Browne at Maastricht University's Department of Data Science and Knowledge Engineering.

\includegraphics[scale=0.3,right]{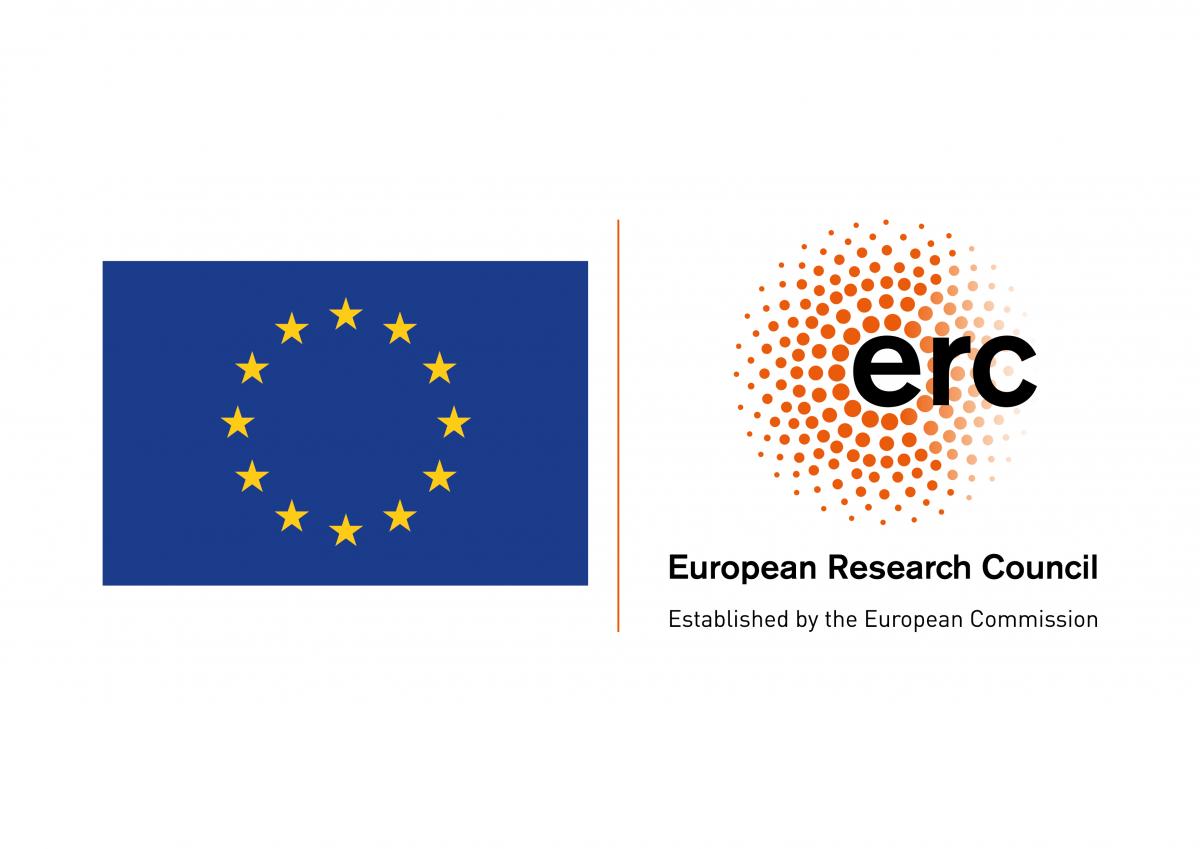}

We would like to thank all the Ludii users who report errors, bugs or possible improvements of Ludii and its language. This work would not be possible without all their support and help.  


\bibliographystyle{apacite}
\renewcommand\bibname{\textcolor{gray!95}{References}}
\bibliography{main}


\end{document}